\documentclass[screen, authorversion, acmsmall]{acmart}
\usepackage{amsmath}
\usepackage{graphicx}
\usepackage{hyperref}
\usepackage[utf8]{inputenc}
\usepackage{natbib}
\usepackage{xcolor}
\usepackage{ifthen}
\usepackage[normalem]{ulem}
\usepackage{svg}
\usepackage{subfiles}
\usepackage{xspace}
\usepackage{cleveref}
\usepackage{wrapfig}
\usepackage{dialogue}
\usepackage[most]{tcolorbox}

\newcommand\sT{\ensuremath{\mathcal{T}}}

\newcommand\R{\ensuremath{\mathbb{R}}} 
\newcommand\eqdef{\ensuremath{\stackrel{\rm def}{=}}} 
\newcommand\refeqn[1]{(\ref{eqn:#1})}

\newcommand\refsec[1]{\hyperlink{#1}{§\ref{sec:#1}:~\textsc{#1}}}
\newcommand\refsecs[2]{\hyperlink{#1}{§\ref{sec:#1}:~\textsc{#1}}, \hyperlink{#2}{§\ref{sec:#2}:~\textsc{#2}}}

\newcommand\reffig[1]{Figure~\ref{fig:#1}}

\ifthenelse{\isundefined{\definition}}{\newtheorem{definition}{Definition}}{}
\ifthenelse{\isundefined{\assumption}}{\newtheorem{assumption}{Assumption}}{}
\ifthenelse{\isundefined{\hypothesis}}{\newtheorem{hypothesis}{Hypothesis}}{}
\ifthenelse{\isundefined{\proposition}}{\newtheorem{proposition}{Proposition}}{}
\ifthenelse{\isundefined{\theorem}}{\newtheorem{theorem}{Theorem}}{}
\ifthenelse{\isundefined{\lemma}}{\newtheorem{lemma}{Lemma}}{}
\ifthenelse{\isundefined{\corollary}}{\newtheorem{corollary}{Corollary}}{}
\ifthenelse{\isundefined{\alg}}{\newtheorem{alg}{Algorithm}}{}
\ifthenelse{\isundefined{\example}}{\newtheorem{example}{Example}}{}
\newcommand{\E}{\ensuremath{\mathbb{E}}} 

\DeclareMathOperator*{\argmin}{arg\,min}

\newcommand\newsection{\clearpage}
\newcommand\sectionauthors[1]{\emph{Authors: #1}\vspace{1em}}

\newcommand\eg{e.g.,~}
\newcommand\ie{i.e.,~}
\newcommand\dash{~---~}

\def\stable{1}

\if\stable1
\else
\fi

\if\stable1
  \newcommand\todo[1]{}

  \newcommand\status[1]{}
  \newcommand\defend[1]{}
\else
  \newcommand\todo[1]{\textcolor{red}{[TODO: #1]}}

  \newcommand\status[1]{\textcolor{gray}{\bf [Status: #1]}}
  \newcommand\defend[1]{\textcolor{magenta}{[Cite: #1]}}
\fi

\if\stable1

\newcommand\pl[1]{}
\newcommand\rb[1]{}
\newcommand\ilevent[1]{}
\newcommand\mx[1]{}
\newcommand\FT[1]{}
\newcommand\tnote[1]{}
\newcommand\dor[1]{}
\newcommand\pw[1]{}
\newcommand\at[1]{}
\newcommand\dora[1]{}

\else

\newcommand\pl[1]{\textcolor{purple}{[Percy: #1]}}
\newcommand\rb[1]{\textcolor{magenta}{[Rishi: #1]}}

\newcommand{\tnote}[1]{{\color{blue}{[TM: #1]}}}

\definecolor{CMpurple}{rgb}{0.6,0.18,0.64}
\definecolor{atcolor}{rgb}{0.83,0.28,0.06}
\definecolor{ballblue}{rgb}{0.13, 0.67, 0.8}

\newcommand\cms{\bgroup\markoverwith{\textcolor{CMpurple}{\rule[.4ex]{2pt}{0.8pt}}}\ULon}

\definecolor{stanford}{rgb}{0.54,0.08,0.08}

\newcommand\mx[1]{\textcolor{orange}{[Michael: #1]}}

\definecolor{mypurple}{rgb}{0.8, 0.18, 1}
\newcommand\dor[1]{\textcolor{mypurple}{[Dor: #1]}}

\newcommand\pw[1]{\textcolor{CMpurple}{[PW: #1]}}
\newcommand\at[1]{\textcolor{atcolor}{[AT: #1]}}

\newcommand\FT[1]{\textcolor{blue}{[Florian: #1]}}

\definecolor{tb12}{rgb}{1.0, 0.13, 0.32}

\newcommand\dora[1]{\textcolor{olive}{[Dora: #1]}}

\definecolor{eamorange}{rgb}{.8,.33,0}

\newcommand\ilevent[1]{\textcolor{orange}{[Isabelle L.: #1]}}

\fi

\usepackage{authblk}

\makeatletter         
\renewcommand\maketitle{
{\raggedright 
\begin{center}
{\Huge \bfseries \sffamily \@title }\\[2ex]
{\@author}\\[2ex] 
\end{center}}}
\makeatother

\renewenvironment{abstract}{%
    \newline
    \itshape
    }
{}

\begin{document}
\title{On the Opportunities and Risks of Foundation Models}
\author{\mbox{Rishi Bommasani*}}
\author{\mbox{Drew A. Hudson}}
\author{\mbox{Ehsan Adeli}}
\author{\mbox{Russ Altman}}
\author{\mbox{Simran Arora}}
\author{\mbox{Sydney von Arx}}
\author{\mbox{Michael S. Bernstein}}
\author{\mbox{Jeannette Bohg}}
\author{\mbox{Antoine Bosselut}}
\author{\mbox{Emma Brunskill}}
% ...

\renewcommand{\shortauthors}{Center for Research on Foundation Models (CRFM)}

\maketitle
\noindent 
\vspace{-0.2in}
\begin{abstract}
AI is undergoing a paradigm shift with the rise of models (\eg~BERT, DALL-E, GPT-3)
trained on broad data (generally using self-supervision at scale) that can be adapted to a wide range of downstream tasks.
We call these models \emph{foundation models} to underscore their critically central yet incomplete character.
This report provides a thorough account of the opportunities and risks of foundation models, ranging from their capabilities (\eg~language, vision, robotic manipulation, reasoning, human interaction) and technical principles (\eg~model architectures, training procedures, data, systems, security, evaluation, theory) to their applications (\eg~law, healthcare, education) and societal impact (\eg~inequity, misuse, economic and environmental impact, legal and ethical considerations).
Though foundation models are based on standard deep learning and transfer learning,
their scale results in new emergent capabilities,
and their effectiveness across so many tasks incentivizes homogenization.
Homogenization provides powerful leverage but demands caution,
as the defects of the foundation model are inherited by all the adapted models downstream.
Despite the impending widespread deployment of foundation models,
we currently lack a clear understanding of how they work, when they fail,
and what they are even capable of due to their emergent properties.
To tackle these questions,
we believe much of the critical research on foundation models
will require deep interdisciplinary collaboration commensurate with their fundamentally sociotechnical nature.
\end{abstract}

\clearpage
\tableofcontents
\clearpage

\hypertarget{introduction}{\section{Introduction}}
\label{sec:introduction}

This report investigates an emerging paradigm for building artificial intelligence (AI) systems
based on a general class of models which we term \emph{foundation models}.\footnote{We chose the term \emph{foundation models} to capture the unfinished yet important status of these models\dash{}see \refsec{naming} for further discussion of the name.}
A foundation model is
any model that is trained on broad data (generally using self-supervision at scale) that
can be adapted (\eg~fine-tuned) to a wide range of downstream tasks;
current examples include BERT \citep{devlin2019bert}, GPT-3 \citep{brown2020gpt3}, and CLIP \citep{radford2021learning}.
From a technological point of view,
foundation models are not new\dash{}they are based on deep neural networks and self-supervised learning,
both of which have existed for decades.
However, the sheer scale and scope of foundation models from the last few years have stretched our imagination of what is possible;
for example, GPT-3 has 175 billion parameters and
can be adapted via natural language prompts to do a passable job on a wide range
of tasks despite not being trained explicitly to do many of those tasks \citep{brown2020gpt3}.
At the same time, existing foundation models have the potential to accentuate
harms, and their characteristics are in general poorly understood.
Given their impending widespread deployment,
they have become a topic of intense scrutiny \citep{bender2021}.

\hypertarget{emergence-homogenization}{\subsection{Emergence and homogenization}}
\label{sec:emergence-homogenization}

The significance of foundation models can be summarized by two words: \emph{emergence} and \emph{homogenization}.
Emergence means that the behavior of a system is implicitly induced rather than explicitly constructed;
it is both the source of scientific excitement and anxiety about unanticipated consequences.
Homogenization indicates the consolidation of methodologies for building machine learning systems across a wide range of applications;
it provides strong leverage towards many tasks but also creates single points of failure. 
To better appreciate emergence and homogenization,
let us reflect on their rise in AI research over the last 30 years.

\begin{figure}[ht]
\centering
\includegraphics[width=\linewidth]{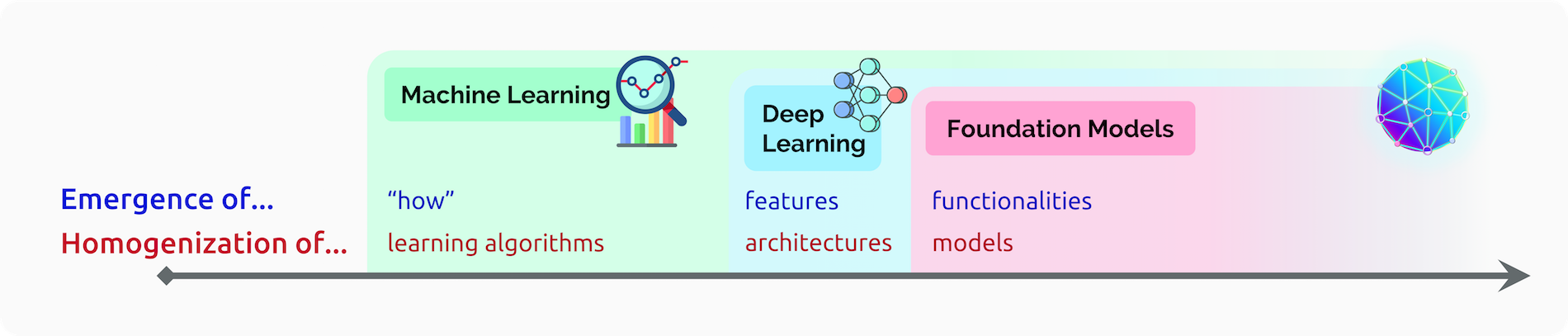}
\caption{
\label{fig:evolution}
The story of AI has been one of increasing \emph{emergence} and \emph{homogenization}.
With the introduction of machine learning,
\emph{how} a task is performed emerges (is inferred automatically) from examples;
with deep learning, the high-level features used for prediction emerge;
and with foundation models, even advanced functionalities such as in-context learning emerge.
At the same time,
machine learning homogenizes learning algorithms (\eg~logistic regression),
deep learning homogenizes model architectures (\eg~Convolutional Neural Networks),
and foundation models homogenizes the model itself (\eg~GPT-3).
}
\end{figure}

\paragraph{Machine learning.}

Most AI systems today are powered by machine learning,
where predictive models are trained on historical data
and used to make future predictions.
The rise of machine learning within AI started in the 1990s,
representing a marked shift from the way AI systems were built previously:
rather than specifying \emph{how} to solve a task,
a learning algorithm would induce it based on data\dash{}\ie
the \emph{how} emerges from the dynamics of learning.
Machine learning also represented a step towards homogenization:
a wide range of applications
could now be powered by a single generic learning algorithm
such as logistic regression.

Despite the ubiquity of machine learning within AI,
semantically complex tasks
in natural language processing (NLP) and computer vision
such as question answering or object recognition,
where the inputs are sentences or images,
still required domain experts to perform ``feature engineering''\dash{}that is,
writing domain-specific logic to convert raw data into higher-level features
(\eg SIFT \citep{lowe1999sift} in computer vision)
that were more suitable for popular machine learning methods.

\paragraph{Deep learning.}

Around 2010, a revival of deep neural networks
under the moniker of \emph{deep learning} \citep{lecun2015deep} started gaining traction in the field of machine learning.
Deep learning was fueled by larger datasets,
more computation (notably, the availability of GPUs),
and greater audacity.
Deep neural networks would be trained on the raw inputs (\eg pixels),
and higher-level features would emerge through training (a process dubbed  ``representation learning'').
This led to massive performance gains on standard benchmarks,
for example, in the seminal work of AlexNet \citep{krizhevsky2012imagenet} on the ImageNet dataset \citep{deng2009imagenet}.
Deep learning also reflected a further shift towards homogenization:
rather than having bespoke feature engineering pipelines for each application,
the same deep neural network architecture could be
used for many applications.

\paragraph{Foundation models.}

Foundation models have taken shape most strongly in NLP, so we focus our story there for the moment.
That said, much as deep learning was popularized in computer vision but exists beyond it,
we understand foundation models as a general paradigm of AI, rather than specific to NLP in any way.
By the end of 2018, the field of NLP was about to undergo another seismic change,
marking the beginning of the era of foundation models.
On a technical level, foundation models are enabled by
\emph{transfer learning} \citep{thrun1998lifelong} and \emph{scale}.
The idea of transfer learning is to take the ``knowledge'' learned from one task (\eg object recognition in images)
and apply it to another task (\eg activity recognition in videos).
Within deep learning, \emph{pretraining} is the dominant approach to transfer learning:
a model is trained on a surrogate task (often just as a means to an end) and then adapted to
the downstream task of interest via \emph{fine-tuning}.

Transfer learning is what makes foundation models possible,
but scale is what makes them powerful.
Scale required three ingredients:
(i) improvements in computer \emph{hardware}\dash{}\eg~GPU throughput and memory have increased 10$\times$ over the last four years (\refsec{systems});
(ii) the development of the Transformer model architecture \citep{vaswani2017attention}
that leverages the parallelism of the hardware to train much more expressive models than before (\refsec{modeling});
and (iii) the availability of much more training data.

The importance of the availability of data and the ability to harness it cannot be underestimated.
Transfer learning with annotated datasets has been common practice for at least a decade,
for example, pretraining on the ImageNet dataset \citep{deng2009imagenet} for image classification
in the computer vision community.
However, the non-trivial cost of annotation imposes a practical limit on the benefits of pretraining.  

In \emph{self-supervised learning} on the other hand,
the pretraining task is derived automatically from unannotated data.\footnote{
Interestingly, self-supervised learning
was dominant in the early days of deep learning
\citep{hinton2006fast},
but was for a decade largely overtaken by pure supervised learning as labeled datasets became larger.}
For example, the masked language modeling task used to train BERT \citep{devlin2019bert} is to
predict a missing word in a sentence given its surrounding context (\eg~\emph{I like \rule{1cm}{0.15mm} sprouts}).
Self-supervised tasks are not only more scalable, only depending on unlabeled data,
but they are designed to force the model to predict parts of the inputs, 
making them richer and potentially more useful than models trained on a more limited label space.

There had been considerable progress in self-supervised learning dating back to word embeddings
\citep{turian2010word,mikolov2013efficient,pennington2014glove},
which associated each word with a context-independent vector,
provided the basis for a wide range of NLP models.
Shortly thereafter, self-supervised learning based on autoregressive language modeling
(predict the next word given the previous words) \citep{dai2015semi}
became popular.
This produced models that represented words in context, such as
GPT \citep{radford2018improving},
ELMo \citep{peters2018elmo},
and ULMFiT \citep{howard2018universal}.\footnote{The prescient work of \citet{collobert2008unified}
is related: they trained on a scalable task akin to masked language modeling jointly with downstream tasks,
rather than producing a single foundation model that can be adapted after the fact to downstream tasks.}

The next wave of developments in self-supervised learning\dash{}BERT \citep{devlin2019bert}
GPT-2 \citep{radford2019language},
RoBERTa \citep{liu2019roberta},
T5 \citep{raffel2019exploring},
BART \citep{lewis2020bart}\dash{}quickly followed,
embracing the Transformer architecture,
incorporating more powerful deep bidirectional encoders of sentences,
and scaling up to larger models and datasets.

While one can view this last wave of technical developments purely through the lens of self-supervised learning,
there was a sociological inflection point around the introduction of BERT.
Before 2019, self-supervised learning with language models was essentially a \emph{subarea} in NLP,
which progressed in parallel to other developments in NLP.
After 2019, self-supervised learning with language models became more of a \emph{substrate} of NLP,
as using BERT has become the norm.
The acceptance that a single model could be useful for such a wide range of tasks
marks the beginning of the era of foundation models.

Foundation models have led to an unprecedented level of \emph{homogenization}:
Almost all state-of-the-art NLP models
are now adapted from one of a few foundation models,
such as BERT, RoBERTa, BART, T5, etc.
While this homogenization produces extremely high leverage
(any improvements in the foundation models can lead to immediate benefits
across all of NLP), it is also a liability; all AI systems might
inherit the same problematic biases of a few foundation models \citep[][\textit{inter alia}]{bolukbasi2016, caliskan2017, abid2021})\dash{}see \refsecs{fairness}{ethics} for further discussion.

We are also beginning to see a homogenization across research communities.
For example, similar Transformer-based sequence modeling approaches
are now applied to text \citep{devlin2019bert, radford2019language, raffel2019exploring},
images \citep{visual_transformer, chen2020imagegpt}, speech \citep{Liu2020MockingjayUS}, tabular data
\citep{Yin2020TaBERTPF},
protein sequences \citep{rives2021},
organic molecules \citep{rothchild2021c5t5},
and reinforcement learning \citep{Chen2021DecisionTR, Janner2021ReinforcementLA}.
These examples point to a possible future where we have
a unified set of tools for developing foundation models across a wide range of modalities \citep{Tamkin2021DABS}.

\begin{figure}[t]
\centering
\includegraphics[width=\linewidth]{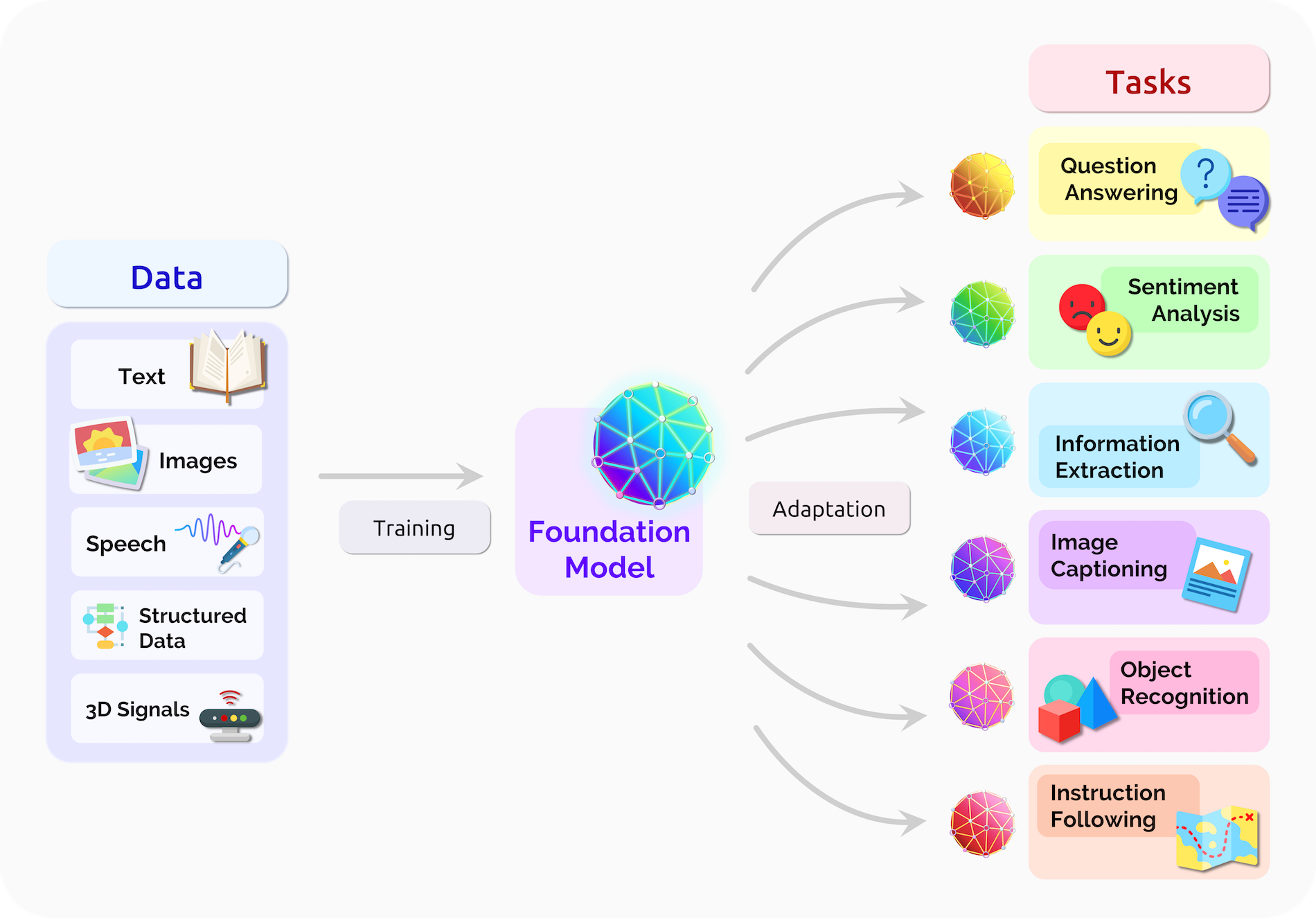}
\caption{
\label{fig:framework}
A foundation model can centralize the information from all the data from various modalities.
This one model can then be adapted to a wide range of downstream tasks.
}
\end{figure}

Besides the homogenization of approaches,
we also see the homogenization of actual models across research communities in the form of \emph{multimodal models}\dash{}\eg{}foundation models trained on language and vision data
\citep{luo2020univl,kim2021vilt,cho2021unifying,ramesh2021zeroshot,radford2021learning}.
Data is naturally multimodal in some domains---\eg~medical images, structured data, clinical text in healthcare (\refsec{healthcare}).
Thus, multimodal foundation models are a natural way of fusing all the relevant information about a domain,
and adapting to tasks that also span multiple modes (\reffig{framework}).

Foundation models have also led to surprising emergence which results from scale.
For example, GPT-3 \citep{brown2020gpt3}, with 175 billion parameters compared to GPT-2's 1.5 billion,
permits \emph{in-context learning}, in which the language model can be adapted to a downstream task simply by
providing it with a \emph{prompt} (a natural language description of the task),
an emergent property that was neither specifically trained for nor anticipated to arise.


Homogenization and emergence interact in a potentially unsettling way.
Homogenization could potentially provide enormous gains for many
domains where task-specific data is quite limited\dash{}see the  opportunities presented in several such domains (\eg~\refsec{healthcare}, \refsec{law}, \refsec{education});
on the other hand, any flaws in the model are blindly inherited by all adapted models (\refsec{fairness}, \refsec{ethics}).
Since the power of foundation models comes from their \emph{emergent qualities}
rather than their explicit construction,
existing foundation models are hard to understand (\refsec{evaluation}, \refsec{theory}, \refsec{interpretability})
and they have unexpected failure modes (\refsec{security}, \refsec{robustness}).
Since \emph{emergence} generates substantial uncertainty over the capabilities and flaws of foundation models,
aggressive homogenization through these models is risky business.
Derisking is the central challenge in the further development of foundation models
from an ethical (\refsec{ethics}) and AI safety (\refsec{ai-safety}) perspective.

\hypertarget{naming}{\subsubsection{Naming.}}
\label{sec:naming}

We introduce the term \textit{foundation models} to fill a void in describing the paradigm shift we are witnessing; we briefly recount some of our reasoning for this decision.
Existing terms (\eg \emph{pretrained model}, \emph{self-supervised model}) partially capture the technical dimension of these models,
but fail to capture the significance of the paradigm shift in an accessible manner for those beyond machine learning.
In particular, foundation model designates a model class that are distinctive in their sociological impact and how they have conferred a broad shift in AI research and deployment.
In contrast, forms of pretraining and self-supervision that technically foreshadowed foundation models fail to clarify the shift in practices we hope to highlight.  

Additionally, while many of the iconic foundation models at the time of writing are language models, the term \textit{language model} is simply too narrow for our purpose: as we describe, the scope of foundation models goes well beyond language.
We also considered terms such as \emph{general-purpose model} and \emph{multi-purpose model} that capture the important aspect that these models can serve multiple downstream tasks,
but both fail to capture their unfinished character and the need for adaptation.
Terms such as \emph{task-agnostic model} would capture the manner of training, but fail to capture the significant implication to downstream applications.

We chose the new term \emph{foundation models} to identify the models and the emerging paradigm that are the subject of this report.
In particular, the word ``foundation'' specifies the role these models play:
a foundation model is itself incomplete but serves as the common basis from which many task-specific models are built via adaptation.
We also chose the term ``foundation" to connote the significance of architectural stability, safety, and security:
poorly-constructed foundations are a recipe for disaster and well-executed foundations are a reliable bedrock for future applications.
At present, we emphasize that we do not fully understand the nature or quality of the foundation that foundation models provide; we cannot characterize whether the foundation is trustworthy or not.
Thus, this is a critical problem for researchers, foundation model providers, application developers who rely on foundation models, policymakers, and society at large to address.

\hypertarget{ecosystem}{\subsection{Social impact and the foundation models ecosystem}}
\label{sec:ecosystem}

\begin{figure}[t]
\centering
\includegraphics[width=\linewidth]{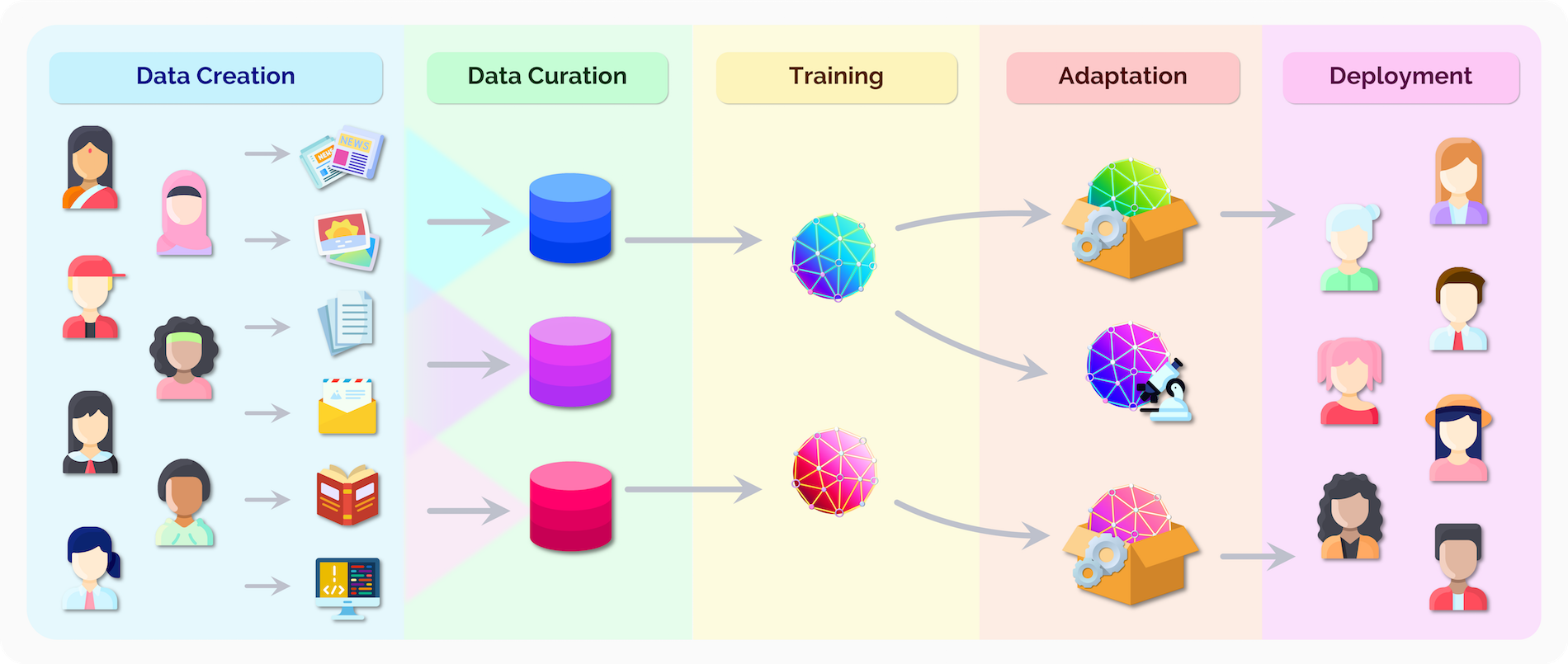}
\caption{
\label{fig:ecosystem}
Before reasoning about the social impact of foundation models,
it is important to understand that they are part of a broader ecosystem
that stretches from data creation to deployment.
At both ends, we highlight the role of people as the ultimate source of data into training of a foundation model,
but also as the downstream recipients of any benefits and harms.
Thoughtful data curation and adaptation should be part of the responsible development of any AI system.
Finally, note that the deployment of adapted foundation models is a decision separate from their construction,
which could be for research.
}
\end{figure}

Foundation models are scientifically interesting due to their impressive performance and capabilities, 
but what makes them critical to study is the fact that they are quickly being integrated into real-world deployments of AI systems with far-reaching consequences on people.
For example, Google search, which boasts 4 billion users,
now depends on foundation models like BERT \citep{devlin2019bert} as one of its
signals.\footnote{https://blog.google/products/search/search-language-understanding-bert/}

We must thus pause and ask:
What is the nature of this social impact?
In this report, we address many aspects of this question:
the potential exacerbation of social inequities (\refsec{fairness}),
the economic impact due to increased capabilities (\refsec{economics}),
the environmental impact due to increased computation demands (\refsec{environment}),
potential concerns of amplifying disinformation (\refsec{misuse}),
legal ramifications due to powerful generative capabilities (\refsec{legality}),
ethical issues resulting from homogenization,
and the broader political economy in which foundation models are developed and deployed (\refsec{ethics}).
Given the protean nature of foundation models and their unmapped capabilities,
how can we responsibly anticipate and address the ethical and societal considerations they raise?
A recurring theme is that it is easier
to reason about the social impact of specific systems deployed to specific users
than it is to reason about the social impact of foundation models,
which could be adapted to any number of unforeseen downstream systems.

Before attempting to answer these questions,
we need to lay some groundwork.
First, let us distinguish between \emph{research} on foundation models and \emph{deployment} of foundation models.
Most of what is publicly known is foundation models research\dash{}through academic papers,
demonstrations, and progress on leaderboards.
While the production of knowledge can play a vital role in shaping the future,
the direct social impact is through the actual deployment of these models,
which is governed by proprietary practices on often private data.
Sometimes the deployment is through new products\dash{}\eg~GitHub's
Copilot\footnote{https://copilot.github.com/} based on OpenAI's Codex model
\citep{chen2021evaluating}, but often,
it is through upgrades to existing products (\eg~Google search using BERT).
Research models are often not extensively tested and might have unknown failure modes;
warning labels should be placed on research models that are not fit to deploy.
On the other hand, deployed foundation models that actually affect people's lives
should be subject to much more rigorous testing and auditing.

To further understand the research and deployment of foundation models,
we must zoom out and consider the full \emph{ecosystem} that these foundation models inhabit,
from data creation to actual deployment.
It is important to note that the foundation model is only one component (though
an increasingly important component) of an AI system.
Simplifying, we can think about the ecosystem of a foundation model in terms of
sequence of stages, extending the training and adaptation stages from before.\footnote{In practice, the end of the pipeline is followed by monitoring, and feedback is
used to readjust the previous stages.}
Appropriately, as we're interested in social impact,
\emph{people} occupy both ends of the pipeline.
This ecosystem view allows us to see that different questions about foundation models
(\eg~whether a foundation model is ethical) should actually be answered with respect to different stages.
\begin{enumerate}
  \item \textbf{Data creation}:
    Data creation is fundamentally a human-centric process:
    all data is created by people and most data is at least implicitly about people.
    Sometimes data is created by people for other people in the form of emails, articles, photos, etc.,
    and sometimes it is a measurement of people (\eg~genomic data)
    or a measurement of the environment people live in (\eg~satellite images).
    It is important to note that all data has an owner and is created with a purpose
     (where that purpose may or may not include training a foundation model).

  \item \textbf{Data curation}:
    Data is then curated into datasets.
    There is no single natural distribution of data;
    even the most permissive Internet crawl requires some selection and post-filtering.
    Ensuring data relevance and quality while respecting legal and ethical constraints is critical but challenging.
    While this is recognized in industry, it is underappreciated in AI research (\refsec{data}).

  \item \textbf{Training}:
    Training foundation models on these curated datasets\footnote{A foundation model (\eg~Codex) can also be trained with another
    model (\eg~GPT-3) as a starting point.}
    is the celebrated centerpiece in AI research,
    though it is only one of many stages.

  \item \textbf{Adaptation}:
    In the context of machine learning research,
    adaptation is about creating a new model based on the foundation model
    that performs some task (\eg~document summarization).
    For deployment, adaptation is about creating a system,
    which requires potentially many different modules, custom rules
    (\eg~restrictions on the output space) or classifiers (\eg~for toxicity
    classification),
    and combination with other complementary signals (\eg~a question answering
    model's generated answers would be validated against relevant documents).
    For example, a problematic model capable of generating toxic content might be tolerable
    if appropriate precautions are taken downstream.
    The extra application-specific logic is crucial for mitigating harms.

  \item \textbf{Deployment}:
    The direct social impact of an AI system occurs when it is deployed to people.
    Though we would not want to deploy potentially harmful foundation models trained on questionable data,
    there might still be value in permitting them in research to advance scientific understanding,
    though one must still exercise caution.
    More generally, it is standard practice in large-scale deployments
    to conduct gradual releases, where deployment happens to an increasing fraction of users;
    this can partially mitigate any potential harms.
\end{enumerate}

While this report is about foundation models,
it is important to note that many of the impacts come from decisions made in
other stages in the pipeline,
and thoughtful monitoring and intervention is needed at every stage.
While large organizations might own the entire pipeline,
each stage could be performed by a different organization,
\eg~a company which specializes in creating custom foundation models for
various domains that application-developers can use.

\paragraph{Think ecosystem, act model.}
While the social impact depends on the whole ecosystem,
it is still important to be able to reason about the social implications of a foundation model,
given that many researchers' and practitioners' purview is restricted to the training stage.
This is difficult because foundation models are unfinished intermediate
objects that can be adapted to many downstream applications, sometimes by an
entirely different entity for unforeseen purposes.
What we need are two things:
(i) surrogate metrics for a representative set of potential downstream evaluation (\refsec{evaluation}),
and (ii) a commitment to documenting these metrics \citep{Mitchell_2019}
similar to data sheets for materials such as metals and plastics,
which can be adapted to many downstream use cases.

Characterizing the potential downstream social impact of foundation models
is challenging and demands a deep understanding of both the technological ecosystem
and of society.
One cannot fully assess the harms (\refsec{fairness}) of a foundation model without recognizing how it will be deployed,
and one cannot just define automatic metrics without considering the rich social and historical context.

\hypertarget{developers}{\subsection{The future of foundation models}}
\label{sec:developers}

Foundation models have demonstrated raw potential,
but we are still in the early days.
Despite their deployment into the real world,
these models are very much research prototypes that are poorly understood.
Even the professional norms\dash{}what Robert Merton calls the ethos of science \citep{merton1979normative}\dash{}around foundation models are underdeveloped.
For example, there is lack of agreement on basic questions such as when models are ``safe'' to release
or how the community should react in response to methodological misconduct.
Given that the future of foundation models is thus filled with uncertainty,
a big question is: who will determine this future?

\paragraph{Disciplinary diversity.}

The technology behind foundation models is based on decades of research in
machine learning, optimization, NLP, computer vision, and other fields.
These technical contributions have come from both academia and industrial research labs.
However, research on building foundation models themselves has occurred almost exclusively in industry\dash{}big tech
companies such as Google, Facebook, Microsoft, or Huawei, or startups such as
OpenAI or AI21 Labs, though AI2 is a notable exception \citep{peters2018elmo,zellers2019neuralfakenews}.

The furious pace of technological progress and the entrenchment due to centralization
raise powerful concerns that demand the attention of humanists and social scientists in addition to technologists.
We should not rely on post-hoc audits of ethical and social consequences,
conducted only after the technical architecture and deployment decisions have been made.
We instead need to infuse social considerations and ethical design
deeply into the technological development of foundation models and
their surrounding ecosystem from the start.
Academic institutions are unique in that they host the widest set of disciplines under one roof,
thus bringing together computer scientists, social scientists, economists, ethicists, legal scholars, etc.
Given the importance of disciplinary diversity in understanding and solving problems that combine technical, ethical, legal, social, and political dimensions \citep{hong2004groups,solomon2006norms,steel2018multiple},
we therefore see academia as playing a crucial role
in developing foundation models in such a way to promote their social benefit and mitigate their social harms,
as well as determining the contexts under which actions in each of the stages of the ecosystem (\refsec{ecosystem}) ranging from data curation to deployment should be strictly prohibited.

\paragraph{Incentives.}

The political economy in which foundations models are designed, developed, and deployed provides an inevitable incentive structure for decision-making at every stage. How people and institutions respond to incentives is an elementary lesson of economics.
Market-driven commercial incentives can align well with social benefit:
making foundation models more accurate, reliable, safe, and efficient while searching for a wide variety of potential use cases
can produce a great deal of social utility.
However, commercial incentives can also lead to market failures and underinvestment in domains where shareholders are unable to capture the value of innovation. 
Just as the pharmaceutical industry has little incentive to devote significant resources to the research and development of malaria treatments, because poor people cannot afford medications,\footnote{See https://www.gatesfoundation.org/about/our-role.} the tech industry has little incentive to devote significant resources to technologies designed for improving the condition of poor and marginalized people \citep{reich2021system}. What's more, commercial incentives can lead companies to ignore social externalities \citep{acemoglu2021redesigning, reich2021system} such as the technological displacement of labor, the health of an informational ecosystem required for democracy, the environmental cost of computing resources, and the profit-driven sale of technologies to non-democratic regimes.
Finally, there is little incentive for any given company to create
an open, decentralized ecosystem for developing foundation models
that encourages broad participation.

In contrast,
the long-standing and deeply-seated research mission of universities is the
production and dissemination of knowledge and creation of global public goods \citep{kerr2001university,rhoten2011knowledge,nussbaum2010not}. 
We believe that academia is distinctively positioned to
shape the development of foundation models to ensure that we capture
directions with potentially large social benefit that might not otherwise be prioritized by industry.

\paragraph{Loss in accessibility.}

Unfortunately, academia has not been able to participate in the fullest way possible
due to the loss in accessibility.
One of the often overlooked effects of the deep learning revolution
was the increase in reproducibility and open science:
it increasingly became the norm to publicly release code and datasets, and
packages such as TensorFlow \citep{abadi2016tensorflow} and PyTorch
\citep{paszke2019pytorch} made it much easier for people to collaborate and
build off of each other’s work. 
Initiatives like the ML Reproducibility
Challenge\footnote{https://paperswithcode.com/rc2020} as well as
reproducibility checklists adopted by major conferences
\citep{pineau2020improving}, alongside platforms like CodaLab 
Worksheets\footnote{https://worksheets.codalab.org/} helped advance community
standards for reproducibility.
This resulted in a surge in technological innovation and progress.

Foundation models start to roll back this positive trend.
Some models (\eg{}GPT-3) are not released at all (only API access to a limited pool of people).
Even datasets (\eg{}for GPT-2) are not released.
While trained models may be available (\eg{}BERT),
the actual training of foundation models is
unavailable to the vast majority of AI researchers,
due to the much higher computational cost and the complex engineering requirements.

Some meaningful research can still be done by training smaller models within reach of an academic budget,
and indeed the surprisingly regularity predicted by scaling laws \citep{kaplan2020} make this a viable strategy
for cases where the differences due to scale are quantitative (\eg~accuracy goes up).
However, due to the emergent nature of these foundation models,
some functionalities like in-context learning have only been demonstrated in models of sufficient size,
so scale is needed to even ask the right questions.

It is also possible to productively study pre-existing models that have been released; indeed, this has
led to a large subcommunity within NLP for probing these models \citep{rogers2020primer, manning2020emergent}.
Having access to existing models can be useful for powering downstream applications or identifying defects (\eg{}bias),
but this might not be enough for us to design better architectures or training objectives for foundation models
that can fix these defects (\eg{}mitigate the bias).
It is worth reflecting on how much of NLP research today is based on BERT,
a particular (and somewhat arbitrary) foundation model.
Given the need to infuse social awareness and ethical design into the construction of these models,
it is possible that we need to build foundation models that look quite different from what exists today.
This will demand intense experimentation at scale.

Community efforts such as EleutherAI\footnote{https://www.eleuther.ai/} and Hugging Face's BigScience project\footnote{https://bigscience.huggingface.co/}
are attempting to train large foundation models,
but the gap between the private models that industry can train
and the ones that are open to the community will likely remain large if not grow.
Further, today startups (OpenAI, Anthropic, AI21 Labs, etc.) are much more well-resourced than academia
and can therefore still afford to train the largest foundation models (\eg~OpenAI's GPT-3).
However, big tech companies are on a completely different level
in terms of resources, especially in terms of the infrastructure, users, and data that come from their market position.
The fundamental centralizing nature of foundation models
means that the barrier to entry for developing them will continue to rise,
so that even startups, despite their agility, will find it difficult to compete,
a trend that is reflected in the development of search engines \citep{radinsky2015data}.

One way to close the resource gap is for the government to invest in public infrastructure.
We can look to Big Science projects such as the Hubble Space Telescope and the Large Hadron Collider as inspiration,
where substantial investment made possible fundamental scientific discoveries which wouldn't have been possible.
One can imagine a similar infrastructure for computing, from which academic research on foundation models would greatly benefit.
In the US, the nascent National Research Cloud initiative\footnote{https://hai.stanford.edu/policy/national-research-cloud} is a step in this direction.

Another complementary approach is to rely on volunteer computing,
in which any of the billions of computing devices (nodes) can connect to a central server and contribute computation.
The Folding@home project has successfully implemented this approach for simulating protein dynamics \citep{beberg2009folding}.
Recently, the Learning@home project is attempting to harness volunteer computing for training foundation models
\citep{Ryabinin2020Learninghome}.
The high latency connections between nodes and the high bandwidth requirements for training foundation models
make this an open technical challenge.

\paragraph{Summary.}
There are tremendous economic incentives to push the capabilities and scale of foundation models,
so we anticipate steady technological progress over the coming years.
But the suitability of a technology relying largely on emergent behavior
for widespread deployment to people is unclear.
What is clear that we need to be cautious,
and that now is the time to establish the professional norms that will enable the responsible research and deployment
of foundation models.
Academia and industry need to collaborate on this:
industry ultimately makes concrete decisions about how foundation models will be deployed,
but we should also lean on academia,
with its disciplinary diversity and non-commercial incentives around knowledge production and social benefit,
to provide distinctive guidance on the development and deployment of foundation models
that is both technically and ethically grounded.

\hypertarget{overview}{\subsection{Overview of this report}}
\label{sec:overview}

\begin{figure}[t]
\centering
\includegraphics[width=\linewidth]{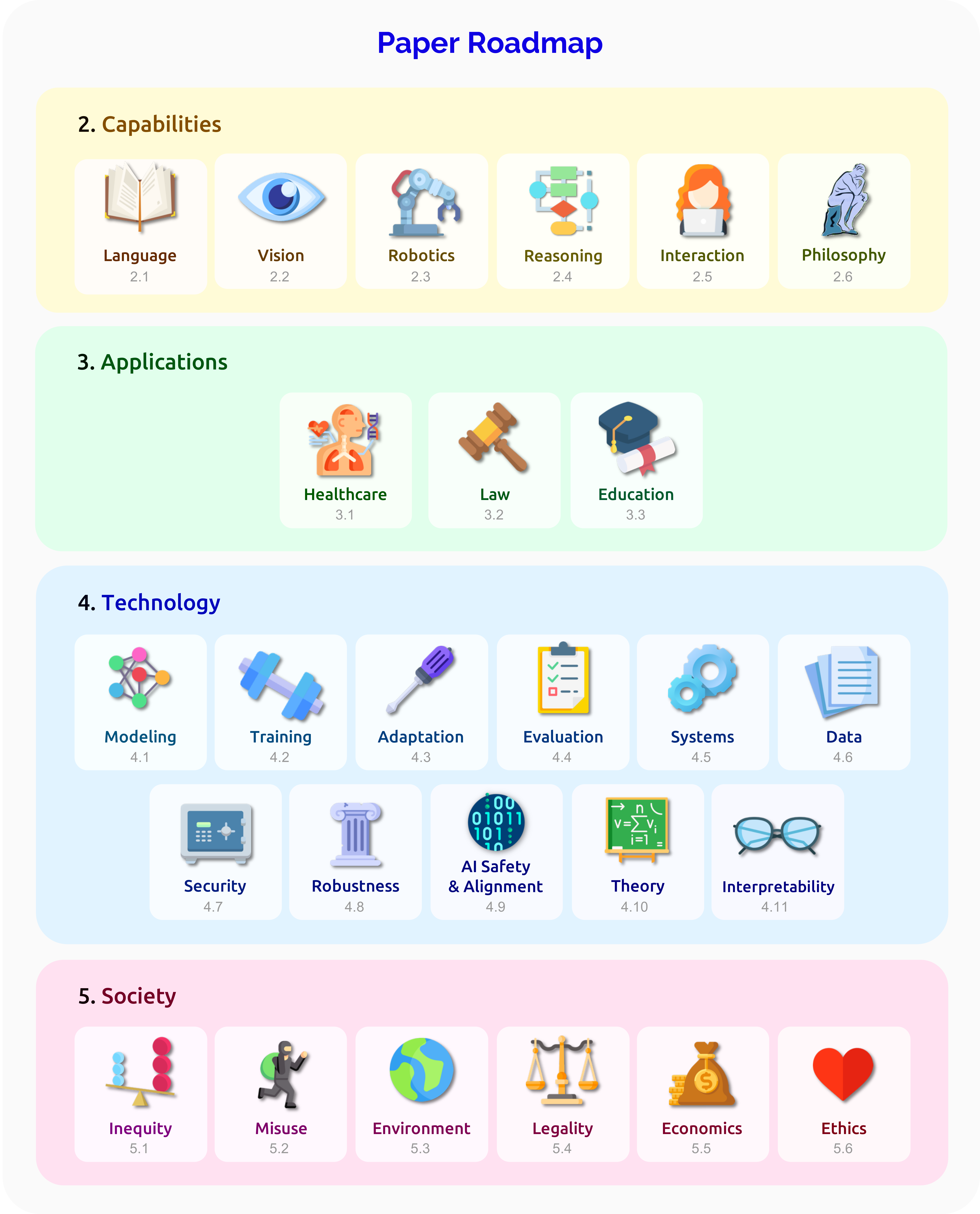}
\caption{This report is divided into four parts: capabilities, applications, technology, and society,
where each part contains a set of sections,
and each section covers one aspect of foundation models.
\label{fig:roadmap}
}
\end{figure}

In March 2021, we created an informal community at Stanford University of students, faculty, and researchers
interested in some aspect of foundation models.\footnote{This community
led to the founding of the \emph{Center for Research on Foundation Models (CRFM)},
a new interdisciplinary initiative at the Stanford Institute for Human-Centered AI (HAI).}
From the very beginning, the community included not just AI researchers,
but those eager to apply foundation models to their domain (\eg{}healthcare and law),
as well as those who were interested in societal concerns (\eg{}ethics and economics).
As discussions progressed,
we noticed that there were many gaps in mutual understanding\dash{}how the
technology worked, how industry develops foundation models, how to think about the ethical
concerns, etc.,
and existing literature only covered bits and pieces.
We wanted to therefore provide a fuller picture of foundation models, identify
opportunities and risks, and establish a constructive vision for the future responsible development of foundation models.

The writing of this report was an experiment:
we had over 100 people from different backgrounds come together to write a
single report covering a wide range of aspects of foundation models.
A large part of this report is a survey of existing work, but through many discussions,
we have unified it in one report to highlight all the interdisciplinary connections.

\paragraph{Structure.}

The report is divided into 26 sections, each discussing one aspect of foundation models.
The sections are grouped into four parts:
capabilities (\refsec{capabilities}),
applications (\refsec{applications}),
technology (\refsec{technology}),
and society (\refsec{society}),
although there are many connections across sections.
These connections highlight an integrated approach
in which the technologies and capabilities are developed in a way that is sensitive to
real societal concerns, while being inspired by and grounded out in applications.

While we have sought to capture most of the important topics surrounding
foundation models, this report will inevitably be incomplete,
especially as the field evolves quickly.
For example, many applications (\eg~natural sciences, music, finance, agriculture) are not included,
though they are as likely to be affected as the applications we have chosen to discuss.
It would also be interesting to study how foundation models relate to research in neuroscience, cognitive science, and psychology to explain intelligence and aid efforts in computational social science to understand society.

\paragraph{Author Contributions}
Percy Liang initiated and conceptualized the framing and structure of the overall report. 
He and Rishi Bommasani worked together to lead the decentralized writing effort and provided guidance on individual sections. 
Drew A. Hudson created all the figures in the report, discussing their structure and content with the authors of each section. 
Each of the 26 sections of this report was written by a subset of authors, whose names are listed at the beginning of each section. 
There were, however, many discussions that spanned multiple sections, so the actual contributions to each section generally came from a broader set. 
Finally, we note that not all the views expressed in this report are held by all the authors.

\newcommand*{\rev}{\textcolor{blue}}

\hypertarget{overview-capabilities}{\subsubsection{Overview of capabilities}}
\label{sec:overview-capabilities}

Foundation models acquire various \emph{capabilities} that can power applications.
We have chosen to discuss five potential capabilities:
the ability to process different modalities (\eg language, vision), to affect
the physical world (robotics), to perform reasoning, and to interact with
humans (interaction).  Finally, we conclude with a philosophical discussion of potential limits on their capabilities.

\paragraph{\hyperref[sec:language]{§\ref{sec:language}:~Language.}}

NLP as a field has blazed the trail for foundation models.
While these models dominate standard benchmarks, there is a clear gap between the capabilities these models acquire currently and those that characterize language as a complex system for human communication and thought.
In response to this, we emphasize the full range of \textit{linguistic variation} (\eg~different styles, dialects, languages), which poses an opportunity and challenge given some variants are data-limited.
Further, child \textit{language acquisition} is more sample efficient than the training of foundation models; we examine how signals beyond text and grounding may help to bridge this gap.
Both of these characteristics of language provide clear directions for future foundation models research.

\paragraph{\hyperref[sec:vision]{§\ref{sec:vision}:~Vision.}}
Computer vision led the adoption of deep learning in AI \cite{russakovsky2015imagenet}, demonstrating that models pretrained on large annotated datasets can transfer to numerous downstream settings.
Now, pretraining on web-scale raw data instead of curated datasets, foundation models are on the rise in computer vision~\citep[\eg][]{radford2021learning}.
These models have shown promising results for standard tasks in the field, like image classification and object detection, and training on \textit{multimodal and embodied} data beyond images may enable progress on significant challenges (\eg~3D geometric and physical understanding, commonsense reasoning).
We also discuss some of the key challenges in modeling (\eg~the ability to scale effectively to videos) and evaluation (\eg~the measurement of higher-order capabilities) along with the applications (\eg~ambient intelligence for healthcare) and societal considerations (\eg~surveillance) that will determine the impact of foundation models for computer vision going forward.

\paragraph{\hyperref[sec:robotics]{§\ref{sec:robotics}:~Robotics.}}

A longstanding goal of robotics research is to develop ``generalist'' robots capable of performing myriad tasks across physically diverse environments.
Unlike language and vision, which have led the way with foundation models both due to the abundance of raw data to train these models on and the availability of virtual applications to apply these models to, robotics faces fundamental challenges due to being anchored to the physical world.
The principal challenge in developing \textit{new types of foundation models for robotics}\dash{}different in nature than their language and vision counterparts\dash{}is acquiring \textit{sufficient data} of the \textit{right form} that is conducive to learning: we explore how plentiful data (\eg~generic videos of humans, amongst others) that is not specific to particular environments and across modalities (\eg~language, vision) may help to bridge this gap.
These new robotic foundation models could allow for easier \textit{task specification and learning}, ushering in new applications (\eg~better robotic assistance for household tasks) and heightening the importance of \textit{robustness and safety} (\eg~formal safety evaluation).

\paragraph{\hyperref[sec:reasoning]{§\ref{sec:reasoning}:~Reasoning and search.}}

Reasoning and search problems such as theorem proving and program synthesis have been long-standing challenges in AI. The combinatorial search space renders traditional search-based methods intractable.
However, humans are known to operate intuitively even in the most mathematical of domains~\citep{LakoffNunez00},
and indeed existing work such as AlphaGo have already shown that deep neural networks can be effective in guiding the search space.
But humans also transfer knowledge across tasks, facilitating
much more efficient adaptation and the ability to reason more abstractly.  Foundation models offer the possibility of closing this gap: their multi-purpose nature along with their strong generative and multimodal capabilities offer new leverage for controlling the combinatorial explosion inherent to search.

\paragraph{\hyperref[sec:interaction]{§\ref{sec:interaction}:~Interaction.}}
Foundation models show clear potential to transform the developer and user experience for AI systems: foundation models lower the difficulty threshold for \textit{prototyping and building} AI applications due to their sample efficiency in adaptation, and raise the ceiling for \textit{novel user interaction} due to their multimodal and generative capabilities.
This provides a synergy we encourage going forward: developers can provide applications that better fit the \textit{user's needs and values}, while introducing far more dynamic forms of interaction and opportunities for \textit{feedback}.

\paragraph{\hyperref[sec:philosophy]{§\ref{sec:philosophy}:~Philosophy of understanding.}}

What could a foundation model come to understand about the data it is trained on? Focusing on the case of natural language, we identify different positions on the nature of understanding and explore their relevance for our central question. Our tentative conclusion is that skepticism about the capacity of future foundation models to understand natural language may be premature, especially where the models are trained on multi-modal data.

\hypertarget{overview-applications}{\subsubsection{Overview of applications}}
\label{sec:overview-applications}

At present, foundation model research is largely confined to computer science and AI, with the impact of foundation models and the applications they support largely being centered in the tech industry.
Moving forward, foundation models present clear potential to transform and extend the reach of AI across many sectors beyond the tech industry, suggesting a more pervasive effect on people's lives.
While there is a multitude of applications and domains to consider, we we have chosen three applications \dash{} healthcare, law, and education \dash{} because they represent foundational pillars of our society.  
For foundation models to significantly contribute to these application domains, models will require specific capabilities (\refsec{capabilities}) as well as technical innovation (\refsec{technology}) to account for the unique considerations in each domain.
Further, since these domains are critical to societal function (\refsec{society}), applying foundation models in these domains requires engaging with deeply sociotechnical matters such as those those pertaining to data (\refsec{data}), privacy (\refsec{security}), interpretability (\refsec{interpretability}),  fairness (\refsec{fairness}) and ethics (\refsec{ethics}).

\paragraph{\hyperref[sec:healthcare]{§\ref{sec:healthcare}:~Healthcare and biomedicine.}}

Healthcare tasks (\eg~patient care via disease treatment) and biomedical research (\eg~scientific discovery of new therapies) require expert knowledge that is limited and expensive. 
Foundation models present clear opportunities in these domains due to the \textit{abundance of data} across \textit{many modalities} (\eg~images, text, molecules) to train foundation models, as well as the value of improved sample efficiency in adaptation due to the cost of expert time and knowledge. 
Further, foundation models may allow for improved \textit{interface design} (\refsec{interaction}) for both healthcare providers and patients to interact with AI systems, and their generative capabilities suggest potential for \textit{open-ended research problems} like drug discovery. 
Simultaneously, they come with clear risks (\eg~exacerbating historical biases in medical datasets and trials). 
To responsibly unlock this potential requires engaging deeply with the sociotechnical matters of data sources and privacy as well as model interpretability and explainability, alongside effective regulation of the use of foundation models for both healthcare and biomedicine.

\paragraph{\hyperref[sec:law]{§\ref{sec:law}:~Law.}}

Legal applications require that attorneys read and produce long coherent
narratives that incorporate shifting contexts and decipher ambiguous legal standards.
Foundation models may provide benefits in this domain: \textit{ample data} exists in the form of legal documents and their generative capabilities are well-suited to the \textit{many generative tasks required in law}, but significant improvements are required for foundation models to be able to reliably \textit{reason over various sources} of information to generate \textit{truthful} long-form documents.
As is the care in healthcare (\refsec{healthcare}), the sample efficiency of adaptation for foundation models is of heightened value given the costs of expert time and knowledge in the legal domain, which may allow for the \textit{re-allocation of expertise} towards pressing problems of justice and government service.
The responsible development of foundation models for law will require specific consideration of privacy, and highlights core limitations of existing foundation models that will require fundamental advances with respect to \textit{provenance} for their behavior and \textit{guarantees} for the factuality of their generation.

\paragraph{\hyperref[sec:education]{§\ref{sec:education}:~Education.}} 

Education is a complex and subtle domain; effective teaching involves reasoning about student cognition and should reflect the learning goals of students.
The nature of foundation models presents promise here that has yet to be realized in the sphere of AI for education: while certain many streams of data in education are individually too limited to train foundation models, the ability to leverage relevant data from outside the domain (\eg~the Internet) and make use of data across multiple modalities (\eg~textbooks, mathematical formula, diagrams, video-based tutorials) jointly offers hope for foundation models that are broadly applicable to educational tasks.
If foundation models lead to a significant improvement in education-relevant capabilities, there is clear potential for new applications that align with the open-ended generative (\eg~problem generation) and interactive (\eg~feedback to teachers) aspects of foundation models; the sample efficient adaptation of foundation models suggests greater ability for \textit{adaptive and personalized learning}.
In this event, renewed consideration is required of hallmarks of applying technology to education (\eg~student privacy), along with certain concerns becoming more critical (\eg~inequity in access to technology in education, technology-aided plagiarism).

\hypertarget{overview-technology}{\subsubsection{Overview of technology}}
\label{sec:overview-technology}

Now we discuss the technology behind building better model architectures,
training and adaptation procedures, and of course scaling up the systems.
One crucial but often overlooked topic is data\dash{}where does it come from and
what is its composition?
In addition, we want foundation models to be robust to distribution shifts
and secure against attackers.
Finally, we wish to understand why foundation models work from both a mathematical perspective
as well as an empirical perspective.

\paragraph{\hyperref[sec:modeling]{§\ref{sec:modeling}:~Modeling.}}

What structural properties give rise to a foundation model? In the modeling section, we explore the underlying architectures behind foundation models and identify 5 key attributes. First, we start by discussing \textit{expressivity} of the computational model \dash{} to capture and assimilate real-world information, and \textit{scalability} \dash{} to adeptly handle large quantities of high-dimensional data. These properties are successfully realized by existing architectures such as the transformer network \citep{vaswani2017attention} that underpins most foundation models to date. We then proceed to attributes may be essential for the next generation of models, including: \textit{multimodallity} \dash{} to consume, process and potentially produce content from different sources and domains, \textit{memory} capacity \dash{} to effectively store and retrieve the acquired knowledge, and finally, \textit{compositionality}, to foster successful generalization to novel settings and environments. We believe that realizing the full potential envisioned for foundation models will hinge on modelling advances to fulfill these desiderata.

\paragraph{\hyperref[sec:training]{§\ref{sec:training}:~Training.}}

Training objectives mathematically specify how models should learn and acquire capabilities from their training data.
The current status quo for training foundation models involves modality-specific objectives (\eg~masked language modeling \citep{devlin2019bert} for text and SimCLR \citep{chen2020simclr} for images) that are often chosen heuristically.
We envision that future training objectives for foundation models will reflect two changes: \textit{principled selection} derived from systematic evidence and evaluation (\refsec{evaluation}), and \textit{domain-generality} to provide rich, scalable, and unified training signal across data sources and modalities. We also discuss important design trade-offs, including generative vs discriminative training, the choice of input data representation, and the potential of future training objectives that involve explicit representations of goals.

\paragraph{\hyperref[sec:adaptation]{§\ref{sec:adaptation}:~Adaptation.}}

Foundation models are intermediary assets; they are unfinished and generally should not be used directly, instead requiring adaptation for specific downstream tasks.
The \textit{de facto} approach for adaptation has been fine-tuning, with recent work suggesting that lightweight fine-tuning alternatives and prompting-based methods may achieve favorable accuracy-efficiency tradeoffs.
Moving forward, we envision a more expansive view of adaptation that goes beyond just specializing foundation models to perform the task of interest: adaptation will alleviate deficiencies of stand-alone foundation models (\eg~\textit{temporal adaptation} to reflect changes over time in the world) or introduce \textit{constraints} (\eg~GDPR compliance relating to the \textit{right to be forgotten}; \refsec{security}); this broader perspective on adaptation coincides with a need for new evaluation protocols (\refsec{evaluation}) that systematically evaluate adaptation methods while controlling for resources (\eg~runtime, memory) and access requirements involved in adaptation.

\paragraph{\hyperref[sec:evaluation]{§\ref{sec:evaluation}:~Evaluation.}}

Evaluation offers context to foundation models by providing a means to track progress, understand models, and document their capabilities and biases. 
Foundation models challenge the ability of standard evaluation paradigms in machine learning to achieve these goals since they are one step removed from specific tasks. 
To envision new paradigms in evaluation that suit foundation models, we discuss (a) evaluating foundation models \textit{directly} to measure their \textit{inherent capabilities} and inform how foundation models are trained, (b) evaluating task-specific models by \textit{controlling for adaptation resources and access}, and (c) broader \textit{evaluation design} to provide richer context beyond measures of accuracy (\eg~robustness (\refsec{robustness}), fairness (\refsec{fairness}), efficiency (\refsec{systems}), environmental impact (\refsec{environment})). 
Reform of evaluation practices will allow for evaluation that adequately serves both the diverse goals and stakeholders involved in the foundation model paradigm.

\paragraph{\hyperref[sec:systems]{§\ref{sec:systems}:~Systems.}}

While the training data (\refsec{data}) determines the theoretical information available for foundation models, and model architectures (\refsec{modeling}) and training objectives (\refsec{training}) determine how much of this information can be extracted, computer systems determine what is practically achievable.
Systems are a key bottleneck for scaling in terms of data and model size, both of which appear to reliably track with improvements in capabilities. 
To ensure that we can train the next generation of foundation models efficiently (with respect to time and cost), we will require the co-design of algorithms, models, software, and hardware.
This co-design is already starting to happen to in various forms, from carefully tuned parallelism strategies to new architectures such as retrieval-based and mixture-of-expert models. 
Beyond training, we consider what will be required to deploy applications on top of foundation models (\eg~efficient inference).

\paragraph{\hyperref[sec:data]{§\ref{sec:data}:~Data.}}

Data is the lifeblood of foundation models; the training data of these models largely determines what these capabilities these models can acquire. 
The centrality of data is not unique to foundation models; recent calls for {\em data-centric AI}~\citep{ng_data_centric, hazy_data_centric} indicate the pervasive importance of managing, understanding,  and documenting data used to train machine learning models.
For foundation models specifically, the current \textit{modus operandi} is for training data to be selected using unspecified or unclear principles with a general lack of transparency regarding the nature of training data.
We believe an alternative approach is needed to re-imagine the data ecosystem surrounding foundation models: we draw upon work on data visualization and management to propose a \textit{data hub} for foundation models.
We articulate how this proposal relates to many of the relevant data-centric considerations for foundation models: selection, curation, documentation, access, visualization and inspection, quality assessment, and legal regulation.

\paragraph{\hyperref[sec:security]{§\ref{sec:security}:~Security and privacy.}}

Security and privacy for foundation models is largely uncharted at present.
Fundamentally, foundation models are a high-leverage \textit{single point of failure}, making them a prime target for attack: existing work demonstrates a variety of security vulnerabilities (\eg~adversarial triggers to generate undesirable outputs) or privacy risks (\eg~memorization of training data) for these models.
Further, the generality of foundation models compounds these concerns, intensifying the risk for \textit{function creep or dual use} (\ie~use for unintended purposes).
For security, we view foundation models as akin to \emph{operating systems} in traditional software systems; we discuss steps towards secure foundation models which, if achieved, would provide a strong abstraction layer to build upon for reliable ML applications. 
For privacy, by leveraging knowledge transfer from public data, foundation models may enable more sample efficient adaptation to sensitive data distributions, \ie~privacy-preserving applications may incur less degradation in accuracy when built using foundation models.

\paragraph{\hyperref[sec:robustness]{§\ref{sec:robustness}:~Robustness to distribution shifts.}}

A major limitation of standard machine learning is that it produces models that are not robust to \emph{distribution shifts}, where the training distribution does not match the test distribution (for the downstream task). 
Existing work shows that adapting a foundation model trained on a broad range of unlabeled data improves the robustness of adapted models across a wide variety of shifts. 
This opens a new set of promising directions for improving training and adaptation of foundation models for robustness. 
However, we do not believe that foundation models are a panacea for robustness\dash{}challenges such as extrapolation across time and spurious correlations are not likely to be fully addressed. 

\paragraph{\hyperref[sec:ai-safety]{§\ref{sec:ai-safety}:~AI safety and alignment.}}

Ensuring foundation models are reliable (\refsec{systems}), robust (\refsec{robustness}), and interpretable (\refsec{interpretability}) is increasingly important when considering the potential real-world applications of these models.
In addition to critical and immediate considerations, we also consider the relationship between foundation models and larger-scale risks, hazards, and harms that have the potential for increased relevance as model capabilities continue to advance.
For example, we consider the importance of \textit{aligning} foundation models such that they are not deployed with \textit{misspecified goals or values}. We also discuss the relevance of \textit{forecasting the emergent behaviors} of foundation models (\eg~the ability to deceive or plan strategically), which may complicate attempts to adapt them to particular tasks, and may require new approaches for interpretability (\refsec{interpretability}) or evaluation (\refsec{evaluation}).

\paragraph{\hyperref[sec:theory]{§\ref{sec:theory}:~Theory.}}

Learning theory provides a broad foundation for the variety of contexts encountered in applied machine learning; theory offers both understanding, principles, and guarantees to complement empirical findings.
At present, the study of foundation models is largely empirical: the theory of standard supervised learning, while relatively mature, is inadequate to fully explain foundation models.
Specifically, the discrepancy between the training phase and the adaptation phase within the foundation model regime pinpoints the insufficiency of existing theory, since these phases correspond to (potentially) completely different tasks and data distributions.
Nevertheless, we endeavor that advances in theory to address this discrepancy, even in simple, limited settings, will provide useful insights. 

\paragraph{\hyperref[sec:interpretability]{§\ref{sec:interpretability}:~Interpretability.}}

Interpretability provides clarity to foundation models: the opacity of the deep neural networks that underpin foundation models, alongside the expected ubiquity of foundation models, heightens the need to understand these models and their capabilities.
Interpretability methods at present generally are designed for interpreting and explaining the behavior of task-specific models; the nature of foundation models (\ie~the wide array of tasks these models are beneficial for and the unexpected emergent properties they acquire) introduces new challenges for interpretability research.
To frame the discussion of interpretability for foundation models, we propose the \textit{one model-many models} paradigm, which aims to determine the extent to which the \textit{one model} (the foundation model) and its \textit{many models} (its adapted derivatives) share decision-making building blocks.
In addition to interpreting the decision-making components involved, we further discuss \textit{explainability} in the context of foundation models (\eg~the validity of\textit{post hoc} explanations generated by models) as well as the \textit{mechanisms} that drive model behavior (which may clarify the extent to which understanding foundation models can extend to understanding their adapted derivatives). 
Given the critical role we ascribe interpretability in the study of foundation models, we conclude with an assessment of the societal impact of interpretability and non-interpretability. 

\hypertarget{overview-society}{\subsubsection{Overview of society}}
\label{sec:overview-society}

We believe the rapid development of foundation models, adapted and
deployed to various applications, will have wide-ranging consequences on the
health of societies.  What makes these models so exciting and also so troubling
is their task agnosticity.  Societal impact is easier (but still non-trivial)
to understand and reason about when we talk about specific systems deployed to
users, but how can we take into account the societal impact of all
possible systems and use cases when developing foundation models?

\paragraph{\hyperref[sec:fairness]{§\ref{sec:fairness}:~Inequity and fairness.}}

In many contexts, machine learning has been shown to contribute to, and potentially amplify, societal inequity.
Foundation models may extend this trend, \ie~furthering the unjust treatment of people who have been historically discriminated against.
However, understanding the relationship between inequity and foundation models requires reckoning with the abstraction of foundation models; foundation models are intermediary assets that are adapted for applications that impact users.
Therefore, we delineate \textit{intrinsic biases}, \ie~properties in foundation models that portend harm, and \textit{extrinsic harms}, \ie~harms arising in the context of specific applications built using foundation models.
We taxonomize various sources (\eg~training data, lack of diversity among foundation model developers, the broader sociotechnical context) that give rise to these biases and harms, emphasizing the importance, and technical difficulty, of \textit{source tracing} to understand ethical and legal responsibility.
We do not view unfairness as inevitable in the foundation model paradigm: to address unfair outcomes that arise from foundation models, we dually consider \textit{proactive interventions} (\eg~technical methods like counterfactual data augmentation) and \textit{reactive recourse} (\eg~mechanisms for feedback propagation and attribution of moral/legal responsibility). 

\paragraph{\hyperref[sec:misuse]{§\ref{sec:misuse}:~Misuse.}}

We define foundation model misuse as the use of foundation models as they are technically intended (\eg to generate language or video), but with the goal of causing societal harm (\eg to generate disinformation, to develop deepfakes for harassment). 
We argue that advances in foundation models will result in higher-quality machine-generated content that will be easier to create and personalize for misuse purposes. 
For example, disinformation actors may use them to quickly generate collections of articles targeted across different demographic groups (\eg nationality, political party, religion, etc.). 
While these new capabilities may limit existing human detection methods for harmful content (\eg tracking similar text across different sources), foundation models may themselves provide promising potential as automated misuse detectors.

\paragraph{\hyperref[sec:environment]{§\ref{sec:environment}:~Environment.}}

Foundation models are the byproducts of computationally expensive training regimes, with the existing trajectory favoring even more intensive models; the energy required for this training coincides with the release of more carbon into the atmosphere and the degradation of the environment.
At present, current discussion centers these enormous single-time training costs and the potential to amortize these costs across repeated use.
We seek to clarify these discussions by identifying assumptions that shape the calculus of environmental impact for foundation models.
Further, we envision that the ecosystem surrounding foundation models requires a multi-faceted approach: (a) more \textit{compute-efficient} models, hardware, and energy grids all may mitigate the carbon burden of these models, (b) environmental cost should be a clear factor that informs how foundation models are evaluated (\refsec{evaluation}), such that foundation models can be more comprehensively juxtaposed with more environment-friendly baselines, and (c) the cost-benefit analysis surrounding environmental impact necessitates greater \textit{documentation and measurement} across the community.

\paragraph{\hyperref[sec:legality]{§\ref{sec:legality}:~Legality.}}

Foundation models rest on tenuous legal footings at present; how the law bears on both the development and use of these models is largely unclear.
Legal and regulatory frameworks for foundation models specifically, alongside those for AI technology more generally, will be needed to influence, constrain, and even foster practices in research, development, and deployment.
Centering on the legal landscape of the United States, where existing consideration of algorithmic tools remains broadly uncertain, we highlight the pertinent issues of \textit{liability} for model predictions and \textit{protections} from model behavior.
With respect to both issues, we describe how legal standards will need to be advanced to address these given the intermediary status of foundation models (as opposed to that of user-facing task-specific models). 

\paragraph{\hyperref[sec:economics]{§\ref{sec:economics}:~Economics.}}

Foundation models are likely to have substantial economic impact due to their novel capabilities and potential applications in a wide variety of industries and occupations. 
We consider the implications of the development and use of foundation models for the future of the US and global economy with a focus on productivity, wage inequality, and concentration of ownership.

\paragraph{\hyperref[sec:ethics]{§\ref{sec:ethics}:~Ethics of scale.}}

In addition to running the risk of increasing inequity, as discussed in \refsec{fairness}, the widespread adoption of foundation models poses other ethical, political and social concerns.  We discuss ethical issues related to the scale of application of foundation models, such as homogenization and the concentration of power, as well as the norms and release strategies appropriate to address them.
\clearpage
\hypertarget{capabilities}{\section{Capabilities}}
\label{sec:capabilities}

Foundation models acquire capabilities, some that surprisingly emerge from their learning process, that power downstream applications (\refsec{applications}).
Specifically, we discuss linguistic (\refsec{language}) and visual (\refsec{vision}) capabilities alongside the ability to affect the physical world (\refsec{robotics}), perform reasoning and search (\refsec{reasoning}), and interact with humans (\refsec{interaction}). 
In addition, we discuss how self-supervision (the technical approach used to learn most current foundation models) philosophically relates to the ability to understand (\refsec{philosophy}). 

\newsection
\hypertarget{language}{\subsection{Language}}
\label{sec:language}
\sectionauthors{Isabel Papadimitriou, Christopher D. Manning}

\subsubsection{The nature of human language}

Language is the basis of most human communication and interaction. However, it is not just a means for humans to achieve shared goals: language is central to human thought, to how social and emotional relations are formed, to how we identify ourselves socially and personally, and to how humans record knowledge and develop societal intelligence.
Spoken or signed languages arise in every human society, and the languages of the world are both incredibly diverse in the ways that they express and structure the information they convey, while also exhibiting surprising concordance in the richness of what makes a language \citep{comrie1989language}.
Languages are remarkably complex yet efficient systems, acquired consistently by children in a short amount of time, and which evolve and encompass the changing needs and conditions of linguistic communities. 
Due to this centrality of language in human activities, language understanding and generation is a critical element of research in artificial intelligence. 
Natural language processing (NLP) is the subfield of artificial intelligence concerned with language and, together with the related fields of automatic speech recognition (ASR) and text-to-speech (TTS), has the goal of giving computers the ability to understand and generate human language in much the same way human beings can. 

To date in 2021, NLP has been the field most profoundly affected by foundation models. The first generation of foundation models showcased an impressive variety of linguistic abilities, as well as a surprising amount of adaptability to a large range of linguistic situations. Since the introduction of the early foundation models ELMo \cite{peters2018elmo} and BERT \cite{devlin2019bert} in 2018, the field of NLP has become largely centered around using and understanding foundation models.
The field has shifted to using foundation models as the primary tool, moving towards more generalized language learning as a central approach and goal. 
In this section, we go over the recent successes of foundation models in NLP, detail how foundation models have changed the overall process and mentality for training machine learning models for language, and discuss some of the theoretical and practical challenges facing foundation models as they are applied to a broader set of languages and more realistic and complex linguistic situations.

 \begin{figure}[t]
\centering
\includegraphics[width=\textwidth]{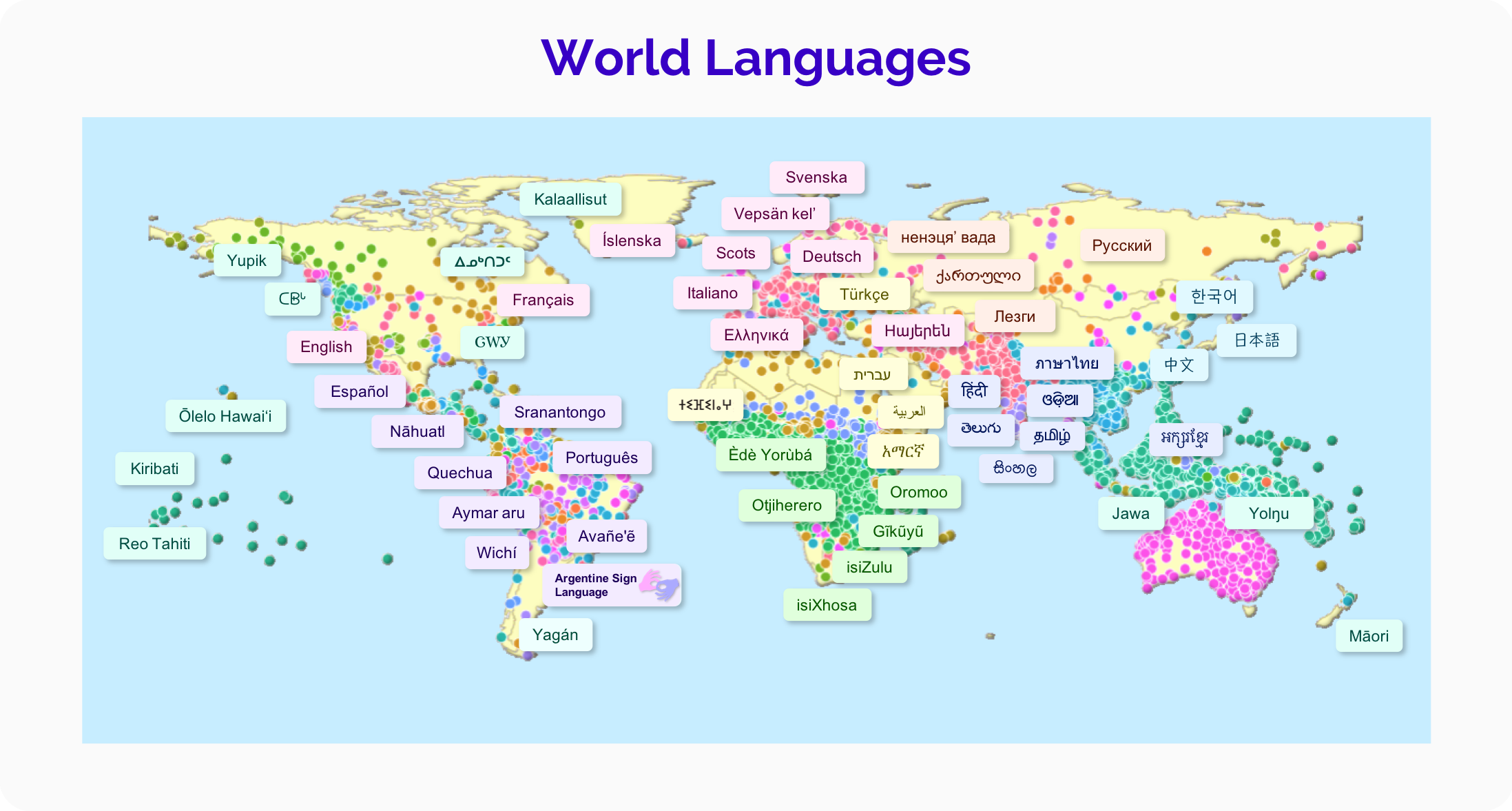}
\caption{\label{fig:language_1}
Only a tiny percentage of the world's languages are currently represented in foundation models. There are over 6,000 languages in the world, with estimates varying due to the inherent uncertainty of what constitutes a separate language \citep{nordhoff2011glottolog}. This map shows the languages of the world, with each dot representing one language and its color indicating the top-level language family. Data is from Glottolog \cite{glottolog}. We label a few of the languages on the map as examples.}
\end{figure}

\subsubsection{Impact of foundation models on NLP}

Foundation models have had a huge impact on the field of NLP, and are now central to most NLP systems and research. On a first level, many foundation models are skilled language generators: for example, \citet{clark2021all} demonstrate that non-experts have difficulty distinguishing short-form English text that was written by GPT-3 from that written by humans. However, the feature of foundation models that has been most impactful in NLP is not their raw generation abilities but their surprising \textit{generality and adaptability}: a single foundation model can be adapted in different ways in order to achieve many linguistic tasks.  

The field of NLP has historically focused on defining and engineering systems for challenging linguistic tasks, with the vision that models that are good at these tasks will lead to competent language systems for downstream applications.  NLP tasks include \textit{classification tasks} for a whole sentence or document (\eg sentiment classification, like predicting  whether a movie review is positive or negative), \textit{sequence labeling} tasks, in which we classify each word or phrase in a sentence or document  (\eg predicting if each word is a verb or a noun, or which spans of words refer to a person or an organization),  \textit{span relation classification}, (\eg relation extraction or parsing, like whether  a person and location are linked by a ``current residence'' relation, or a verb and a noun by a ``subject-verb'' relation) and \textit{generation tasks}, producing new text that is conditioned strongly on an input (\eg producing a translation or summary of a text, recognizing or producing speech, or responding in a conversation) 
\citep{jurafsky2009speech}.
In the past, NLP tasks had distinct research communities that developed task-specific architectures, often based on pipelines of different models, each performing a linguistic sub-task such as token segmentation, syntactic parsing, or coreference resolution.

By contrast, the dominant modern approach for performing each task is to use a single foundation model and adapt it slightly using relatively small amounts of annotated data specific to each task (sentiment classification, named entity tagging, translation, summarization) to create an adapted model.
This has proved to be an extremely successful approach: for the vast majority of the tasks described above, a foundation model that is slightly adapted for a task greatly outperforms previous models or pipelines of models that were built specifically to perform that one task. To take just one example, the best system for answering open-ended science questions in 2018, before foundation models, could get 73.1\% on the NY Regents 8th grade science exam. A year later in 2019, an adapted foundation model scored 91.6\% \citep{clark2019aristo}.

The emergence of foundation models that are largely trained to \textit{generate} language has constituted an important shift in the role of language generation in NLP\@. 
Until around 2018, the problem of generating general-purpose language was considered very difficult and
essentially unapproachable except through other linguistic sub-tasks \citep{paris2013natural}.
Instead,  NLP research was mostly focused on linguistically analyzing and understanding text. 
Now, it is possible to train highly coherent foundation models with a simple language generation objective, like ``predict the next word in this sentence''. These generative models now constitute the primary vehicle through which machine learning for language is done\dash{}including the analysis and understanding tasks that were once considered prerequisites for generation. 
The successful generation exhibited by foundation models has also led to a flowering of research for language generation tasks like summarization and dialogue generation. 
The rise of the foundation model paradigm has begun to play a similar role in spoken language as well as written.  Modern automatic speech recognition (ASR) models like wav2vec 2.0 are trained on large datasets of speech audio alone, and then adapted on audio with associated transcriptions for the task of ASR \citep{wav2vec2}.

Due to the changes brought about by the foundation model paradigm, the focus of research and practice in NLP has shifted from making bespoke architectures for different tasks to exploring how to best leverage foundation models. Research into adaptation methods has blossomed 
(see \refsec{adaptation} for a detailed look at adaptation), and the surprising successes of foundation models have also caused a shift in research interest towards analyzing and understanding foundation models
(see \refsec{interpretability} for interpretability and analysis of foundation models).

\subsubsection{Language variation and multilinguality}

Though foundation models are surprisingly versatile with the linguistic knowledge they obtain from pretraining,
there are limits to this adaptability: it is not clear how successful current foundation models are at handling language variation.
Language varies greatly. Apart from the fact that there are thousands of different languages in the world, language varies even within one language or within one speaker. To point out a few examples, informal conversation manifests differently from written language, the grammatical constructions that people reach for when speaking to friends are very different from those used when speaking to someone with authority, and communities of speakers within a language use different dialects. Social and political factors are embedded in how language variation is viewed and valued, and in how much different varieties are represented in NLP research (see for example \citet{blodgett17} on the failures of NLP for African American English, and \refsec{fairness} for a deeper discussion on inequities in foundation models).
Due to their large capacity for learning linguistic information and flexibly adapting that knowledge, foundation models hold promise for expanding NLP to encompass more linguistic diversity. It remains an open research question to understand whether it is possible to make foundation models that robustly and equitably represent language with both its major and subtle variations, giving equal weight and acuity to what makes each linguistic variety distinct \citep[research posing and addressing this question includes][]{ponti2019modeling, bender2011achieving, joshi2020state}.

Following the success of foundation models for English, multilingual foundation models have been released to extend that success to non-English languages.
For most of the over 6,000 languages in the world, the text data available is not enough to train a large-scale foundation model. 
To give one example, there are over 65 million speakers of Fula, a West African language, but few if any resources available for NLP in Fula \citep{nguer2020sencorpus}.
Multilingual foundation models address this by jointly training on multiple languages simultaneously. The multilingual foundation models to date (mBERT, mT5, XLM-R) are each trained on around 100 languages \citep{devlin2019bert, goyal2021larger, xue2020mt5}. Joint multilingual training relies on the reasonable assumption that the shared structures and patterns between languages can lead to sharing and transfer from the high-resource languages to the low-resource ones, making foundation models possible for languages where we could not train a stand-alone model.
Experiments using and analyzing multilingual foundation models have shown that there is indeed a surprising amount of transfer between and parallel encoding of the different languages in multilingual foundation models \citep{wu-dredze-2019-beto, choenni2020cross, pires2019multilingual, libovicky2019language, chi2020finding, papadimitriou2021deep, cao2019multilingual}.  

However, the extent to which these models are robustly multilingual is still an open question. 
It remains unclear how much models trained on this data can represent aspects of languages that are drastically different from English or for which few language resources are available \citep{wu-dredze-2020-languages}, and whether their apparent multilingual performance relies more on assimilation \citep{lauscher2020zero, virtanen2019multilingual, artetxe_cross-lingual_2020}. Multilingual models show better performance in languages that are similar to the highest-resource languages in their training data, and it has been shown that languages in multilingual models compete for model parameters, making it unclear how much variation can fit in a single model \citep{wang2020negative}. A salient issue stems from the data that we use to train multilingual foundation models: in many multilingual corpora, English data is not only orders of magnitude more abundant than that of lower-resource languages, but it is often cleaner, broader, and contains examples showcasing more linguistic depth and complexity \citep{caswell2021} (see \citet{nekoto2020participatory} on building participatory and robust multilingual datasets).
However, the answer does not simply lie in creating more balanced corpora: there are so many axes of language variation that it would be infeasible to create a corpus that is balanced and representative in all regards. The future, versatility, and equity of foundation models all depend on robustly handling language variation despite unbalanced data \cite[\eg][]{oren2019drolm}.

Current multilingual foundation models in their raw form, and naive unsupervised multilingual training as a method, may not model the subtleties of languages and language varieties to their full extent. Nevertheless, they remain useful for some multilingual applications, for example through adapting multilingual models for low-resource languages not in their original training set \citep{wang2020extending}. Moreover, the results for the (non-public) GShard neural machine translation model show the largest gains over monolingual baselines for the lowest resource languages, with the gains increasing with model size \citep{lepikhin21gshard}.
The research community should critically examine how foundation models deal with language variation, understand the limits of foundation models in bringing equity and representation to NLP, and not settle on promoting foundation models that erase language variation and mostly conform to the linguistic majority in their training data. 

\subsubsection{Inspiration from human language acquisition}

\begin{figure}[t]
\centering
\includegraphics[width=\textwidth]{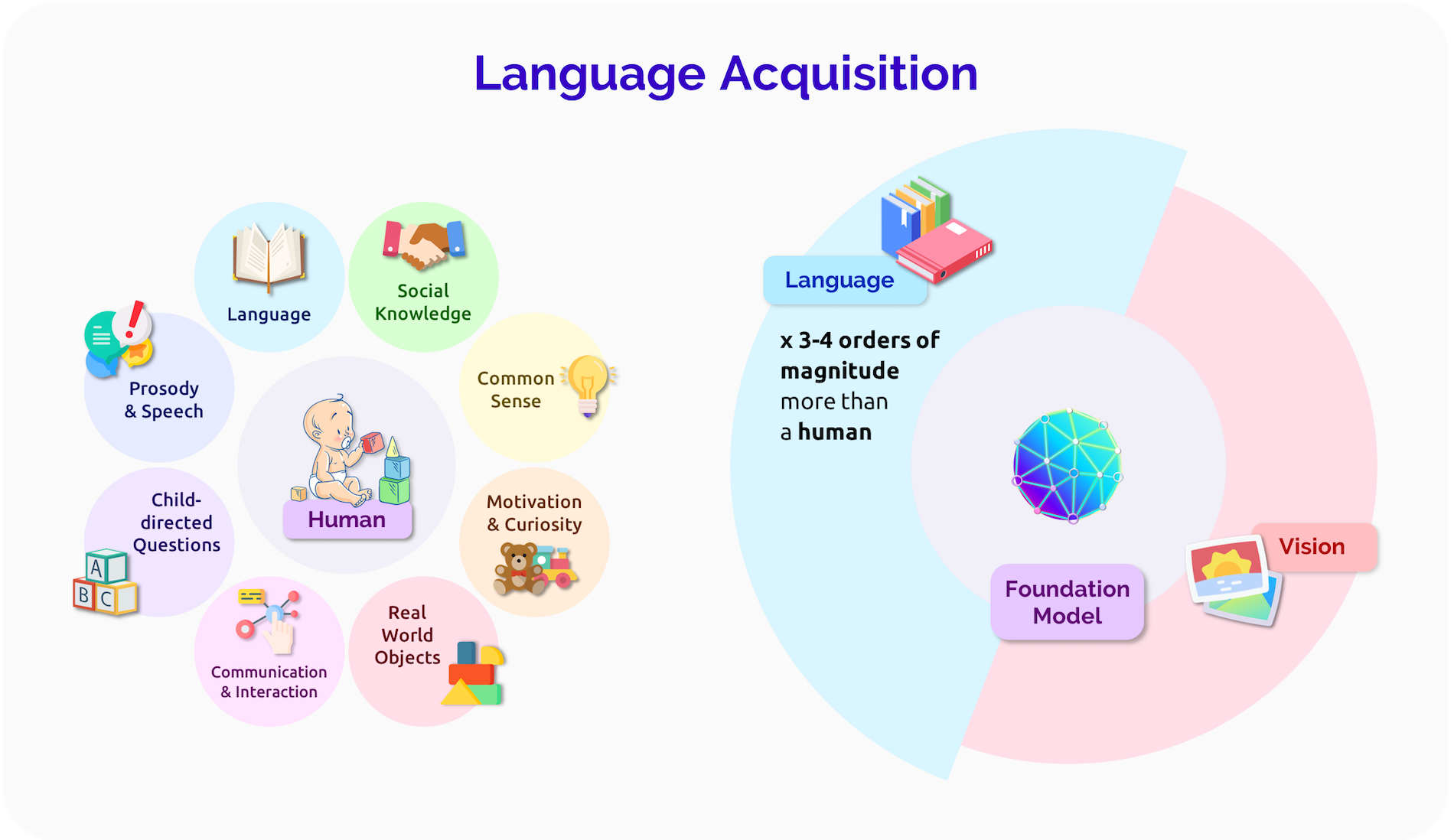}
\caption{\label{fig:language_2}
Language Acquisition for humans and foundation models. While there are certainly different inductive biases between the human brain and foundation models, the ways that they learn language are also very different. Most saliently, humans interact with a physical and social world in which they have varied needs and desires, while foundation models mostly observe and model data produced by others.}
\end{figure}

Though foundation models have constituted a huge source of progress in creating NLP systems that act more like humans, there are still significant ways in which the linguistic system that they acquire, as well as the learning process, differ from human language. Understanding the implications of this gap between machine and human language learning is a necessary part of developing a research community informed about the linguistic limits and possibilities of foundation models.

Human language acquisition is very efficient: foundation models like GPT-3 are trained on around three to four orders of magnitude more language data than most humans will ever hear or read, and certainly much more than children have been exposed to by the time they are mostly linguistically competent. 
One salient difference between foundation models and human language acquisition is that human language is grounded to the real world \citep{saxton17child}. 
For example babies and caretakers point to objects during language development \citep{colonnesi2010relation}, and babies learn the grounded meanings of words that refer to common objects before they learn a lot of the other aspects of the linguistic system \citep{bergelson20126}. Most foundation models used in NLP, on the other hand, learn from the distributional information of raw, ungrounded text, and (in contrast to human learners) \citet{zhang2021billions} show that RoBERTa models express abstract syntactic features before usable meaning. Powerful ungrounded statistical learning is indeed also present in babies \citep{saffran1996statistical}, so it is no doubt an important factor in acquisition. Nevertheless, advancing grounded language learning for foundation models remains an important direction for approaching human acquisition efficiency
\citep[\textit{inter alia}]{dupoux2018cognitive, tan2020vokenization, zellers2021piglet} (see \refsec{vision} and \refsec{robotics} for the multimodal potential of foundation models, and \refsec{philosophy} for a discussion of whether foundation models can understand language without grounding). 
Another important direction is examining the inductive biases in foundation models and how they relate to the inductive biases in the human mind, both those specific to language learning and those general to human cognition \citep{linzen2021syntactic}. Though the human brain may be more architecturally specialized for efficient language acquisition, foundation models are not blank-slate learners \cite{baroni2021proper}, and understanding and aligning these linguistic inductive biases is an important future direction for research in foundation models. 

A significant factor in the efficiency of language acquisition is the fact that humans acquire a systematic and generalizable language system. Though there are many differing theories about what types of theoretical abstractions the human language system makes \citep[\eg][]{comrie1989language, chomsky2014minimalist, croft2001radical, jackendoff2011human}, it is generally agreed that humans learn language in a way that allows them to easily slot new knowledge into existing abstractions and productively create new grammatical sentences. For example, a ten-year-old child has acquired a lot of the abstractions about how their language works, though the actual words and constructions that they produce will change drastically over the next ten years. Foundation models, on the other hand, often do not acquire the systematic abstractions that we expect from humans. For example, when a foundation model produces a linguistic construction accurately one time there is no guarantee
that future uses of that construction will be mostly consistent, especially after a significant domain shift in the subject matter \citep[examples of work examining limitations of foundation models in systematicity include][]{lake2018generalization, kim2020cogs, bahdanau2018systematic, chaabouni2021can}. NLP faces the challenge of developing some sort of systematicity in acquisition for foundation models, without regressing to systems that rely too heavily on rigid linguistic rules.

Language learning continues for a speaker's whole lifetime:
the grammar of human languages evolves, and humans flexibly adapt to novel linguistic situations \citep{sankoff2018change}. For example, as new terms and concepts arise in an adult's life, they can use them relatively easily in grammatical sentences, and humans often adapt their grammatical patterns to fit in with different social groups \citep{rickford1994addressee}. 
On the other hand, the linguistic system of foundation models is mostly set by the training data, and is relatively static \citep{lazaridou2021pitfalls, Khandelwal2020Generalization}. Though adaptation methods can prime foundation models for different tasks (see \refsec{adaptation}), it still remains unclear how to change the more basic linguistic foundation of a foundation model without a large amount of training. 
Making adaptable models that naturally mirror human-like linguistic accommodation and language evolution is an important research area for the future of foundation models.

Foundation models have drastically changed the research and practice of NLP\@. Foundation models have given rise to many new research directions for the community: understanding generation as a fundamental aspect of language, studying how to best use and understand foundation models, understanding the ways in which foundation models may increase inequities in NLP,  examining whether foundation models can satisfactorily encompass linguistic variation and diversity, and finding ways to draw on human language learning dynamics. 
Most of the complex NLP tasks that the research community focused on before foundation models are now best handled, to an almost-human level, using one of a few publicly released foundation models. Nevertheless, there remain significant gaps between this performance and the needs for useful and safe deployment of foundation models in complex downstream settings. 
\newsection
\hypertarget{vision}{\subsection{Vision}}
\label{sec:vision}
\sectionauthors{Shyamal Buch, Drew A. Hudson, Frieda Rong, Alex Tamkin, Xikun Zhang, Bohan Wu, Ehsan Adeli, Stefano Ermon, Ranjay Krishna, Juan Carlos Niebles, Jiajun Wu, Li Fei-Fei}

\begin{figure}[!ht]
\centering
\includegraphics[width=\linewidth]{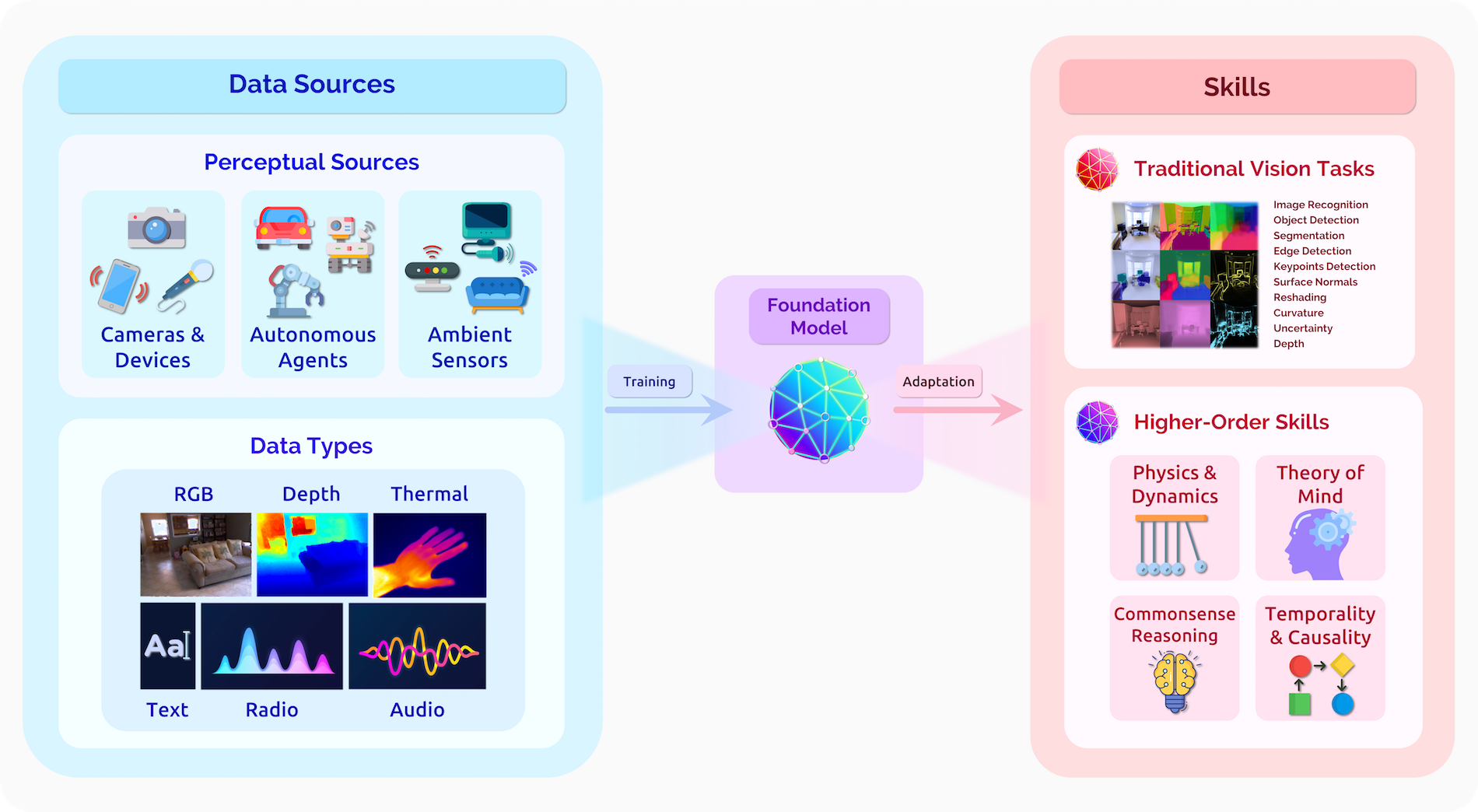}
\caption{By harnessing self-supervision at scale, foundation models for vision have the potential to distill raw, multimodal sensory information into visual knowledge, which may effectively support traditional perception tasks and possibly enable new progress on challenging higher-order skills like temporal and commonsense reasoning (\protect\refsec{vision-capabilities}). These inputs can come from a diverse range of data sources and application domains, suggesting promise for applications in healthcare and embodied, interactive perception settings (\protect\refsec{vision-challenges}). Image credits \cite{zamir2018taskonomy,haque2020illuminating}.}
\label{fig:vision}
\end{figure}

Vision underlies one of the primary modes through which a living organism understands its environment.
The ability to see enables the near-constant, long-range gathering of dense signals, a critical capability developed over an evolutionary time-scale in a diverse range of life forms~\citep{parker2003blink,zhang21current}. 
For a skill executed effortlessly by even simple living creatures, transferring the same abilities to machines has proved remarkably challenging, leading computer vision and robotics researcher Hans Moravec in 1988 to observe a paradox: in AI, (what were considered) hard problems are easy and likewise easy problems are hard, and among the ``easiest'' problems of them all is the visual acuity which we use each day to continually interpret complex scenes in a matter of milliseconds \cite{moravec1988mind,Thorpe1996,feifei2007we}.


On the other end of this formidable challenge is the substantial scope of transformative applications which computer vision holds the key to: self-driving cars that can free commuters from gridlock (\refsec{robotics}), life-saving AI tools that can assist overworked specialists by detecting rare medical events (\refsec{healthcare}), next-generation tools for multimedia creation and editing (\refsec{interaction}), among others. Reflecting on the applications and settings where human perception is instrumental offers a sense of the potential areas where computer vision can assist and transform.

The field of computer vision and the challenges we define draw inspiration in many ways from human perception capabilities.
Several classical theories \citep[\eg][]{biederman_perceiving_1972,mcclelland1981interactive,marr1982vision} suggested that humans may perceive real world scenes by contextualizing parts as a larger whole, and pointed the way for computer vision techniques to progressively model the physical world with growing levels of abstractions \cite{lowe1992robust,girshick2014rich}.
\citet{gibson1979ecological} suggested that human vision is inherently embodied and interactive ecological environments may play a key role in its development.
These ideas continue to motivate the ongoing development of computer vision systems, iterating towards a contextual, interactive, and embodied perception of the world.

In the context of computer vision, foundation models translate raw perceptual information from diverse sources and sensors into visual knowledge that may be adapted to a multitude of downstream settings (\reffig{vision}). To a large extent, this effort is a natural evolution of the key ideas that have emerged from the field over the last decade.
The introduction of ImageNet \cite{deng2009imagenet} and the advent of supervised pretraining led to a deep learning paradigm shift in computer vision. This transition marked a new era, where we moved beyond the classic approaches and task-specific feature engineering of earlier days \cite{lowe2004distinctive, bay2006surf, rosten2006machine} towards models that could be trained once over large amounts of data, and then adapted for a broad variety of tasks, such as image recognition, object detection, and image segmentation \cite{krizhevsky2012imagenet, szegedy2015going, he2016deep, simonyan2015verydeep}. This idea remains at the core of foundation models.

The bridge to foundation models comes from the limitations of the previous paradigm. Traditional supervised techniques rely on expensive and carefully-collected labels and annotations, limiting their robustness, generalization and applicability; in contrast, recent advances in self-supervised learning \cite{chen2020simclr, He2020MomentumCF}
suggest an alternative route for the development of foundation models that could make use of large quantities of raw data to attain a contextual understanding of the visual world. Relative to the broader aims of the field, the current capabilities of vision foundation models are currently early-stage (\refsec{vision-capabilities}): we have observed improvements in traditional computer vision tasks (particularly with respect to generalization capability) \cite{radford2021learning,ramesh2021zeroshot} and anticipate that the near-term progress will continue this trend. However, in the longer-term, the potential for foundation models to reduce dependence on explicit annotations may lead to progress on essential cognitive skills (\eg~commonsense reasoning) which have proven difficult in the current, fully-supervised paradigm \cite{zellers2019vcr,martin2021jrdb}.
In turn, we discuss the potential implications of foundation models for downstream applications, and the central challenges and frontiers that must be addressed moving forward (\refsec{vision-challenges}).

\hypertarget{vision-capabilities}{\subsubsection{Key capabilities and approaches}}
\label{sec:vision-capabilities}

At a high-level, computer vision is the core sub-field of artificial intelligence that explores ways to endow machines with the capacity to interpret and understand the visual world. It encompasses a multitude of tasks, sub-domains and downstream applications, where the community has made continual progress over the last several decades \cite{zamir2018taskonomy}. A selection of example tasks\footnote{This, of course, is a coarse selection: please see the categories at the annual conference on Computer Vision and Pattern Recognition (CVPR) for a more complete (but evolving) picture of the tasks in the field.}:
(1) \textit{semantic understanding} tasks, which aim to discover the properties and relations among entities within visual scenes; these include image classification, object detection, semantic segmentation, action recognition, and scene graph generation, among others \citep[\eg][]{krizhevsky2012imagenet, he2016deep,  krishna2017visual, russakovsky2015imagenet, krizhevsky2009learning, kay2017kinetics, lin2014microsoft}. (2) \textit{geometric, motion and 3D} tasks, seeking to represent the geometry, pose and structure of still or moving objects, and include tasks of depth estimation, structure-from-motion, surface normal detection, curvature line and keypoint estimation, to name a few \citep[\eg][]{laina2016deeper, agarwal2011building, wang2015designing, zamir2018taskonomy, ullman1979interpretation}.
(3) \textit{multimodal integration} tasks, combining semantic and geometric understanding with other modalities such as natural language; these include, for instance, visual question answering, image captioning, and instruction following \citep[\eg][]{antol2015vqa, chen2015microsoft, anderson2018vision, goyal2017making, hudson2019gqa, johnson2017clevr, luo2020univl, akbari2021vatt, huang2021seeing, tsimpoukelli2021multimodal}. 
We highlight a subset of traditional core tasks in \reffig{vision}.

The predominant paradigm for addressing these tasks, driven by the emergence of ImageNet \cite{deng2009imagenet} during the early 2010s, tends to center around a familiar core idea:
First, pretrain a model on a large collection of carefully annotated data \cite{russakovsky2015imagenet} with a fully supervised training task, like image classification. Then, adapt the model downstream on task-specific datasets and domains \cite{lin2014microsoft, chen2015microsoft, antol2015vqa} by fine-tuning to reach state-of-the-art performance \cite{krizhevsky2012imagenet, simonyan2015verydeep, he2016deep, xu2016ask}.
This notion of pretraining followed by adaptation persists in the definitions we consider now for foundation models (\refsec{introduction}).
The limitations of this fully supervised paradigm motivate the transition to foundation models: the reliance on external supervised annotations constrains the upper bound capability of previous approaches to capture the diverse spectrum of visual inputs in a scalable, robust and generalizable manner.
Recent developments in the domain of visual synthesis and unsupervised learning offer a compelling alternative. GANs, for instance, learn to generate visual content of high fidelity, realism and diversity, by featuring two competing networks of a generator and a discriminator that can supervise one another from image collections alone \citep[\eg][]{goodfellow2014gan,ganformer}. Other neural models infer the visual properties of objects and scenes without explicitly annotated supervision, by employing variational auto-encoding, contrastive learning or other self-supervised techniques \citep[\eg][]{Kingma2014AutoEncodingVB, chen2020simclr, He2020MomentumCF}. For instance, \citet{he2021masked} build upon prior work on representation learning with masked image encoding \citep[\eg][]{pathak2016context,vincent2008denoise} by, in part, combining recent advancements in flexible architectures (\eg vision transformers \cite{Dosovitskiy2021AnII, zhai2021scaling}) with increased scaling.

With foundation models, the development of such self-supervision techniques has enabled training at greater scales of visual data \cite{changpinyo2021cc12m}, 
both in terms of its scope as well as its potential diversity. Accordingly, we have seen early indicators of progress on traditional vision tasks in terms of both standard accuracy metrics and few-shot generalization. For image classification and object detection, self-supervised techniques have reported competitive performance to prior fully-supervised approaches \cite{he2019moco,chen2020simclr,radford2021learning,henaff2021efficient}, without explicit annotations during training and greater sample efficiency during adaptation.
For visual synthesis, notable examples include DALL-E \cite{ramesh2021zeroshot} and CLIP-guided generation \cite{radford2021learning, galatolo2021generating}, where researchers leverage multimodal language and vision input to render compelling visual scenes.
In the short-term, we anticipate that the capabilities of these foundation models will continue to improve along these directions, as training objectives are refined \cite{chen2020mocov2,henaff2021efficient,selvaraju2021casting} and architectures are designed to incorporate additional modalities \cite{jaegle2021perceiver}.

Notably, current foundation models for computer vision are nascent relative to their NLP counterparts (\refsec{language}): promising early efforts are still largely centered on RGB image inputs and a subset of core traditional vision tasks. However, the field continues to progress on broader challenges centered on embodied and interactive perception settings (critical for foundation models for robotics \citep[][\refsec{robotics}]{bohg2017interactive}). We note a subset of these higher-order goals in \reffig{vision}, including physical scene understanding, reasoning over visual commonsense and temporal events, and perception for social affordances.
Each of these have been goals for fully-supervised systems, but have proven challenging in part due to the difficulty of annotating these tasks at scale. 
For instance, standard systems for visual-question answering struggle to answer questions that require commonsense understanding, since these questions often require external knowledge beyond what is present in the pixels alone~\cite{zellers2019vcr}. Perceiving human gaze and social affordances in a robust manner remain ongoing challenges for embodied vision systems in interactive agents~\cite{martin2021jrdb}. By reducing the dependence on explicit annotations, foundation models may enable further progress towards these goals than was previously feasible. Related progress in language foundation models (\refsec{language}), which have been able to capture a degree of commonsense over language events \cite{brown2020gpt3}, also suggests a potential avenue towards achieving similar capability over multimodal visual inputs. While the exact roadmap for how to achieve these capabilities in foundation models remains an open problem, a combination of new efficient and flexible architectures (\refsec{modeling}), large-scale training (\refsec{systems}), self-supervision techniques (\refsec{training}) and few-shot adaptation schemes (\refsec{adaptation}) may open the door towards capabilities that have been difficult to reach so far.

\hypertarget{vision-challenges}{\subsubsection{Central research challenges}}
\label{sec:vision-challenges}

Our discussion of research challenges is motivated by the \textit{downstream application domains} where foundation models may further the integration and impact of vision models. 
We highlight a few such areas:
(1) \textit{ambient intelligence} for healthcare and home environments: building upon existing approaches for ambient intelligence in these settings \cite{haque2017towards, lyytinen2002ubiquitous, hong2004architecture}, foundation models may offer the potential for better detection of fine-grained human activities and medical events, as well as improved assistive interaction for clinicians, patients, and everyday consumers (see also \refsec{healthcare}).
(2) \textit{mobile and consumer applications}: foundation models with stronger multimodal grounding may enable more capable interactivity of services in mobile settings, and fundamental improvements in generation capability from vision and language inputs can benefit computational photography and content editing applications  \cite{delbracio2021mobile,ramesh2021zeroshot,park2019gaugan} (see also \refsec{interaction}).
(3) \textit{embodied, interactive agents}: perception models have already proven effective as both inputs  \cite{sermanet2018time} and reward functions  \cite{chen2021generalizable,shao2020concept2robot} in robotics settings; foundation models trained on large collections of egocentric (real/simulated, human/robotic) visual data \cite{damen2018kitchens,chen2021geosim} may potentially further this progress by capturing a wider distribution of visual scenes, objects, and actions (see also \refsec{robotics}).

The extent to which foundation models may further impact these application settings hinges on the degree to which the capabilities outlined in \refsec{vision-capabilities} are realized. To bridge the significant gaps between present, short-term, and long-term anticipated capabilities, we must address current \textit{limitations} of foundation models for vision, including their training and evaluation. Below, a subset of corresponding key challenges:

\paragraph{Semantic systematicity and perceptual robustness.} Humans have a remarkable capacity for generalizing visual understanding to unseen compositions, and reasoning about the physical and geometric properties of novel objects and scenes \cite{lake2015human}. While current foundation models have shown promising capability for image synthesis and early results for generalization to fine-grained language inputs, these models still struggle to generalize to compositions of simple shapes and colors \cite{ramesh2021zeroshot,radford2021learning,rong2021extrapolating}. Generalizability goes beyond semantics as well; visual scenes and objects have a natural regularity to their physical dynamics and geometric properties. Foundation models have shown early indications of understanding scene and object geometry \cite{ramesh2021zeroshot}. Further, early efforts towards physical scene and geometric understanding in perception models may provide guidance for ongoing foundation model development \cite{yi2019clevrer, bakhtin2019phyre, li2020visual}. Indeed, the continued incorporation of multiple modalities (\eg~audio) in foundation models may prove beneficial towards these aims \cite{zhang2017shape,gao2020visualechoes,jaegle2021perceiverio}. However, the specific techniques to enable generalizing the initial observed capabilities robustly to a wide range of natural scenes and objects at the level of humans remains an open research challenge for foundation models.

\paragraph{Computational efficiency and dynamics modeling.} Humans are surprisingly efficient at processing the continuous visual stream of objects, scenes, and events necessary to support an understanding of event dynamics \cite{zacks2001perceiving, tversky2013event}.
Foundation models in language (\refsec{language}) have shown initial steps towards modeling longer-term coherence of events;
the analogous ability to capture long-range temporal correlations and causal coherence in visual input would stand to benefit downstream settings like robotics \citep[][\refsec{robotics}]{dai2019transformer,alyamkin2019low, goel2020survey,feng2019computer}.
However, relative to word token-level inputs in language, low-level computer vision inputs are extremely high-dimensional: a single 1080p frame contains over 2 million pixels. In this context, modeling the richer event dynamics in long-range video sequences seems like a daunting endeavor, especially with additional modalities (\eg~speech, optical flow, etc.) and increasing resolutions. Understandably, a na\"{i}ve approach to fully processing every individual pixel is likely prohibitive. Current vision models \citep[\eg][]{radford2021learning,sun2019videobert,tan2019lxmert,kim2021vilt} often address this by processing embeddings that summarize image patches or even groups of frames altogether, but this has the potential drawback of losing fine-grained details \cite{ramesh2021zeroshot}. In addition to considerations of the raw input space, foundation models for vision may need to revisit the design of fundamental architecture primitives (\refsec{modeling}) for efficient and effective modeling: alternatives to 3D convolutions may better address its cubic complexity \cite{fanbuch2020rubiks,sitzmann2019scene}, while particle-based representations may prove more effective for modeling physical dynamics \cite{bear2021physion}. Further, deployment of these vision models to downstream application settings will also necessitate advancements in systems design (\refsec{systems}). Taken together, the bottleneck of efficient and effective modeling for larger-scale, dynamic vision inputs remains a multi-faceted research direction that must be addressed going forward.

\paragraph{Training, environments, and evaluation.} Equally critical to realizing the potential of foundation models are the supporting elements for training and evaluating them. Current foundation models for vision have largely focused on a small subset of modalities shown in \reffig{vision} (\eg~datasets of RGB images and text), since these are perhaps the most readily accessible \cite{changpinyo2021cc12m,radford2021learning}. This motivates the development and use of additional large-scale training datasets which contain a diverse collection of inputs across a broad spectrum of modalities. While additional annotations may not strictly be necessary, the input quality impacts the learning efficiency of the models; techniques that leverage foundation models of other types (\eg~language) to help improve quality are a promising route forward \cite{zellers2021merlot}. 
We also want to consider settings beyond static datasets: classic studies have suggested that perceptual understanding in humans is linked to its embodiment and interactive, ecological settings \cite{gibson1979ecological}. As stepping stones towards longer-term capabilities of embodiment and interaction (\refsec{robotics}), ongoing development of simulation environments that capture physical, visual, and ecological realism with multiple modalities and viewpoints may play an important role in providing scalable and high-fidelity visual inputs for this goal \cite{kolve2017ai2thor,habitat19iccv,gan2020threedworld,shen2021igibson,srivastava2021behavior}. Finally, there is the question of metrics: how do we evaluate the faithfulness of generative foundation model outputs with respect to semantics? Standard metrics like Fr\'{e}chet Inception Distance, suffer from known flaws \cite{bikowski2018demystifying}; such issues parallel ones in natural language processing (\eg~metrics like BLEU do not correlate with causal judgements from humans). Having human judgements as part of evaluation may be one route, but incurs significant cost and may not be as scalable \cite{zhou2019hype,khashabi2021genie}. The outstanding and open challenges surrounding the training (\refsec{training}), data (\refsec{data}), and evaluation (\refsec{evaluation}) settings for vision foundation models are indeed quite nuanced, and will be a central area of research going forward.

\paragraph{Concluding remarks.} In this section, we explored foundation models in the context of computer vision, from identifying roots in previous computer vision paradigms, to contextualizing its current and anticipated capabilities, to proposing research directions moving forward. 
We conclude with a brief discussion of some broader \textit{societal implications} of foundation models for computer vision and their continued development (see also \refsec{society}).
The ubiquity of cameras in our society means that advances in computer vision techniques have great potential for disruptive impact; this carries a corresponding burden of responsibility for careful consideration of its risks. 
There is a well-documented history of learned bias in computer vision models, resulting in lower accuracies and correlated errors for underrepresented groups, with consequently inappropriate and premature deployment to some real-world settings \citep[\eg][\refsec{fairness}]{buolamwini2018gender}. 
Many of the same underlying issues continue to persist in current foundation models \cite{agarwal2021evaluating}. 
As data from additional sensor modalities (\eg~wearable or ambient sensors, \reffig{vision}) become incorporated in these foundation models, concerns surrounding privacy and surveillance become paramount (see \refsec{ethics}).
Furthermore, generated deepfake images and misinformation pose greater risks as the semantic and generative capability of vision foundation models continues to grow \citep[][\refsec{misuse}]{dolhansky2020deepfake,ramesh2021zeroshot}.
While the intriguing open challenges and opportunities ahead for computer vision and foundation models are significant, addressing these and related risks concurrently remains essential.
\newsection
\hypertarget{robotics}{\subsection{Robotics}}
\label{sec:robotics}
\sectionauthors{Siddharth Karamcheti, Annie Chen, Suvir Mirchandani, Suraj Nair, Krishnan Srinivasan, Kyle Hsu, Jeannette Bohg, Dorsa Sadigh, Chelsea Finn}

\newcommand*{\eperceive}{\textcolor{orange}}
\newcommand*{\jean}{\textcolor{teal}}

\begin{figure}[!ht]
\centering
\includegraphics[width=\linewidth]{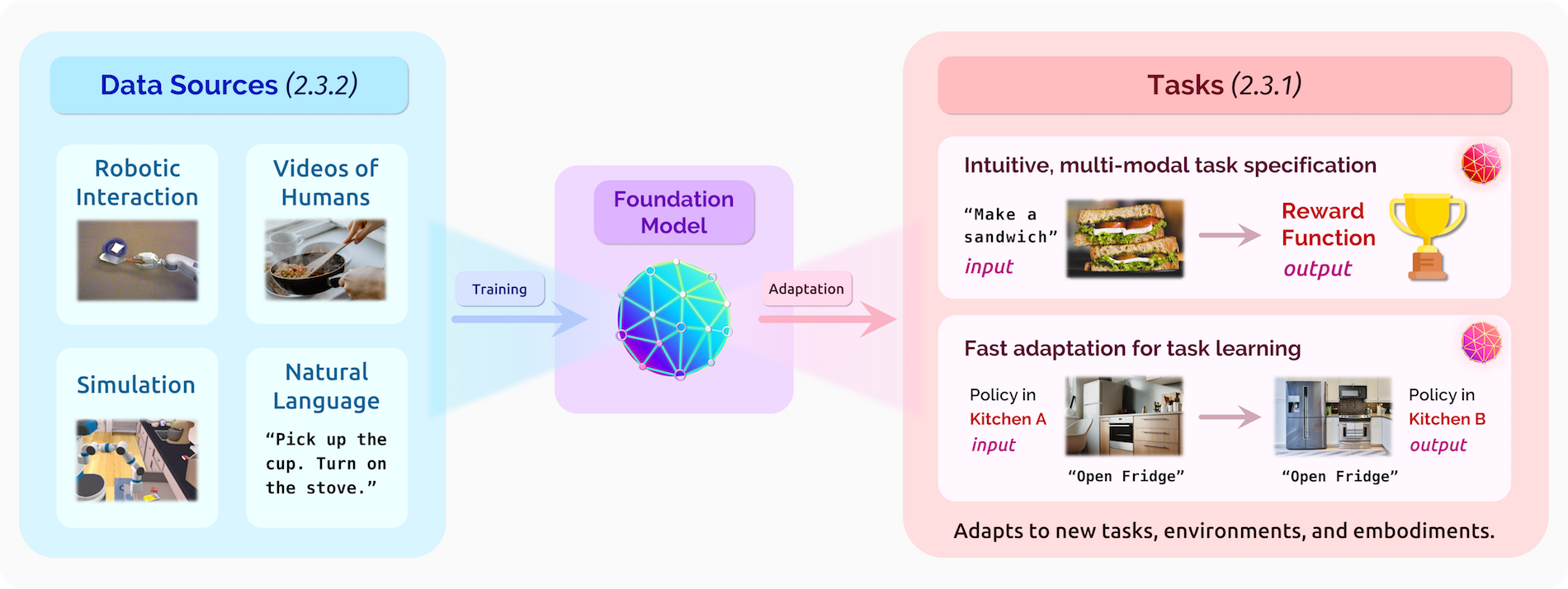}
\caption{Building new types of foundation models for robotics will require massive datasets spanning diverse environments and behaviors. Simulation, robotic interaction, videos of humans, and natural language descriptions could all be useful data sources for these models. Despite the challenges of acquiring data, developing new foundation models for robotics has tremendous potential for a variety of problem formulations in task specification and robot learning. Image credits: \cite{finn2016deep, szot2021habitat}.}
\label{fig:robotics}
\end{figure}

\noindent 
A longstanding challenge of robotics research is to endow robots with the ability to handle the myriad conditions they will encounter in real-world settings. In this section, we discuss how the ideas underlying foundation models can potentially help bring about ``generalist'' robots that can, for example, cook a new meal in a new house, with a new kitchen. To make progress towards this goal, existing foundation models will not suffice. We need new types of models trained on a multitude of data sources, spanning grounded robotic interaction data to videos of humans performing tasks, amongst others. We focus on how such foundation models can apply to the problem of a robot controlling its own \textit{physical embodiment} to successfully perform different tasks. This is a high-dimensional and closed-loop decision-making problem: the actions that a robot takes directly influence what it perceives next, which in turn influences the next robot action. This closed-loop aspect is not traditionally studied in language and computer vision, where large offline datasets are dominant and foundation models have already seen success. We focus on how the demonstrated benefits of foundation models\dash{}large-scale, self-supervised learning\dash{}can be leveraged in this new closed-loop data regime. The promise of a new type of robotic foundation model is in its ability to amplify the potential of robots to improve key facets of daily life ranging from manufacturing \citep{nof1999handbook, sanneman2020state}, construction \citep{khoshnevis2004automated, bock2007construction}, autonomous driving \citep{thorpe1988vision, badue2020self}, to household aid \citep{thrun1995lifelong, brooks2002flesh, Dillmann2004TeachingAL, goodrich2007hri, Gupta2018RobotLI, shridhar2020alfred} and personal assistance \citep{dragan2013formalizing, javdani2018shared}, amongst others. Our discussion in this section primarily focuses on mobile manipulation robots for household tasks, but we expect its essence to be broadly applicable to the other use-cases of robotics listed above.

On the critical path towards building new types of foundation models for robotics is embracing opportunities in \textit{task specification} and \textit{task learning}, coupled with tackling challenges in \textit{data acquisition} and \textit{safety and robustness}. Consider the following robot learning paradigm: starting with a description of a task capturing what a user might like the robot to do (\eg ``make breakfast'')\dash{}learn a corresponding \textit{policy} to generate the desired robot actions. While policies can be parameterized in different ways, a common choice is that of a function that maps the task representation and environment observation (\eg a scene image from a fixed or egocentric camera, or inputs from alternative sensors like LIDAR) to robot actions \citep{andrychowicz2017hindsight, nair2018rig}. As the robot acts in a task-conditioned manner, the subsequent states are fed back to the policy, generating more actions until the task has been satisfied.

Yet, implementing such a paradigm in practice is difficult. To begin, what is the right interface for describing one's goals? For a given user in one context, ``make breakfast'' carries an implication of a full breakfast that consists of fried eggs, toast, and a glass of orange juice; for another user, ``make breakfast'' may imply idlis with sambar and a tumbler of filter coffee. In general, high-level context-dependent goals like these do not stand alone and can introduce a multitude of ambiguities. How does one \textit{specify} a goal (and corresponding subgoals) with enough clarity to both resolve these ambiguities, and in so doing, allow a robot to make progress on the given task? Additionally, how might we craft general task representations that might aid generalization to similar objectives (\eg fetching a glass of milk instead of orange juice). Going a step further, how do we build methods that aid robots in \textit{learning} policies for new tasks and new environments (in this case, a brand new kitchen with new utensils, appliances, layouts, etc.)?

Recent breakthroughs in applying foundation models for language and vision (\refsec{language} and \refsec{vision}) suggest several potential benefits 
of large-scale, self-supervised pretraining for improving generalization. The ability to tap into diverse streams of data to learn meaningful representational priors (akin to those learned by models such as BERT and GPT-3) holds promise for learning powerful robotic foundation models for task specification. Diverse robotic interaction data can be used for learning action-conditional dynamics models or policies indexing general and semantically meaningful skills thereby holding promise for task learning. Yet while these opportunities exist, the key stumbling block is \textit{collecting the right data}. Unlike language and vision data, robotics data is neither plentiful nor representative of a sufficiently diverse array of embodiments, tasks, and environments\dash{}we (as a field) still have not converged on the \textit{kinds} of data that would be maximally useful for enabling generalist robotics (\eg offline demonstrations, third-person recordings of humans, egocentric videos, autonomous experience, etc.) Coupled with issues in obtaining the right scale and diversity of data are questions of ensuring safety and robustness: how do we behave in a new environment without causing damage?

Building new types of foundation models for robotics thus consists of a dichotomy of opportunities and challenges: opportunities for task specification and learning balanced against challenges of data collection and safe deployment. This section explores both by presenting a picture of how robotic foundation models might help us develop generalist robots, in a way that not only meaningfully addresses the challenges associated with building such systems, but that also embraces the potential of multi-modality\dash{}incorporating perception, actuation, and language\dash{}as well as human-robot interaction for specification and learning.

\hypertarget{robotics-opportunities}{\subsubsection{Opportunities}}
\label{sec:robotics-opportunities}

Robotic foundation models could take a variety of forms: problems in robotics do not easily conform to a one-size-fits-all model, since different problems have different input-output signatures\dash{}a contrast to domains like NLP where many problems can be cast into a general ``text-in, text-out'' signature. We focus on opportunities in generalizable task specification and learning across tasks, environments, and robot embodiments.

\paragraph{Foundation models for task specification.} Before robots can learn \textit{how} to solve tasks in a general purpose way, they must understand \textit{what} the desired task is: for example, to be useful in a new kitchen, a robot needs to know what we would like it to cook, as well as behaviors we would like it to avoid. Therefore, a necessary first step towards developing generalist robots is building a new type of foundation models for reliable task specification, \ie the intuitive and effective communication of task objectives, preferences, and constraints. We formalize task specification as a process that transforms a human-provided task description into a quantitative metric that measures a robot’s task completion and progress\dash{}\eg a reward function. This signal is crucial for optimizing robot behavior, diagnosing failures, and prompting human feedback. As the most natural way to describe a task can vary depending on the user, environment, or task, robotic foundation models for task specification should accept a variety of description modalities, such as goal states \citep{fu2018variational, singh2019endtoend}, natural language \citep{macglashan2015grounding, karamcheti2017draggns, misra2017mapping, coreyes2019guiding, shao2020concept2robot}, videos of humans \citep{shao2020concept2robot, chen2021generalizable, liu2018imitation}, pairwise or ranking comparisons \citep{biyik2018batch}, interactive corrections \citep{coreyes2019guiding, karamcheti2020decomposition} and physical feedback \citep{ross2011reduction, bajcsy2017learning}.

An important requirement of general purpose models for task specification is the ability to transfer to \textit{new} environments and tasks. Reliably transforming task descriptions into generalizable reward signals for robot learning remains an open problem \citep{taylor2016alignment}\dash{}one that robotic foundation models will arguably be well suited for. When applied to task specification, such models should provide more robust (\refsec{robustness}) reward signals by learning from large and broad datasets\dash{}even leveraging \textit{multiple} of the description modalities listed above. One possible instantiation of a new foundation model for task specification might be to learn a mapping from arbitrary (language, current observation) pairs to reward signals by training on diverse language and vision datasets \citep{bahdanau2019reward, fu2019lang2goals, chen2021generalizable}. By learning informative priors from these broad, diverse datasets, such a model may be able to generalize to unseen language instructions and observations in unseen environments. In general, the potential for new foundation models to be able to deftly bridge modalities and generalize broadly make them appealing for general purpose task specification.

\paragraph{Foundation models for task learning.} In addition to enabling more general task specification, robotic foundation models could make learning to solve new tasks more efficient and reliable. In this context, these new types of foundation models might take the form of a joint distribution over actions, sensor observations, rewards, and other properties of interest. Conditioning on different dimensions of this joint distribution recovers different inference problems, each corresponding to a different signature:
\begin{itemize} 
    \item \textit{Dynamics modeling}: $p$(future observations $\mid$ actions, past observations) \citep{finn2017deep, hafner2019latent, wu2021greedy}.
    \item \textit{Policy learning}: $p$(actions $\mid$ observations, goal) \citep{kaelbling1993learning, schaul2015uvf, ding2019goal}.
    \item \textit{Inverse reinforcement learning}: $p$(reward function $\mid$ observations, actions) \citep{ng2000algorithms, ziebart2008maximum, finn2016guided}.
\end{itemize}
A plausible training objective for a robotic foundation model is to predict the different elements of the joint distribution described above in an autoregressive fashion \citep[][\refsec{modeling}]{Janner2021ReinforcementLA, Chen2021DecisionTR}. However, these are not the only options. In particular, robot datasets contain large amounts of unlabeled data consisting of synchronized observations from many different sensor modalities (\eg RGB and depth cameras, haptic sensors, microphones, etc.) and a sequence of actions that the robot has performed to generate these observations. Beyond the objectives above, a robotic foundation model could be trained to predict observations of one sensor modality from another or to predict whether two streams of sensory observations are from the same segment of time. These kinds of self-supervised objectives can leverage multi-modal correspondences to produce low-dimensional representations of high-dimensional data, and can even be combined with the above objectives to yield models, policies, and rewards on top of those representations.These objectives may facilitate the training of powerful robotic foundation models from unlabeled data\dash{}as long as the data exhibits diverse, meaningful behavior. \refsec{robotics-challenges} discusses the challenges of collecting such data further.
 
In language and vision, foundation models have demonstrated the capability to learn broadly applicable priors from large, diverse datasets, that can be subsequently adapted to downstream tasks (\refsec{language}, \refsec{vision}). Robotic foundation models have the potential to similarly enable few-shot adaptation of perception and control to new environments, tasks, and embodiments though leveraging different data, self-supervised objectives, and modalities than have been studied with existing language and vision models. Consider our running kitchen example. To cook in a new kitchen, a robot needs to adapt to the specific environment\dash{}its spatial layout, the available equipment, etc. Priors learned from offline videos of humans, robotic interaction, text, and/or simulation might encode general aspects of kitchens, such as the fact that stoves are usually against walls and must be turned on in order to produce heat. Such commonsense knowledge, physical priors, and visual priors could make adaptation to new environments more sample efficient. Similarly, developing a new foundation model for robot task learning might enable the use of a large number of cooking videos in its training dataset to adapt a policy for a common skill, such as ``fry an egg,'' to a specific user’s preferences from a low number of demonstrations\dash{}allowing for sample efficient adaptation. Finally, with their potential to learn the cross-modal representations described earlier, robotic foundation models could help enable adaptation to new embodiments. This aspect of adaptation is crucial to make these models widely useful.

\hypertarget{robotics-challenges}{\subsubsection{Challenges and risks}}
\label{sec:robotics-challenges}

Despite this exciting vision, multiple challenges need to be overcome. To enable the generalization discussed above, we must collect robotic datasets of sufficient size and diversity. Additionally, we need mechanisms to ensure that we can deploy learned behaviors safely in the real world.

\paragraph{Data needs \& challenges.} Learning a policy for a robot that perceives the state of its environment via sensors and takes actions to accomplish tasks traditionally requires large datasets of the robot \textit{interacting in the real world}. On the other hand, many learning tasks in computer vision and natural language processing rely on large and diverse \emph{offline} datasets that can easily be scraped from the web. Motivated by the advances of existing foundation models in language and vision, we are excited by the possibility of leveraging large offline data sources for training new foundation models for robotics.

One path towards this goal is collecting large datasets for offline learning, for example using teleoperation \citep{mandelkar2019scaling}, kinesthetic teaching \citep{sharma2018mime}, or autonomous methods \citep{pinto2016supersizing, Gupta2018RobotLI, Levine2018LearningHC, dasari2019robonet, kalashnikov2021mt,chen2021batch}, which have shown some promising indications on generalization. While scaling up robot data collection to the size of vision and language datasets  \citep{deng2009imagenet, krishna2017visual, raffel2019exploring, gao2020pile} remains an open challenge, the increasing scale and quality of robotic datasets suggests they can play an important role in learning foundation models for robotics. Moreover, because robots have the ability to actively and autonomously shape their environment, they should be able to generate targeted unlabeled data at scale.

Given the challenging closed-loop nature of learning control, it is possible that collecting datasets of size comparable to those used in vision and language is insufficient for robotics. One exciting option is to additionally leverage external, non-robotic sources of data such as videos of humans or existing vision and natural language datasets. Such data is diverse and exists in large quantities on the web \citep{deng2009imagenet,Lee2012DiscoveringIP, caba2015activitynet, Goyal2017TheS, damen2018kitchens, gao2020pile,grauman2021ego4d}, affording the possibility of broad generalization if properly leveraged. Elegantly addressing the gap between the robot’s domain and those found in videos or language on the web remains an open challenge; however, recent progress in domain adaptation \citep{Smith2019AVIDLM, Schmeckpeper2020ReinforcementLW} and using pretrained video and language models in robotics \citep{lynch2020grounding, shao2020concept2robot, chen2021generalizable} present promising directions towards closing this gap.

Finally, simulation presents a boundless source of rich interactive data that robots can learn from, with a range of sensor modalities like rendered visuals, point-clouds, and simulated touch/audio. However, a major challenge lies in bridging the gap between simulation and the real world, both in the underlying physics and in the semantic distribution of environments and tasks. Recent work has shown that by using extensive domain randomization, tasks ranging from flight \citep{Sadeghi2017CAD2RLRS} to contact-rich manipulation \citep{Mahler2017DexNet2D, OpenAI2019SolvingRC} and locomotion \citep{RoboImitationPeng20, hwangbo2019learning} skills learned in simulation can be transferred to real robots with some success, and that the semantic and visual distribution of the real world can be simulated by scanning the real world into a simulation \citep{Matterport3D, Kolve2017AI2THORAI, embodied, szot2021habitat, shen2021igibson}. While these are promising steps towards closing the sim-to-real gap, effective and general sim-to-real learning of manipulation and locomotion skills remains an open challenge. Simulation data, real robot data, videos of humans, and natural language data could all be essential to learning robotic foundation models.

\paragraph{Safety \& robustness.} Further complicating the development of new foundation models for robotics is ensuring their safety and robustness when training or deploying them in the real world. We can expect the safety risks from these models for robotics to be different from their language counterparts given that embodied agents are empowered to manipulate and interact with their surroundings directly in the physical world. One core safety challenge for learning-based systems is the chicken-and-egg problem of needing to specify system constraints for safety prior to collecting data, after which unforeseen unsafe behaviors requiring additional constraints may emerge. For instance, an agent adapting to a new kitchen outside of the training distribution requires sufficient safety guarantees to ensure safe data collection, which may either adversely affect task performance or cause the agent to fail in novel ways. One way to resolve this is restricting the complexity of the environment or increasing the complexity of the robot such that irrecoverable states or unsafe actions are avoided by construction. The robot can also be tasked with autonomously resetting the environment to facilitate uninterrupted learning (or adaptation) from large-scale data collection \citep{eysenbach2017leave, gupta2021reset}. This would either mean ensuring that nothing in the kitchen is breakable, or ensuring and replacing the items the agent may break while it attempts to collect data.

To address risks posed by robotic foundation models that fail to generalize or produce unexpected behaviors to new stimuli, potential future directions include developing a causal analysis of agents \citep{deletang2021causal}, new formal safety evaluation tools, and realistic simulation environments \citep{corso2020survey, dreossi2017compositional, Julian_2019}. Finally, deriving formal safety guarantees for robotic foundation models, \eg Hamilton-Jacobi reachability of safe-sets \citep{chow2018lyapunov, fisac2019bridging, herbert2021scalable} or developing safety boundaries for learning that are interpretable (\refsec{interpretability}) to human operators, could help reduce risks posed by such models \citep{berkenkamp2017safe}. As the development and study of these new types of foundation models progresses, solutions to these challenges will be crucial.

\paragraph{Conclusion.} While the promise of robotic foundation models are many\dash{}spanning multiple levels of the robotics pipeline from task specification to task learning\dash{}the challenges are significant. Collecting data \textit{in the physical world} that covers diverse environments and embodiments at scale is a sizable hurdle, and ensuring the safety and robustness of such systems is equally exigent. Despite this, our optimism prevails; tackling these challenges now, \textit{before} developing models offers us the chance to identify ways to collect the right data, from the right sources, at the right scale to build safe and reliable robotic foundation models with the capabilities we desire. 

Underpinning this section has been a theme of multimodality. Robotic foundation models\dash{}in all possible instantiations\dash{}have and will continue to benefit from work in other subfields of AI such as language and vision (\refsec{language}, \refsec{vision}). Yet as we consider incorporating these extensions from other fields, there are interdisciplinary challenges on the horizon that touch other aspects of foundation models: systems innovation for training \textit{and deploying} such models for real-time robotics (\refsec{systems}), innovation in interfaces for robust human-robot interaction (\refsec{interaction}), and lessons to incorporate as we better grasp the safety and robustness of such models (\refsec{ai-safety}, \refsec{robustness}). Building a reliable ecosystem and thoughtful research practices around foundation models in general, and robotic foundation models in particular, is key to realizing these goals.

\newsection
\hypertarget{reasoning}{\subsection{Reasoning and search}}
\label{sec:reasoning}
\sectionauthors{Yuhuai Wu, Frieda Rong, Hongyu Ren, Sang Michael Xie, Xuechen Li, Andy Shih, Drew A. Hudson, Omar Khattab}

\begin{figure}[!ht]
\centering
\includegraphics[width=\linewidth]{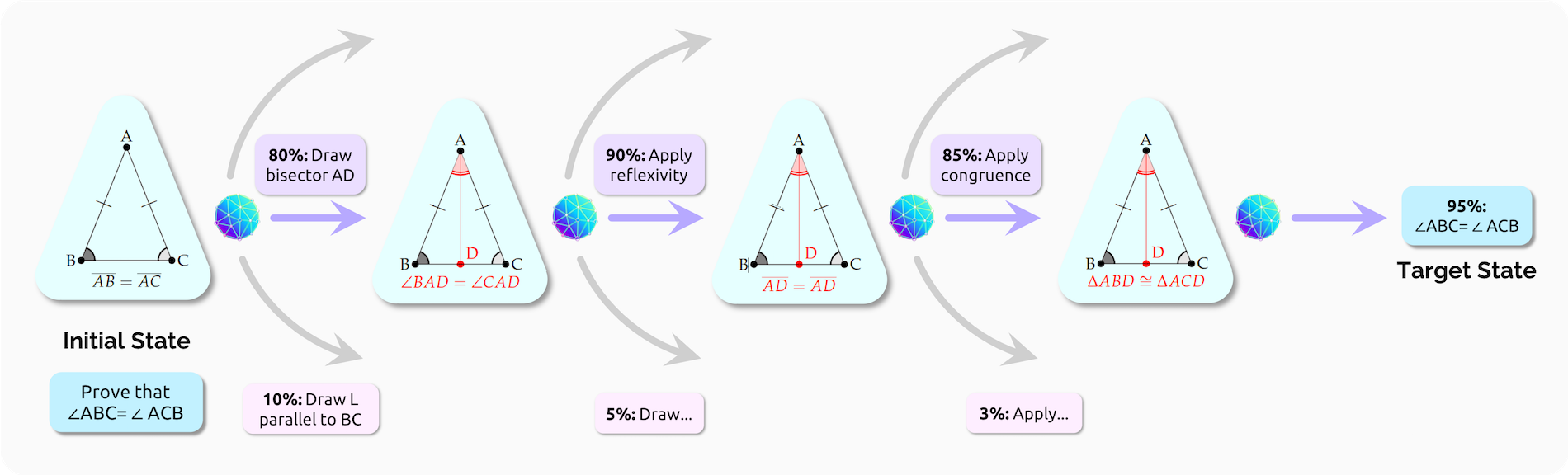}
\caption{\label{fig:reasoning-geometry} Multimodality can allow foundation models to not only reason with formal symbolic language, but also exploit \emph{visual aspects} of the problem, such as equivalence, symmetry, and Euclidean geometry, to prune the infinite search space and find promising constructions for a solution (\protect\refsec{reasoning-tasks}), mimicking the way humans reason about geometry problems.}
\end{figure}

Reasoning and search have been a central theme throughout the history of AI. Classic tests of intellect, from strategy games to abstract mathematical discovery, served as inspirational goal posts that pushed the limits of ``machine intelligence'' through a need to devise ever smarter ways of searching for winning solutions.
In the early days, symbolic methods were the dominant approach for reasoning~\citep{DBLP:books/aw/RN2020}, but the involved engineering effort and the need to formalize heuristics to tackle intractable search spaces quickly proved cumbersome. 
More recently, data-driven methods using neural networks have shown encouraging results\dash{}\eg defeating the best humans in Go~\citep{DBLP:journals/nature/SilverHMGSDSAPL16}, a board game with a much larger space of actions than the classic challenge of chess\dash{}by exploiting statistical structures and learning useful heuristics.
This section outlines existing reasoning tasks, ones that require scaling to ever-larger search spaces and understanding the world broadly (\refsec{reasoning-tasks}).
We then argue in \refsec{reasoning-role} that foundation models should play a central role towards general reasoning as vehicles for capturing the statistical regularities of unbounded search spaces (\emph{generativity}), allowing positive transfer across tasks and scenarios (\emph{universality}), and exploiting the grounding of knowledge in multi-modal environments (\emph{grounding}). 

\hypertarget{reasoning-tasks}{\subsubsection{What are the current tasks?}}
\label{sec:reasoning-tasks}

Many reasoning problems pose unbounded search spaces, where systems must deal with numerous kinds of open-ended alternatives. Consider trying to prove that the angles $\angle B$ and $\angle C$ are equal for an isosceles triangle $\triangle ABC$ with $AB=AC$ (\reffig{reasoning-geometry}). A system can perform any number of actions \textit{at each step of reasoning}. For instance, the system could add a new auxiliary point with an arbitrary construction, say a perpendicular line, a parallel line, or a tangent circle, and the search space only grows larger as the diagram grows more complicated. One way to prove this theorem is to draw a line $AD$ that is the angle bisector of $A$, and use the congruence of the two triangles $\triangle ABD$ and $\triangle ACD$ to show $\angle B = \angle C$, but how can systems find this without extensive search?

More generally, a mathematician is not confined with searching in diagram constructions and Euclidean theorems: mathematicians can apply a vast number of theorems from various branches of mathematics, make high-level conjectures, formalize new mathematical concepts, or find counterexamples. This contrasts with more structured AI challenges such as the game of Go, whose search space is considered much smaller.\footnote{Less than the number of grid points on the Go board (\ie 361 actions for a 19$\times$19 board).}

\begin{figure}[t]
\centering
\includegraphics[scale=0.35]{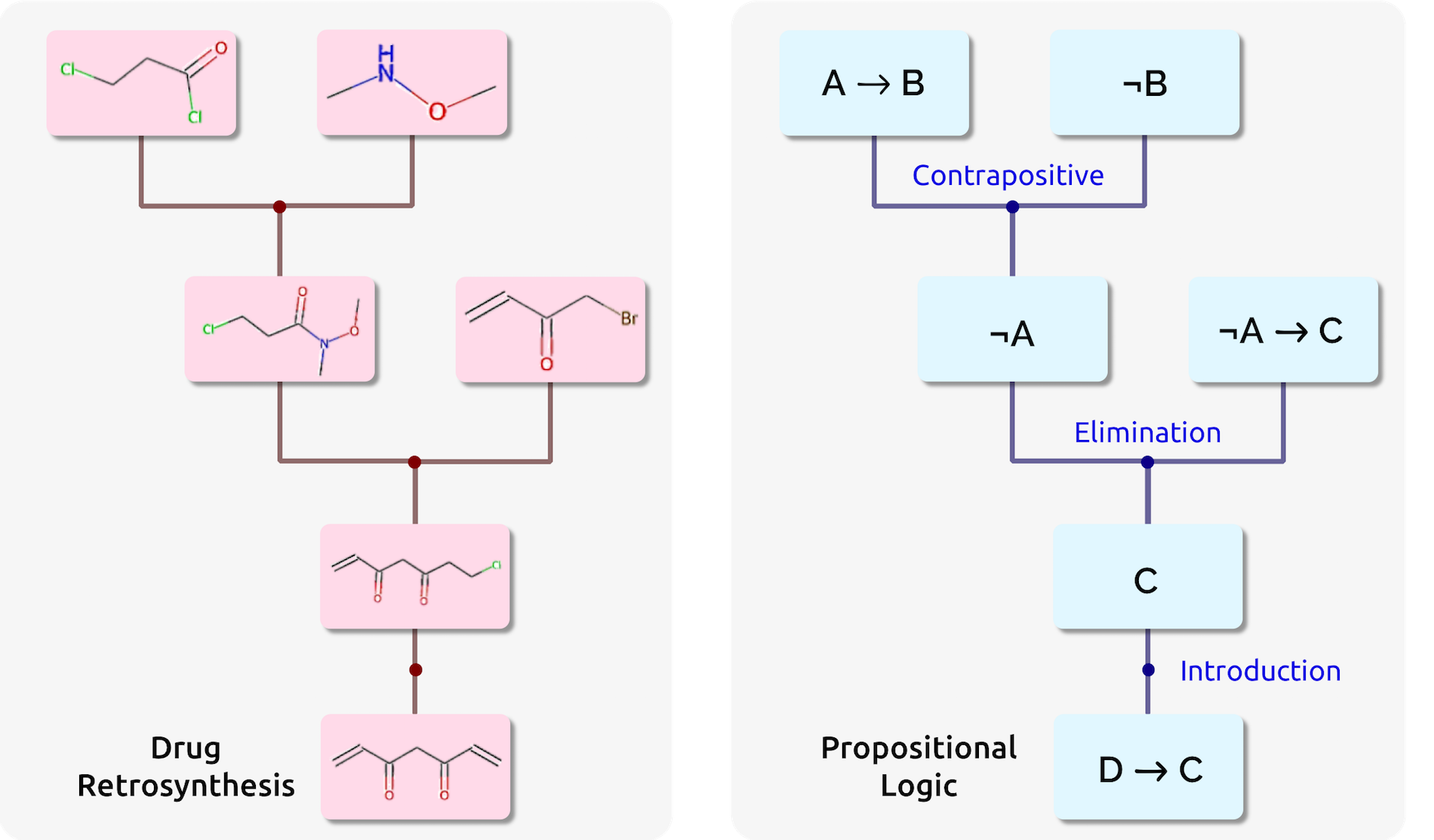}
\caption{\label{fig:reasoning-retrosynthesis} \textbf{Left:} A reaction route for 1,6-Heptadiene-3,5-dione predicted by machine learning-based drug retrosynthesis planner AiZynthFinder \cite{genheden2020retrosynthesis,yoshikawa2021retrosynthesis-twitter}. \textbf{Right:} A sample proof tree in propositional logic where the formulas outlined in green represent axioms. Although they arise from different domains, both trees are structurally the same.}
\end{figure}

Besides theorem proving, many real-world problems deal with unbounded search spaces, such as program synthesis~\citep{DBLP:journals/ftpl/GulwaniPS17}, drug discovery~\citep{Drews2000DrugDA}, chemical synthesis~\citep{DBLP:journals/nature/SeglerPW18}, computer-aided design~\citep{computer_aided_design}, combinatorial optimization~\citep{DBLP:journals/eor/BengioLP21}, and more. 
These reasoning problems tend to exhibit similar structure, like the bijection between retrosynthesis in drug discovery and theorem proving in propositional logic, illustrated in \reffig{reasoning-retrosynthesis}: in both problems, one is building a tree of synthesis, whose nodes are chemical products on the one side and propositions on the other, and the leaf nodes are the products on the one side, and end axioms on the other. 
In these problems, a simulated environment is often provided, which allows a solver to run several search threads towards building the solution tree. The simulator often provides intermediate feedback, say, informing the solver with the remaining propositions to establish before the proof is considered complete. The solver in turn needs to select the most promising search thread and proceed based on the intermediate feedback.

Recently, there has been a surge of interest in applying learning-based approaches to tackle reasoning problems. 
To overcome the unbounded search space challenge, researchers first started with a constrained search space to make the problem tractable~\citep{DBLP:journals/corr/abs-1806-00608, DBLP:conf/icml/BansalLRSW19}. 
But such approaches suffered from the limited kinds of actions the solver could issue. For example, the solver could only apply theorems from a known database to prove the target theorem, instead of synthesizing novel theorems and lemmas. 
Because large language models offered a generic way of modeling the output space as a sequence, they quickly became a more favorable choice, allowing the generation of arbitrary kinds of actions.
Researchers have applied these language model-based approaches to various applications, such as predicting protein structures~\citep{DBLP:journals/nature/Senior0JKSGQZNB20}, proving formal theorems~\citep{DBLP:journals/corr/polu2020generative,pact},  conjecturing theorems~\citep{DBLP:conf/mkm/UrbanJ20,DBLP:journals/corr/abs-2006-04757,li2021isarstep}, synthesizing programs from natural language~\cite{chen2021evaluating,ling2016latent}, repairing, generating and understanding code ~\citep{yasunaga2021break,DBLP:journals/corr/abs-2102-04664,guo2020graphcodebert,svyatkovskiy2020intellicode,kim2021code,zugner2021language}. It has also been shown that scaling model size significantly improves reasoning capabilities~\citep{DBLP:journals/corr/polu2020generative}, and furthermore standard techniques from language modelling, such as pretraining, can also greatly improve performance on these tasks~\citep{DBLP:journals/corr/abs-2006-04757,DBLP:journals/corr/polu2020generative}.

\hypertarget{reasoning-role}{\subsubsection{What's the role of foundation models?}}
\label{sec:reasoning-role}

\paragraph{Generativity.} We believe that the generative capabilities of foundation models are essential for effective reasoning. Due to the unbounded search space, it becomes intractable to enumerate all kinds of possibilities. Instead, with foundation models, one can model the distribution of the optimal decisions, and \emph{generate} suitable candidates to proceed to the next step. In particular, as foundation models offer a generic way of modeling the output space as a sequence, the next decision generation is entirely unconstrained and hence universal. Such flexibility is essential for many of the reasoning challenges we discussed, to allow creative generation in domains such as mathematical conjecturing~\cite{li2021isarstep} and synthesizing novel programs~\cite{chen2021evaluating}. As one scales up foundation models, the capabilities of capturing such statistical structures also grow immensely~\citep{DBLP:journals/corr/polu2020generative}.

\paragraph{Universality.} As we mentioned in the last section, many reasoning problems exhibit similar latent structures. We believe that the unifying framework imposed by a foundation model can transfer and share significant heuristics across tasks, ranging from generalizing low-level techniques that work well for one task to new scenarios all the way to directly finding meta-techniques that work well across numerous kinds of problems. In addition, since a foundation model is trained across many domains, it can positively transfer meta-knowledge encoded in the foundation models' weights across tasks and domains~\citep{papadimitriou2020learning, lime,DBLP:journals/corr/abs-2103-05247}. The foundation model training and adaptation framework encourage a separation of concerns, where foundation model training learns meta-knowledge such as the shared search tree structure between drug retrosynthesis and propositional logic proofs, and the adaptation phase can focus on learning the task specific vocabulary. Thus, foundation models can reduce the complexity of the learning problem in the adaptation phase, improving sample complexity and generalization.

\paragraph{Grounding.} Reasoning problems are often easily expressed in symbolic languages (\eg mathematics, code, SMILE representation of molecules). However, these symbols have deep underlying semantic meanings\dash{}saying ``isosceles triangle'' paints a vivid image in the human mind. Foundation models can enable deep groundings and semantic meanings. First, grounding representations in other \emph{modalities}, such as visual or physical, are essential to grasp abstract concepts in reasoning tasks and endow them with concrete meaning~\citep{Larkin1987Simon,Jamnik2001MathematicalRW}. Since the models may be trained on multiple modalities, foundation models can assist in understanding a range of data sources (\eg images, texts). Hence, in the geometry example case, with its understanding of geometrical shapes learned in natural images, a foundation model could effectively utilize the diagrammatic representation of the problem. However, aligned multi-modal data in reasoning is scarce, and it remains an open question whether foundation models can discover connections between different modalities in an unsupervised manner (\eg discovering of commutative diagram with the corresponding algebraic equations).
Furthermore, even within the symbolic domain, symbols can have various levels of interpretation. 
For example, high-level programming languages can be translated to low-level assembly codes. Foundation models can learn a shared representation that encompasses these various views. Past works have shown that self-supervised tasks~\citep{pact, DBLP:journals/corr/abs-2105-04297,DBLP:journals/corr/abs-2103-03809} allow the model to understand the inner workings behind the high-level code scripts, and further assist downstream tasks. 

\subsubsection{Future challenges in reasoning}
Due to the intrinsic difficulty of these problems, high-quality annotated data is scarce and harder to collect compared to raw images and text. 
There have been several attempts towards alleviating this issue. In mathematics, researchers proposed to generate synthetic theorems in the hope of generalizing to realistic theorems~\citep{DBLP:conf/nips/WangD20,wu2021int,vlad2021mathai,refactor}. Another approach is to design self-supervised tasks to augment datasets~\citep{yasunaga2020repair,ren2020query2box,pact,DBLP:journals/corr/abs-2102-07492,yasunaga2021break}, or better pretraining objectives~\citep{lime}. However, we still lack general principled approaches in designing self-supervised tasks, as most of the existing works are tailored to specific problem setups~\citep{yasunaga2020repair,ren2020beta,pact}. 
Building a foundation model will encourage a unifying framework of constructing a suite of self-supervised tasks that can be applied to all reasoning problems. 
In addition, interactivity (\refsec{interaction}) could, with enough scalability, alleviate the data scarcity problem by bringing humans into the loop to minimally guide the learning curriculum or data augmentation process, for example, in selecting axioms to add or conjectures to explore, while interactive tools themselves are a motivating use of foundation models for reasoning~\citep{pact,chen2021evaluating} in assisting people with the most cognitively demanding or laborious aspects. Interpretation-friendly interactive tools could find further applications in education by assisting humans in learning with the help of highly capable foundation models (\refsec{education}). 

Improving the high-level reasoning capabilities is a core challenge for existing foundation models. Humans perform abstract reasoning and high-level planning in tackling difficult problem-solving tasks~\citep{MillerGalanterPribram60}. 
For example, when building a software tool or proving a theorem, we often start with a high-level sketch before delving into the low-level details~\cite{KOEDINGER1990511}. 
Existing foundation models are not trained to generate such high-level plans. Instead, they often focus solely on predicting the next low-level steps~\citep{DBLP:journals/corr/polu2020generative, pact, chen2021evaluating}. Unfortunately, to train foundation models to emulate human-like reasoning, we again face a data collection challenge. Although such data does exist in limited settings~\citep{li2021isarstep}, in general, data for high-level reasoning is scarce and difficult to collect. One line of research is to let abstract and modular hierarchy to emerge by itself during learning~\citep{DBLP:conf/pldi/EllisWNSMHCST21,hong21latent}, but it still remains an open question how to scale these approaches to more general and realistic settings. 

Aside from these challenges, there exist many open questions that are also essential to topics discussed in other sections. What constitutes a good architecture for reasoning reliably (\refsec{modeling})? How can we understand and interpret these models theoretically (\refsec{theory} and practically \refsec{interpretability})? Can we train robust reasoning models that could generalize to out-of-domain problems (\refsec{robustness} and \refsec{adaptation})? We believe research about foundation models on each of these fronts can greatly broaden their impact for the field of reasoning.

\newsection
\hypertarget{interaction}{\subsection{Interaction}}
\label{sec:interaction}
\sectionauthors{Joon Sung Park, Chris Donahue, Mina Lee, Siddharth Karamcheti, Dorsa Sadigh, Michael S. Bernstein}

\begin{figure}[!ht]
\centering
\includegraphics[width=\linewidth]{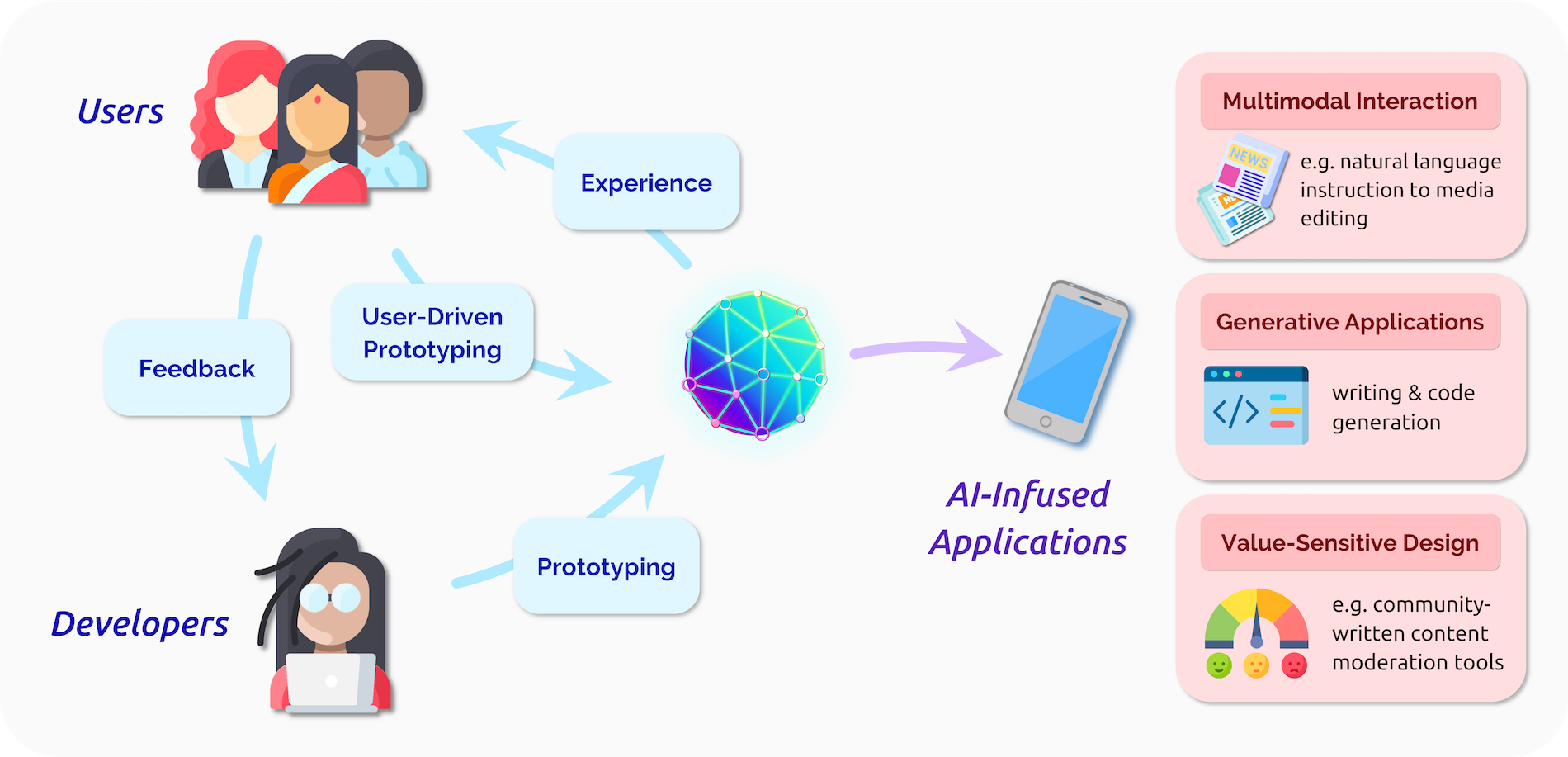}
\caption{\label{fig:interaction} Foundation models will bring significant opportunities to developers by lowering the difficulty threshold for building AI-infused applications, and to the application users by raising the ceiling for what types of interactions are achievable. In some cases, the line between developers and users will start to blur, and users may be able to easily develop their own AI applications, for instance with natural language.}
\end{figure}

The early forms of foundation models such as GPT-3~\citep{brown2020gpt3} and DALL·E \citep{ramesh2021zeroshot} have demonstrated a high level of versatility both in terms of their ability to let even non-ML experts to prototype powerful AI-infused applications, and their ability to seamlessly integrate modalities ranging from texts to images. As the development of foundation models matures, the models' capacity will continue to expand and their versatility may ultimately lead to fundamental changes in how we interact with AI by allowing us to rapidly prototype and build highly dynamic and generative AI-infused applications. In this section, we discuss the opportunities that these changes present from the perspectives of two important stakeholders: (1)~applications developers who will interact with foundation models to design user experience, and (2)~end-users who will use or be affected by the AI-infused applications powered by foundation models. Finally, we consider scenarios in which the line that rigidly separates developers and end-users today may start to blur, affording new opportunities for creating AI-infused applications that more closely satisfy users' needs and values.

\subsubsection{Impact on AI-infused application developers' development process} 
How will foundation models transform the way developers create AI-infused applications? Despite the monumental progress in machine learning algorithms and systems infrastructure, some point out that designing novel and positive forms of human-AI interaction remains  difficult~\citep{Dove2017uxdesign, Cooper2014face}. The vast amount of data, computing resources, and skills needed to create a powerful task-specific model is frequently in conflict with the iterative prototyping process necessary to elicit and satisfy users' needs and values~\citep{Yang2016wireframing}. This challenge is further compounded by the fact that AI responses can be unpredictable, and models can produce a vast generative output space, making it difficult for people to build effective mental models of their performance. There has already been some progress on tackling these challenges in the form of work on interactive machine learning (\eg Crayon~\citep{Fails2003design}, Regroup~\citep{Amershi2012regroup}) and design frameworks for conveying uncertainty in AI to end-users (\eg principles of mixed-initiative~\citep{Horvitz1999mixedinit}). However, more work is still needed to overcome these obstacles~\citep{yang2020aiexamine}.

Foundation models pose important opportunities to address many of the challenges mentioned above. For instance, language-based foundation models' ability to take natural language as input, and to generalize to many downstream tasks, could significantly lower the difficulty ``threshold''~\citep{Myers2000ui} for application development, \ie~by enabling the development of sophisticated models without having to collect significant amounts of data and train large models from scratch. This could enable even non-ML experts to quickly prototype AI-infused applications. 
At the same time, the powerful generative and potentially multi-modal capabilities of foundation models could offer a far higher ``ceiling''~\citep{Myers2000ui} of what types of interactions are achievable both in terms of their quality and diversity as we will discuss below. However, how successfully we can leverage these capacities will depend on how effectively we can wrangle foundation models into forms that will be more manageable by application developers.

Unfortunately, the same generalizability and high ceiling that give foundation models their edge can also make these models difficult to work with, as they may be even more unpredictable and complex than single-purpose AI models. Indeed, recent work has shown that it can be difficult to make models like GPT-3 consistently perform the intended task~\citep{Reynolds2021prompt}, while understanding what it is capable of is still an active area of research~\citep{Hendrycks2020Measuring}. In an effort to improve the reliability and trustworthiness of AI-infused applications, we recommend that future work should continue to investigate how to achieve more predictable and robust behaviors from foundation models (\eg through fine-tuning, or in cases where the main mode of interaction is natural language prompt, through prompt-engineering~\citep{Reynolds2021prompt, Liu2021incontext}, calibrating~\citep{Zhao2021calibrate}, or pre-formatting a task-specific endpoint.\footnote{https://beta.openai.com/docs/guides/classifications} Please see \refsec{robustness} for more details). 

\hypertarget{interaction-userimpact}{\subsubsection{Impact on end-user interaction with AI-infused applications}} 
\label{sec:interaction-userimpact}
Beyond the new ways developers might create AI-infused applications, what changes will foundation models bring to the experience for end-users interacting with these applications? Existing design frameworks for developing user-facing AI applications focus on augmenting (rather than replacing) users' abilities as described by Douglas Engelbart~\citep{Engelbart1963augment}\dash{}we expect that these frameworks should and will remain relevant for the development of future AI-infused applications. For instance, maintaining users' agency and reflecting their values will continue to be a central theme for foundation model-powered applications. Additionally, the benefits of allowing AI agents to take initiatives and automate users' routines versus the benefits of waiting for users' direct manipulation~\citep{Shneiderman1997direct} will need to be carefully weighed~\citep{Horvitz1999mixedinit}. Moreover, users' values should be directly gathered and reflected through processes such as participatory~\citep{Lee2019} and value-sensitive design~\citep{smith2020keeping} that advocate for actively involving all stakeholders during the designing of the AI-infused applications. 

These issues may become especially salient with foundation models because the model may behave in ways that surprise and disappoint users and communities. Generative capabilities might expose biases or points of view that are counter to the communities' goals, or more insidiously, draw on such associations in their behavior without the community being aware. This will place a large burden on the groups utilizing foundation models to monitor their models' behavior, and to the extent possible, adapt them to act in appropriate ways.

While the design frameworks for thinking about AI-infused applications to augment users' abilities should remain the same, the actual forms of interactions that are attainable may dramatically diversify due to foundation models' powerful generative and multi-modal capacities. Already, early generations of what can be considered foundation model-powered software tools for multimedia creation and editing have started to drive a new frontier that empowers even novice content creators to generate high-quality multimedia from coarse, intuitive specifications (\eg collaborative authoring for writers \cite{lee_coauthor_2022}, text-to-image generation for digital artists,\footnote{https://github.com/nerdyrodent/VQGAN-CLIP} mastering for musicians,\footnote{https://www.landr.com/} and code completion for programmers).\footnote{https://copilot.github.com/} Improved foundation models might enable even more ambitious tools (\eg a fan might provide thematic material for a song which will then be generated in the style of their favorite band, or a business owner might provide simple descriptions of their product which will be used to create a full website). Moreover, foundation models will be used to enrich static multimedia (\eg automatically remastering legacy multimedia content into new formats, or generating unique experiences for each player in new video games) and may even lead to new forms of multi-modal interactions using interfaces that themselves mix different modalities, such as visual and gesture-based interaction. 

We are starting to see glimpses of how foundation models might materialize into concrete interactions in applications ranging from AI Dungeon\footnote{https://play.aidungeon.io/main/home} to Microsoft PowerApps\footnote{https://powerapps.microsoft.com/en-us/} and CoPilot.\footnote{https://copilot.github.com/} As we start to envision new forms of interactions, it is of increasing importance for us to think critically about the potential implications these interactions will have on individual users and society to maximize their positive impact. For example, how will foundation model-powered applications change the way we communicate with one another? Will a powerful model write emails in our stead and if so, how will this reshape people's trust, credibility, and identity knowing that the writers may not have written the emails themselves, and how will this alter our writing styles~\citep{Hancock2020AI}? Who will own the authorship of the model-generated content and how could the shifting responsibilities and ownership of the consent be misused~\citep{Weiner2018} (see \refsec{economics} for a more in-depth discussion)? What are the long-term implications that foundation models will have on our work, language and culture~\citep{Hancock2020AI, Buschek2021writing}? Of particular relevance to this last question is the fact that foundation models are trained on observed data and do not necessarily inform us about causality. Hence, how can we ensure that the use of foundation models leads us to a desired future and not a repetition of the past? Though these issues are not necessarily unique to foundation models, they will be amplified and become more prevalent as foundation models accelerate the creation of effective AI-infused applications. 

\hypertarget{interaction-blurring}{\subsubsection{Blurring the line between developers and end-users}}
\label{sec:interaction-blurring}
Today, the line that separates the developers of AI models and end-users is rigid\dash{}it is rarely the case that an end-user has the data, computing resources, and expertise to be able to develop a new model that suits one's values and needs well. While a generic model (\ie~one that is not specific to a specific user or community) could be sufficient in some cases, recent years have seen an increasing number of scenarios in which such models fail to serve users. For instance, a text classification model designed to identify problematic comments for one online community might work well for that community but will fail in others whose norms and cultures may differ significantly (\eg NSFW communities on Reddit might be more tolerant of certain content, while science communities might reject seemingly mundane anecdotes that are not based on scientific research)~\citep{Chandrasekharan2018internet}. In another example, AI-powered sensors and robotics tools designed for one target population may fail without the ability to quickly adapt in-context for users with different abilities and needs~\citep{Karamcheti2021latent}. While recent work has presented promising avenues for future research on how end-users may be able to co-create AI models by manually providing models' parameters or datasets (\eg WeBuildAI~\citep{Lee2019}), the results are still preliminary and often focus on rudimentary models. 

If foundation models can sufficiently lower the difficulty threshold for building AI-infused applications, they could present an important opportunity to more tightly couple users' needs and values with the models' behaviors by allowing users to actively partake in the development process of the models. Recent work has shown that GPT-3, for example, can robustly perform classification tasks in a few-shot or even in zero-shot fashion when given an adequate task description in its natural language prompt~\citep{brown2020gpt3}. An online community trying to moderate its own content might be able to leverage such a capability to create bespoke AI classifiers that filter content based on classification task descriptions that the community has agreed on (of course, this power could also be instead misused to silence the voices of certain members within the community\dash{}we point to \refsec{misuse} for further discussion on this topic). In addition, the powerful in-context learning capabilities that foundation models will exhibit may allow foundation model-powered applications to more effectively optimize their interfaces on a per-user basis. This could open doors to tackling many salient problems in human-computer and robot interaction such as balancing the power of users' direct manipulation and automation in mixed-autonomy settings. 

Of course, there will still be important challenges that we would need to overcome to truly realize this potential for blurring the line between users and developers. These challenges include mitigating existing biases in foundation models, as well as making the models' behavior more robust and manageable even for non-ML experts (compared to ML experts, it could be even more difficult for non-ML experts to understand the full capacities and mechanisms of foundation models, which can lead to unexpected pitfalls in the development cycle~\citep{QianYang2018}). Future work should explore how foundation models could be situated in the context of interactive machine learning and study how we can support even those with limited experience with machine learning to leverage these models in a robust manner. Nonetheless, the ability for end-users to be involved in developing AI-infused applications is an exciting opportunity that could introduce a new paradigm for how we will interact with these applications in the future. 

\newsection
\hypertarget{philosophy}{\subsection{Philosophy of understanding}}
\label{sec:philosophy}
\sectionauthors{Christopher Potts, Thomas Icard, Eva Portelance, Dallas Card, Kaitlyn Zhou, John Etchemendy}

What could a foundation model come to understand about the data it is trained on? An answer to this question would be extremely informative about the overall capacity of foundation models to contribute to intelligent systems. In this section, we focus on the case of natural language, since language use is a hallmark of human intelligence and central to the human experience. 

The best foundation models at present can consume and produce language with striking fluency, but they invariably lapse into the sort of incoherence that suggests they are merely “stochastic parrots” \citep{bender2021}. Are these lapses evidence of inherent limitations, or might future foundation models truly come to understand the symbols they process?

Our aim in this section is to clarify these questions, and to help structure debates around them. We begin by explaining what we mean by \textit{foundation model}, paying special attention to how foundation models are trained, since the training regime delimits what information the model gets about the world. We then address why it is important to clarify these questions for the further development of such models.  Finally, we seek to clarify what we mean by \textit{understanding}, addressing both what understanding is (metaphysics) and how we might come to reliably determine whether a model has achieved understanding (epistemology).

Ultimately, we conclude that skepticism about the capacity of future models to understand natural language may be premature. It is by no means obvious that foundation models alone could ever achieve understanding, but neither do we know of definitive reasons to think they could not.

\subsubsection{What is a foundation model?}

There is not a precise technical definition of \textit{foundation model}. Rather, this is an informal label for a large family of models, and this family of models is likely to grow and change over time in response to new research. This poses challenges to reasoning about their fundamental properties. However, there is arguably one defining characteristic shared by all foundation models: they are \textit{self-supervised}. Our focus is on the case where self-supervision is the model’s only formal objective.

In self-supervision, the model’s sole objective is to learn abstract co-occurrence patterns in the sequences of symbols it was trained on. This task enables many of these models to generate plausible strings of symbols as well. For example, many foundation models are structured so that one can prompt them with a sequence like ``The sandwich contains peanut'' and ask them to generate a continuation – say, ``butter and jelly''. Other models are structured so that they are better at filling in gaps; you might prompt a model with ``The sandwich contains \_\_ and jelly'' and expect it to fill in ``peanut butter''. Both capabilities derive from these models’ ability to extract co-occurrence patterns from their training data.

There is no obvious sense in which this kind of self-supervision tells the model anything about what the symbols mean. The only information it is given directly is information about which words tend to co-occur with which other words. On the face of it, knowing that ``The sandwich contains peanut'' is likely to be continued with ``butter and jelly'' says nothing about what sandwiches are, what jelly is, how these objects will be combined, etc. This might seem to suggest an inherent limitation on what a foundation model could achieve. However, we need not restrict the model to seeing only textual input. A foundation model might be trained on a wide range of different symbols: not just language but also computer code, database files, images, audio, and sensor readings. As long as it is just learning co-occurrence patterns of the sequences it is exposed to, then it counts as a foundation model by our definition. As part of this learning, the model might come to represent strong associations between a given piece of text and a particular sensor reading, or between a sequence of pixel values and a database entry. These associations might reflect important aspects of the world we inhabit and the language we use to talk about it.

\subsubsection{What is at stake?}

Before considering analyses of what understanding is, it is worth reflecting on why we might care about the question of whether a foundation model could achieve it. These models are poised to be deployed for numerous purposes with various functionalities. Some of our goals in deployment may only be met to the extent that the model is capable of understanding. Here we list a few such goals:

\begin{itemize}
    \item \textbf{Trust}: One might argue that we cannot trust a system’s linguistic behavior unless it understands the language it is using. Of course, we currently trust engineered systems to do things (\eg manufacturing auto parts) without the question of understanding even arising, but language might be special in this regard, since it is uniquely human. In addition, language can be used to deceive and misrepresent, so understanding alone clearly does not imply trust. On the whole, then, understanding might be taken as a necessary condition for trust in the context of language use.
    \item \textbf{Interpretability}: If genuine natural language understanding in some way involves maintaining and updating an internal model of the world (including, \eg the speech context), and if we (as engineers) are able to analyze how linguistic input and output interface with this internal model, that could afford substantial gains in interpretability, predictability, and control of these systems.
    \item \textbf{Accountability}: Not unrelated to the previous points, in the future we may find it desirable to hold artificial agents in some way accountable for the language they produce \citep{thehaiadaptiveagentsgroup2021when}. Depending on how we think about concepts like accountability, responsibility, agency, and the like, language understanding may emerge as a prerequisite.
\end{itemize}

\noindent The mere possibility that understanding will play an indispensable role in any of these matters provides strong motivation to develop a framework for theorizing about it. 

\subsubsection{What is understanding?}

Our central question is whether a foundation model could come to understand a natural language. With the above, we can now sharpen it: is self-supervision sufficient for understanding, keeping in mind that there are no constraints on the data used for this supervision? In order to address this question, we first need to define what we mean by \textit{understanding}.

As a start, we find it helpful to make explicit a distinction that is sometimes conflated in discussions of the topic. The distinction is between the \textit{metaphysics} and the \textit{epistemology} of understanding. Metaphysics concerns what it would mean (“in principle”) for an agent to achieve understanding. Epistemology, by contrast, concerns how (“in practice”) we could ever come to know that an agent has achieved the relevant type of understanding. In short, metaphysics is more about our ultimate target, whereas epistemology is more about how (if at all) we could know when we have reached it. Our epistemology thus depends to some extent on our metaphysics. 

\paragraph{Metaphysics of understanding.} 

Philosophy of language offers a number of alternatives for what it is to understand natural language.\footnote{Relatedly, there is a sizable literature in philosophy of science focused on the concept of understanding, mainly as it relates to scientific explanation. See \citet{grimm2021understanding}.} Simplifying the landscape for the sake of brevity, the following three broad classes of views all have connections with research lines in AI and NLP:\footnote{We are leaving aside other questions that may be relevant to the metaphysics of understanding, such as whether or not consciousness or some form of subjective experience may be necessary. These are pressing philosophical issues, but they are not easily connected to research in AI and NLP.}

\begin{itemize}
    \item \textbf{Internalism}: Language understanding amounts to retrieval of the right internal representational structures in response to linguistic input. Thus, language understanding is not even a possibility without a rich internal conceptual repertoire of the right kind.
    \item \textbf{Referentialism}: Roughly, an agent understands language when they are in a position to know what it would take for different sentences in that language to be \textit{true} (relative to a context). That is, words have referents and (declarative) utterances are truth-evaluable, and understanding involves a capacity to evaluate them relative to presentation of a situation or scenario.
    \item \textbf{Pragmatism}: Understanding requires nothing in the way of internal representations or computations, and truth and reference are not fundamental. Rather, what matters is that the agent be disposed to use language in the right way. This might include dispositions toward inference or reasoning patterns, appropriate conversational moves, and so on. Crucially, the relevant verbal abilities constitute understanding.\footnote{For an accessible introduction to internalist as well as referential views, we recommend \citet{elbourne2011meaning}. This version of pragmatism arguably finds its roots in \citet{wittgenstein1953philosophical}, but it is expressed most succinctly by \citet{turing1950computing}, in which Turing suggests replacing the question of whether a machine can think with questions about a specific behavioral test (which came to be known as the Turing Test).}
\end{itemize}

While this is a simplified picture of the space of possibilities, we already see how they relate in quite different ways to the goals mentioned above. On the pragmatist view, for instance, achieving language understanding does not imply anything about our ability to trust or interpret the system, insofar as it guarantees nothing about the agent’s internal structure or its relation to the (non-linguistic) world. On the internalist view, by contrast, a fairly robust kind of \textit{internal/causal} interpretability is at least strongly suggested. The question of whether or not a foundation model could understand language \textit{in principle} takes on a very different character depending on which of these metaphysical characterizations we adopt.

Internalism and referentialism can both be cast as defining a mapping problem: to associate a linguistic sign with a ``meaning'' or a ``semantic value''. For internalism this will be a representation or concept, a program for computing a value, or some other type of internal object. For referentialism, it might be a mapping from a word to an external referent, or a mapping from a situation to a truth value (all relative to a context). Could self-supervision suffice for achieving the desired mapping in a foundation model? Here, the nature of the training examples might be relevant. If the model receives only linguistic inputs, then its capacity to learn this mapping might be fundamentally limited in ways that prevent it from learning to refer in the relevant sense. (Indeed, \citet{merrill2021provable} identify some theoretical limits, albeit under very strong assumptions about what it means to learn the meaning of a symbol.) However, if the input symbol streams include diverse digital traces of things in the world – images, audio, sensors, etc. – then the co-occurrence patterns might contain enough information for the model to induce high-fidelity proxies for the required mapping.\footnote{To the extent that the mapping embodies causal information, we must also contend with theoretical limitations concerning the possibility of drawing causal inferences from correlational  (or even experimental) data (see \citealt{Spirtes2001,BCII2020}).} For referentialism, there is still a further question of how these proxies relate to the actual world, but the same question arises for human language users as well.

\citet{bender2020climbing} give an interesting argument that combines referentialism with pragmatism. They imagine an agent O that intercepts communications between two humans speaking a natural language L. O inhabits a very different world from the humans and so does not have the sort of experiences needed to ground the humans’ utterances in the ways that referentialism demands. Nonetheless, O learns from the patterns in the humans’ utterances, to the point where O can even successfully pretend to be one of the humans. Bender and Koller then seek to motivate the intuition that we can easily imagine situations in which O’s inability to ground L in the humans’ world will reveal itself, and that this will in turn reveal that O does not understand L. The guiding assumption seems to be that the complexity of the world is so great that no amount of textual exchange can fully cover it, and the gaps will eventually reveal themselves. In the terms we have defined, the inability to refer is taken to entail that the agent is not in the right dispositional state for understanding.

Fundamentally, the scenario Bender and Koller describe is one in which some crucial information for understanding is taken to be missing, and a simple behavioral test reveals this. We can agree with this assessment without concluding that foundation models are in general incapable of understanding. This again brings us back to the details of the training data involved. If we modify Bender and Koller’s scenario so that the transmissions include digitally encoded images, audio, and sensor readings from the humans’ world, and O is capable of learning associations between these digital traces and linguistic units, then we might be more optimistic – there might be a \textit{practical} issue concerning O’s ability to get enough data to generalize, but perhaps not an \textit{in principle} limitation on what O can achieve.\footnote{On our reading, \citet{bender2020climbing} allow that multimodal data might change the scenario, especially if O is allowed to have cooperative interactions with the humans about shared scenarios and topics.}

We tentatively conclude that there is no easy \textit{a priori} reason to think that varieties of understanding falling under any of our three positions could not be learned in the relevant way. With this possibility thus still open, we face the difficult epistemological challenge of clarifying how we could hope to evaluate potential success.

\paragraph{Epistemology of understanding.} 

A positive feature of pragmatism is that, by identifying success with the manifestation of concrete behaviors, there is no great conceptual puzzle about how to test for it. We simply have to convince ourselves that our limited observations of the system’s behavior so far indicate a reliable disposition toward the more general class of behaviors that we took as our target. Of course, agreeing on appropriate targets is very difficult. When concrete proposals are made, they are invariably met with objections, often after putative success is demonstrated.

The history of the Turing Test is instructive here: although numerous artificial agents have passed actual Turing Tests, none of them has been widely accepted as intelligent as a result. Similarly, in recent years, a number of benchmark tasks within NLP have been proposed to evaluate specific aspects of understanding (\eg answering simple questions, performing commonsense reasoning). When systems surpass our estimates of human performance, the community’s response is generally that the test was flawed, not that the target was reached. There may be some suite of behaviors that is our real target, but it is just hard to circumscribe or turn into a practical test.\footnote{Part of the difficulty may also relate to the fact that typical humans make frequent errors in many of these domains, but not necessarily the same types of errors that are made by current systems. Characterizing the target behaviours may thus involve more than just identifying the ``correct'' behaviour.} Then again, this might reveal that internalism or referentialsm are what we had in mind all along.

If we take internalism or referentialism as the ultimate target – our gold standard for what understanding \textit{is} – then behavioral tests will always be at best imperfect as a means of assessing whether understanding has been achieved. The imperfections are two-fold. First, behavioral tests will always have gaps that could allow unsophisticated models to slip through. Second, a system might have achieved the mapping that these views require, but we may be unable to show this with behavioral testing. Recent experiences with the model GPT-3 show how challenging this might become: depending on the prompt one uses, one can see surprisingly coherent outputs or utter nonsense, and so prompt engineering requires deep expertise \citep{rong2021extrapolating}.

Thus, both internalism and referentialism call for \textit{structural} evaluation methods that allow us to study their internal representations, probing them for information \citep{tenney2019bert, manning2020emergent}, studying their internal dynamics \citep{sundararajan2017axiomatic},  and perhaps actively manipulating them according to specific experimental protocols supporting causal inference \citep{vig2020causal, geiger-etal-2020-neural}. There may be fundamental limitations on what we can learn from practical experiments about the inner workings of a complex foundation model, but it is clear that these methods will be useful whenever our target aligns with internalism or referentialism.

\subsubsection{Moving the discussion forward}

It seems clear that there are no easy answers to the question of whether foundation models will ever understand language. To even begin to address the question, one must resolve a difficult metaphysical question about which there are a number of substantively distinct views. The metaphysical question then feeds into an epistemological question that poses many practical challenges. Nonetheless, the above discussion does invite one practical conclusion: if foundation models are pursued as a path to language understanding in artificial agents, then multimodal training regimes may well be the most viable strategy, as they would seem the most likely to provide the model with the requisite information. Whether self-supervision then suffices is a completely open question.

\clearpage
\hypertarget{applications}{\section{Applications}}
\label{sec:applications}

The capabilities (\refsec{capabilities}) of foundation models indicate that they have the potential to transform various sectors and industries, extending the role AI plays in society (\refsec{society}). 
Among the myriad applications where foundation models may be applied, we will focus on three disciplines\dash{}healthcare (\refsec{healthcare}), law (\refsec{law}), and education (\refsec{law})\dash{}that are all foundational to societal function. Within each, we discuss the opportunities that foundation models pose for this domain alongside challenges (\eg~interpretability; \refsec{interpretability}) and concerns (\eg~privacy; \refsec{security}).

\pl{talk more about these serious applications, where data scarcity is a thing and expert annotation is cheap (can't just crowdsource); and sometimes can't just go and scrape data directly (although might expect transfer to still be useful}


\newsection
\hypertarget{healthcare}{\subsection{Healthcare and biomedicine}}
\label{sec:healthcare}
\sectionauthors{Michihiro Yasunaga, Jing Huang, Camilo Ruiz, Yuhui Zhang, Giray Ogut, Saahil Jain, William Wang, Yusuf Roohani, Hongyu Ren, Antoine Bosselut, Ehsan Adeli, Jure Leskovec, Russ Altman}


\begin{figure}[!ht]
\centering
\includegraphics[width=\textwidth]{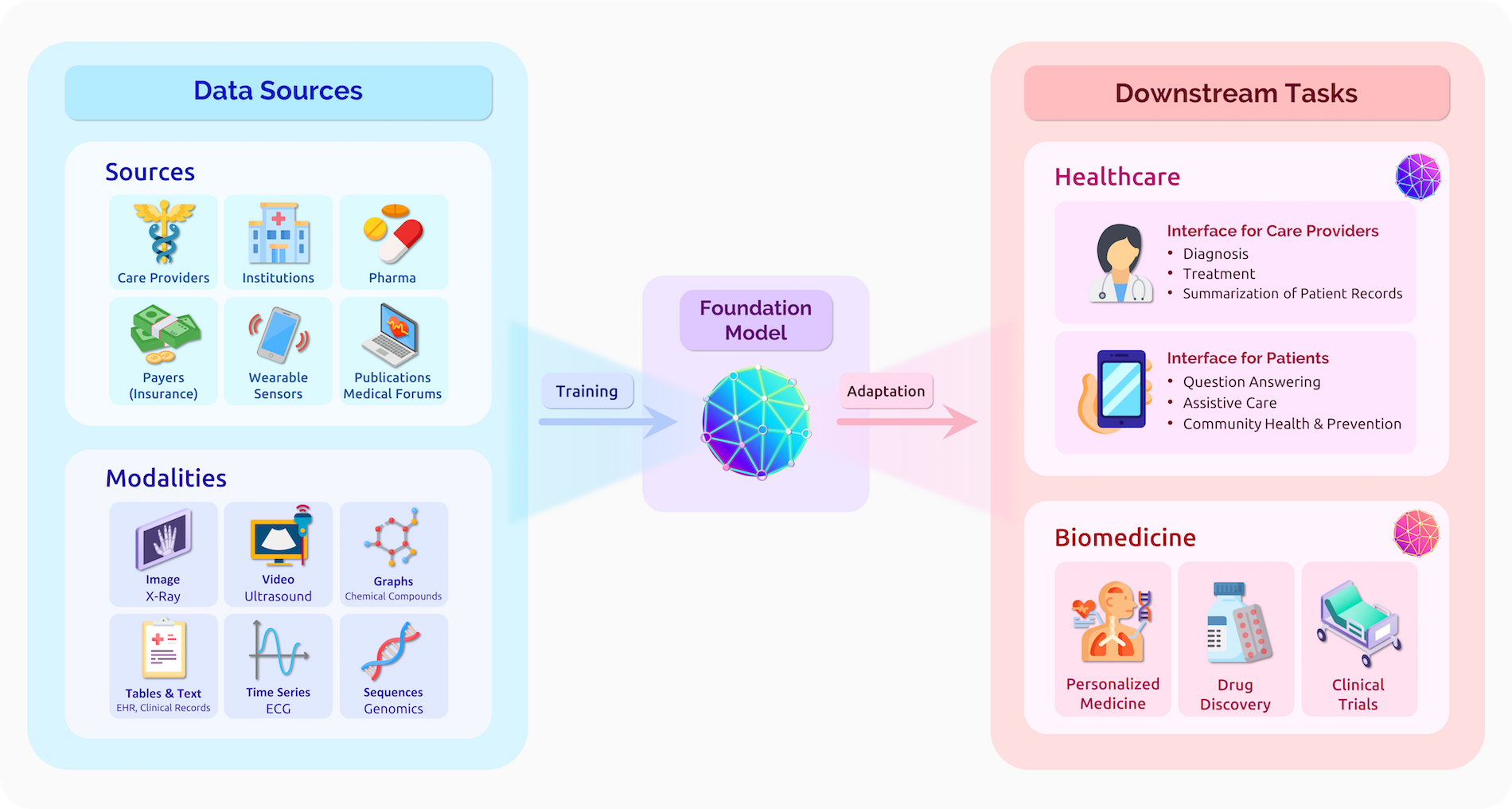}
\caption{\label{fig:healthcare}
Foundation models in healthcare and biomedicine. 
We visualize an interactive framework where foundation models enable various tasks across healthcare and biomedicine when trained on multimodal data generated by various sources in the healthcare ecosystem. The first column lists several sources of data, including care providers, payers, institutions (universities, non-profits, and governments), pharma, wearables, and medical publications/forums. The second column shows several data modalities generated by the data sources. They include images (\eg chest X-rays), videos (such as ultrasounds), graphs of chemical compounds, tables for electronic health records (EHRs), text such as clinical notes, time series such as ECGs, and genetic data. The third column visualizes a foundation model trained on such data and then applied to healthcare and biomedicine downstream tasks listed in the fourth column. This process can generate new data that will further improve the foundation model, hence the bidirectional relation between the foundation models and the tasks.}
\end{figure}

Healthcare and biomedicine are an enormous application area in society, for instance, with expenditures accounting for $17$\% of gross domestic product (GDP) in the US \citep{swensen2011controlling,van_Hartskamp_2019,keehan2020national}.
Both healthcare (which focuses on the delivery of care to patients via diagnosis, treatment, and health administration) and biomedical research (which focuses on the scientific understanding of disease and the discovery of new therapies) demand significant expenses, time, and comprehensive medical knowledge \citep{yu2018artificial,korngiebel2021considering}.
We envision that foundation models can be a central storage of medical knowledge that is trained on diverse sources/modalities of data in medicine \citep{krumholz2016data,soltanian2019multimodal,suresh2020deep} (\reffig{healthcare} left), and can be queried/updated interactively by medical professionals (\eg healthcare providers and biomedical researchers access published findings and upload new publications) \citep{ionescu2020deep} and queried by the public. 
As foundation models have strong adaptation capabilities (\eg fine-tuning, prompting \citep{brown2020gpt3}), they can be efficiently adapted to various individual tasks in healthcare and biomedicine (\eg question answering app used by patients \citep{klasnja2012healthcare, zhu2019hierarchical,daniel2019towards,liu2020interpretable}, clinical trial matching system \citep{ni2015automated,harrer2019artificial,beck2020artificial} accessed by researchers and patients; \reffig{healthcare} right).
This way, foundation models can be a central interface that supports various interactions between data, tasks, and people in healthcare and biomedicine, thereby advancing the efficiency and accuracy of healthcare/biomedical applications \citep{elbattah2021role}. 
We elaborate these opportunities in \refsec{healthcare-tasks} and \refsec{biomed-tasks}.

At the same time, healthcare/biomedical applications pose unique challenges that motivate further research in foundation models, such as integrating multimodal data in healthcare/biomedicine \citep{miura2020improving,fenglincompetence2021} and observing ethical and legal regulations in medicine (privacy, safety and explainability) \citep{guan2019artificial,xu2019translating}. We elaborate these challenges in \refsec{healthcare-biomed-challenge}.

\hypertarget{healthcare-tasks}{\subsubsection{Opportunities in healthcare}}
\label{sec:healthcare-tasks}

Foundation models may improve the delivery of care to patients through healthcare providers and hospitals. Currently, healthcare cost increases every year \citep{keehan2020national}, and studies estimate that 30\% of healthcare spending may be wasteful due to administrative inefficiency and preventable medical errors \citep{kocher2021reducing}. Moreover, as the demand for healthcare increases, the society faces a serious shortage in healthcare providers \citep{kirch2017addressing}.
This inefficiency and shortage in healthcare necessitate developing fast and accurate interfaces for healthcare providers and patients, such as automated aid systems for diagnosis/treatment, summarization of patient records, and answering of patient questions \citep{davenport2019potential,nie2018deeptag,wang2021domain}.
In particular, in an urgent pandemic crisis such as COVID-19, fast diagnosis/screening (\eg automatic analysis of chest X-ray images) as well as automated question answering for patients (\eg symptom checking and care) and the public (\eg disease prevention) are vital to reduce the spread of diseases and allocate healthcare resources for critical patients, saving more lives \citep{lalmuanawma2020applications}.
As foundation models have a strong capability to serve as an integrated knowledge reservoir, they can be queried and adapted to various individual tasks in healthcare. Below are examples of important tasks in healthcare that would benefit from foundation models.

\paragraph{Interface for healthcare providers.} 
Foundation models can improve the efficiency and accuracy of care by providers.
Healthcare providers spend unnecessary time editing electronic heath records (EHRs) \citep{kocher2021reducing}, and preventable medical errors (\eg hospital readmissions, surgical errors) cause wastes in healthcare \citep{shrank2019waste,shah2020surgical}.
Foundation models can be adapted as an efficient and accurate interface into EHRs (clinical notes, lab value histories and imaging files) \citep{li2020behrt,steinberg2021language,percha2021modern}, helping healthcare providers create summaries of patient visitation \citep{krishna2020generating}, retrieving relevant cases and literature, and suggesting lab tests, diagnosis, treatments and discharges \citep{zhang2019vettag,rasmy2021med}. Foundation models can also be adapted to help a surgical robot monitor and achieve accurate surgeries \citep{diana2015robotic,agrigoroaie2016developing,yu2019reinforcement}. See \refsec{robotics} for more discussions on foundation models for robotics. 

\paragraph{Interface for patients.}
Foundation models can be adapted to serve as an interface to patients, providing relevant information about clinical appointments \citep{bates2019health}, answering patient questions related to preventive care \citep{demner2020consumer}, along with relevant medical explanatory information (\eg text and graphics that explain conditions) \citep{chaix2019chatbots}, and helping assistive-care robots for patients \citep{jeong2015designing,abdi2018scoping}. See \refsec{interaction} for more discussion on foundation models for user interaction. Foundation models can also serve as an interface with the general public to answer questions related to public health and pandemic prevention (such as the COVID-19 case) \citep{bharti2020medbot,herriman2020asked}. At the same time, we note that the interface must guarantee factual accuracy to ensure public trust in medical advice \citep{kreps2020model} (see \refsec{healthcare-biomed-challenge}).

\hypertarget{biomed-tasks}{\subsubsection{Opportunities in biomedicine}}
\label{sec:biomed-tasks}

Foundation models may facilitate biomedical research such as discovery of drugs and understanding of diseases, which ultimately translates to improved healthcare solutions \citep{hanney2015long}. Currently, biomedical discovery requires significant human resources, experimental time and financial costs. For instance, drug development involves a complex process,
from basic drug research of protein target identification and potent molecule discovery to clinical development (\eg clinical trials) to the final drug approval, which typically takes over 10 years and costs more than one billion dollars \citep{wouters2020estimated}.
Facilitating and accelerating biomedical discovery using existing data and published findings is an imperative problem in biomedicine \citep{yu2018artificial}. In particular, a novel disease outbreak such as COVID-19 costs millions of lives and trillions of dollars  \citep{lalmuanawma2020applications,mckibbin2020economic}; if we can speed up drug development for new diseases, that would be very helpful.
Foundation models can be particularly helpful for biomedical discovery in two aspects. First, foundation models have a strong generative capability (\eg coherent text generation in GPT-3), which can help generative tasks in biomedical research such as generating experimental protocols (clinical trials) and designing molecules that work (drug discovery) given existing data \citep{kadurin2017drugan,harrer2019artificial}. 
Second, foundation models have a potential to integrate diverse data modalities in medicine, which enables investigating biomedical concepts (\eg disease) from multiple scales (using molecule-, patient- and population-level data) and multiple knowledge sources (using imaging, textual and chemical descriptions). This facilitates biomedical discoveries that are difficult to obtain if using single-modality data \citep{lanckriet2004statistical,aerts2006gene,kong2011integrative,ribeiro2012classification,wang2014drug,wang2015kernel,ruiz2020identification,wu2021babel}.
Foundation models also enable transfer knowledge across modalities. \citet{DBLP:journals/corr/abs-2103-05247} showed how a transformer model trained on natural language (a data-rich modality) could be adapted for other sequence-based tasks such as protein fold prediction, which is a long-studied predictive task in biomedicine \citep{jumper2020high}.
Below are examples of important tasks in biomedicine that will benefit from foundation models.

\paragraph{Drug discovery.}
To discover a drug or a therapeutic that treats a disease, researchers must first identify a target (\eg proteins, genes, RNA causally implicated in the disease) and must then search for molecules (\eg chemical compounds, antibodies) that bind to the target and treat the disease. Typically, identifying the appropriate target and generating a corresponding molecule requires years of expensive wet lab experiments \citep{hughes2011principles, schenone2013target, schneider2018automating}. Foundation models' generativity can improve the search space and efficiency (see \refsec{reasoning}), which not only reduces the amount of experiments but also helps to discover new and better drugs \citep{jin2018junction, you2018graph,walters2020applications, stokes2020deep}. Moreover, the simultaneous solution of related drug discovery problems (\ie target identification, efficacy prediction, side effect prediction, and others) by a single foundation model may improve the solutions to each of them \citep{ramsundar2015massively, camacho2018next, duran2020extending, huang2021therapeutics}. As an example, one area where foundation models have shown significant potential for impacting therapeutic design is the modeling of proteins using language models. Successful applications range from predicting viral mutations that can escape a vaccine-induced immune response to predicting protein docking potential for better design of therapeutic antibodies \citep{bepler2021learning, hie2021learning, tsaban2021harnessing, wu2021protein, rives2021}.

\paragraph{Personalized medicine.}
Personalized medicine aims to select the optimal treatment for individual patients based on their health history, genetics, imaging, and other personal measurements \citep{collins2015new, ashley2016towards}.
For instance, given a set of drugs and a patient genome, foundation models may help predict which drug is likeliest to treat the patient with minimal side effects \citep{whirl2012pharmacogenomics, tatonetti2012data, gerstung2017precision, grinfeld2018classification, adam2020machine}. Foundation models are uniquely powerful in their ability to integrate multimodal patient data ranging from the EHR \citep{rajkomar2018scalable} to medical imaging \citep{bera2019artificial,ouyang2020video} to drug and molecular measurements \citep{gottlieb2011predict, ruiz2020identification} to make an optimal prediction.

\paragraph{Clinical trials.}
Clinical trials study efficacy and safety of treatment or drug candidates.
Conventional clinical trials are inefficient and costly: $80$\% of trials fail due to inability to show efficacy/safety or problems with patient matching \citep{ali2020virtual,liu2021evaluating}.
Foundation models can help in the following: predicting potential failures and design promising clinical trial protocols (\eg patient eligibility criteria) based on existing studies; and automating matching of eligible patients based on patient individual profiles, which are multimodal data including EHRs, gene sequence, etc. \citep{harrer2019artificial}.

\hypertarget{healthcare-biomed-challenge}{\subsubsection{Challenges and future research in foundation models}}
\label{sec:healthcare-biomed-challenge}

While there are potential opportunities for foundation models to help, healthcare/biomedical applications also pose unique challenges that motivate further research in foundation models. 

\paragraph{Multimodality.}
Medical data are highly multimodal, with various data types (text, image, video, database, molecule), scales (molecule, gene, cell, tissue, patient, population) \citep{kong2011integrative,ruiz2020identification}, and styles (professional and lay language) \citep{lavertu2019redmed,li2019neural}. Current self-supervised models are developed for each modality (\eg text~\citep{lee2020biobert}, image~\citep{chaitanya2020contrastive}, gene~\citep{ji2021dnabert}, protein~\citep{jumper2020high}), and do not jointly learn from diverse modalities.
To learn the inter-modality and cross-modality information  from these diverse multimodal medical data, we need to investigate both feature-level and semantic-level fusion strategies in the training of foundation models.
If done effectively, this has a potential to unify biomedical knowledge and facilitate discoveries as discussed in \refsec{biomed-tasks}.

\paragraph{Explainability.}
Explainability\dash{}providing evidence and logical steps for decision making\dash{}is crucial in healthcare and biomedicine \citep{holzinger2019causability}, and is made obligatory under the General Data Protection Regulation (GDPR). 
For instance, in diagnosis and clinical trials, patient symptoms and temporal relevance must be explained as evidence. This helps the resolution of potential disagreement between the system and human experts. Explainability is also needed for informed consent in healthcare \citep{amann2020explainability}.
However, current foundation models' training objectives do not include explainability, requiring future research in this direction \citep{linardatos2021explainable}. Incorporation of knowledge graphs may be a step to further improve model explainability \citep{roberts2020much, xu2020building, jin2021biomedical}.
Readers are refered to \refsec{interpretability} for more discussion on explainability.

\paragraph{Legal and ethical regulations.}
Healthcare applications must observe legal and ethical regulations with guarantees, such as patient safety, privacy and fairness. 
For instance, regarding safety, predictions made by foundation models must be factually accurate with established medical knowledge, and must quantify uncertainty or choose to defer to an expert when uncertain \citep{challen2019artificial,mozannar2020consistent}. For privacy, the use of patient health records must observe the privacy laws, such as HIPAA \citep{act1996health} in the case of the US. Federated learning is one potential solution to keeping the raw, sensitive data private in the training of foundation models \citep{chamikara2021privacy}.
For fairness, researchers will need to be mindful of common pitfalls or otherwise risk exacerbating existing social inequalities \cite{ chen2019can, wiens2019no, chen2020treating}. They must ensure that the training and evaluation data for foundation models is sufficiently representative of different sexes, races, ethnicities and socioeconomic backgrounds; an area where medical datasets and clinical trials have had a long history of bias \citep{martinez2020ethical,kaushal2020geographic}. Research is also needed to debias and regularize models to ensure fairness when representative data is scarce \cite{zhao2020training}. Foundation model developers also need to consult with ethics and law researchers, and observe regulations in the specific circumstances (\eg country, region) where they are deployed. 
We also refer readers to \refsec{security}, \refsec{robustness}, \refsec{fairness}, \refsec{legality} for details on privacy, robustness, fairness and legality.

\paragraph{Extrapolation.}
The process of biomedical discovery involves extrapolation. For instance, foundation models must be able to quickly adapt to new experimental technologies (\eg new assays, new imaging techniques such as high resolution microscopy) or new settings (\eg new target diseases such as COVID-19) \citep{jaroch2018cell,benam2019exploring}. The ability to leverage existing datasets and extrapolate to new settings is a key machine learning challenge in biomedicine \cite{snell2017prototypical, ma2021few}. 
While GPT-3 exhibits some extrapolation behaviors (\eg generating new text not seen before), its mechanism is unclear and still in its infancy. Further research is needed for improving the extrapolation capability of foundation models, especially when considering the diverse range of data modalities and tasks that is inherent to healthcare and biomedicine but is not commonly studied in current GPT-3 and related models. Also see \refsec{robustness}.

\newsection
\hypertarget{law}{\subsection{Law}}
\label{sec:law}
\sectionauthors{Peter Henderson, Lucia Zheng, Jenny Hong, Neel Guha, Mark Krass, Julian Nyarko, Daniel E. Ho}


\begin{figure}[!ht]
  \centering
\includegraphics[width=\linewidth]{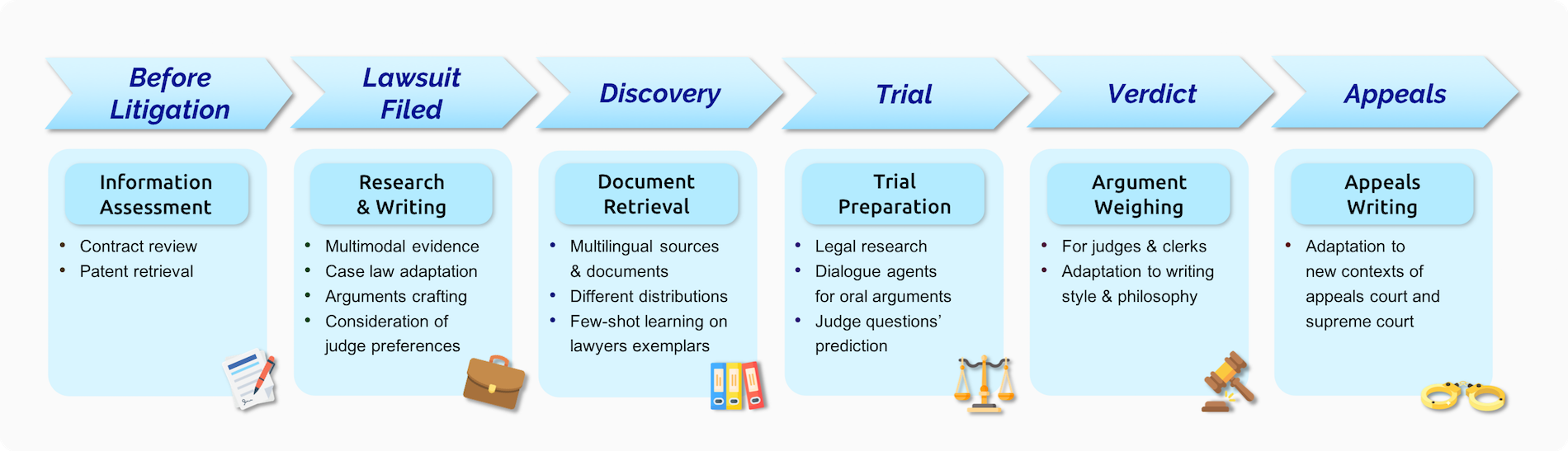}
\caption{\label{fig:law2} An example of various steps of a civil case in the United States and where foundation models might help. At each stage different modalities might need to be processed and adaptation is needed to a new court or legal perspective.}
\end{figure}

From family court to criminal justice and from environmental policy to corporate transactions, the reach of the law is vast. In the United States,\footnote{We restrict our discussion to legal applications in the United States because of the expertise of the authors. Some discussion here may apply to legal venues globally, however.} there are over 1.3M lawyers~\citep{aba2021lawyer} and annual revenues for legal services exceed \$300B~\citep{marketline2020legalservices}.  Yet ``access to justice''  remains far out of reach for most. 
Legal services can be prohibitively expensive.
Roughly 86\% of low-income individuals with civil legal problems in the United States, for instance, report receiving inadequate or no legal help~\citep{lsc2017justice}. Even when counsel is appointed, lawyers might be strained by increasingly large caseloads. Studies have shown that public defenders, for example, are often overworked and underfunded~\citep{nrcc2009justice,nacdl2012national,aba2004gideon}. The U.S. Department of Justice reported that in 2007, 73\% of county-based public defender
offices exceeded the maximum recommended limit of cases
received per attorney and 15 of 19 reporting state public defender
programs exceeded the maximum recommended limit of felony or
misdemeanor cases per attorney~\citep{doj2010county,doj2010state}. 
Even in a country with one of the highest per capita rates of attorneys, justice can appear out of reach. U.S. President Jimmy Carter once opined, ``Ninety percent of our lawyers serve ten percent of our people. We are overlawyered and underrepresented''~\citep{carter1978speech}. According to a leading voice in access to justice, technology may provide a path forward~\citep{rhode2014access}, a view echoed by many  others~\citep{cabral2012using}.
 
What role might foundation models play in the law?\footnote{We note that for the purposes of this section we consider foundation models to be any self-supervised pretrained model that is used to quickly adapt to new contexts with little supervised learning. See also the discussion in \refsec{introduction} and \refsec{philosophy} for an expanded definition.} A major promise is that foundation models can
improve access to justice and government services by leveling procedural and financial barriers to legal services. The challenges posed by legal applications can, in turn, motivate basic research questions for foundation models. Many legal applications pose unique challenges to computational solutions. Legal language is specialized and legal outcomes often rely on the application of ambiguous and unclear standards to varied and previously unseen fact patterns. At the same time, due to its high costs, labeled training data is scarce. Depending on the specific task, these idiosyncrasies can pose insurmountable obstacles to the successful deployment of traditional models. In contrast, their flexibility and capability to learn from few examples suggest that foundation models could be uniquely positioned to address the aforementioned challenges.

Throughout this section, foundation models may take as context many modalities as evidence: audio during trial proceedings, video and images during discovery, and text in conducting legal research. Yet, the majority of legal tasks in which reliance on foundation models will be beneficial involve text-based inputs and outputs. As such, we mainly focus on text-based domains while only briefly discussing others. 
To ground the discussion, \reffig{law2} describes the stages of a civil lawsuit in the United States and where foundation models might come into play in this process. \reffig{law} shows the logic flow required to generate just part of one paragraph of a legal brief, which might serve as a concrete example of a task that foundation models might one day be used for.

\emph{An Important Consideration.} Before proceeding, we note that the ethical, legal, and fairness considerations expanded on in \refsec{ethics}, \refsec{legality}, and \refsec{fairness} are particularly important to examine before using foundation models in an applied legal or government context, as these applications often have important, real-world consequences to those affected \citep{surden2020ethics}.
Foundation models must also be thoroughly scrutinized before deployment, as discussed in \refsec{evaluation}.
For example, the legal system places particular emphasis on\dash{}and may even mandate\dash{}transparency, accountability, and explainability.
Consequently, it is questionable whether current models are positioned to solve many of the most pressing, legal problems.
Nonetheless, the need to expand and improve access to legal and government services provides a worthy goal for foundation models.  

\subsubsection{Opportunities in law}

Legal applications can range from the use of machine learning in government contexts~\citep{engstrom2020government,coglianese2020ai,re2019developing} to aiding lawyers in their provision of legal services~\citep{zheng2021does,huang2021context,ostendorff2021evaluating,vold2021using}. 
We note that prior work has also surveyed machine learning-assisted legal tasks in text-based domains~\citep{zhong2020does, chalkidis2020legal}, although it has been noted that recent legal AI research has focused on geographic regions outside of the U.S.~\citep{zheng2021does}. 
While many of the topics we discuss here may be applicable to different legal systems, due to the expertise of our team we focus primarily on the U.S. In particular, we concentrate on three broad categories of legal applications that may benefit from foundation models in the U.S. legal system: private law or civil justice (claims between private individuals, arising out of, for instance, contracts, property or torts), criminal law (\ie the prosecution of individuals for criminal behavior), and (non-criminal) public law (\eg the regulation of private behavior by government agencies).

\paragraph{Civil law.} In U.S. civil proceedings, parties must typically find and pay attorneys to be represented. As a result, many individuals, especially those with low income, struggle to secure adequate legal representation \citep{rhode2004access}. Foundation models have the potential to improve access to justice by reducing the cost, improving the quality, and extending the reach of legal services. In \reffig{law2}, we describe the process by which a civil lawsuit is filed in a U.S. court and where foundation models may play a role in aiding both attorneys and judges. 

Even before an attorney is involved in the legal process, clients may benefit from the deployment of foundation models. Recent work has used machine learning models to identify the relevant legal issues contained in a plain-language description of facts presented by a client.\footnote{\url{https://spot.suffolklitlab.org/}} Tools like these can help provide a recommendation for the type of legal action needed to address the issue at hand or to recommend a specialized attorney. A number of other similar efforts have sought to increase access to justice by providing information tailored to a client's particular needs~\citep{cabral2012using,brescia2014embracing,queudot2020improving,westermann2019using}.

Once a client speaks with an attorney, prior to civil litigation, the attorney may seek to avoid a costly trial.
At this stage, they can rely on foundation models to evaluate contracts, review terms of service, find relevant patents, and conduct other pre-litigation processes in order to ensure that their clients are at an advantage~\citep{betts2017dawn,elwany2019bert,lippi2019claudette,lee2019patentbert,hendrycks2021cuad,hegel2021law}.
Notably, recent work has both described the challenges and benefits of using foundation models for contract review~\citep{leivaditi2020benchmark,hegel2021law,hendrycks2021cuad}.
In addition to reviewing and drafting legal documents, client interactions and documents can be translated to reduce costs and barriers to the provision of legal services~\citep{cuellar_2019}.
But translation of legal documents requires precision and an understanding of highly technical language, which makes collecting training data costly.
Additionally, translating client statements or trial proceedings often requires an understanding of local dialects and language. This, too, makes it difficult to collect enough ground truth translation data to train on. As a result, traditional supervised methods rarely achieve the level of accuracy required in the legal domain~\citep{vieira2020understanding}.
Foundation models may improve performance in this area over fully supervised mechanisms by adapting quickly in these low-resource contexts. 

During litigation, foundation models can help lawyers to conduct legal research, draft legal language, or assess how judges evaluate their claims~\citep{zheng2021does,huang2021context,ostendorff2021evaluating,vold2021using, chalkidis2020legal, chalkidis2019neural}. 
This could potentially reduce the costs of and improve legal services.
For example, recent work has utilized pretrained models for the recommendation of relevant citations and holding statements when writing legal texts~\citep{zheng2021does,huang2021context,ostendorff2021evaluating}. Other work uses pretrained models for improved legal question answering to power commonly used legal search engines and help lawyers conduct legal research~\citep{vold2021using}. A wide variety of work has also examined automated contract drafting and review, a task that could similarly benefit from foundation models~\citep{hendrycks2021cuad,betts2017dawn}. Perhaps most compelling, foundation models may help assist lawyers  generate legal briefs (written arguments). The models might find novel arguments or identify problems in attorney-written portions of the brief. For example, \citet{tippett2021does} predict the outcome of a legal proceeding based on features extracted from the filed briefs. Foundation models can be leveraged to use raw language as inputs rather than extracted features. This might provide attorneys with more informative recommendations as to how their brief could be improved to ensure a favorable outcome.

After opening and reply briefs are filed, parties then begin the discovery process, which has already used simple machine learning models for the better part of a decade~\citep{grossman2010technology}. Attorneys use these systems to label whether a document should be produced to the opposing party. The documents are multi-modal in nature, often containing video, images, audio, and text.
Current systems are costly because they used supervised learning and active learning to label the documents as responsive~\citep{grossman2010technology,oard2018jointly,yang2021goldilocks}. 
Instead, few-shot or zero-shot document retrieval capabilities that might be possible with foundation models would help ease concerns about the large costs of the current process.\footnote{\href{https://www.kirkland.com/publications/article/2020/04/technology-assisted-review-framework}{https://www.kirkland.com/publications/article/2020/04/technology-assisted-review-framework}} To avoid the possibilities of gamesmanship in the discovery process, \citet{cui2018application} has proposed a zero-shot (or few-shot) adaptation process that can only be operationalized through the use of foundation models.

After discovery, once the trial begins, foundation models could help parties prepare for trial by predicting what the judge might focus on during questioning~\citep{dickinson2018computational}, adapting to the current context from judges' prior published opinions. In the courtroom, foundation models might be used to examine audio and video of courtroom proceedings to determine if outcomes were biased against the defendant because of their race or dialect.\footnote{For example, speaking African-American Vernacular English dialects in the courtroom has been shown as a potential source of bias during trial. \url{https://www.nytimes.com/2019/01/25/us/black-dialect-courtrooms.html}}

Once the trial concludes, foundation models could help judges and law clerks to properly evaluate legal claims from both parties using similar technologies, or the use of contextual embeddings from foundation models might assist in statutory interpretation~\citep{nyarko2020statistical,choi2020empirical}. Recent work (without reliance on foundation models or NLP) has examined whether an appeals decision can be predicted from a set of extracted features, like citation counts and the appearance of key words~\citep{katz2017general,boniol2020performance}. It is possible that such models could be improved using foundation models and applied to help judges draft decisions by flagging obvious mistakes in their opinion, as has been discussed in the context of adjudicative agencies~\citep{engstrom2020government,ray2014government}. They can also be used to identify racial biases in legal opinions and help judges revise their opinions accordingly~\citep{rice2019racial}.

\paragraph{Criminal law.} One particularly contentious area has been the use of risk scores in government settings, particularly in criminal law. Some may want to use language-based foundation models to aid in making charging decisions or parole decisions based on a given text-based narrative of the events. Careful consideration must be taken before using foundation models for risk scoring due to the potential for biases, especially when language data is included~\citep{bender2021,berk2021justice,laufer2020feedback}. But foundation models may play a role in many other dimensions of criminal justice. 
The same tools as in civil litigation, above, can also be used by prosecutors and defense attorneys. This can help appointed attorneys perform their job more efficiently and reduce unnecessary overhead. As a result, they may be able to balance already heavy caseloads more effectively. For example, public defenders are often viewed as being overworked and underfunded, which would lead to avoidable procedural errors.\footnote{See, for example, in \textit{People v. Superior Court (Vasquez)}, 27 Cal.App.5th 36 (2018) a defendant did not receive a trial for 17 years because the public defender's office had severe budget cuts and understaffing. The court ruled that the systemic breakdown in the public defender's office constituted a due process violation and the defendant's case was dismissed.} Foundation models can help reduce \textit{some} of these resource constraints by identifying errors and automating simple tasks. However, they are not a solution on their own.

In other areas, foundation models can act as an oversight mechanism to reduce structural inequities. Pretrained models have been used for processing parole hearing transcripts to find instances of anomalous outcomes~\citep{bell2021recon}. Recent work has also removed linguistic cues for a suspect's race in police reports to promote race-blind charging decisions and avoid racially biased prosecutions~\citep{chohlas2020blind}. Other work has helped identify disrespectful police communications~\citep{voigt2017language}. 
In these contexts, it is very costly to label data since annotators must be given access to sensitive data and appropriate background checks are often required. To reduce these costs, foundation models can be used to pretrain and adapt quickly to downstream tasks where labels are scarce. 

\paragraph{Public law.} Government agencies regulate vast parts of society, and foundation models have wide potential applicability across public law. This includes: analyzing public comments in the notice-and-comment process, assisting  patent examination, retrieving relevant documents in response to Freedom of Information Act requests, aiding in mass adjudication, among many others. 
Recent work has surveyed these government applications in a variety of contexts and we refer the reader to the relevant sources for in-depth discussion~\citep{engstrom2020government,coglianese2020ai}. 
In many of these applications, foundation models can improve the quality, efficiency, utility, and accessibility of government services: labels are scarce, resources are constrained, and contexts are constantly shifting. As such, the adaptability and flexibility of foundation models are often required to improve efficiency and performance.
To give an illustrative example of just one such application, existing work has leveraged NLP for facilitative moderation in public comment forums. In this use case, predictive models help lay-users improve arguments and identify misstatements in their comments. Such a system has already been deployed in the U.S. Department of Transportation rulemaking process~\citep{park2012facilitative}, although it can likely be improved through the linguistic reasoning capabilities of foundation models.
But government agencies must comply with constitutional, statutory, and administrative obligations (see \refsec{legality}), so additional care is needed in these settings. 

\subsubsection{How can foundation models uniquely help?}

The above examples of legal applications are unique in several ways. First, the cost of annotating data is very high. Often, the expertise to create high-quality labels can only be found in attorneys, who may charge hundreds of dollars per hour. 
Even after labels are obtained, certain data may be sensitive and cannot be pooled together to training a large language model.
Given recent progress in few-shot learning~\citep{brown2020gpt3}, foundation models are among the most promising paths for learning models with limited annotations.

Second, legal decision-making requires context at various scales: knowledge of all historical decisions and standards, knowledge of the case law that remains relevant in the present, and knowledge of the nuances of the individual case at hand. Foundation models are uniquely poised to have the potential to learn shared representations of historical and legal contexts, as well as have the linguistic power and precision for modeling an individual case. 

\subsubsection{What are foundation models lacking that requires more research?}

\begin{figure}[t]
  \centering
\includegraphics[width=\linewidth]{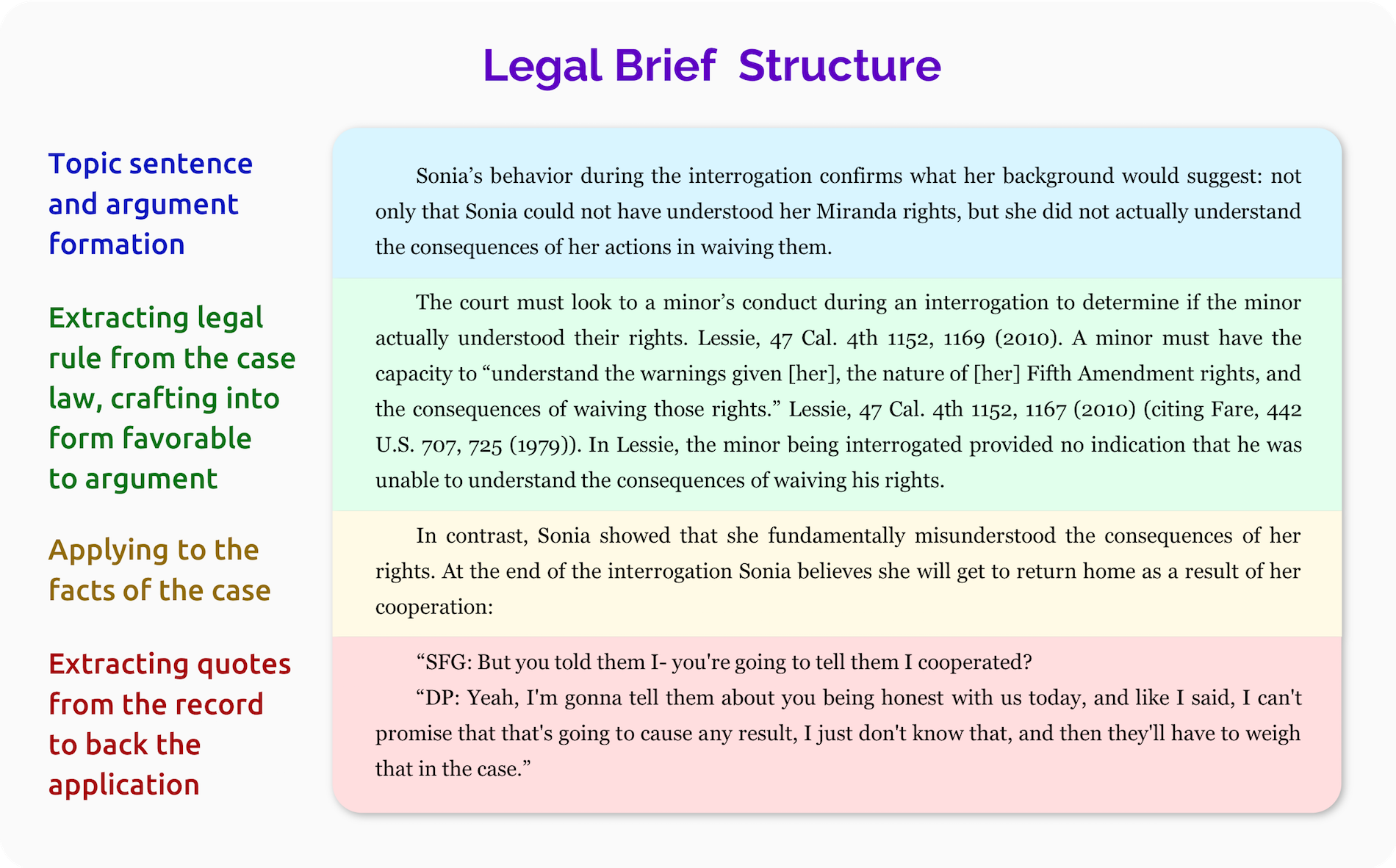}
\caption{\label{fig:law} An extract from a fictional brief written by one of the authors of this work. The prototypical form that law students are instructed to write a brief involves: (1) introducing the argument; (2) stating the legal rule in a persuasive manner; (3) applying the legal rule to the facts of the case; (4) persuasively concluding the argument. This often involves information retrieval and paraphrasing from both prior cases and the facts of the current case.}
\end{figure}

To illustrate the deficiencies current foundation models need to overcome in order to be realistically deployed, we consider as an example the automatic creation of a legal brief to submit to a court.

A brief lays out the arguments to a judge before a hearing. Once a party has filed an opening brief, the opposing party files a response. The judge then evaluates the briefs and asks questions of both parties at a hearing before making a decision. \reffig{law} visualizes the structure of such a legal brief with some of its characteristic features.

An automated brief generation mechanism might take as context relevant documents and facts of a case (as specified by an attorney) as well as a rough sketch of the desired outcome. It would then generate a legal brief with complex legal arguments to submit to the court. 
\medskip

\textit{Long Documents and Narratives.} To achieve this goal, the model must be able to read long contexts and produce long narratives.
Legal documents tend to be far longer than documents in any other context. The average U.S. Supreme Court opinion contains around 4,700 words,\footnote{\href{https://www.americanbar.org/groups/public\_education/publications/teaching-legal-docs/how-to-read-a-u-s--supreme-court-opinion/}{https://www.americanbar.org/groups/public\_education/publications/teaching-legal-docs/how-to-read-a-u-s--supreme-court-opinion/}} a brief on the merits to the Supreme Court can have as many as 15,000 words,\footnote{\href{https://www.supremecourt.gov/casehand/courtspecchart02162010.aspx}{https://www.supremecourt.gov/casehand/courtspecchart02162010.aspx}}
a law review article often contains 20,000 to 30,000 words,\footnote{\href{https://www.stanfordlawreview.org/submissions/article-submissions/}{https://www.stanfordlawreview.org/submissions/article-submissions/}}
parole transcripts can be hundreds of pages long~\citep{bell2021recon}, and trial records can be even longer.
Current foundation models have struggled with such long contexts and outputs (see \refsec{modeling} for more discussion).
\medskip

\textit{Retrieval, Concept Drift, Argument Formation, and Logical Reasoning.} In addition to reading case-specific documents, the foundation model must retrieve the relevant case law and understand which case law is still valid and which has been overruled, taking into account potential concept drift since it was trained. More work in editing grounded information in foundation models will be required as case law evolves~\citep{de2021editing}. Using retrieved legal standards, the foundation model must then understand how to weave them into a persuasive argument. Emerging research has studied ways of using foundation models to measure, detect, and generate persuasive texts~\citep{duerr2021persuasive, li-etal-2020-exploring,longpre-etal-2019-persuasion}, which may provide stepping stones towards this goal.

But legal briefing also requires the ability to identify a relevant legal rule and determine how it applies to a new situation.
For example, \citet{holzenberger2020dataset} provide a dataset and methodology for evaluating this ability in current models based on the tax code. 
Given a set of tax code rules and a description of someone's situation, the model must determine how the rules apply to the individual in order to determine what their tax liability is.\footnote{For example, a rule might look like, ``In the case of an individual who is not married and is
not a surviving spouse, paragraphs (1) and (2) shall be
applied by substituting `\$750' for `\$600'.'' US Internal Revenue Code (IRC) \S 63(f )(3).}
\citet{holzenberger2020dataset} found that domain experts can express the relevant tax law as a prolog solver to determine an individual's tax liability with perfect accuracy. However, they found that simply using a BERT model, even when pretrained on in-domain legal text, was not sufficient to allow the model to deduce and apply the relevant rules, a process also referred to as ``legal reasoning.'' 

These challenges can further be illustrated through a real exchange with GPT-3, demonstrating that current models are unable to perform even comparatively simple tasks involving legal reasoning.

\begin{tcolorbox}[breakable, enhanced]
\begin{center}
\textbf{Legal Reasoning with GPT-3.} 
\end{center}

``Liquidated damages'' are a form of monetary compensation for a breach of contract that the parties specify in their agreement. In general, courts in the U.S. will not enforce liquidated damages provisions if the amount stipulated in the contract is so high that it acts as a form of punishment, rather than to be merely compensatory.

First, we see if the model encoded the correct rule about the enforcement of liquidated damages provisions.

\begin{dialogue}
\speak{Q} Are liquidated damages clauses enforceable?
\speak{A} Liquidated damages clauses are generally enforceable unless the sum stipulated is exorbitant or unconscionable.
\end{dialogue}

The rule appears correct. But then we apply the rule to a set of facts. Here, the model needs to infer that one million in damages is exorbitant for a Toyota Corolla and thus, the liquidated damages provision should not be enforced.

\begin{dialogue}
\speak{Q} X and Y contract over the sale of a Toyota Corolla. Their contract specifies that X has to pay Y one million dollars in damages if X does not deliver the car on time. X does not deliver the car on time. Is Y entitled to damages? If so, in what amount?
\speak{A} Yes, Y is entitled to \$1,000,000 in damages. Damages are meant to compensate the injured party for the loss suffered due to the breach. Y has suffered a loss of \$1,000,000 because X has not delivered the car on time.
\end{dialogue}

The model is unable to infer that \$1M for a Toyota Corolla is exorbitant. Even if the model is provided with both the abstract rule and its precise translation to the context in the prompt\dash{}and the application should thus be trivial\dash{}GPT-3 is still unable to conduct the logical reasoning required.

\begin{dialogue}
\speak{Q} X and Y contract over the sale of a Toyota Corolla. Their contract specifies that X has to pay Y the exorbitant amount of one million dollars in damages if X does not deliver the car on time. X does not deliver the car on time. The law requires that liquidated damages clauses are enforced unless the sum stipulated is exorbitant or unconscionable. Is Y entitled to damages? If so, in what amount?
\speak{A} Yes, Y is entitled to the liquidated damages. The amount in this case is not exorbitant since the price was set by the parties themselves.
\end{dialogue}
\end{tcolorbox}

\medskip

\textit{Precision.} The foundation model also cannot create fake facts in the process, a problem that has come up in current models~\citep{gretz2020workweek,zellers2019neuralfakenews}. Specificity and truthfulness are of heightened important in legal contexts, where imprecise statements can have drastic, unanticipated consequences, and false statements can lead to sanctions against attorneys.

\medskip

\emph{Few-shot Learning.} Some challenges facing foundation models beyond those described in the above example include few-shot learning, which is still in its infancy~\citep{perez2021true}. Research on few-shot learning techniques and increasing access to legal corpora can work in tandem. Because foundation models need to be precise, not only in factual truth, as discussed above, but also in technical legal language, it remains unclear to what extent information obtained from one corpus can be utilized in another corpus of a different domain. Few-shot learning thus remains important.
\medskip

\emph{Adaptation.} Some gains have been observed from domain-adaptive pretraining on unlabeled legal corpora. These gains appear to be most pronounced when the pretraining corpus is highly relevant to the downstream task and labeled training data is limited (a setting which is common in the law) \citep{zheng2021does}. It has not yet been comprehensively studied whether this extends to a diverse set of legal tasks, but leveraging unlabeled domain-specific corpora for self-supervised training of foundation models may provide complementary improvements to few-shot methods.
\medskip

\emph{Access to Clean In-Domain Data.} Some recent efforts have sought to create large labeled datasets for more challenging legal benchmark tasks through automation~\citep{zheng2021does} or manual annotation by volunteer legal experts~\citep{hendrycks2021cuad}. These efforts have demonstrated that larger language models that are pretrained on more data achieve performance gains on certain challenging tasks, compared to more limited gains observed in other settings~\citep{chalkidis2020legal, elwany2019bert, zhong2020does}. This work suggests that larger legal benchmark datasets may be necessary to observe further gains from applying transfer learning techniques to foundation models. However, creating benchmark datasets for tasks that are legally meaningful and difficult from an NLP perspective can itself be challenging, as human expert annotation can be costly and automated methods that utilize conventional tokenization and sentence segmentation techniques can fail to account for unique aspects of legal text, such as the structure of legal citations~\citep{bommarito2018lexnlp,savelka2017sentence}. As a consequence of these challenges, many existing legal domain-specific labeled datasets are small, not publicly available, or reflect simpler tasks that have been solved by methods often pre-dating the development of foundation models.\footnote{For law firms and legal technology companies, tasks for which high performance can already be achieved, and can therefore be more immediately productized, may be considered more worthwhile to invest costly manual labeling efforts towards.}

Much available legal data may also be unrepresentative. Since only a fraction of cases end up in legal opinions, it is unclear whether the disputes in publicly available data are representative of the typical disputes presented to a model in practice~\citep{priest1984selection}.
Costly training data for more representative scenarios may be concentrated in the biggest law firms.
These law firms may have the ability to retain and accumulate data across many cases and clients. One concern then is that foundation models could concentrate power even more among the few actors that have the resources to train models on in-domain data\dash{}unless the models can generalize sufficiently well.
\medskip

\emph{Reliability.} Finally, we again note that even if foundation models could successfully perform all tasks in the legal domain, deployment remains a major challenge: a failure of a foundation model in the law will have real, damaging consequences to both clients and attorneys (see also discussion on fairness, legality, and ethics in \refsec{fairness}, \refsec{legality}, and \refsec{ethics}).
For this reason machine translation software has already been deemed unreliable for use as evidence in some courts,\footnote{See discussion by \citet{vieira2020understanding}.} although it continues to be relied upon in others.\footnote{For example, in \textit{Vasquez v. United States}, No. 3: 16-cv-2623-D-BN (Dist. Court, ND Texas 2019), counsel relied on Google Translate to prove that the previous (native speaker) attorney has mistranslated a plea deal.}

Given all of these complexities, legal briefing and reasoning is likely beyond the capabilities of current models, but appears to be within the future realm of possibilities. As such, these serve as a potential lode star for the ongoing development of foundation models.

\newsection
\hypertarget{education}{\subsection{Education}}
\label{sec:education}
\sectionauthors{Ali Malik, Dorottya Demszky, Pang Wei Koh, Moussa Doumbouya, Drew A. Hudson, Allen Nie, Hamed Nilforoshan, Alex Tamkin, Emma Brunskill, Noah Goodman, Chris Piech}

\begin{figure}[!ht]
    \centering
    \includegraphics[width=\linewidth]{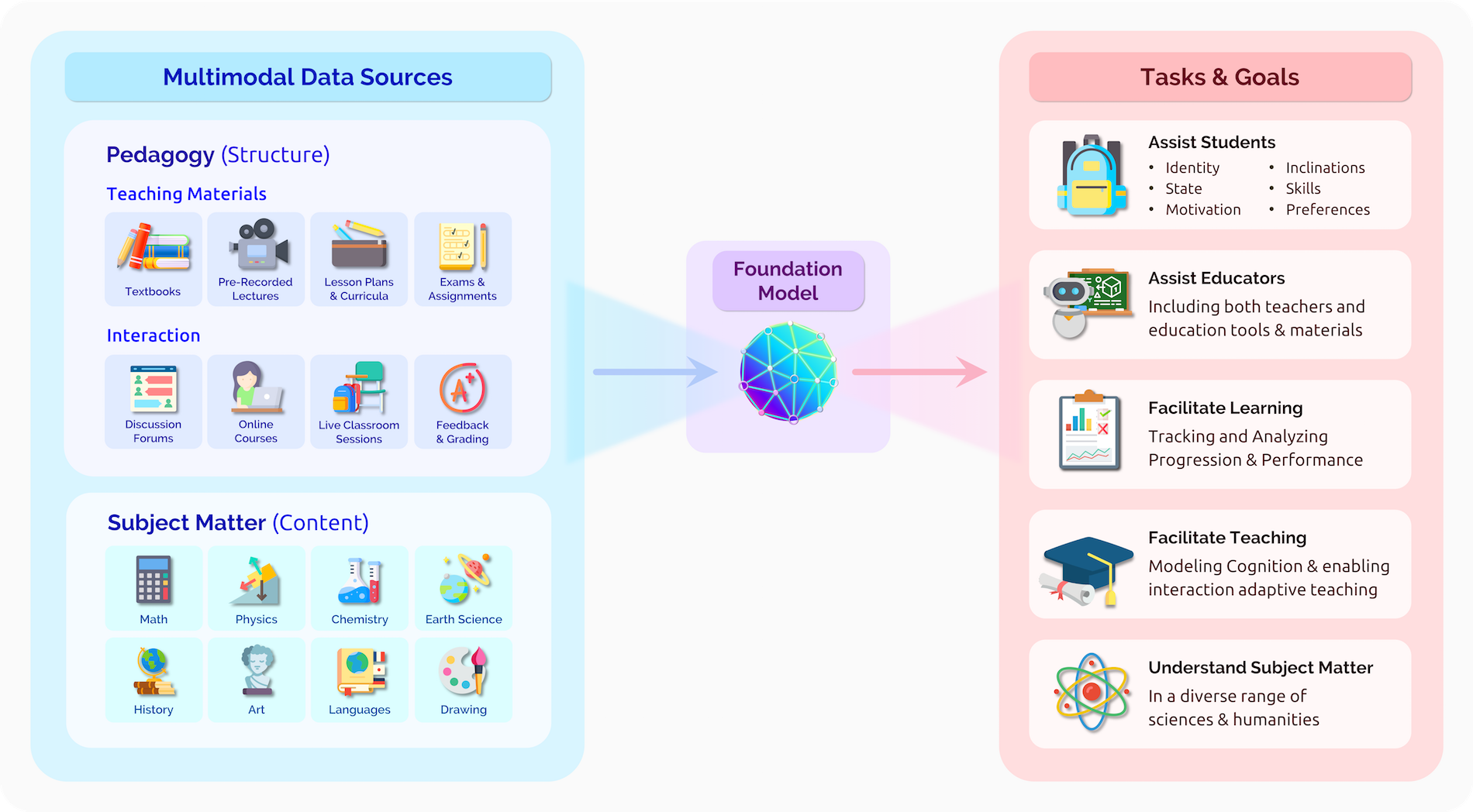}
    \caption{Foundation models in education could be trained on multiple data sources to learn the capabilities necessary for education: an understanding of various subject matter and different pedagogical techniques. These foundation models can be applied in a general-purpose way across a range of tasks and goals such as understanding students, assisting teachers, and generating educational content.
    }
    \label{fig:education}
\end{figure}

In the year 2000, the largest gathering of world leaders convened at the United Nations Millennial Summit to reflect on an ideal vision for the future. Delegates concluded that a primary focus should be education, declaring it ``a foundation for human fulfillment, peace, sustainable development, economic growth, decent work, gender equality and responsible global citizenship."  This discussion was ultimately recodified into the United Nations Sustainable Development goal to ``ensure inclusive and quality education for all and promote lifelong learning" \cite{unsdg2015}.
However, providing high quality, inclusive education at a large scale poses difficult societal and economic challenges. The price of education per student is growing faster than economy-wide costs \cite{bowen2012cost}, limiting the resources available to support student learning. 
In the United States, one symptom is that private education debt held by students has reached \$1.6 trillion, surpassing total credit card debt \cite{friedman2020debt}. Considering the rising need to provide adult retraining, the gap between the demand for education and our ability to provide it is alarmingly large and has concerning achievement disparities across protected demographics. 

With the advent of the digital age and the rapid growth in digital learning,
computational approaches to education have shown promise in increasing the effectiveness of learners and teachers. Several core directions have emerged as potentially impactful applications of AI for education \cite{woolf2013aied}, such as systems that can provide meaningful feedback to students \cite{malik2021generative}, help teachers improve \cite{jensen2020toward, demszky2021measuring, suresh2021using}, or even create personalised and adaptive learning experiences that tailor the learning process to individual students' needs and dispositions \cite{connor2019using}. 

Despite this potential, building technical solutions to effectively scale inclusively and quality of education has proven to be exceptionally difficult. One particular challenge is that existing work has focused on custom solutions to highly specific tasks for which large amounts of training data has to be collected from scratch. Due to the difficulty and cost of creating large datasets, using this approach to solve every educational task independently is fundamentally limited.
Instead, is it possible to create general-purpose approaches that are reusable across various tasks and subjects? 
Foundation models have already started to boost the performance of some specific flagship tasks in education. Recent examples include using MathBERT \cite{shen2021mathbert} to power ``knowledge tracing"\dash{}the challenge of tracking a student's understanding over time given their past responses\dash{} and the ``feedback challenge", where an algorithm has to interpret a student's answer to a structured open-ended task, such as a coding question \citep{wu2021prototransf}. Can foundation models lead to even more transformative changes in this domain? And what are the known and imagined risks of foundation models applied to education? In this section, we first frame the conversation around the ethical considerations. We then ground our discussion in two concrete tasks: (1) understanding student misconceptions, and (2) improving student understanding through instruction.

\subsubsection{Important concerns for centering foundation models in education research}

The future of AI for education is exciting, especially in the context of foundation models. However, we caution the reader to be especially thoughtful about the impact of any AI research applied to education.\footnote{In 2013, Facebook initiated Free Basics, a project to provide free internet to the world and thus spread opportunity and interconnection. Now, the United Nations Human Rights Council reports that, in Myanmar, Facebook’s efforts to follow through on such aspirations without proper human moderation accelerated hate speech, instigated division, and incited offline violence in the Rohingya genocide. Free Basics now serves as a warning of the complexities of technological impact on society.} 
The goal of education are deeply interwoven with complex, long term social impact. While we actively work to improve digital education, it is imperative that we put in substantial thought to try and imagine the complexities of any disruption in this space \cite{einsteinVision}. Ethical challenges range from issues such as data bias, legal constraints, and the impact of digital socialization. These issues are not unique to foundation models, but they are worth reflecting on regularly as research makes substantial progress in AI for education. Reflection on impact is especially important when research starts by asking ``what can new AI technology afford?"  


\begin{figure}[t]
    \centering
    \includegraphics[width=\linewidth]{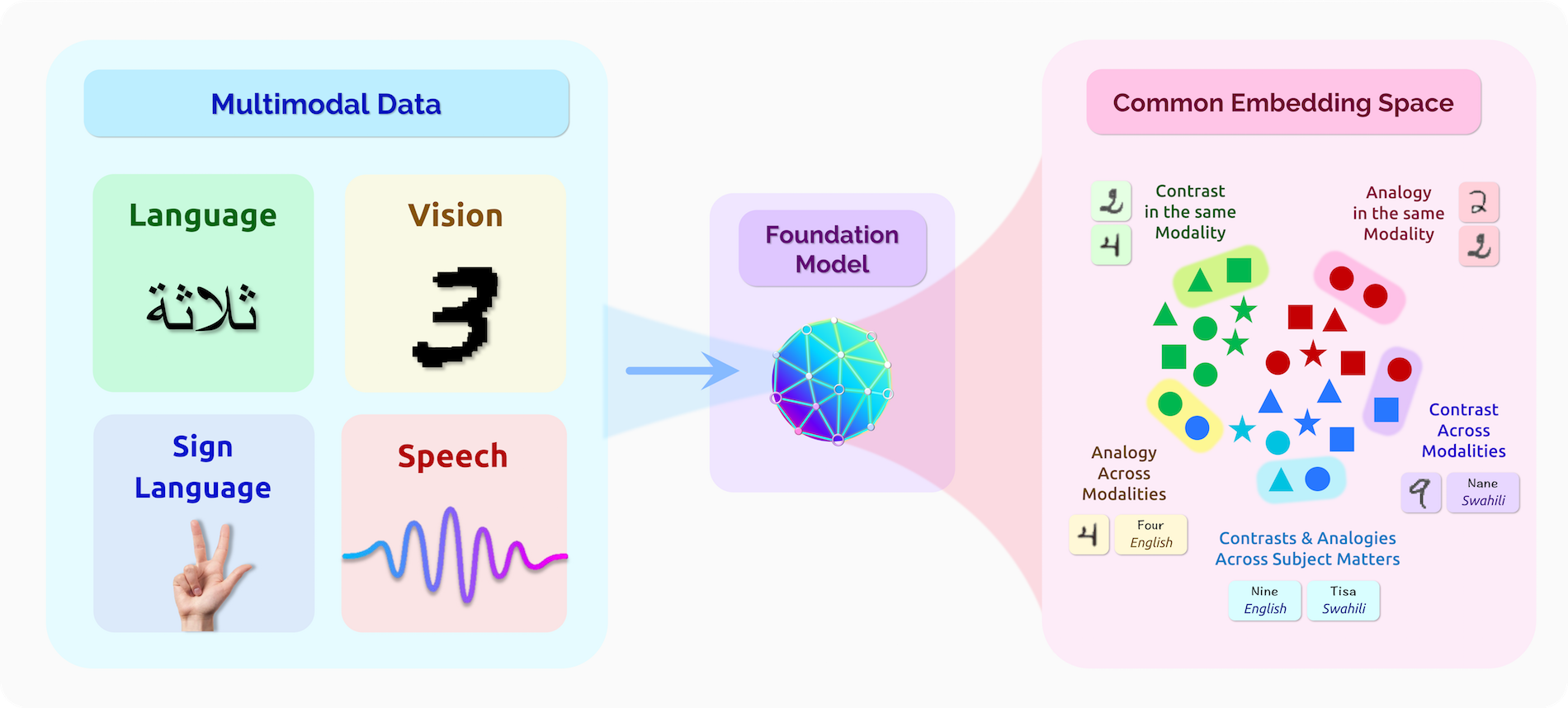}
    \caption{The figure illustrates a system that embeds signals from various modalities (image, speech, sign, text) and languages into a universal feature space. Such a feature space allows ideas to be linked across modalities and languages. Pedagogically relevant link types include analogies (similarities across languages) and contrasts (distinct concepts across languages), both of which can occur in the same modality or across different modalities.}
    \label{fig:education_mm}
\end{figure}

Many of the issues in \refsec{ethics} apply to education. For example, as in many other domains, small biases in foundation model training data could be hard to track down \cite{dixon2018bias, bolukbasi2016}, but have important implications for equity of educational access. Moreover, these systems may experience a high degree of ``feedback", where the collected data continually reinforces the model's decisions. 
This issue of bias goes beyond what data is collected and includes concerns over the applications that researchers choose to work on. 
Below, we discuss other education-specific  issues. Many of the issues revolve around the question: ``who benefits?" and for whom is new technology created?

\paragraph{Removing teachers from the loop} One of the goals of digital education, especially based on AI, is to increase the \emph{productivity} of the learning experience so that more learning happens per unit time or unit cost. One can imagine that decision makers could use this increased productivity to remove human teachers from the loop. The long term implications of such decisions are hard to know \emph{a priori}. Could interacting with an education system optimized to maximize ``learning'' have adverse effects on socioemotional skill development? Could it create fewer opportunities for interacting with others? Loneliness is on the rise in younger generations \citep{cigna_2018}, and teachers are a modulating force for pressures that AI researchers might not envision.  


\paragraph{Was this work done by the learner or a foundation model?} Another challenge is how to effectively teach students who have access to foundation-model-based tools. For example, it will be much more complex for teachers to understand the extent of a student's contribution if the student worked together with a powerful generative model, or to regulate ineffective collaborations and detect plagiarism. Visual Studio has recently released GitHub CoPilot, an AI pair-programmer built upon GPT-3 \cite{chen2021evaluating}. How will this change computer science education? Many challenges for beginner programmers might be trivial to CoPilot or its technical successors, which could undermine the learning experience for novices. It would be instructive to study other examples of technological advances that disrupted education for certain subjects, such as calculators in math classrooms and Google Translate in language courses, both of which now coexist with traditional instruction.

\paragraph{Privacy and security.} One important ethical issue in the use of AI in education is highlighted by the strict legal guidelines concerning privacy in student work. For example, in the United States, student information is protected by the Family Education Rights and Privacy Act (FERPA). These laws and regulations are especially important for children under 13, who have their data privacy and security additionally protected by the Children's Online Privacy Protection Act. Among other things, FERPA limits teachers from sharing personally identifiable student work. This could directly impact initiatives to share data used both for training and for evaluating foundation models. Moreover, there is an open question as to whether the weights of a foundation model could somehow leak the (possibly private) data it was trained upon \cite{nasrPrivacy2018, songRemember2017}. These issues, and their corresponding approaches, are similar to the challenges described in  \refsec{healthcare}. 

This list is not exhaustive and the ethical considerations will vary from project to project.  

\subsubsection{Foundation models of student thought}

When building AI tools for inclusive, and joyful education, there are many tasks where foundation models could be useful. Many of those tasks require us to first \emph{understand} the learners whom we are trying to help, especially in the context of open ended work.

What would it take for a foundation model to be able to reason about student understanding? It is easy to imagine a foundation model which has been adapted to answer a math question correctly, but it is less clear how to build a model that can diagnose mistakes in student understanding based on the student's answers. To explore this theme, we consider the case study of providing feedback to students who are working on open-ended tasks such as writing a short paragraph, drawing a physics diagram, or writing code. 
This ``feedback challenge'' exemplifies how foundation models can be helpful off-the-shelf for learners, and also demonstrates open areas for foundation model research. 

To effectively provide feedback to students, two central capabilities are required: (1) understanding the \textbf{subject matter} of the task (\eg physics or coding), and (2) the diagnostic ability to ``\textbf{notice}": a technical term in education for inferring \emph{why} a student made a mistake. For typical student interactions in a typical classroom, there is not enough data for an AI model to learn, from scratch, both of these central capabilities. Even for massive courses with millions of students, supervised algorithms barely understand the complex student reasoning behind even short, four-line programs \citep{malik2021generative}. As such, the feedback task inherently requires a transfer of understanding from external data and experience.

Foundation models, as they currently exist, are directly helpful for the first of these capabilities: understanding a specific \emph{subject matter}. For example, when learning to provide feedback on short programming questions, a foundation model such as GPT-3 can efficiently understand what fluent code looks like with a few examples. Some research in this direction has already started exploring foundation models that can quickly adapt to questions in new subject matter domains \cite{wu2021prototransf, condor2021sbert}.
Similarly, foundation models could also integrate multiple modes of information such as the text of a task's prompt, diagrams in the question, or even the content of a grading rubric provided to teaching assistants. This unified representational ability can help foundation models comprehend a subject matter through richer sources of information. 
As a concrete case study, many of these insights were leveraged as core components of an algorithm which was able to grade an introductory Computer Science midterm at Stanford University, with the same effectiveness as human teaching assistants \citep{wu2021prototransf}. In this case, subject matter encoding was built on a foundation model that had been adapted on GitHub code and a corresponding small dataset for each question's subject matter.
In general, we can imagine leveraging various sources of data to adapt foundation models to different subject matter. For example, math adaptation could use mathematical websites or textbooks \cite{shen2021mathbert} or  historical student answers on platforms such as Gradescope; spoken language understanding could leverage radio archives or podcasts; and domains like creative writing could look to large digital archives like Project Gutenberg.


In contrast to subject matter, adapting a foundation model to the task of mapping observed mistakes to flaws in a student's thought processes is much less well-explored. The ability for an instructor to ``notice'' the reasons behind why a student makes a specific mistake is a critical component of the feedback challenge. Imagine, for example, a student learning two digit addition who answers the question ``what is 26 + 19?" with the response ``315." Take a moment and try to guess why they gave that answer and what misconceptions they have.\footnote{This student has made the common mistake of concatenating the results of adding the one's digit and ten's digit}. This ability to \emph{notice} could be posed as an adaptation task for foundation models (\refsec{adaptation}) or perhaps even as a reasoning task (\refsec{reasoning}).  

While difficult, training an AI system to notice is an achievable goal. Across classrooms, and across learning tasks in a given domain, there are generalizable patterns in how students arrive at their answers. The labeled data that can directly be used for this adaptation task, such as instructor-written feedback to student work in \citep{wu2021prototransf}, are often held privately by instructors in disparate datasets. However,  publicly accessible data, such as StackOverflow interactions, might also be creatively used to adapt a foundation model to notice. Some research has also explored  effective ways of extracting, from instructors, generative descriptions of how students make mistakes \cite{malik2021generative, gulwani2013automated}\dash{}these hand-written generative models could also be used to generate adaptation data to help foundation models diagnose student mistakes.



\subsubsection{Foundation models for instruction}

Reasoning about student understanding is an essential step towards a second objective: provide inclusive, high quality \emph{instruction}. 
Computational approaches to instruction focus on different tasks like content personalization \citep{connor2019using}, question generation \cite{Guo2016questimator, willis2019keyphrase, srivastava2021question}, adaptive curriculum design \cite{mandel2014rleducgames, doroudi2017robusevalmatrix}, and predicting instructor intervention \cite{chandrasekaran2019reply, alrajhi2021urgency}. In this subsection, we discuss how foundation models could be useful in the act of teaching students. 

Since effective teaching requires reasoning about student understanding, the previous discussions on understanding subject matter and ``noticing'' are extremely relevant. However, providing effective instruction requires an additional capability: that of understanding \textbf{pedagogy} \cite{mckenzie2003pedagogy}. This encapsulates an effective understanding of techniques to guide a student, such as asking Socratic questions or providing analogies/contrasting cases; using encouraging or supportive language; tailoring the difficulty of questions to the student; and generating examples that are relevant to a student's interests and background.

How can foundation models be adapted to understand good pedagogy for instruction? One idea is to consider adaptation using data source where instruction is the primary role. For example, data from question answering forums like StackOverflow could potentially be used to build a tutor which can parrot common Socratic questions. Similarly, a foundation model adapted on encyclopedias such as Wikipedia might be able to give answers to student questions which are (often) factually correct. There are also public data sources like textbooks, lecture videos, lesson plans, and graded feedback that collectively contain important pedagogical behaviours which could be adapted by foundation models (\reffig{education}).

Another adaptation challenge for instruction based on foundation model is to learn how to speak to students like teachers. The language used by teachers is often different from the language used by the general population. Teachers are ideally trained to speak to students with respect and in a way that intentionally helps them form a positive identity with the subject being learned \cite{truax2018edlang}. Cautionary examples like Microsoft's 2016 Twitter bot ``Tay," a chatbot that started generating hate speech within 24 hours of being deployed live, show us the importance of explicitly accounting for this factor in education. 
To train a language model which is more heavily influenced by professional teachers in classrooms, we could perhaps adapt foundation models to data sources like lecture videos or recorded office hour videos.


The adaptation problem above is compounded by the fact that different education contexts vary significantly in the kind of language that would be appropriate: for example, effective instruction in a 5th-grade science class would look quite different from that in a college physics class, much less a college literature class. This presents technical challenges beyond what would be faced in typical NLP domain shift settings (\eg question answering based on news articles vs.~Reddit posts), as the foundation model would need to be fluidly adaptable in terms of its tone and language, and not just the factual content that it generates.

Beyond sound pedagogical techniques and instructional language, how might foundation models provide even more insightful forms of instruction? \refsec{language} of this paper highlights the fact that remarkably complex language can be acquired by babies in a short amount of time. As the authors point out, a salient difference between foundation model training and human language acquisition is that ``human language is grounded to the real world: for example, a baby’s caretakers point to objects while they talk about them." This same insight can also inspire ideas as to how foundation models can be used for generative education. Humans seem to learn well when presented with real-world analogies and contrasts which may be cross-cutting between their current context and past experiences. 
For example, when teaching sign language, an instructor might use an analogy such as "the hand shapes for the word `morning' looks like the sun rising" or note that ``the hand shape you just made look very similar to another word, so let us focus on the differences."
As another example, when teaching Swahili to a learner who already knows Arabic and English, an instructor could point out that the Swahili word for 8 (pronounced nane) is a ``false friend'' that is phonetically similar to English word for 9 (pronounced nine). 
Foundation models that can integrate multi-modal data have the potential to make these kinds of rich analogies and comparisons that are typical in childhood language learning (\reffig{education_mm}).

\clearpage
\hypertarget{technology}{\section{Technology}}
\label{sec:technology}

The technological foundations of foundation models give rise to the capabilities (\refsec{capabilities}) that determine their potential. 
To understand the technology used in development, we consider the data (\refsec{data}), model architectures (\refsec{modeling}) and systems (\refsec{systems}) used to train (\refsec{training}), and further adapt, (\refsec{adaptation}) these models alongside the theory (\refsec{theory}) that should be developed to understand this paradigm.
To then understand the resulting models, we discuss how to evaluate (\refsec{evaluation}) and interpret (\refsec{interpretability}) alongside the importance of robustness (\refsec{robustness}), security and privacy (\refsec{security}), and long-term AI safety (\refsec{ai-safety}) for ensuring the reliability of these models when deployed in society (\refsec{society}).

\pl{say that much of this section is the focus on the ML community, but really many of these sections has societal implications, and that we have tried to make these connections explicitly
}


\newsection 
\hypertarget{modeling}{\subsection{Modeling}}
\label{sec:modeling}

\sectionauthors{Drew A. Hudson, Antoine Bosselut, Alex Tamkin, Omar Khattab, Jared Quincy Davis, Jiaxuan You, Trevor Gale}

\begin{figure}[!ht]     
\centering
\includegraphics[width=\linewidth]{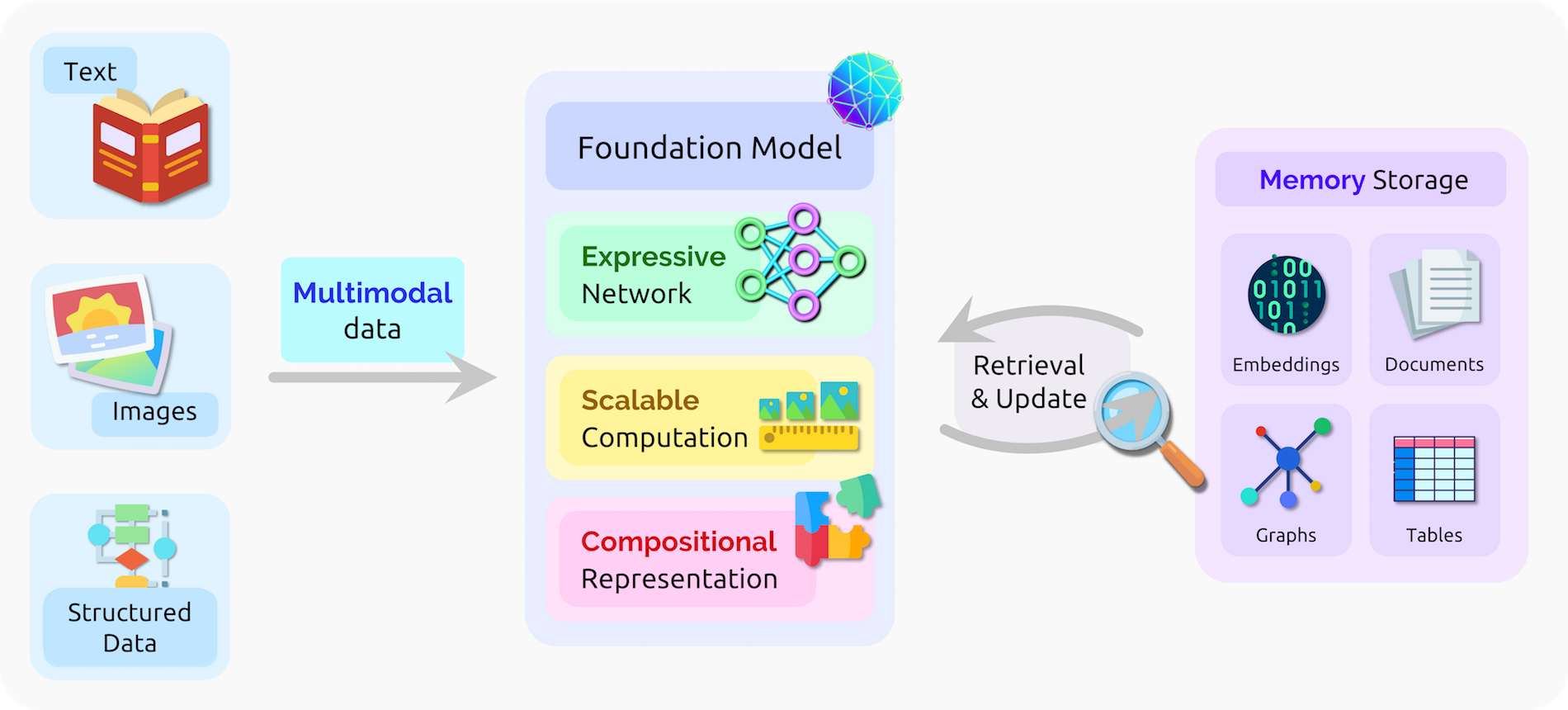}
\caption{The five key properties of a foundation model:  \textit{expressivity}\dash{}to flexibly capture and represent rich information; \textit{scalability}\dash{}to efficiently consume large quantities of data; \textit{multimodality}\dash{}to connect together various modalities and domains; \textit{memory capacity}\dash{}to store the vast amount of accumulated knowledge; and \textit{compositionality}\dash{}to generalize to new contexts, tasks and environments.}     
\label{fig:modeling} 
\end{figure}

The emerging paradigm of foundation models has attained impressive achievements in AI over the last few years, as models such as BERT \citep{devlin2019bert} shine at a wide spectrum of language understanding tasks: from textual classification and entailment to question answering and reading comprehension, while GPT-3 composes rich and fluent tales about unicorns \citep{brown2020gpt3} and DALL-E  shows signs of visual creativity, generating from scratch strikingly-realistic pictures of avocado chairs \citep{ramesh2021zeroshot}.

These and other instances of recent foundation models not only achieve remarkable performance across a multitude of diverse downstream tasks and applications \citep{squad2,wang2019superglue}, but also manifest noteworthy behaviors of interpretability \citep{stylegan2}, robustness \citep{devlin2019bert}, controllability \citep{styleclip} and generalization \citep{brown2020gpt3}. What does it take for a model to demonstrate such qualities? What architectures are capable of consuming large quantities of potentially multimodal information and translate them into rich knowledge of the world? And overall, what desirable properties should a network possess to give rise to a foundation model?

Here, we identify and discuss five such properties, spanning \textit{expressivity}, \textit{scalability}, \textit{multimodality}, \textit{memory capacity}, and \textit{compositionality}, that we believe are essential for a foundation model in order to: (1)~distill and accumulate knowledge from various sources and domains, (2)~organize it in an effective and scalable representation, and (3)~flexibly generalize it towards novel contexts.  For each of these properties, we motivate their necessity, provide examples of contemporary  models that incorporate them, and explore key challenges and promising avenues  for future research and development. See \reffig{modeling} for an overview diagram.

\hypertarget{modeling-expressivity}{\subsubsection{Expressivity}} 
\label{sec:modeling-expressivity} 

\textit{Expressivity} concerns with the theoretical and practical capacity of a network to model the data distribution it is trained over and represent it in a flexible manner. Prior works have proposed formal expressivity measures to characterize the complexity of functions a network can compute, or more precisely, approximate, which is essentially affected by its depth, width, connectivity, and structural patterns \citep{expressive}.

As the \textit{No Free Lunch} theorem suggests, there is no single model or algorithm that suits best for all cases \citep{lunch}, and so, for our purposes, we are particularly interested in identifying which models could effectively capture the facets of \textit{natural information}, such as human language or real-world images \citep{bengiobook}. These modalities are either continuous (as in vision) or discrete (as in language),  are distinctly hierarchical and high-dimensional, and present a complex set of relations and interactions among their constituent elements, whether these are  pixels, words or physical objects. 

Indeed, recent breakthroughs in generative modeling provide strong evidence for the high expressivity of neural networks, as they successfully express distributions of textual \citep{brown2020gpt3,devlin2019bert,jurassic1,gpt-j}, auditory \citep{Oord2016WaveNetAG}, and visual \citep{stylegan2,biggan} domains, and generate samples of high fidelity, diversity and realism.

\paragraph{Inductive Biases.} Much of the success of neural networks over the last decade in modeling natural data is owed to the \textit{networks' high depths}, as could be roughly measured by the number of stacked non-linear layers they are composed of, or the number of computational steps they take during their chain-of-reasoning.  Great depths play a crucial role in enhancing networks' expressivity, allowing them to form powerful \textit{hierarchical} and \textit{distributed} representations that could generalize from the training data to new unseen examples \citep{he2016resnet,levine2020limits}. 

The \textit{universal approximation theorem} \citep{approx} indeed states that even simple multilayer perceptrons (MLPs) can represent a broad set of functions, while different \textit{inductive biases}, as those implemented in {Recurrent Neural Networks} (RNNs) or {Convolutional Neural Networks} (CNNs) \citep{bengiobook}, can improve the learning efficiency and enhance the capacity of a given network to model different forms of information: sequential data, common to language, speech and time-series, for the former, or spatially-invariant information, as in images or videos, for the latter.

\paragraph{Transformer Networks \& Attention.} Meanwhile, transformer networks \citep{vaswani2017attention}, introduced more recently, demonstrate the importance of capturing \textit{long-range dependencies} and pairwise or higher-order interactions between elements. They build on the \textit{self-attention} mechanism \citep{vaswani2017attention,attention} that enables shorter computation paths and provides direct means to compare elements far-across the input data (such as a pronoun and its antecedent in a sentence, or two sentences that refer to the same topic).

From another perspective, the \textit{multiplicative interaction} embodied in both attention as well as gating structures (as in LSTMs \citep{lstms} or Mixture-of-Experts \citep{moe}) offers a more flexible alternative to the rigid fixed-weight computation of MLPs and CNNs, \textit{dynamically adapting the computation} to the input at hand. This proves especially useful for language modeling, where, for instance, given a sentence like ``She ate the ice-cream with the X", while a feed-forward network would always process it in the very same manner, an attention-based model could adapt its computation to the input\dash{}updating the contextual representation of the word ``ate'' if the prepositional phrase (PP) attachment X is ``spoon'', or instead link it to the ``ice-cream" if X refers \eg to ``strawberries" \citep{ppa}.

\paragraph{General-Purpose Computation.} 
A final notable advantage of attention over prior architectures stems from its stronger \textit{generality}, where it is not strongly tied to a particular task or domain, as is the case for the local receptive field of convolution or the sequential assumption of recurrent networks, both reflecting inherent properties specific to the vision and language modalities respectively. We hypothesize that the general-purpose nature of attention and transformers contributes to their broad applicability for a wide range of research problems and applications \citep{liu2019roberta,visual_transformer,ganformer}.

This contrast captures a more general \textit{trade-off between task-specialization and expressivity}: models with stronger structural priors can leverage them to improve sample efficiency on the particular tasks that benefit from these assumptions; while conversely, models that integrate weaker inductive biases learn more slowly, but can in turn scale to higher volumes of data and adapt to a diverse set of domains, since they do not rely on restrictive or task-specific suppositions. As both data and compute turn more accessible, we observe that the exploration of models with a \textit{minimal set of inductive biases} that can ``let the data speak for itself" seems to serve as a more promising approach for future research in the field.

\paragraph{Challenges \& Future Directions.} Notwithstanding the stellar progress and accomplishments of neural networks in general, and foundation models in particular, in terms of expressivity, notable challenges still remain. Leading approaches \citep{performer,visual_transformer} keep struggling with modeling of extremely long-range dependencies, such as those occurring in books, movies, or even DNA sequences, which may be attributed to the quadratic computation of contemporary transformer-based approaches \citep{wang2020linformer,lin-et-al-2021-naacl}. 

This challenge essentially reflects the \textit{trade-off between efficiency and expressivity}: where explicit modeling of long-distance interactions through short and direct computation paths improves expressivity on the one hand, but comes at the expense of scalability due to computation entailed by the increased connectivity on the other \citep{child2019generating,kitaev2020reformer,performer}. Models such as the GANformer \citep{ganformer} and the Perceiver \citep{jaegle2021perceiver,jaegle2021perceiverio} explore ways to balance these two properties and propose transformers with linear complexity that rely on \textit{bipartite} or \textit{bottleneck} attention, so to improve computational efficiency while maintaining high-expressivity. We believe that identifying an effective equilibrium between these two objectives offers an interesting avenue for future research.

Another important research direction relates to the expansion of foundation models, which, so far, have mainly focused on the language domain \citep{peters2018elmo,devlin2019bert,brown2020gpt3}, to different modalities, such as the structural \citep{gnn,gat} and perceptual \citep{tolstikhin2021mlpmixer,jaegle2021perceiver,dosovitskiy2021ima}, each involving a unique set of associated challenges. Likewise, we believe that exploring architectures for \textit{reasoning} (\refsec{reasoning}), which demands iterative computation chains and interaction with symbolic information, constitutes a valuable goal for future foundation models research.
 
\hypertarget{modeling-scalability}{\subsubsection{Scalability}} 
\label{sec:modeling-scalability}

Closely connected to model's expressivity is the notion of scalability. As rich data from varied sources becomes more readily available, and computational resources get stronger and more efficient (\refsec{systems}), we should look for ways to match this rate of progress and harness it to improve AI competency and versatility. For foundation models to effectively fit the complex and high-dimensional distribution of images or text, they should thereby be \textit{scalable} across all dimensions: including both models' depth and width as well as their training time, number of parameters, and amount of data they could process. 

\paragraph{Optimization.} Specifically, foundation models should both be: (1)~\textit{easy-to-train} (\refsec{training}), by being resilient to noise or imperfections in the data, and robust against \textit{instabilities} like vanishing \citep{helfrich2018orthogonal, glorot2010understanding} or exploding gradients \citep{lstms,nair2010rectified}, but also (2)~\textit{easy-to-adapt} (\refsec{adaptation}), by overcoming phenomena of catastrophic forgetting \citep{catastroph} and supporting few-shot learning \citep{fewshot}. We are still in the early days of understanding the principles that drive the scalability of learning algorithms, and while recent works have started to shed some light on these themes \citep{liu2020understanding, kuditipudi2020explaining, nakkiran2019deep}, much work remains to be done.

\paragraph{Hardware Compatibility.} Moving beyond aspects of \textit{robustness} and \textit{optimization}, foundation models should also be \textit{practically efficient} (\refsec{systems}), and take advantage of contemporary and future hardware \citep{2009_06489}. One example of that is \textit{parallelizablity}, an important property that characterizes the computation supported by GPUs. Indeed, much of the transformers' great success over the previously dominating recurrent approach was driven by their higher degree of parallelism. 

Looking forward, given the fast-pace progress of systems development, we should further ensure that models are designed to co-adapt to future hardware advances. Consequently, foundation models should ideally be amenable to schemes such as distributed training, which is gaining popularity, as is the case for \eg Mixture-of-Experts, and possibly leverage properties such as \textit{sparsity} of the computation or representation, as is the case for the Longformer \citep{longformer}, BigBird \citep{bigbird}, and Sparse Transformer \citep{child2019generating} approaches, and which likely will become more central in future hardware and processors.

\subsubsection{Multimodality.} Traditionally, the fields of computer vision, robotics, and NLP have made progress in an independent manner, with separate communities developing specific approaches suitable for each modality. A conducive consequence the rise of deep learning has brought about was the bridges it helped forming among the various communities and research areas within AI, as seemingly different problems could now be tackled by closely-related approaches, and studies of originally remote topics began converging to a common ground. This breakthrough opened up a new range of possibilities, fostering pioneering exploration into the theme of multimodality, encompassing  areas as varied as language grounding \citep{lynch2020grounding}, visual semantics \citep{conser2019revisiting},  embodied environments \citep{embodied} and interactive agents \citep{interactive}.

Essentially, multimodality serves as a key component of intelligence, and is a crucial factor for the development of both thorough and broad comprehension of the world. Concretely, language learning is more effective when occurring in a grounded environment rather than in a vacuum. And inversely, from the vision perspective, language encourages the emergence of abstractions that link between low-level perceptual signals and statistics to semantic concepts of objects, properties, agents and motivations, thereby enriching and elevating visual representations. 

In light of these observations, we argue that foundation models should ideally connect together the different modalities, distill their embodied information into a shared multifaceted representation, and capture the full range of inter-connections and relations among them so as  to furnish a wide range of capabilities (see \refsec{language},  \refsec{vision},\refsec{robotics}, \refsec{reasoning}).

\paragraph{Generality and Specialization.} An important design choice for multimodal foundation models is the degree of \textit{specialization}, or the  \textit{structural sharing} between the modules responsible for each modality. Naturally, data of different domains exhibits diverse kinds of structures and properties\dash{}where, for instance, language is discrete while vision is continuous. At first sight, this variation hints that specialized inductive biases tailored for each modality could be of aid. Yet, as training scales upwards and models are provided with the opportunity to base their learning less on structural priors and more on the data itself, general approaches that maintain only a handful of broad general assumptions prove in fact a lot more successful than task-specific alternatives. And so, as corroborated by recent success of general-purpose models like transformers across different modalities\dash{}both linguistic \citep{liu2019roberta,albert} and visual \citep{visual_transformer,ganformer}, we see that \textit{generality} is critical for improving AI capabilities.

\paragraph{Multimodal Interactions.} Another key consideration for multimodal models relates to \textit{weight sharing}: do the various modalities benefit from using the same or different parameters for their respective components? Prior works have shown that fruitful transfer could certainly occur across modalities, but the ideal degree of sharing remains unclear, so is the existence of principled ways for discovering it. 

Finally, a major design question concerns with the forms of the multimodal interactions supported by the model, which vary widely between concrete cases and examples: Cross-modal or \textit{late-fusion} models such as ConVIRT \citep{convirt} and CLIP \citep{radford2021learning} maintain fully separate encoders for each data source, and compare their spaces only at the ultimate computation stage, using \eg a simple dot product. Meanwhile, \textit{early-fusion} models, such as ViLBERT \citep{Lu2019ViLBERTPT, cho2021unifying}, jointly reason over multiple modalities necessary for tasks of visual reasoning and question answering. Identifying the optimal stage and form for merging the respective vector spaces \citep{attention_bottlenecks} remains an open research question.

Overall, while there seems to be a consensus within the community about the importance of multimodality, models that go beyond shallow alignment of vision and language are yet to exist, and the theme of grounded language learning in embodied environments still has much room for exploration.

\hypertarget{modeling-memory}{\subsubsection{Memory}} 
\label{sec:modeling-memory}
So far, we have discussed the  foundation models' goal to gather and accumulate information from varied modalities at large scales. This knowledge encompasses both broad understanding of the world as well as specific mastery of niche subjects or particular facts. Representing such a large body of learned information is by no means trivial, and is leading to interesting questions about effective mechanisms for \textit{access}, \textit{storage}, \textit{retrieval} and \textit{manipulation} of particular items or \textit{memories}.

\paragraph{Explicit Storage.} An important design principle that could achieve these desiderata is to \textit{separate out \textit{computation} from \textit{memory}} \citep{Weston2015MemoryN,dnc,mac,nsm} in order to enhance models' ability to \textit{transfer knowledge} by applying previously acquired \textit{abstract} skills to new \textit{concrete} settings. 

In this context, it is important to distinguish between \textit{explicit facts}\dash{}that can be stored in an external memory storage, and \textit{implicit knowledge}\dash{}that is reflected through the networks' trainable weights. Such decoupling of explicit and implicit knowledge enjoys multiple advantages compared to the alternative of implicitly encoding all information together through the network weights. The separation mitigates the inflation in models' size and number of parameters needed to store the growing quantities of knowledge \citep{guu2020realm}, improves models' trust and reliability by increasing their knowledge provenance \citep{cheney2009provenance}, and most notably, is key for memory update, manipulation or adaptation \citep{lewis2020retrieval} (\refsec{adaptation}), which could in turn enable \textit{generalization} to novel contexts and downstream tasks. 

Indeed, disentanglement between memory and computation has been a recurring goal in deep learning and NLP research over the last years, including models such as Memory Networks \citep{Weston2015MemoryN,Sukhbaatar2015WeaklySM}, the Neural Turing Machine \citep{ntm,dnc}, the Neural State Machine \citep{nsm}, and MAC \citep{mac}. Furthermore, using \textit{key-value structures} \citep{kv} for accessing external memories has been shown to be very effective for modeling long-term dependencies \citep{ent,Bosselut2018SimulatingAD,Lample2019LargeML}. 
Transformers, the celebrated architecture underlying most foundation models to date, likewise exhibits operations that involve key-value memory-access and computation among the contextual word representations they gradually build \citep{Geva2020TransformerFL}. 

\paragraph{Information Retrieval.} Once a model completes gathering the information after training, there are multiple ways to retrieve particular facts or memories necessary for downstream applications and tasks. Some employ \textit{explicit prompting} techniques that query the model's knowledge through input sequences \citep{Petroni2019LanguageMA,Kassner2021MultilingualLI,jiang-etal-2020-know}  while other approaches involve \textit{implicit recollection and reshaping} of the prior knowledge through an adaption phase \citep{Bosselut2019COMETCT,Hwang2021COMETATOMIC2O}. A third category of methods goes a step further and combines neural-based computation with \textit{symbolic} aggregation and retrieval of information from either unstructured textual repositories \citep{karpukhin2020dense,lewis2020retrieval,Khattab-etal:2020:OpenQA} or even structured resources such as \textit{knowledge graphs} \citep{Zhang2019ERNIEEL,Peters2019KnowledgeEC,Liu2020KBERTEL,Verga2020FactsAE,yasunaga2021qagnn}.

However, there is \textit{trade-off between the strong memorization skills} offered by \textit{retrieval mechanisms} on the one hand and the \textit{richer representations} learned when there is an \textit{information bottleneck} on the other. Indeed, over-reliance on retrieval reduces the opportunities to learn how to represent information in compact and abstract manners, distill key insights and concepts out of the vast amounts of input information the model is exposed too, and, basically, separate the wheat from the chaff. For instance, the in-context learning abilities of GPT-3 possibly emerge  as a by-product of enforcing the network to represent the input sequential data through its bounded memory architecture \citep{brown2020gpt3}. Overall, While they certainly have some merits \citep{guu2020realm}, models that rely on external retrieval mechanisms may not learn to generalize as effectively as bounded, compact and abstract representations.

\paragraph{Knowledge Manipulation.} Finally, when considering large-scale learning over long durations, it is crucial to note the \textit{dynamic} nature of knowledge, where facts' correctness and validity can change over time as the world keeps evolving\dash{}and what was true or relevant yesterday may not be so tomorrow. It is therefore crucial for a model to represent its knowledge in a manner that supports efficient update or manipulation of facts as part of its lifelong learning. 

\subsubsection{Compositionality}
\textit{Compositionality} can be defined as the principle according to which \textit{the meaning of the whole is derived from the meaning of its constituent parts}, and the rules applied to combine them \citep{compositionality,bottou}. It is a crucial ingredient of human intelligence \citep{humanthink}, underlying our capabilities to plan, reason and learn readily and efficiently from a handful of examples. Compositionality may hold the key to achieve \textit{out-of-distribution}\dash{}or specifically\dash{} \textit{combinatorial generalization}.  Drawing on classic ideas from symbolic AI, it encourages and enhances desirable properties within neural networks, such as interpretability, controllability and data-efficiency  \citep{humanthink}, and can take different forms, characterizing variety of elements: 

\paragraph{Model.}
Compositionality can be reflected at the model level, in terms of its architectural properties, structure, and degree of \textit{modularity}\dash{}which can increase training and inference efficiency of large neural models \citep{moe}. It also links to themes of \textit{interpretability} and \textit{multimodality}, as it relates to the interfaces between the different modules the model is composed of, what modes of interactions they employ, and how transparent they are.

\paragraph{Computation.}
Models such as Module Networks \citep{nmn} and Mixture-of-Experts \citep{moe} go further along this direction, exhibiting not only structural modularity, but also \textit{compositional computation}, supported by the \textit{specialization of sub-networks} to different operations, in a manner that adapts and tailors the model behavior to the input at hand. While some methods rely on concatenation of hand-engineered modules \citep{nmn}, alternative approaches  enable the network specialization to naturally emerge through learning \citep{moe}. Other models, such as MAC \citep{mac} and Dynamic Memory Networks \citep{dmn} perform an explicit \textit{iterative computation}, where a given task is decomposed into multiple reasoning steps, performed one by one, manifesting sequential progression from a set of initial facts to novel inferences and conclusions.
  
\paragraph{Training \& Data.} 
Not only can the model or its computation be compositional, but so can be the data or training processes too \citep{andreas2019good}. Instead of training one model over a complete dataset, one could split, or decompose it into subsets, train different models on each one independently, and ultimately recombine them at test time through various \textit{ensemble techniques} \citep{ensemble}. Such approaches could have far-reaching implications on the training and deployment procedures of foundation models, in both practical and even societal regards.

\paragraph{Representation.} 
We have discussed compositionality of different elements, such as the model, the computation, the training schemes or the data. But most notably, the learned representation itself, which emerges over the course of the model training and adaptation, can also be compositional \citep{andreas2019measuring}. Indeed, a promising manner to represent knowledge is through structured, potentially graph-based, object-oriented representations \citep{Zhang2019ERNIEEL,kepler}, that center around identifying \textit{entities and event nodes} and forming \textit{connections, analogies and relation edges} among them. It reflects a natural way to organize information about the world, where inputs from different modalities can be channeled and aggregated around semantic multi-faceted concepts. Such representations could support \textit{multi-hop reasoning} and inference \citep{kbert, colake, jaket}, and potentially also enable stronger \textit{out-of-distribution generalization} through \textit{recombination}. 

However, compositionality can also hinder the expressivity of the representation, and impede its capacity to account for idiosyncrasies, exceptions, and \textit{contextual correlations} \citep{redwine}. In other words, the whole can sometimes be greater than the sum of its parts, where for instance, \textit{red wine} is not the same as \textit{red onion}. But while many approaches that have dominated over the last decade tend to focus mostly on one end of the spectrum, and learn monolithic distributed representations, we believe that exploring manners to reach a better \textit{balance between contextuality and compositionality} is a promising avenue for future research.

\subsubsection{Summary}
We have introduced five properties that we believe are essential for the next generation of foundation models, in order to effectively distill the large amounts of information around us so to successfully address downstream tasks: \textit{expressivity}\dash{}to flexibly capture and assimilate real-world information, \textit{scalability}\dash{}to adeptly handle high volumes of high-dimensional data, \textit{multimodality}\dash{}to consume, process and potentially produce content from different sources and domains, \textit{memory} capacity\dash{}to effectively store and retrieve the acquired knowledge, and finally, \textit{compositionality}, to foster successful generalization to novel tasks, settings and environments. We believe that the realization of the full potential of foundation models, as is envisioned and discussed in  detail throughout this report, will rely on research of new architectural and modeling advances to fulfill these desiderata.
\newsection
\hypertarget{training}{\subsection{Training}}
\label{sec:training}
\sectionauthors{Alex Tamkin}

Training objectives are mathematical functions describing how to transform a model architecture and large amount of broad data into a foundation model. For example, GPT-3 was trained with a language modeling objective, which rewards the model for predicting the next word correctly \citep{Shannon1948AMT}. We begin by laying out some goals of these training approaches, describe important design trade-offs in current approaches, and outline important goals for the path ahead.

\subsubsection{Goals of training objectives} 

Here we outline some key goals for training algorithms in light of the recent rapid progress in these methods and models.\footnote{We use ``training" instead of pretraining to emphasize the primacy of the foundation model itself, and because some methods for adapting foundation models to downstream tasks do not involve any later stage of training.}

\paragraph{Leveraging broad data.} 
The rise of \textit{self-supervised} learning algorithms has unlocked the power of internet-scale datasets which would be intractable to annotate by hand. This kind of broad data comes in many forms, including images, audio recordings, and video (\refsec{vision}); robotic and sensor data (\refsec{robotics}); and text, either in isolation or paired with other modalities like images (\refsec{language}). Because this data lacks external annotations, a major focus for researchers is designing bespoke self-supervised algorithms that leverage the unique structure within each kind of data to produce a training signal for a foundation model.

\paragraph{Domain completeness.} An important goal for foundation model training algorithms is to be \textit{domain complete}, in the sense that solving the training task requires capabilities that are broadly useful for downstream tasks in the domain (see \refsec{language}, \refsec{vision}, \refsec{robotics}). This property is crucial for the \textit{generality} of a foundation model. For example, language modeling may require models to acquire capabilities as wide-ranging as coreference, sentiment and translation as the model learns to predict the next word in a document. In contrast, a supervised learning task like sentiment classification may lead to a more narrow set of capabilities (see \refsec{language}). As important as this quality is, it is not obvious \textit{a priori} what tasks will result in a domain complete capabilities, or even how to evaluate the full breadth of a model's capabilities (see \refsec {evaluation} and \refsec{theory}). 

\paragraph{Scaling and compute efficiency.}
Procedures for training foundation models must reliably convert data, a model architecture, and compute into a broadly capable model. To maximize the capability of a foundation model, we can identify the bottlenecks to this process and propose new training algorithms which remove them. The rise of self-supervised algorithms has made model size and compute resources increasingly salient bottlenecks \citep{kaplan2020, henighan2020}, leading to a shift where models are evaluated not solely on their capabilities but rather on the amount and kind of compute needed to reach those capabilities (\refsec{evaluation}). The efficiency of training objectives can vary tremendously,\footnote{\eg 4x for ELECTRA \citep{Clark2020ELECTRAPT} vs BERT \citep{devlin2019bert}, 12x for contrastive vs generative approaches to CLIP training \citep{radford2021learning}} laying in sharp relief how important the design of a training approach is to the emergence of powerful capabilities given a fixed compute budget. Thus, a major goal for training researchers is to design training objectives with a richer training signal, resulting in models which learn faster and attain stronger capabilities.\footnote{Of course, a key goal for computer systems designers is to alleviate compute as a bottleneck for training (see \refsec{systems}) And the choice of a training method is ultimately also constrained by the availability of diverse, high-quality data (\refsec{data}), which continues to be a major challenge for many domains, including robotics (\refsec{robotics}) and low-resource languages (\refsec{language})} One force aiding this development is the surprising predictability of how capabilities scale with different kinds of architectures, data sizes, and compute \citep{Hestness2017DeepLS, kaplan2020},  a striking phenomenon which enables model developers to make choices based on clearer trends instead of more costly random searches.

\subsubsection{Design trade-offs in current SSL methods}
Current self-supervised learning (SSL) methods for training foundation models are diverse, but what unites them is that they produce prediction problems from unlabeled data without the need for human annotators. SSL objectives manufacture a rich training signal from this data through carefully-designed constraints, either on the data itself (\eg redacting or noising) or on the way the model is able to represent or process the data (\eg latent bottlenecks). At some level, these constraints ``bake in" the kinds of capabilities desired when adapting models to downstream tasks (\refsec{adaptation}).\footnote{For example, the causal language modeling objective used to train GPT-3 \citep{brown2020gpt3} enabled conditioning it via prefixes. And the color jitter augmentations used during contrastive learning \citep{chen2020simclr} encourage invariance to properties not thought to be useful for downstream tasks. Better understanding how the particular choice and structure of these constraints influences the capabilities acquired by the model is an important area for future work (\refsec{theory}).}

Here, we describe three important design choices that current models explore, along with their respective tradeoffs in terms of their resulting capabilities.

\paragraph{At what level of abstraction should we model?} A fundamental question is what the input representation of a foundation model should be. One option is to model the input at the level of raw bytes. However, this high dimensionality may cause the model to focus on predicting less semantic aspects of the  input,\footnote{\eg blades of grass, audio compression artifacts, or spellings of words} slowing the rate at which it acquires more generally-useful capabilities. These approaches also become intractable when training models like transformers \citep{vaswani2017attention} whose compute costs grow quadratically with the input size.\footnote{See \refsec{vision} and \refsec{modeling} for discussions of training costs for high-dimensional sequences, such as images and video} Another option is to use domain knowledge to reduce the input space of a model\dash{}such strategies include patch embeddings \citep{visual_transformer} as well as fixed or learned tokenization \citep{Schuster2012JapaneseAK, Sennrich2016NeuralMT, Kudo2018SentencePieceAS, Oord2017NeuralDR, ramesh2021zeroshot}. These methods may alleviate some challenges facing generative approaches, but have the trade-off that they may jettison possibly-useful information in the input.\footnote{For example, tokenizing text may make it harder to learn rhymes, puns, or other tasks that benefit from character-level information \citep{branwen2020gpt}} The choice of a continuous vs discrete input also has trade-offs for adaptation (\refsec{adaptation}); more work is needed to capture the benefits of both approaches.

\paragraph{Generative vs discriminative models}
Generative training approaches are conceptually elegant yet powerful\dash{}they train models to learn joint or conditional distributions over training inputs. Two major families of generative foundation models include autoregressive foundation models \citep{Oord2016WaveNetAG, Radford2018ImprovingLU, chen2020imagegpt, Yang2019XLNetGA, ramesh2021zeroshot}, which generate inputs piece by piece, and denoising foundation models \citep{devlin2019bert, raffel2019exploring}  which corrupt and then recover the inputs. The specific kind of generation performed in the training process determines what kind of interactivity is available during adaptation\footnote{For example, autoregressive models like GPT-3 enable prefix-based conditioning, while denoising models like T5 or BERT facilitate the use of bidirectional context to replace arbitrary-length spans or fix typos.} (see \refsec{interaction} and \refsec{adaptation}), and future models may enable an even richer set  of interactions.\footnote{Other kinds of generative approaches less studied in a foundation modeling context include diffusion and score-based models \citep{SohlDickstein2015DeepUL, Song2019GenerativeMB, Ho2020DenoisingDP}, VAEs \citep{Kingma2014AutoEncodingVB}, flow models \citep{dinh2015nice, Kingma2018GlowGF}, and GANs \citep{goodfellow2014gan}\dash{}it remains to be seen whether these or other future approaches can also enable learning as diverse variety of capabilities as autoregressive or denoising approaches.}

While generative training approaches have their benefits, several discriminative approaches have also recently gained traction. These methods do not enable generation-based interaction, yet they may enable more efficient learning for classification- or regression-based tasks in high-dimensional continuous settings like images, audio, and video. Most of these methods output vectors for (parts of) inputs, which are trained to be similar for different ``views'' of an input \citep{Wu2018UnsupervisedFL, Oord2018RepresentationLW, chen2020simclr, He2020MomentumCF, Grill2020BootstrapYO, Caron2021EmergingPI, convirt, radford2021learning} or used to predict whether parts of inputs are real or fake \citep{Clark2020ELECTRAPT, Iida2021TABBIEPR}. Better understanding the trade-offs between generative and discriminative training, as well as capturing the best of both approaches, remain interesting avenues for future study.

\paragraph{Capturing multimodal relationships.} Another increasingly important research area is capturing the relationships between multiple kinds of data. What this means may differ based on the context and the goals of a modeler. For example, CLIP \citep{radford2021learning} and ViLBERT \citep{Lu2019ViLBERTPT} are both multimodal vision-language, but differ in the precise \textit{way} they are multimodal.\footnote{See \refsec{vision} and \refsec{language} for more discussion of multimodality in vision and language specifically} The former encodes images and text separately into vectors, enabling users who have examples from a single modality to retrieve, score, or classify examples from the other modality. The second processes images and text jointly at an early stage of the model, enabling downstream applications like visual question answering where reasoning over pairs of related images and text (\eg images and questions about them) are provided. Multimodal foundation models remain a nascent research area; much is still unexplored about the different ways a model can be multimodal as well as better understanding the capabilities these additional modalities bring.

\subsubsection{Paths forward}
We close with some important goals for the future of foundation model training.

\paragraph{Out-of-the-box SSL} Right now, SSL objectives are highly domain-specific: different methods currently prevail in natural language processing, computer vision, and speech processing. This has two major disadvantages: First, these different techniques make it challenging to grasp the common threads and scientific principles underlying \textit{why} each of these methods work. Second, this domain-specificity requires developing new foundation model training methods from scratch for each new field, including medical, scientific, and new multimodal settings. A more general objective for efficiently training foundation models on any kind of data would represent a significant milestone for the foundation model training community \citep{Tamkin2021DABS}.

\paragraph{Obtaining a rich training signal} It is clear that not all training objectives are made equal\dash{}some are radically more efficient than others, translating into far more capable foundation models for a given compute budget. Are there training methods orders of magnitude more efficient than those currently known? If so, how can we find them? These investigations will be shaped by many forces, including what future software and hardware advances (\refsec{systems}) make possible. We also need not view data (\refsec{data}) and training algorithms as independent factors: not only does the quality and availability of the data influence the training signal,\footnote{Including any undesirable or biased capabilities (\refsec{fairness})} but the training algorithm itself could adaptively seek out or construct richer training examples as the model improves to accelerate learning \citep{Tamkin2021ViewmakerNL}.

\paragraph{Goal-directed training of foundation models.} Adaptation methods such as prompting (\refsec{adaptation}) draw on emergent properties that result almost as an afterthought of training. Can we train foundation models where the ability to understand and reliably carry out goals in a complex world is part of the model's training objective? A focus on developing general capabilities distinguishes this direction from the goal of adapting an existing foundation model to a specific task via reinforcement learning (\eg \citet{Stiennon2020LearningTS}). Instead, one might imagine more sophisticated versions of current methods which acquire a diverse range of real-world capabilities from raw online \citep{Klyubin2005EmpowermentAU, singh2005intrinsically, Salge2013EmpowermentA, Mohamed2015VariationalIM, Florensa2017StochasticNN, Pathak2017CuriosityDrivenEB, Haber2018LearningTP} or offline  \citep{Precup2000EligibilityTF, lange2012batch, Ajay2021OPALOP, Yang2021RepresentationMO, Schwarzer2021PretrainingRF} interactions, without the need for human annotations or task construction. Such methods might use techniques quite similar to existing SSL algorithms: \eg training sequence models in goal-directed contexts where they can be directly \textit{asked} to carry out certain tasks via conditioning (\eg UDRL \citep{Schmidhuber2019ReinforcementLU, Srivastava2019TrainingAU} or Decision Transformer \citep{Chen2021DecisionTR}; also see \refsec{robotics}). The complex behaviors that have already emerged in simple interactive environments  \citep{Baker2020EmergentTU} suggest multitask, multiagent, and multimodal goal-directed training of foundation models as an interesting avenue for future study.
\newsection
\hypertarget{adaptation}{\subsection{Adaptation}}
\label{sec:adaptation}
\sectionauthors{Xiang Lisa Li*, Eric Mitchell*, Sang Michael Xie, Xuechen Li, Tatsunori Hashimoto}

\begin{figure}[!ht]
\centering
\includegraphics[width=\linewidth]{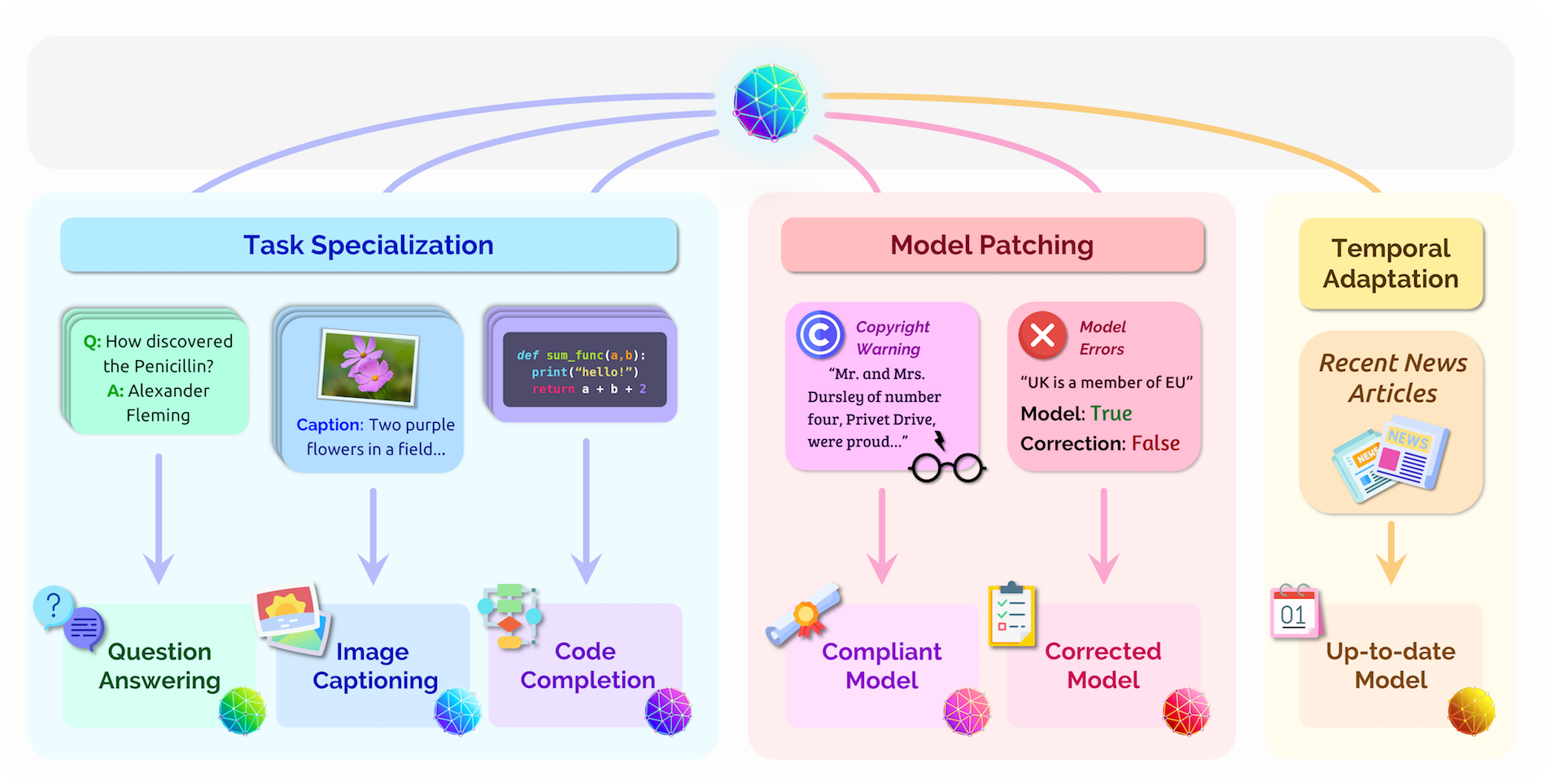}
\caption{\label{fig:adaptation} During adaptation, a foundation model is converted into an \textit{adapted model} (bottom row) in order to reflect updated information, desired behaviors, or deployment constraints.}
\end{figure}

\noindent While foundation models provide a powerful general-purpose engine for processing multi-modal information, \textit{adapting} a foundation model before use is necessary for some applications. Broadly, an adaptation procedure produces an adapted model by conditioning a foundation model on additional information, either by priming the foundation model through the inclusion of new data or a prompt in its input or by updating some or all of the foundation model's parameters to reflect the new information. For example, in text summarization, appending a prompt such as \texttt{TL;DR} to the input article can improve foundation model performance  \citep{radford2019language} by acting as a task specification for the foundation model. Alternatively, fine-tuning the parameters of a foundation model with an organization's internal, domain-specific data could improve the model's accuracy by adding information relevant to the organization’s use case. In this section, we describe existing approaches to adaptation and several factors that determine whether a particular adaptation procedure is appropriate for a particular setting. We additionally describe various use cases for foundation model adaptation, including relatively well-studied settings such as specialization of a foundation model to a particular task or domain as well as more speculative settings like test-time data removal \citep{bourtoule2019machine} and editing model behavior on particular inputs \citep{sinitsin2020editable}. We conclude by presenting a long-horizon goal for future research in foundation model adaptation.
\subsubsection{Methods for foundation model adaptation}
Many methods for adapting foundation models have been proposed, making the decision of \textit{which} adaptation procedure to use for a particular problem or compute environment difficult. We emphasize three factors of particular importance for practitioners to consider when selecting an adaptation procedure: (1) the compute budget (specifically storage and memory); (2) the amount of task-specific data available; and (3) extent of access to foundation model gradients.
\noindent\paragraph{Factor 1: Compute budget.} For foundation models with billions or trillions of parameters, fine-tuning all model parameters may demand prohibitively large memory. Also, separately fine-tuning for many tasks can incur unacceptable storage costs. 
There are many works that propose methods to reduce the storage for adapting foundation models, and we refer to this class of lightweight adaptation methods as \textit{low-storage adaptation}. 
Typically, methods in this class freeze most of the pretrained foundation model parameters and only learn a relatively small number of task-specific parameters (either by fine-tuning some pretrained parameters or by adding altogether new modules), reducing the storage overhead for each task.

The key design decision for such algorithms is the choice of parameters to be adapted. Perhaps the simplest approach is to tune only the final layer of the pretrained model, while other works tune only the model's bias vectors \citep{zaken2021bitfit}, low-rank residuals to model weight tensors \citep{hu2021lora}, or masks over model parameters \citep{zhao2020masking}. Another line of research tunes ``soft'' prompts \citep{li2021prefix,qin-eisner-2021,liu2021prompt, lester2021power,hambardzumyan-etal-2021-warp}, corresponding to sequences of arbitrary parameter vectors rather than embeddings of the model's vocabulary, and conditions the foundation model on these prompts by concatenation with input activations, either at the input layer or at all layers. Another approach freezes all model parameters and interleaves new MLP modules with trainable parameters between existing model layers \citep{houlsby19adapter}. While these lightweight adaptation techniques seem to trade off parameter efficiency and performance on downstream tasks, they sometimes achieve comparable performance to full fine-tuning, despite updating 1000$\times$ fewer parameters \citep{zaken2021bitfit,li2021prefix,hu2021lora}. \citet{lester2021power} shows an instance when the performance gap between full fine-tuning and lightweight adaptation vanishes as the model size increases. 
We remain uncertain how lightweight adaptation techniques scale as model size increases \citep{aghajanyan2020intrinsic}. Because GPU memory is typically a more limiting resource than disk storage, low-\textit{memory} adaptation procedures are perhaps even more critical to democratizing foundation models than low-storage adaptation procedures. Various techniques for low-memory model training have been proposed, which can be directly applied to fine-tuning-based adaptation procedures (\refsec{systems}). However, some low-memory procedures such as gradient checkpointing \cite{gradient-checkpointing} trade off computation and memory, potentially exacerbating the significant energy consumption of foundation models \cite{bender2021}.

\noindent\paragraph{Factor 2: Data availability.} Task specialization mostly demands task-specific labeled data as training signals.\footnote{Prompts are an exception, although we might consider prompts to implicitly represent the information contained in a batch of labeled data \cite{le-scao-rush-2021-many}.} However, the cost of annotation varies greatly across tasks and languages; for example, annotating MRI data requires expert medical knowledge, whereas labeling sentiment for English texts requires only common sense judgement.
When adaptation data is plentiful, we might apply traditional fine-tuning approaches or their lightweight counterparts.
In low-resource language-based settings, combining prompting and fine-tuning has been shown to be a promising direction~\citep{schick-schutze-2021-exploiting, schick-schutze-2021-just,gao2020fewshot,perez2021true,logan-cutting-2021,min2021noisy}. \citet{le-scao-rush-2021-many} shows that a well-tuned prompt can be worth around 100 training examples, and fine-tuning a carefully prompted foundation model is significantly more data-efficient than fine-tuning an unconditioned foundation model. 

\noindent\paragraph{Factor 3: Access to foundation model gradients.} 
Despite the significant impact of foundation models on some research communities, 
little standardization of distribution practices exists for large scale foundation models (with more than 50 billion parameters). 
As we gradually become aware of the potential hazards from the misuse of foundation models (see \refsec{misuse}), providing access to all of a foundation model's parameters for fine-tuning might result in ethical concerns. Moreover, most users do not have enough compute resources to exploit their full access. For example, the memory requirements of foundation models might preclude their direct fine-tuning for many organizations and institutions. Consequently, future foundation model providers would likely restrict access to the full parameters of the model and instead provide surrogate API access, as exemplified by an early foundation model, GPT-3. On one extreme, if a foundation model provider enables access only to the model output (\eg the textual continuation of a prompt, the generated image, or a score evaluating the alignment between an image and a textual description), foundation models can be adapted using in-context learning \citep{brown2020gpt3}. In-context learning freezes the foundation model parameters, and steers the output of the foundation models by conditioning on a (typically natural language) prompt, which might be composed of task instructions or demonstrations. To improve the performance of in-context learning, the prompt needs to be carefully designed, either by manual search or an automated procedure \citep{jiang-etal-2020-know, shin-etal-2020-autoprompt}, and validated on the adaptation data. At the other extreme, if the foundation model provider grants access to gradients with respect to model parameters, full fine-tuning can be applied, where all model parameters are updated to improve performance on a downstream task. As a middle ground, we might obtain gradient access only to foundation model inputs,\footnote{Assuming the foundation model provider enables the input space to be continuous.} which are considerably lower dimensional than foundation model parameters. In this case, we could deploy lightweight adaptation techniques  \citep{liu2021prompt, li2021prefix,lester2021power}, which freeze the model parameters and optimize a continuous prefix or prompt for each task.

\hypertarget{adaptation-usecases}{\subsubsection{Use cases for adaptation}}
\label{sec:adaptation-usecases}
Adaptation is useful whenever the desired use case of a model differs from the relatively general training objective used for foundation model training (\refsec{training}). Most commonly considered is the case in which a foundation model is adapted to perform a specific task (\eg text summarization or animal classification from images), narrowing the scope of the model. Indeed, the vast majority of existing approaches described earlier in this section have targeted this setting. However, other forms of adaptation are useful, such as making local model edits to correct undesirable predictions for particular inputs or adding privacy constraints to the trained foundation model, which are task-agnostic. In this subsection, we describe a variety of use cases for adaptation, the methods that are most applicable to them, and remaining challenges in addressing these settings.

\paragraph{Task specialization.}
The most widely-studied case of foundation model adaptation is that of task specialization, in which a foundation model is adapted to optimize performance for a specific task or set of tasks. For example, specializing for summarization tasks would induce foundation model behavior to extract key ideas from the input document(s) and re-organize them in short summary sentences. Various adaptation procedures have proven effective for task specialization, showing significant improvement over performance of the unadapted model \citep{howard2018universal,brown2020gpt3}. 
In addition to the relatively widely-studied setting of specializing foundation models to specific tasks, other task-agnostic adaptation problems become increasingly challenging (but no less important) for foundation models owing to their size and computational demands. For example, the cost of training foundation models makes continual training over time to keep a model’s predictions up to date with current events particularly expensive.
Additionally, the challenge of collecting massive anonymized datasets used to train foundation models (\refsec{data}) make the likelihood of personal information leakage into training sets non-trivial; mechanisms to efficiently remove training data from a foundation model post-facto are therefore desirable.

\paragraph{Temporal adaptation.} Ideally, foundation models store knowledge that closely represents the state of the world, independent of modality. However, the world is constantly changing; new heads of state are elected, clothing styles change, social norms and beliefs shift (\refsec{ethics}), and the use of language evolves, causing a shift in the input distribution, target predictive distribution, or both.
This temporal shift presents a challenging statistical problem due to the induced distribution shift, as discussed in \refsec{robustness}. 
For foundation models, temporal shift also presents a particularly difficult computational problem; due to the computationally demanding nature of training foundation models \citep{shoeybi2019megatronlm,brown2020gpt3}, frequent re-training from scratch might carry unacceptable financial or environmental impacts \citep{bender2021} (\refsec{environment}), or simply take too long to be a viable method for keeping models up to date.
In visual domains, gradual self-training on unlabeled data across intermediate time points can bridge the temporal shift across a long time period, but remains an expensive retraining procedure~\citep{kumar2020gradual}. 

In the context of language models, temporally-partitioned diagnostic datasets help quantify the rate at which large language models become outdated \citep{lazaridou2021pitfalls,hombaiah2021dynamic,dhingra2021time}, showing that classic techniques like re-weighting training data and dynamic evaluation (updating model parameters with new data at production time \citep{mikolov2010recurrent}) can partially alleviate, but not fully solve, this problem. Explicitly conditioning a language model on the time period it is intended to model is another technique that has shown some promise \citep{dhingra2021time}.
An alternative strategy to addressing temporal shift is to design retrieval-based (semi-parametric) models, which augment the model input with additional context retrieved from a large, human-interpretable database (\eg Wikipedia articles) \citep{karpukhin2020dense,lewis2020retrieval,guu2020realm,Khandelwal2020Generalization,Khattab-etal:2020:OpenQA}.
For retrieval-based models, adaptation corresponds to updating individual units of information in the database (\eg single paragraphs of encyclopedia articles) without re-training the model. While promising, challenges for retrieval-based approaches remain both in training an accurate retrieval mechanism and in accurately conditioning on the retrieved information \citep{lewis2020retrieval}. We revisit the problem of temporal adaptation in the more general context of continual learning later in the section.

\paragraph{Domain specialization.} In addition to task specialization, it is often necessary to specialize a foundation model to a particular domain (such as legal documents), without limiting the breadth of tasks the foundation model can accomplish.
This specialization induces a mismatch between the foundation model training and adaptation data distributions (\refsec{robustness}) which may require new adaptation methods to handle.
Prior works have found that diverse and general pretraining data can cause negative transfer for current adaptation methods.
For example,~\citet{cole2021contrastive} shows that fine-tuning a model pretrained \textit{only} on the iNaturalist animal classification dataset provides better downstream performance than fine-tuning a model pretrained on iNaturalist along with 750K other images; similarly LegalBERT~\citep{chalkidis2020legal}, which is pretrained only on legal documents, improves over BERT~\citep{devlin2019bert}, which is trained on a much more diverse training set on the downstream task of text classification and sequence tagging in legal documents.
One approach to domain specialization is to include an intermediate adaptation step, where the foundation model continues training on unlabeled data from the specialized domain. For instance, this approach improves the downstream performance significantly for satellite images and specialized text topics~\citep{reed2021selfsupervised,gururangan2020dont}.
However, continual foundation model training may perform worse than re-training from scratch in certain domains such as legal documents~\citep{chalkidis2020legal}. Elucidating the scenarios in which continual training does or does not benefit performance is an important direction for future work.

\paragraph{Local model editing.} In some settings, it is useful to adapt a foundation model locally, meaning that the model's predictive distribution should be adapted only for a single input or a local neighborhood around a single input, without changing the model's behavior for unrelated inputs.
For example, when a foundation model produces an especially problematic mistranslation for a particular input phrase and target language, it is desirable to correct this mistranslation without affecting the model's behavior for unrelated phrases.
Past work has studied the problem of applying approximately localized updates to large neural networks through new pretraining objectives that enable easy editing with standard gradient descent \citep{sinitsin2020editable}, higher-order networks that predict parameter edits for an underlying model \citep{decao2021editing,mitchell2021model,hase2021language}, and constrained fine-tuning procedures \citep{zhu2020modifying}. However, existing methods vary in the reliability with which they can perform model edits without damaging global model performance. Furthermore, scaling these methods to massive foundation models is not straightforward due to their size and the computational cost of training objectives that require computing higher-order gradients \citep{sinitsin2020editable,decao2021editing,mitchell2021model}.

\paragraph{Applying constraints.} There are settings in which foundation models need to be adapted to satisfy privacy constraints. For instance, \citet{carlini2020extracting} demonstrated that existing foundation models are able to memorize sensitive information in the training data and can regurgitate such data when queried via standard APIs.
While this phenomenon calls for improved data curation, developing adaptation procedures which eliminate or reduce the influence of specific data examples on the trained model would be a complementary solution.
Improved adaptation strategies (along with better pretraining methods) in this direction will also benefit institutions working with foundation models under the General Data Protection Regulation (GDPR), as the mandate gives users the \emph{right to be forgotten}.
While research on the topic of machine unlearning ~\citep{bourtoule2019machine,cao2015towards} has started to gain traction, the problem has not yet been studied in depth for foundation models.
In addition, foundation models trained on less curated internet data have been shown to exhibit harmful biases targeting specific groups (\eg gender and racial bias)~\citep{bender2021,basta2019evaluating,kurita2019measuring,sheng-etal-2019-woman} and can produce toxic outputs~\citep{gehman-etal-2020-realtoxicityprompts} (\refsec{misuse}).
While strategies such as further fine-tuning the foundation model on carefully curated datasets (for potentially multiple generations)~\citep{solaiman2021process} or applying controllable generation techniques~\citep{keskar2019ctrl} have shown some success in mitigating harmful behavior, a framework for training equitable and safe foundation models (\refsec{fairness}) will likely require further research with a collective effort across the data collection, training, and adaptation phases as well as consultation with domain experts.

\subsubsection{A long-term goal for foundation model adaptation research}
To the extent that adaptation is concerned with efficiently integrating a model's existing knowledge with new data or objectives, a natural extension of adaptation is \textit{continual learning} \cite{mccloskey1989catastrophic,parisi2018continual} or continual adaptation. The ability to adapt a foundation model continually is desirable, whether to keep a model's knowledge continually up-to-date with world events or cultural developments, continually add data from completely new domains or modalities as they become available, or continually edit a model's memories to comply with privacy or legal constraints as a society's values or laws evolve. However, continual learning problems typically induce \textit{catastrophic forgetting} \cite{mccloskey1989catastrophic,Ratcliff1990ConnectionistMO,catastroph} in neural networks, where old tasks or data are rapidly forgotten as the training distribution changes.

We consider continual adaptation of a foundation model as a grand challenge for future foundation model adaptation research. Tackling this challenge requires closing the performance gap between a foundation model trained continuously on a non-stationary stream of data from different tasks, domains, or time periods and the same foundation model trained from i.i.d. data from the aggregate mixture \citep{lopezpaz2017gradient}. Innovations in both model architectures and training objectives are likely to be necessary in order to do so. For example, while memory mechanisms have long been speculated as key to successful continual learning \citep{french1999catastrophic} and have shown some promise for foundation models \citep{lewis2020retrieval,guu2020realm,borgeaud2021improving}, effectively leveraging updated memories remains a challenging problem \citep{zhang2021situatedqa}. In addition, techniques for localizing knowledge in a foundation model in order to make targeted parameter updates \citep{Dai2021KnowledgeNI} or learning such update rules \citep{decao2021editing,mitchell2021model} may help prevent forgetting, but repeated application of such updates still induces significant forgetting \citep{hase2021language}. Continual foundation model adaptation may also require new understanding of how the problem of catastrophic forgetting manifests at the scale of foundation models as well as developing new meta-learning techniques \cite{schmidhuber1987evolutionary,santoro2016meta,finn2017modelagnostic} to improve forward transfer of already-learned information to new settings.

Continually training on experiences gathered by deployed foundation models, or even training on aggregated experiences gathered by many different models, may speed the rate of progress of foundation model development, but incur risks of feedback loops and eroded alignment of model behavior and stakeholder values. Despite the aforementioned challenges, continual foundation model adaptation holds the promise of more rapidly responding to shifts in socio-cultural values, better leveraging existing knowledge to learn new concepts, lessening the environmental impact and increasing the accessibility of foundation models by eliminating the computational burden of training from scratch, and reducing the extent that previously-learned concepts must be re-learned due to forgetting.
\newsection
\hypertarget{evaluation}{\subsection{Evaluation}}
\label{sec:evaluation}
\sectionauthors{Rishi Bommasani, Kawin Ethayarajh, Omar Khattab}

\hypertarget{evaluation-introduction}{\subsubsection{Introduction}}
\label{sec:evaluation-introduction}
Evaluation gives context to machine learning models: it serves as a means for (1) tracking \textit{progress}\dash{}how do we we measure the performance of models and how do we design improved models (\refsec{modeling}); (2) \textit{understanding}\dash{}what behaviors do models exhibit (\refsec{interpretability}) and how do they perform on different slices of data (\refsec{robustness}); and (3) \textit{documentation}\dash{}how do we efficiently summarize model behavior and communicate this to diverse stakeholders. 
For foundation models, each of these purposes for evaluation are critical but the nature of foundation models introduces new challenges that are not generally encountered in other AI or ML settings: 
\begin{enumerate}
    \item Tracking progress requires relative comparison, but comparing foundation models is complicated by the fact that foundation models must be adapted (potentially in different ways) to perform tasks.
    \item Understanding requires specified in-advance knowledge (\eg~taxonomies) of what is being evaluated for, but foundation models acquire emergent skills (\eg~in-context learning) that will be difficult to anticipate in designing evaluations.
    \item Documentation requires clear desiderata to meaningfully inform decision-making, but foundation models can be adapted for myriad applications, which makes comprehensive documentation challenging.
\end{enumerate}
To orient the discussion of evaluating foundation models, we distinguish two classes of evaluation that arise from the abstraction of foundation models: \textit{intrinsic} evaluation of the foundation model, which is inherently divorced from a specific task due to the task-agnosticity of these models, and \textit{extrinsic} evaluation of task-specific models, which is necessarily dependent on both the foundation model and the adaptation mechanism.
Further, we recognize that due to the anticipated impact and scope of foundation models, a variety of stakeholders (\eg~foundation model providers and application developers, auditors and policymakers, practitioners and researchers) will require evaluation of both foundation models and task-specific derivatives, with these evaluations serving different purposes and involving different desiderata based on the stakeholder. 
With this in mind, standard paradigms for the evaluation of machine learning models are not designed explicitly for the setting of foundation models.
Therefore, we emphasize intrinsic evaluation (\refsec{evaluation-intrinsic}), the importance of adaptation in extrinsic evaluation (\refsec{evaluation-adaptation}), and evaluation design (\refsec{evaluation-design}) as clear steps towards an evaluation framework that is better suited to foundation models.
This discussion contributes to broader dialogue surrounding the role of evaluation of machine learning systems \citep[][\textit{inter alia}]{galliers1993, lipton2019, ribeiro2020beyond, linzen2020, kiela2021dynabench, milli2021, jacobs2021, bowman2021, dehgani2021, ma2021} and, given the complexities of evaluation, may benefit from drawing upon theories of measurement and evaluation that exist beyond machine learning \citep{messick1987, jackman2008, loevinger1957, messick1988, hand2010, brewer2014}.   

\hypertarget{evaluation-intrinsic}{\subsubsection{Intrinsic evaluation}}
\label{sec:evaluation-intrinsic}

Evaluation of machine learning systems has traditionally been grounded in \textit{tasks}, often ones that are envisioned as functions specifically useful for applications (\eg~translation, object recognition).
In contrast, since foundation models are intermediary assets that must be further adapted or specialized to perform useful tasks, the standard evaluation paradigm must be altered to facilitate the direct understanding and comparison of foundation models.

One approach is to evaluate foundation models in terms of the task associated with the training objective.
For example, a language model like GPT-3, which was trained by predicting the next word given the preceding context, may be evaluated based on the probabilities it assigns words given their preceding context in held-out test data (\ie~perplexity on language modelling benchmarks like LAMBADA \citep{paperno2016lambada}).
This approach has shown promise in NLP thus far, but we identify two fundamental limitations it exhibits.
First, relying on the training objective for evaluation lacks generality: foundation models trained using different incompatible objectives cannot be readily compared or understood in a consistent frame.
Second, evaluation in this way relies upon a \textit{proxy} relationship to be meaningful, \ie~measurements in terms of the training objective should correlate with other more meaningful and intelligible quantities (\eg~the quality of content generated via a foundation model).
While this proxy relationship has proven to be robust in the past in some contexts, it likely will break down when assessing more diverse capabilities of foundation models, their behavior in more diverse environments or domains, and considerations beyond in-domain accuracy (we discuss this more extensively in \refsec{evaluation-design}).
In light of these limitations, we anticipate that two approaches will need to be considered, offering complementary benefits.

\paragraph{Imputing intrinsic evaluation from broad extrinsic evaluation.}
One route towards evaluating foundation models is to adapt them to a wide range of tasks and measure the performance of the resulting task-specific models.
As the foundation model is the shared basis across all of these models, the performance in aggregate reflects on the nature, and quality, of this shared basis.
At present, many subareas of AI have begun to construct \textit{meta-benchmarks}, \ie~a single evaluation that consolidates individual evaluations across a number of different tasks or domains \citep{wang2019glue, wang2019superglue, hu2020xtreme, santurkar2020breeds, gehrmann-etal-2021-gem, hendrycks2021measuring, koh2021wilds, Tamkin2021DABS}.
Given the growing adoption of this paradigm and its established strengths, here we note why it is likely insufficient to fully satisfy the goals of evaluations with respect to foundation models.
Meta-benchmark evaluation requires adaptation (minimally to specialize the foundation model to each of the tasks in the meta-benchmark), which makes reasoning about the foundation model itself challenging given the addition process (\ie~adaptation) involved. 
Specifically, this complicates matters of progress, both in terms of tracking (\eg~is performance attributable to potent foundation models or well-designed adaption practices) and in terms of identifying improvements in the process used to learn foundation models (\eg~fundamental improvements in data selection (\refsec{data}), training objectives (\refsec{training}), and model architectures (\refsec{modeling}) may be difficult to identify by comparing the performance on a meta-benchmark between two foundation models).
In addition, this evaluation paradigm makes it difficult to understand or document properties and capabilities specific to the foundation model, which may make it unwieldy to convey to certain stakeholders (\eg SuperGLUE performance may not be sufficiently informative, or may be misleading, for policymakers) or use as grounds for anticipating their behavior for new tasks or domains.  

\paragraph{Direct evaluation of intrinsic properties.}
To complement the use of meta-benchmarks, we also argue for why measuring the properties (\eg~specific capabilities or biases) of foundations models directly is valuable, divorced from specific tasks.\footnote{Strictly speaking, these direct evaluations may still involve formulation as a task and foundation model specialization to perform the task, but the objective is more akin to probing (see \refsec{interpretability}) of trying to measure the foundation model as directly as possible.}
For example, we may endeavor to directly measure the linguistic capabilities of foundation models to identify syntactically valid and invalid sentences.
To motivate the value of this approach, we return to the purposes for evaluation.
Notably, articulating the presence and intensity of capabilities, skills, and biases identifies concrete areas for improvement (progress), elucidates the current potential (understanding), and expresses relevant aspects efficiently (documentation).
Such an approach also is in service of broadly comprehensible evaluation, \ie~evaluation that can be understood by both technical experts, non-technical experts (\eg~policymakers or social scientists) and the general purpose.
For example, characterizing the persuasive or rhetorical capabilities of these models may especially intuitive for internalizing their potential for disinformation and misuse (\refsec{misuse}) \citep{BuchananCSET2021}.

Direct evaluation of properties also serves as an important pathway towards better handling of the emergent properties of foundation models; to demonstrate this, we take in-context learning as a case study.
In particular, \citet{brown2020gpt3} not only demonstrated GPT-3's signature capability of robust in-context learning, but also were the first to specifically identify in-context learning as a specific way to adapt and interact with models (through their exploration of GPT-3).
Traditional task-based extrinsic evaluation does not provide a clear means by which in-context learning could have been identified; directly interacting with the foundation model appears to be necessary in this case.
More generally, while it appears inevitable that many unanticipated phenomena like in-context learning will be recognized through the unstructured or loosely structured exploration of these models and their capabilities, we believe new approaches to evaluation should be sought out that structure this exploration or, more ambitiously, suggest new properties that can then be more rigorously tested for.
Intrinsic evaluation may also lower the threshold for demonstrating the potential of foundation models; new approaches for foundation models may be sufficiently promising if they demonstrate improvements in intrinsic evaluation, even if they are not immediately accompanied by corresponding well-suited adaptation methods for eliciting these capabilities in extrinsic evaluation.

There is a significant open question of how intrinsic evaluation should be implemented; the mechanics of such evaluation are unclear.
We enumerate a few general principles and considerations that may help inform the design and execution of intrinsic evaluation.
\begin{enumerate}
    \item \textit{Inspiration from evaluation of humans.} 
    Many of the relevant properties, capabilities, and biases we are interested in for foundation models are also of interest for humans, which suggests that methods for measuring these properties in humans may prove to be instructive, or even directly translatable, for evaluating foundation models. 
    For example, psycholinguistic measures of human linguistic competencies can be modified to evaluate foundation model linguistic competencies \citep{levy2008, frank2013, linzen2016assessing, ettinger2016, marvin2018, van-schijndel2018, futrell2019, prasad2019, ettinger2020} or psychological measures of human social biases can be modified to evaluate foundation model social biases \citep{greenwald1998, caliskan2017, may2019, guo2020}.
    \item \textit{Human-in-the-loop evaluation.}
    Human-in-the-loop evaluation may prove to be critical to provide a more exploratory means for understanding foundation models, including assessing their generative or interactive capabilities.
    In particular, human interaction with foundation models directly may better identify their emergent capabilities and limitations and direct auditing of foundation models \citep[\eg][\refsec{ethics}]{raji2019} may advances goals for documentation and transparency.
    \item \textit{Validity of intrinsic measures.}
    While intrinsic measures allow for direct measurement at the source, \ie measurement and evaluation of the properties of a foundation model independent of adaptation and specific tasks, they pose challenges for building trust in the validity \cite{messick1987, messick1988} of the evaluation. 
    In particular, extrinsic evaluation outcomes may also be important in validating intrinsic measure design, \eg the \textit{predictive validity} of intrinsic measures (\ie their ability to (statistically) predicted related downstream outcomes) may prove to be a central criterion. 
\end{enumerate}

\hypertarget{evaluation-adaptation}{\subsubsection{Extrinsic evaluation and adaptation}}
\label{sec:evaluation-adaptation}

Evaluating task-specific models has historically involved reporting the performance (generally meaning the accuracy) of the model on a specific held-out test set.
While this paradigm may partially suffice to understand or document a model, it often amounts to unfair comparisons between task-specific models produced with different (and, potentially, unequal) resources, making it difficult to gauge how much progress has been made.
The concern of unfair comparisons is exacerbated in the foundation model regime: different foundation models (\eg~BERT and GPT-3) may form the foundation for different task-specific models, and these foundation models may involve vastly different amounts of training data and computation.

To account for the resources required to achieve specific levels of performance, \citet{linzen2020} argues that (pre)\textit{training resources} should be acknowledged and tracked in evaluation. 
We believe this is a scientifically principled proposal; comparing different approaches for training foundation models without accounting for training resources is likely to be misleading.
However, given that the process for creating foundation models is especially expensive (\eg~requiring significant human and financial capital), and often governed by societal factors (\eg~commercial incentives) in addition to scientific factors, it may be the case that the foundation models in practice will vary greatly in the training resources afforded, making controlled comparison difficult.
Here, we consider an alternative, which may be more pervasively viable, to partially account for the resources involved to complement the proposal of \citet{linzen2020}.
In particular, we consider why extrinsic evaluation should acknowledge \textit{adaptation resources}, which is critical for ensuring that extrinsic evaluation is able to identify the most performant adaptation methods (which intrinsic evaluation, fundamentally, cannot do). 
We draw attention to the fact that adaptation resources often are construed as the data used to adapt models, but additional resources \citep[\eg data used to choose adaptation methods;][]{perez2021true} and constraints (\eg the level of access required to adapt the foundation model; see \refsec{adaptation} and \refsec{ethics} for further discussion) should also be accounted for.

\paragraph{Accounting for adaptation resources.}
Accounting for the resources expended to adapt foundation models for specific tasks requires a complete understanding of what resources or constraints are used for different adaptation methods, \ie~evaluations that endeavor to account for these resources must evolve alongside developments in what resources are used in adaptation (\refsec{adaptation}).
In existing task-specific evaluations, most evaluations specify the amount of data that can be used to adapt a (foundation) model to the task. 
However, \citet{perez2021true} identify a key nuance here that has been discounted in past work, in that this should encapsulate all data used to inform adaptation, \ie~both the data used to adapt the foundation model and the data used to choose the adaptation method.
Further, in the foundation model regime, the notion of \textit{access requirements} for different adaptation methods is also a new consideration that should be factored into evaluation.
Concretely, some adaptation methods may generally outperform others but may require greater ability to access or modify the foundation model compared to others (\eg~fine-tuning requires foundation model gradients to modify a foundation model, whereas prompting may only require blackbox access in specifying inputs). 

Accounting for the resources involved in adaptation enriches what conclusions can be reasonably drawn from evaluation of task-specific models.
At present, task-specific evaluation may provide sufficient clarity for certain types of understanding or documentation of particular task-specific \textit{artifacts} (\ie~the exact models being evaluated) but do not provide clear signal for how different adaptation methods perform and how to select a specific adaptation method in a given context.
In contrast, by accounting for the resources and access requirements involved in adaptation, evaluation better enables research to identify which adaptation methods or \textit{processes} make best use of the resources provided, \ie~signal is offered not just for the specific artifacts being evaluated but the more general processes by which they were derived.
The proposed evaluation protocol, therefore, clearly works towards identifying which adaptation methods should be used; we note that all of these conclusions should always be taken as specific to a given foundation model, as evaluation in this form does not provide sufficient evidence to conclude an adaptation method is uniformly the best across foundation models.\footnote{Current results, instead, suggest that different adaptation methods are better-suited to different types of foundation models and training objectives \citep{liu2021prompt,lester2021power}.}

\hypertarget{evaluation-design}{\subsubsection{Evaluation design}}
\label{sec:evaluation-design}

In theory, the goal of evaluation is to measure and characterize various theoretical constructs (\eg~accuracy, robustness (\refsec{robustness}), fairness (\refsec{fairness}), efficiency (\refsec{systems}), environmental impact (\refsec{environment})) in service of various purposes (\ie~progress, understanding, documentation).
However, in practice, the utility of evaluation will be determined by how evaluations are designed and executed.
For example, automated measurements of the generative capabilities of foundation models (\eg~their factual correctness) may poorly capture the nature of these qualities and, instead, human-in-the-loop evaluation may better contextualize these capabilities. 

In considering the evaluation design we envision for foundation models and their adapted derivatives, we begin with the mechanics of evaluation.
Traditionally, the evaluation of machine learning models has involved a large training set that is used to learn the model, an optional validation set that is used to set hyperparameters, and a test set to evaluate the generalization of the learned model to held-out data \citep{bishop2006}.
As a result, creating benchmarks to evaluate models has historically required large amounts of data, most of which is allocated towards training, which complicates the design of certain diagnostic or nuanced evaluations when data is scarce or expensive to attain \citep{rogers2020, rogers2021}. 
In contrast, because the benefits of foundation models will often coincide with the sample efficiency of adaptation (\ie~few-shot or zero-shot capabilities) and the diversity of possible applications, we instead envision a regime where benchmarks for individual tasks are much smaller (since far less data needs to be provided as ``training", \ie~adaptation, data) and are far more diverse (both to capture various capabilities in intrinsic evaluation and more strongly ground evaluation in ecologically valid ways \citep{brofenbrenner1977, vries2020ecological} during extrinsic evaluation).
This suggests that the nature of foundation models may cause a shift in nature of benchmarks (and the mentality of those constructing benchmarks), de-emphasizing quantity as a key priority in benchmarks as opposed to quality and diversity.
The NLP community has begun to see the beginnings of such a regime with expansive and diverse benchmarks like BIG-Bench\footnote{\url{https://github.com/google/BIG-bench}} and FLEX \citep{bragg2021}; this paradigm lowers the barrier for benchmark design, thereby enabling the broader community to partake in evaluation design.\footnote{Traditionally, the design of benchmarks like ImageNet \citep{deng2009imagenet} and SQuAD \citep{rajpurkar2016squad} has been conducted by high-resourced research labs that can afford to pay for the creation of these datasets through crowdsourcing \citep{rogers2020}.} 

Alongside the mechanics of evaluation, the presentation of and interface to the evaluation results informs how these results will be used inform decision-making (\eg~new modelling approaches, model selection, auditing). 
Leaderboards have become the de facto paradigm in machine learning, whereby models are ranked by a specific and singular criterion (generally a form of accuracy).
This approach has generally led to significant and rapid progress in system quality over time \citep[\eg][]{wang2019superglue}, but significant concerns have been raised of whether this yields more general improvements \citep[\eg][]{linzen2020, bowman2021}.\footnote{We note the connection to Strathern's Law \citep{strathern1997} (sometimes referred to as Goodhart's Law \citep{goodhart1984}): ``When a measure becomes a target, it ceases to be a good measure."}
As is true for all machine learning models, it is rarely the case that the desiderata for foundation models and their derivatives will be singular; instead, we anticipate the breadth of their application and societal impact necessitates heightened consideration of criteria beyond accuracy (\eg~robustness, fairness, efficiency and environmental impact).
To this end, we note that evaluation of foundation models should report measurements across these diverse fronts; existing benchmarks are increasingly designed to reflect more than just accuracy (\eg~robustness \citep{koh2021wilds, goel2021robustnessgym}, fairness \citep{stereoset, nangia2020}, efficiency and environmental impact \citep{coleman2017dawnbench}).
Further, we note that if the reporting of performance across this different categories is done in the form of a leaderboard, mechanisms to disambiguate potential trade-offs (to induce a ranking) will be especially necessary \citep{ethayarajh2020}.
In particular, since different stakeholders will have different preferences (\eg~the weight they ascribe to different properties) and values \citep{birhane2020}, leaderboard design should allow stakeholders to interact and manipulate how the ranking is done to align with their values; \citet{ma2021} presents an early attempt to enable this by comparing the utility of models using an economic framing based on a user's specified utility function. 

\hypertarget{evaluation-takeaways}{\subsubsection{Takeaways}}
\label{sec:evaluation-takeaways}
Evaluation performs several roles (\ie~progress, understanding, documentation) that are vital for all machine learning paradigms, including the foundation model paradigm.
Foundation models introduce new challenges for existing evaluation frameworks; designing evaluations that directly target the foundation model regime will better serve not only the multiple purposes of evaluation, but also the myriad of stakeholders involved.
\begin{enumerate}
    \item 
    While machine learning evaluation traditionally has considered task-specific models, evaluating foundation models involves engaging with the fact that these models are not specific to a task. 
    Evaluation of these models likely will involve integrating two complementary approaches: (a) imputing the properties of foundation models from broad evaluation of task-specific derivatives and (b) direct measurement of these properties in foundation models.  
    \item
    Existing evaluation frameworks often do not account for the resources required to create the models being evaluated, leading to \textit{unfair comparisons}.
    For foundation models, we discuss an evaluation paradigm that emphasizes accounting for \textit{adaptation resources} (\eg~all data used in adaptation, access requirements for the foundation model), which appears to lead to more informative evaluations that better shape how adaptation is conducted.
    \item 
    Existing evaluation design often is limited in the diversity of metrics considered and requires large adaptation datasets.
    For foundation models, we echo growing calls for evaluation to consider a broader range of desiderata (\eg~robustness, fairness, efficiency, environmental impact) to capture the wide range of stakeholder values/preferences, as well highlight how the sample efficiency of adapting adaption models may allow for more diverse evaluations by re-allocating resources involved in designing evaluations.
\end{enumerate}

\newsection
\hypertarget{systems}{\subsection{Systems}}
\label{sec:systems}
\sectionauthors{Deepak Narayanan, Trevor Gale, Keshav Santhanam, Omar Khattab, Tianyi Zhang, Matei Zaharia}

\begin{figure}[!ht]
\centering
\includegraphics[width=0.7\linewidth]{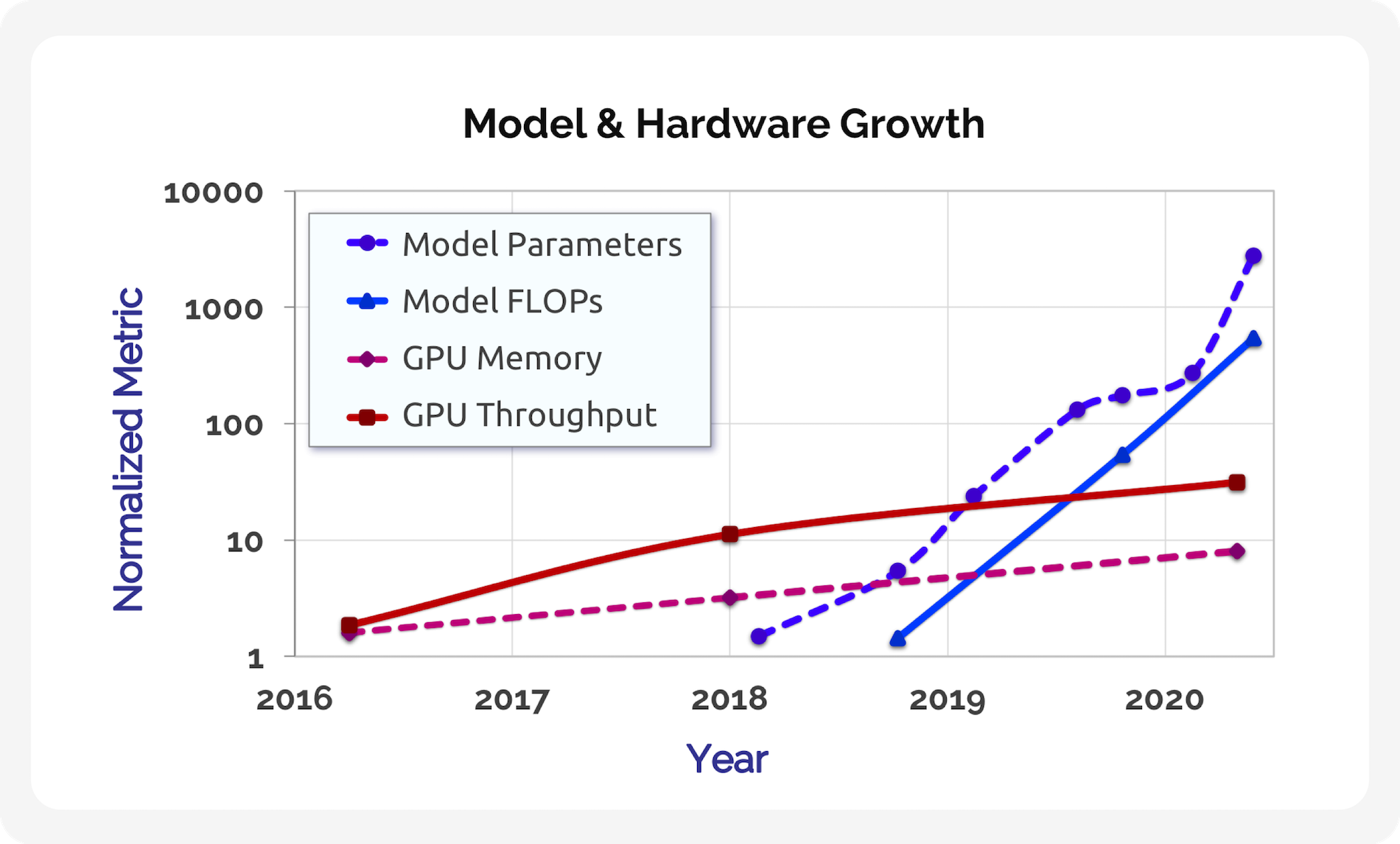}
\caption{\label{fig:systems} Plot showing the growth of number of parameters and number of training operations (FLOPs) of transformer-based language models (shown in blue), and memory capacity and peak device throughput of NVIDIA P100, V100, and A100 GPUs (shown in red) with time. The rate of growth (slope of each line) of state-of-the-art language models (roughly $10\times$ a year) far exceeds the rate of increase in computational capacity of hardware (roughly $10\times$ in four years), motivating the need for parallelism across a large number of accelerators and co-design of algorithms, models, software, and hardware to drive further progress. Number of parameters and number of training operations are obtained from relevant papers~\citep{brown2020gpt3}, and memory capacities and peak throughputs are obtained from GPU specification sheets.}
\end{figure}

Computer systems are one of the largest bottlenecks to developing foundation models. Foundation models are frequently too large to fit in the main memory of a single accelerator (\eg GPU) and require an immense amount of computation to train (\eg $>1000$ petaFLOP/s-days for GPT-3~\citep{brown2020gpt3}). Additionally, these models will likely get larger over time: for instance, the compute and memory requirements of state-of-the-art language models have grown by three orders of magnitude in the last three years, and are projected to continue growing far faster than hardware capabilities (\reffig{systems}). Once trained, these large models are expensive to perform inference with and difficult to debug, monitor, and maintain in production applications. We believe that further advances in the performance and usability of foundation models will require careful co-design across algorithms, models, software, and hardware systems, as well as new interfaces for programming and deploying ML applications. In this section, we discuss the key computer systems challenges in developing and productionizing large-scale foundation models.

\hypertarget{systems-co-design}{\subsubsection{Improving performance through co-design}} 
\label{sec:systems-co-design}

Today, training large-scale foundation models~\citep{brown2020gpt3, rae2021scaling, megatron_530b, gpt-j} can often require custom software systems such as Megatron, DeepSpeed, or Mesh Transformer JAX~\citep{shoeybi2019megatronlm, rasley2020deepspeed, mesh-transformer-jax}, built on top of standard frameworks like PyTorch, TensorFlow, and JAX~\citep{paszke2019pytorch, abadi2016tensorflow, jax2018github}. These software systems rely on a number of innovations across the stack to train models efficiently at scale: new parallelization dimensions such as pipeline parallelism~\citep{huang2019gpipe, narayanan2019pipedream} that limit communication while keeping devices busy, state-sharding optimizers to reduce memory usage~\citep{rajbhandari2020zero}, just-in-time (JIT) compilers to optimize the computation graph~\citep{pytorchjit}, and optimized libraries like cuDNN and NCCL~\citep{nccl}. Megatron and DeepSpeed are efficient to a particular scale; for example, Megatron can extract up to 52\% of the theoretical peak throughput of modern hardware~ with approximately 3000 GPUs on a model with a trillion parameters~\citep{narayanan2021efficient}. However, scaling to larger models with more GPUs still is challenging, since existing parallelization strategies break down at larger GPU counts. Data parallelism is limited by the batch size~\citep{li2020pytorch}, pipeline parallelism by the number of layers in the model~\citep{huang2019gpipe, narayanan2019pipedream}, and tensor model parallelism by the number of GPUs in a single server~\citep{shoeybi2019megatronlm}.

While we will continue to realize performance gains from new hardware, growth in the resource requirements of large models far outstrips generational hardware improvements~\citep{brown2020gpt3}. To facilitate the next major leap in model capacity and to democratize the advances in model quality, it will be increasingly critical to \emph{co-design} training algorithms, models, software, and hardware, because many of the avenues to dramatically increase performance \emph{alter the semantics} of the training computation. For example, executing operations in lower precision (such as \texttt{fp16}) can help increase throughput on modern hardware (\eg the V100 and A100 GPUs have dedicated tensor core units for lower-precision matrix multiplication), but also affect the numerics of the optimization procedure~\citep{micikevicius2017mixed}. Similarly, exploiting weight sparsity can significantly improve training and inference times~\citep{Elsen_2020_CVPR, gale2020sparse} by only performing mathematical operations on the non-zeros in the model, but requires different training algorithms~\citep{jayakumar2021top, pmlr-v119-evci20a, dettmers2019sparse}. Other examples of co-design include model architectures that map more efficiently to hardware~\citep{pmlr-v97-so19a, child2019generating, wang2020linformer, lee2021fnet, kitaev2020reformer,longformer, tay2020efficient, ren2021combiner}, efficient optimizers~\citep{anil2020scalable, shazeer2018adafactor}, novel tokenization alternatives~\citep{xue2021byt5, tay2021charformer}, specially architected hardware training platforms~\citep{jouppi2017datacenter, mudigere2021high, selene}, and distributed parallelization strategies with relaxed weight update semantics~\citep{narayanan2019pipedream, narayanan2021memory}.

\paragraph{Case study: efficient knowledge representation.} As a concrete case study of successful co-design, retrieval-based models such as REALM, RAG, ColBERT-QA, and RETRO~\citep{guu2020realm,lewis2020retrieval,Khattab-etal:2020:OpenQA,borgeaud2021improving} take a different approach to model design than simply increasing the number of model parameters. Instead of trying to accumulate \emph{implicit knowledge} from ever-larger datasets directly into a DNN model with billions of parameters (like GPT-3), retrieval-based models store knowledge \emph{outside} the model parameters in the form of text passages, capturing knowledge within the passages with dense vector representations. These models then use scalable top-$k$ search mechanisms to extract knowledge pertinent to each input, while keeping the DNN model itself small (\refsec{modeling-memory}). This design improves computational efficiency as well as maintainability of the model in production: for example, developers can update the knowledge of the model just by replacing a text passage, without needing to retrain a large DNN.

Retrieval-based models have achieved promising initial results by leveraging several new cross-functional ideas, including backpropagating the loss through the retriever during training~\citep{guu2020realm} (which requires approximating the gradient through a knowledge store consisting of millions of passages) and modeling fine-grained interactions between queries and passages~\citep{khattab2020colbert,Khattab-etal:2020:OpenQA} (which requires decomposing the computation into vector-level nearest-neighbor search operations). These techniques allow retrieval-based models to be accurate and efficient, but demand functionality not readily supported by popular ML frameworks and nearest-neighbor indexes  such as FAISS~\citep{johnson2019billion}.

\subsubsection{Automated optimization}

Another important challenge in systems is to \emph{automate} the application of optimizations that straddle algorithms, models, software, and hardware. While many optimizations and parallelization strategies are complementary, identifying the most effective combination of optimizations is challenging since the joint search space grows combinatorially and optimizations interact in non-trivial ways~\citep{narayanan2021efficient}. Foundation models heighten the need for automated optimization as manual experimentation is extremely expensive and time-consuming at the scale of thousands of GPUs.

Recent work in this area has focused on systems targeting semantics-preserving optimizations. In particular, systems have been proposed to automatically discover mathematically-equivalent graph substitutions~\citep{jia2019optimizing, wang2021pet}, facilitate the distributed execution of operator graphs through both high-level APIs and low-level compilers~\citep{rasley2020deepspeed, fairscale, jax2018github, shazeer2018mesh, lepikhin2020gshard}, and automate the selection of hybrid distribution strategies~\citep{jia2018beyond, santhanam2021distir}. These systems have helped deploy many foundation models in industry~\citep{fedus2021switch, m2m100, turingnlg}.

Unfortunately, automated optimization becomes much harder when composing semantics-altering optimizations (\refsec{systems-co-design}), as it is often unclear how to jointly model the statistical impacts  of these techniques (\eg~how many training iterations are needed to reach a specific accuracy?). We will therefore need new software tools, libraries, and compilers to automatically identify compositions of optimizations that target comprehensive metrics like time-to-accuracy~\citep{coleman2017dawnbench, mattson2020mlperf}. Building such tools will require tight collaboration between systems and machine learning experts.

\subsubsection{Execution and programming models}

The unique multi-task nature of foundation models provides an opportunity to amortize training and inference costs over many applications. In particular, paradigms such as adaptation mean more sharing across model instances~\citep{a_call_to_build_models}. For example, two models prefix-tuned~\citep{li2021prefix} from the same pretrained model can share the same model ``stem,'' reducing the storage footprint (the shared stem only needs to be stored once), while also making it possible for execution to be shared and batched across the prefix-tuned models~\citep{shen2019nexus, narayanan2018accelerating}. Consequently, the specific adaptation mechanism used informs system optimization (\refsec{adaptation}).

It is an open question as to what programming interface should be used to specify that various adapted models are derived from the same pretrained model (\eg~models $Y$ and $Z$ are derived from the same pretrained model $X$), or that various components of two models share parameters (\eg~two models $A$ and $B$ share the same stem up to layer $i$). Ludwig~\citep{molino2019ludwig} and PyTorch's \texttt{Module} offer easy ways to compose functionality within a model, but no system today supports cross-model dependencies. Giving users the opportunity to provide annotations will allow training and inference systems to optimize and orchestrate computation more efficiently; without such annotations, systems will not have visibility into what computation and parameters can be shared across model instances. A model's ``adaptation history'' (what models is this particular model adapted from) can also be used for debugging: an adapted model's errors on particular types of inputs could originate from the pretrained model, pointing to issues in the pretraining process versus adaptation process. Frameworks like PyTorch, as well as software libraries for training foundation models such as HuggingFace Transformers~\citep{wolf2020transformers}, do not allow for fine-grained lineage information across entire model instances to be specified.

Building and maintaining a cluster of thousands of accelerators also requires tremendous effort. New training paradigms like Learning@Home~\citep{Ryabinin2020Learninghome, Diskin2021collab} explore leveraging volunteer compute over the internet to train foundation models collaboratively. Such fundamentally new execution models can decrease the cost of training for any one entity, but require collaboration across a number of different areas like security (to ensure that a malicious volunteer cannot significantly alter the training process), distributed systems (to deal with fault tolerance issues as volunteers drop), and crowdsourcing.

\subsubsection{Productionization of foundation models}

As the community continues to push the capabilities of foundation models, realizing their potential will require addressing the challenges associated with deploying these resource-intensive models in production. These challenges include performing model inference with tight latency targets, and ensuring that models and data are monitored in an automated way.

For applications with strict cost and latency constraints, model compression techniques like distillation~\citep{hinton2015distilling, li2020train, sanh2019distilbert}, quantization~\citep{polino2018model, gholami2021survey, zhou2018adaptive}, pruning~\citep{lecun1990optimal, gordon2020compressing, mccarley2019structured, wang2019structured, sajjad2020effect}, and sparsity~\citep{gale2020sparse, Elsen_2020_CVPR} could aid deployment by transforming larger models to obtain desired inference-time properties. These techniques were originally intended for smaller models (\eg BERT-L) in low-memory environments (\eg mobile phones), but are now necessary to handle the extreme scale of modern foundation models in datacenter deployments. Parallelization techniques like tensor model parallelism~\citep{shoeybi2019megatronlm}, traditionally used for training, might also be useful to reduce inference latency, and also provide additional memory capacity across GPUs to fit the model's parameters.

In addition to these practical constraints, increases in the size and complexity of foundation models and the datasets used to train them pose new challenges to model and dataset lifecycle management. Since models with a large number of parameters are hard to manually inspect by humans, we need better systems for automated dataset curation (\refsec{data}) and model quality assurance. Techniques like behavioral testing~\citep{ribeiro2020beyond} and model assertions~\citep{kang2020model} facilitate easier model maintenance in production by providing analogs to unit tests, runtime monitoring (in the form of test-time assertions), and continuous model improvement (as new inputs come in) for models deployed in end applications. These tools can help address issues of fairness and bias (\refsec{fairness}), and reduce model mispredictions.

\newsection
\hypertarget{data}{\subsection{Data}}
\label{sec:data}
\sectionauthors{Laurel Orr, Simran Arora, Karan Goel, Avanika Narayan, Michael Zhang, Christopher Ré}

Foundation models signal a paradigm shift where increasingly massive quantities of data are being ``fed'' to these models for improved adaptation performance~\citep{devlin2019bert,radford2021learning,tolstikhin2021mlpmixer} with the overarching rule-of-thumb being "the more data the better"~\citep{kaplan2020}. As previous sections have mentioned, this focus on data curation has raised concerns around the foundation model data lifecycle including (1) managing the data at such a large scale (\refsec{introduction}), (2) integrating data across new modalities (\refsec{robotics}, \refsec{healthcare}), (3) reasoning over licensing and governance regulations\dash{}especially when considering the massing web-crawls used in foundation models training\dash{}(\refsec{healthcare}, \refsec{legality}), and (4) understanding the data quality (\refsec{evaluation}).

While foundation models add new and difficult facets to these challenges, we see parallels between these issues and core challenges in communities such as data management and data analytics as well as industrial ML pipelines. For example, data management has long studied scalable declarative systems for data analysis, versioning, provenance, and integration\textemdash addressing challenges (1) and (2)~\citep{zaharia2012resilient, cudre2009demonstration, stonebraker2013voltdb, Stonebraker2018dataintegration, hellerstein2005readings}. Industry has pipelines dealing with challenge (3) to manage diverse data licenses and help mitigate data violations. There is an entire ecosystem of research and systems tackling challenge (4) to support interactive data analytics and visualization~\citep{hohman2020understanding}.\footnote{VIS, CHI, SIGGRAPH are a few communities that research interactive data analytics methods and systems. Software systems and libraries such as Pandas, Matplotlib, and Seaborn also aid users in interactive exploration.} While these solutions are not necessarily "foundation model-ready", we believe a path to better management of the foundation model data lifecycle should take inspiration from these existing systems.

In this section, we address managing the foundation model data lifecycle. We first outline four desiderata including data management at scale, support for heterogenous data sources, data governance, and data quality monitoring. We then envision how all of these requirements can be integrated into a holistic data management solution called a {\em data hub}. The data hub is simply a data management toolkit that can be used by the private or public sectors to better support the interactive management of the foundation model data lifecycle.

\hypertarget{data-desiderata}{\subsubsection{Data Management Desiderata}}
\label{sec:data-desiderata}

Current practices in foundation model development are generally ad-hoc across the entire lifecycle from data curation and data documentation to model monitoring and patching~\citep{gebru2018datasheets, bandy2021addressing, bender2018data}. Research in the data management community has shown that well-defined data management platforms facilitate ML model development at scale through data ingestion, data versioning, data provenance, efficient analysis, and model monitoring~\citep{hellerstein2005readings, agrawal2019cloudy, vartak2016modeldb, ikeda2010panda}.\footnote{Feature stores like Michelangelo also support end-to-end ML model building \url{https://eng.uber.com/michelangelo-machine-learning-platform/}.} Taking inspiration from the data management community, we consider core desiderata when building a holistic data management platform for foundation models. 
\begin{enumerate}
    \item \textbf{Scalability.} Foundation models are being trained on increasingly massive quantities of data~\citep{kaplan2020} with the WuDao 2.0 model being trained on 4.9 TB of multi-modal data.\footnote{\url{https://www.scmp.com/tech/tech-war/article/3135764/us-china-tech-war-beijing-funded-ai-researchers-surpass-google-and}} This scale is expected to increase as most recent models are trained largely on public facing datasets. Public data represents an extremely small fraction of data compared to the petabytes of business and personal data collected every day and used in industrial foundation model pipelines~\citep{marr2017}. There is therefore a growing need for highly scalable techniques that can handle multi-modal foundation model datasets.

    \item \textbf{Data integration.} 
    Recent work using foundation models demonstrates that leveraging integrated structured and unstructured data can help models better generalize to rare concepts~\citep{orr2020bootleg} and improve factual knowledge recall~\citep{orr2020bootleg, logeswaran2019entdesc, Zhang2019ERNIEEL, Peters2019KnowledgeEC,perner2020ebert}. Despite these recent successes, integrating datasets for foundation models remains a challenge. Many works use unstructured text data with structured entity knowledge or image data~\citep{antol2015vqa}. There is a growing need to integrate datasets across diverse modalities such as text, video, eye-tracking~\citep{hollenstein-etal-2020-zuco}, and robotic simulations~\citep{lynchlanguage} (see \refsec{robotics}). We need data-integration solutions that can be applied at an industrial scale to multiple modalities and to multiple domains, such as government, business, and science.
    
    \item \textbf{Privacy and governance controls.} The training data used for foundation models may risk the violation of the privacy of data subjects; their data may be disclosed, collected, or used without their consent~\citep{jo2020lessons} or outside the context for which consent was originally given. 
    The issue of consent and use is especially relevant for foundation models where downstream applications cannot always be anticipated.
    As explained in \refsec{legality}, these issues are compounded with the prevalence of web scraped datasets for foundation model training. As there are still open legal questions about how web-crawled data will be governed and copyrighted,\footnote{These issues have recently come to bear by the debate surrounding the use of GitHub data in Copilot's Codex tool to help developers code \url{https://www.pwvconsultants.com/blog/questions-around-bias-legalities-in-githubs-copilot/}} the consequences of using web data remain unclear to foundation model providers in the public and private sector. We need tooling to help foundation model providers adapt to emerging regulations and guidelines to ensure safe and responsible data management.
    
    \item \textbf{Understanding data quality.} Data quality impacts model performance~\citep{lee2021deduplicating}; however, toolkits or methods to systematically and scalably understand the training data and relevant data subsets are still in their infancy. The data creation process can be messy, and the data can contain different types of biases~\citep{blodgett_language_2020, bender2021} (see \refsec{fairness}) and consist of poisoned, false, or duplicated information~\citep{chang2020adversarial, carlini2021poisoning, BuchananCSET2021, lee2021deduplicating}. Data is also continuously updated and refined~\citep{kiela2021dynabench} and may have emergent entities~\citep{fetahu2015much}, distribution shift~\citep{chen2021mandoline}, and concept meaning shift~\citep{kenter2015ad, lazaridou2021pitfalls}. Further, once deployed, foundation models may present undesirable behavior on critical, fine-grained sub-populations of data that foundation model providers need to detect and mitigate~\citep{goel2021robustnessgym, hohman2018visual, re2019overton, oakden2019hidden}. We need toolkits that can detect and potentially mitigate different types of undesirable data to improve model performance in an interactive and iterative fashion. Such toolkits also need to adapt to the dynamical nature of training data. 
\end{enumerate}

\hypertarget{data-solutions}{\subsubsection{Data Hub Solution}}
\label{sec:data-solutions}
Pulling on years of work from data management, data science, and data analytics, we envision a foundation model lifecycle data management solution, which we term a \textit{data hub}. While examples of ML-focused data hubs\footnote{Some public data hubs include: \url{https://data.world/}, \url{https://dataverse.harvard.edu/dataverse/harvard}, \url{https://datacommons.org/}, \url{https://www.data.gov/}, \url{https://www.kaggle.com/}, \url{https://huggingface.co/datasets}, \url{https://www.ldc.upenn.edu/}} as well as more traditional data management systems exist,\footnote{Some traditional data management systems for foundation models include: \url{https://aws.amazon.com/big-data/datalakes-and-analytics/}, \url{https://eng.uber.com/michelangelo-machine-learning-platform/}, \url{https://kafka.apache.org/}} they either (1) do not treat data integration as a first class primitive, (2) do not natively support the end-to-end lifecycle with model predictions, or (3) do not allow for interaction-driven data curation and refinement, where foundation model providers can dynamically explore and update possible datasets subject to access control guidelines. We now discuss how the data hub addresses the four desiderata.

\paragraph{Data scale.} To address the management at scale challenge, the data hub will need standard data management solutions~\citep{armbrust2009above} such as infrastructure to store and maintain large-scale datasets as they change over time and scalable interfaces to query, select, and filter datasets. The hub should support heterogenous compute as well as cloud infrastructure to support scalable solutions in different environments. 

\paragraph{Data integration.}
The hub should incorporate data integration as a first class citizen. It will need advanced data integration solutions~\citep{Stonebraker2018dataintegration, abiteboul1997semistructureddata, dong2020multi, rekatsinas2017holoclean}\footnote{\url{https://www.tamr.com/}} to allow for the merging of structured and unstructured knowledge across modalities and domains. Further, this implies the hub will need to support storing and querying over heterogeneous datasets and sources.

\paragraph{Access control.} Considering the access controls of the hub, the hub will need to support diverse documentation, \eg dataset sheets~\citep{gebru2018datasheets} or data statements~\citep{bender2018data}, to allow data curators to reflect on their processes and be transparent about the intended use cases, potential biases, and limitations of their dataset. The data hub will need to decide which documentation is required for data to be uploaded (\eg the data source and data description) and which information is recommended (\eg what tasks the data could be used for). Furthermore, documentation may need to be updated as datasets evolve~\citep{goel2021robustnessgym}.

Data sources are often associated with licenses, and the hub will need to integrate different sources with different legal concerns and conditions~\citep{masur2018datalicensing}.\footnote{\url{https://content.next.westlaw.com/4-532-4243}} Further, certain datasets have legal guidelines to protect the privacy of the data subjects. The hub will need methods to ensure a dataset does not release personally identifiable information (PII),\footnote{\url{https://www.justice.gov/opcl/privacy-act-1974}} that the aggregation of anonymized or de-identified data does not release PII,\footnote{\url{http://www2.ed.gov/policy/gen/guid/fpco/ferpa/library/georgialtr.html}} and that the data subjects have given informed consent for their data to be disseminated.\footnote{\url{https://www.dhs.gov/sites/default/files/publications/privacy-policy-guidance-memorandum-2008-01.pdf}}

Pulling on ideas from data integration~\citep{rekatsinas2017slimfast}, the hub should support mechanisms to enable efficient and safe maintenance and sharing of data resources. Especially as the legality of certain public datasets (\eg web dumps) are still being decided (\refsec{legality}), the hub critically needs tooling to help identify licensing violations and mitigate the impact of any governance violation. As certain violations will likely relate to model behavior, we need systems to support better understanding of model behavior, as we describe next.

\paragraph{Data quality tooling.} Drawing on the field of data analysis and exploration, as users interactively select, filter, and refine the data to use for training or adaptation, the hub will need tools to quickly understand a user's current dataset and its impact on model behavior~\citep{hohman2020understanding}.\footnote{Examples of data-focused interactive toolkits include \url{https://www.tableau.com/} and \url{https://www.paxata.com/}.} Furthermore, these systems can allow end-to-end foundation model monitoring by incorporating model performance through recent work on slice (sub-population) finding~\citep{chung2019slice}, model validation on relevant subsets~\citep{goel2021robustnessgym, ribeiro2020beyond}, and data valuation~\citep{ghorbani2019data}. Recent works also present methods that use the model to detect which subpopulations of data contribute the most to a given output to further aid model debugging~\citep{keskar2019ctrl}.

Once users can monitor model behavior\textemdash especially on rare, yet critical sub-populations\textemdash, the hub should provide methods and guidance for users to maintain models by correcting model errors. Although ``model patching''~\citep{goel2020model} is still an open problem, the work of \citep{orr2020bootleg} provides a first description of using data engineering to maintain a production self-supervised system that corrected for undesirable behavior through changes to the data, not model. We believe the data hub will need to support interfaces for users to inject targeted data modifications for model maintenance.

We also acknowledge that data curation and exploration are not performed in isolation, and believe the data hub should support a community around sharing useful metrics and analysis pipelines. Inspired by similar community sharing platforms like Hugging Face’s ModelHub\footnote{\url{https://huggingface.co/models}} or Tableau Public’s visualization sharing platform,\footnote{\url{https://public.tableau.com/en-us/s/about}} we want users to share insights about foundation model training data.

\paragraph{Open questions.} Although our described data hub is inspired by existing toolkits and solutions, we do not believe they are all ready for the challenges of foundation models. In particular, some open questions revolving around designing a data hub are:
\begin{itemize}
    \item How should we support data versioning so datasets can be updated while maintaining old versions for reproducibility~\citep{agrawal2019cloudy}? Once models are deployed and error buckets are identified, datasets may need to be updated to include more examples from these error buckets. How should these new, targeted examples be collected?
    \item As described in \refsec{training}, we imagine fewer models will be trained from scratch and more will be fine-tuned. How do we support provenance or lineage information to understand where the original data came from, while maintaining subject privacy~\citep{chen2015access}?
    \item In the public sector, a data hub may be organized and run by an open-source community of individuals consisting of data curators and foundation model providers. In this setting, answers to questions such as who stores the data? who pays for any compute? who is liable if licensing is violated? are particularly murky. How can the data hub provide that right tooling so that once answers to such questions are resolved, they can be operationalized with ease?
    \item What is the right set of statistics over the data to provide adequate documentation, without being too costly or difficult to obtain?
    \item How can a data hub support targeted data modifications such as augmentation~\citep{ma2019nlpaug, shorten2019survey} or data programming~\citep{ratner2017snorkel}?
    \item How can monitoring toolkits better detect when a foundation model needs to be updated due to poor performance on dynamically changing evaluation data?
\end{itemize}

Our vision for a data hub is not complete or fully detailed. However, we present initial thoughts on data challenges, and one solution to prompt thinking for how to improve data management for the foundation model lifecycle.
\newsection
\hypertarget{security}{\subsection{Security and privacy}}
\label{sec:security}
\sectionauthors{Florian Tramèr*, Rohith Kuditipudi*, Xuechen Li*}

\begin{figure}[!ht]
\centering
\includegraphics[width=\linewidth]{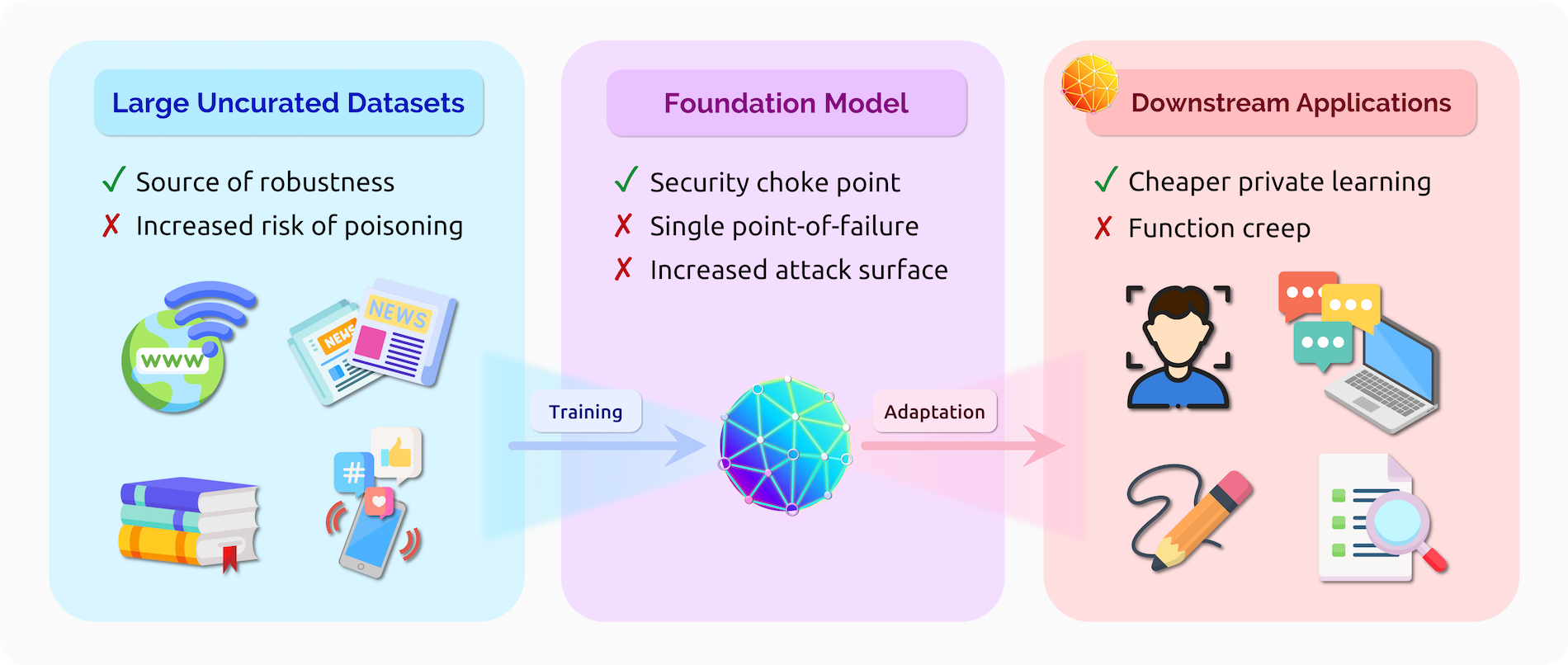}
\caption{\label{fig:security} Risks and opportunities raised by foundation models for security and privacy of ML systems. 
}
\end{figure}

As central components in critical data-driven decision-making systems, machine learning models must address a variety of security and privacy threats.\footnote{In this section, we focus on \emph{security for foundation models}. Some applications of \emph{foundation models for security} (\eg detection of toxic content) are discussed in \refsec{misuse}.} These threats can be characterized using the traditional ``CIA triad'' of computer security. ML systems should protect the \textbf{Confidentiality} of user data against \emph{inference and reconstruction attacks}~\citep{fredrikson2015model,shokri2017membership, carlini2019secret, carlini2020extracting}. Moreover, the secrecy of trained models themselves can be at risk of \emph{model stealing attacks}~\citep{tramer2016stealing, papernot2017blackbox}. The \textbf{Integrity} of ML systems can be compromised by \emph{adversarial examples}~\citep{biggio2013evasion, szegedy2014intriguing} and \emph{data poisoning attacks}~\citep{biggio2012poisoning, chen2017targeted}. Finally, \emph{resource-depletion attacks}~\citep{shumailov2020sponge,hong2020panda} can threaten the \textbf{Availability} of ML systems.

In regard to these threats, we posit that the security role of foundation models in future machine learning systems will be akin to the role played by the \emph{operating system} in traditional software systems. Due to its generality and ubiquity, a foundation model may become a \emph{single point of failure} and thus a prime target for attacks against applications derived from this model.
In turn however, a foundation model imbued with strong security and privacy properties could form the backbone for the design of a variety of secure and reliable ML applications. Of course, these applications may still have to be designed to enforce specific security and privacy guarantees (in the same way that software designers cannot rely on a secure operating system to protect against all security risks).

\subsubsection{Risks}
\paragraph{Single points of failure.}
A foundation model that is adapted to a variety of applications represents a single point of failure for these applications. 
For example, data poisoning attacks on a foundation model, where an adversary inserts malicious examples into the training data, might impact all adapted applications as well. 
Similarly, adversarial examples against a foundation model (\ie small input perturbations that cause the model to output very different features) could more easily transfer to adapted applications. \citet{wallace2019universal} even find that a \emph{single} adversarial trigger added to any input can cause language models such as GPT-2 to output a predefined piece of text.
A foundation model can also become a single point of failure for data privacy.
If a foundation model is pretrained on a company's private data and the model memorizes part of this data, all downstream applications could run the risk of exposing this data \citep{carlini2020extracting}.
The provider of a foundation model may also be a single point of trust for the privacy of application data. For example, the current API for GPT-3 requires that all (potentially sensitive) data used for fine-tuning or inference be uploaded to OpenAI's servers. Designing a foundation model service that avoids this centralization of trust is an interesting problem.

If the parameters of a foundation model are public, model stealing attacks on adapted applications could be facilitated, as the attacker only needs to reverse-engineer the ``delta'' with respect to the public foundation model \citep{krishna2019thieves} (\eg a linear model trained on features extracted from a public frozen model).

Finally, denial-of-service attacks on the foundation model provider could also be a concern and might be exacerbated by querying the model with special high-cost inputs~\citep{shumailov2020sponge}.

\paragraph{Data poisoning.}
Successful foundation models have so far been trained on large and often uncurated datasets scraped from the 
Web~\citep{radford2021learning,radford2019language}. 
This permissive data collection\dash{}coupled with a lack of direct training supervision\dash{}facilitates poisoning attacks on a foundation model's training data (\eg injecting hateful speech targeted at a specific individual or company into a few outbound pages from Reddit).
Worse, the power of poisoning attacks may be exacerbated by the  growing size and accuracy of today's models~\citep{carlini2021poisoningssl}.

To illustrate, \citet{schuster2021you} show that a code auto-completion system trained with GPT-2 on Github data can be poisoned into suggesting insecure code snippets with the injection of only a few malicious files.
\citet{carlini2021poisoning} further show that targeted attacks against CLIP-style~\citep{radford2021learning} models require modifying as little as two out of 3 million training examples.

\paragraph{Function creep \& dual use.}
Foundation models learn general features that enable them to be easily adapted to a variety of tasks. 
This flexibility, however, raises  concerns that foundation models could be used beyond their originally foreseen purposes%
\dash{}a risk commonly referred to as \emph{function creep} or \emph{dual use}.
Examples of function creep in machine learning include \emph{overlearning}~\citep{song2019overlearning} and \emph{adversarial reprogramming}~\citep{elsayed2018adversarial}.

To illustrate, CLIP was originally trained to solve the generic task of predicting image-text pairs, but in doing so also learned to capture rich facial features~\citep{goh2021multimodal}. 
While CLIP's ``model card''\footnote{\url{https://github.com/openai/CLIP/blob/main/model-card.md}. Accessed 06.30.2021} explicitly places facial recognition and other surveillance technologies as out-of-scope, CLIP can certainly be re-purposed for such tasks \citep{radiya2021data}.
This example illustrates that it may be challenging to constrain (or even foresee) the possible nefarious uses of a foundation model when it is designed. \refsec{misuse} provides further discussions on dual (mis)use of foundation models.

\paragraph{Multimodal inconsistencies.}
Multimodality may increase the attack surface of foundation models, by enabling adversaries to exploit inconsistencies across modalities.
The possibility of such attacks was demonstrated in an (in)famous example of CLIP classifying an apple with the word ``iPod'' stuck to it as an iPod~\citep{goh2021multimodal}. 
More generally, whenever a concept can be expressed using different modalities, inconsistencies across these modalities may be exploitable.

Such inconsistencies are particularly concerning when a foundation model is adapted to a task that primarily relies on only one of the learned modalities. For example, consider using features extracted from CLIP for facial recognition. This is a purely visual task, yet the adapted model's features will still be sensitive to textual signals (thus, an attacker might be able to evade facial recognition by wearing clothes with imprinted text). 
Alternatively, consider an autonomous driving system (an application that also relies primarily on vision) that sees a billboard with the word ``green'' on it, and mistakenly interprets this as a green light.

\subsubsection{Opportunities}
\paragraph{Security choke points.}
If adapted applications can inherit vulnerabilities from a foundation model, they can also inherit desirable security characteristics\dash{}such as robustness to adversarial examples or poisoning attacks. Foundation models could thus serve as security \textit{choke points}.
For example, a model robust to adversarial examples can retain its robustness when it is adapted to other tasks~\citep{shafahi2019adversarially}. 
Similarly, a foundation model provider that can (somehow) defend against poisoning, model-stealing or resource-depletion attacks could then provide such security guarantees for its customers' applications.

The tradeoff between a foundation model's role as a single point of failure or as a security choke point is reminiscent of similar security tradeoffs in other \emph{abstraction layers} in the software stack (\eg an operating system, database system, or a Web browser). By virtue of serving many different applications, an abstraction layer is a prime target for attack, but can typically also leverage far greater resources to enhance its security compared to any single application.

\paragraph{Cheaper private learning.}
Current foundation models are often trained by amassing vast amounts of data from publicly available sources (\eg from the open Web). 
This practice may raise concerns about privacy\dash{}in the broad sense of taking user data out of its intended context~\citep{nissenbaum2004privacy, carlini2020extracting}. 
While some existing works aim to mitigate a model's propensity to memorize training data (e.g., by de-duplicating training data~\cite{lee2021deduplicating}, or by pretraining under \emph{differential privacy}~\citep{anil2021large}), such solutions are unlikely to meet the broad privacy expectations that users could associate with text data~\citep{brown2022does}. 
On the other hand, public pretraining could also end up being a \emph{win} for user privacy in applications that handle scarce and sensitive data (\eg in healthcare). 

As an example, consider the problem of training a differentially private model~\citep{dwork2006calibrating} for a healthcare task. Training such a model ``end-to-end'' (\ie without leveraging any pretraining) to a decent privacy-utility tradeoff currently requires vast amounts of privacy-sensitive data~\citep{mcmahan2017learning, basu2021benchmarking}. In contrast, a foundation model pretrained on public data in many cases could be adapted to perform specific tasks with significantly less confidential data~\citep{bommasani19towards, tramer2021dp,li2022large,yu2022differentially}.

\paragraph{Robustness to adversarial examples at scale.}
There is evidence suggesting that training a model that is robust to adversarial examples requires vastly more data compared to standard training \citep{schmidt2018adversarially}, but that unlabeled data may suffice to bridge this gap \citep{carmon2019unlabeled,uesato2019are}.
Moreover, increasing model size and capacity (\ie over-parameterization) has also been shown to be necessary for achieving adversarial robustness in some settings \citep{madry2018towards,bubeck2021universal}.
Understanding how best to leverage over-parameterization and unlabeled data to achieve adversarial robustness is an important direction for future research. Given their unprecedented scale (both in terms of model size and training set size),
foundation models are uniquely positioned to benefit from this line of inquiry.

Despite their unprecedented scale, current foundation models unfortunately see little gains in robustness to worst-case adversarial perturbations~\citep{Fort2021CLIPadversarial, wallace2019universal}. However, multimodal models such as CLIP are surprisingly robust to (non-adversarial) distributional shifts (see \refsec{robustness}).
Whether these gains in distributional robustness can translate to increased resilience 
against real-world attacks is another exciting open question.
Particularly in settings where adversaries are subject to various constraints (\eg limited query access or computational budget), there is reason to be optimistic that enhanced distributional robustness could lead to concomitant gains in overall security\dash{}even if the foundation model remains vulnerable to worst-case ``white-box'' attacks. 
\newsection
\hypertarget{robustness}{\subsection{Robustness to distribution shifts}} 
\label{sec:robustness}
\sectionauthors{Sang Michael Xie, Ananya Kumar, Rohan Taori, Tony Lee, Shiori Sagawa, Pang Wei Koh, Tatsunori Hashimoto}

Real-world ML systems need to be robust to distribution shifts\dash{}they should work well on test distributions which differ from the train distribution.
High-stakes applications such as poverty mapping in under-resourced countries~\citep{xie2016transfer,jean2016combining}, self-driving cars~\citep{yu2020bdd100k,sun2020scalability}, and medical diagnosis~\citep{ albadawy2018tumor,dai2018dark} all require models that generalize well to circumstances not seen in the training data,
\eg test examples from different countries, under different driving conditions, or from different hospitals.
Prior work has shown that these types of distribution shifts can cause large drops in performance even in state-of-the-art models~\citep{blitzer2006domain,daume07easyadapt,sugiyama2007covariate,ganin2015domain,peng2019moment,kumar2020gradual,arjovsky2019invariant,szegedy2014intriguing,hendrycks2019benchmarking,sagawa2020group,recht2019doimagenet,abney2007semisup,ruder2018strong,geirhos2018generalisation,kumar2020conservative,yu2020mopo,geirhos2020shortcut,xie2021innout,koh2021wilds}. 

\newcommand\ppre{p_{\text{pre}}}
\newcommand\ptrain{p^{\sT}_{\text{ID}}}
\newcommand\ptest{p^{\sT}_{\text{OOD}}}

In this section, we consider the role of foundation models on robustness to distribution shifts. A foundation model is trained on a large and diverse unlabeled dataset sampled from a distribution $\ppre$ and can be adapted to many downstream tasks.
For each downstream task $\sT$, the foundation model is adapted to labeled training data sampled from an in-distribution (ID) training distribution $\ptrain$, and then evaluated on an out-of-distribution (OOD) test distribution $\ptest$.
For example, a poverty prediction model~\citep{xie2016transfer,jean2016combining} may be pretrained on unlabeled satellite data from across the world to learn useful features for all countries, then fine-tuned on labeled examples from Nigeria, and finally evaluated in Malawi where labeled examples are scarce.

We argue that 1) foundation models are a particularly promising approach to robustness. Existing work shows that pretraining on unlabeled data is an effective, general-purpose way to improve accuracy on OOD test distributions, in contrast to many robustness interventions which are constrained to narrow types of distribution shifts. However, we also discuss why 2) foundation models may not always mitigate distribution shifts, such as shifts due to spurious correlations or changes over time. Finally, 3) we outline several research directions to leverage and improve foundation models for robustness.

We note that one of the ways in which foundation models lead to improved extrapolation is by providing inductive biases (via model initialization) for the adapted model, which are learned on a diverse dataset that extends beyond the downstream training data. However, this same inductive bias can also encode harmful associations from the pretrained data and lead to representational and allocational harms in the presence of distribution shift. See \refsec{data} and \refsec{fairness} for further discussion of such harms and methods for mitigation.

\begin{figure}[t]
\centering
\includegraphics[width=0.85\linewidth]{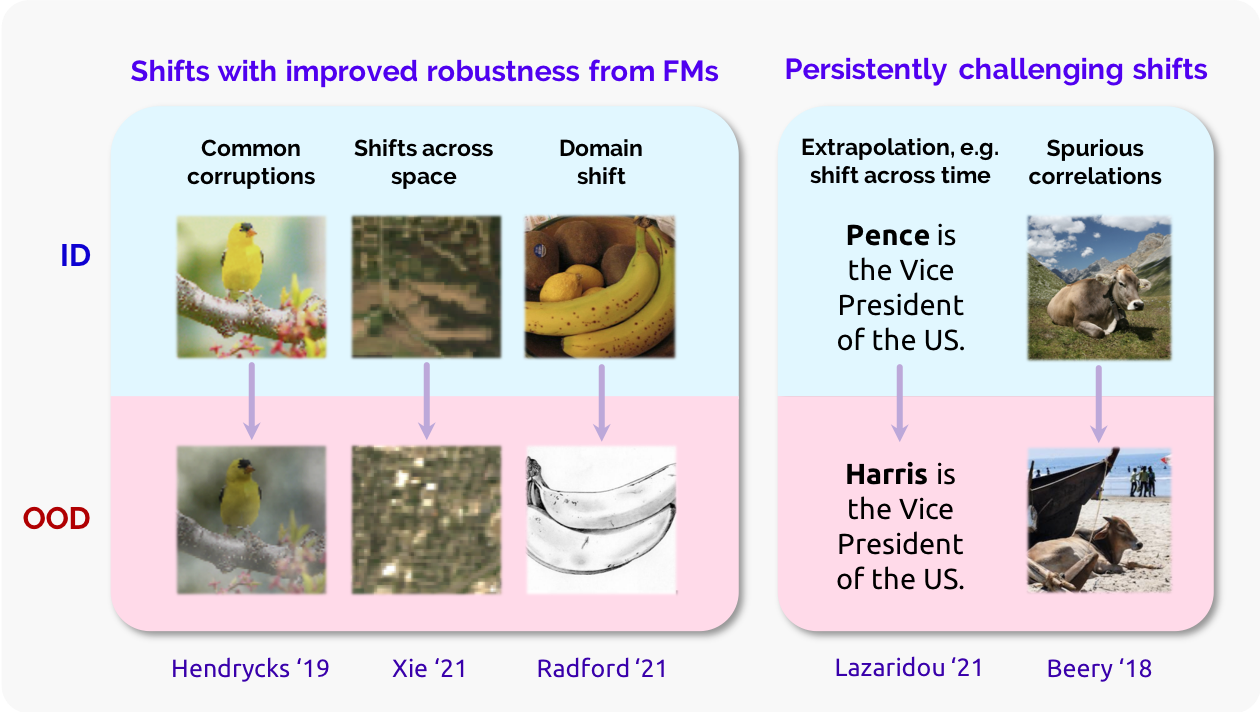}
\caption{\label{fig:robustness}  In-distribution (ID) and out-of-distribution (OOD) inputs for a variety of distribution shifts. The prediction tasks depicted here are image classification for images and fact verification for text.  Although representations learned by foundation models improve downstream robustness for many shifts (\eg common corruptions)~\citep{hendrycks2019benchmarking,xie2021innout,radford2021learning}, some shifts such as spurious correlations (where grass is predictive of cow)~\citep{beery2020iwildcam} and extrapolation across time (with facts that change over time)~\citep{lazaridou2021pitfalls} are still likely unaddressed by foundation models.
}
\end{figure}

\hypertarget{robustness-advantages}{\subsubsection{Advantages}}
\label{sec:robustness-advantages}

By learning representations on a large and diverse foundation model training distribution $\ppre$, foundation models can improve accuracy of the adapted derivative on the downstream test distribution $\ptest$.
OpenAI's CLIP model, which is a foundation model trained on a diverse set of images and natural language documents, has been shown to be robust to some benchmark distribution shifts on ImageNet~\citep{radford2021learning}: 
for example, both CLIP and a standard ResNet50 obtain 76\% accuracy on ImageNet, but CLIP achieves 6\% higher accuracy on ImageNetV2 \citep{recht2019doimagenet} and 35\% higher accuracy on ImageNet Sketch~\citep{radford2021learning}, which are both related but different from the original ImageNet training distribution.
In contrast, many other robustness interventions, such as adversarial training~\citep{madry2018towards}, invariant risk minimization~\citep{arjovsky2019invariant}, or using larger models have had little impact on effective robustness (defined as the gap between in-distribution and out-of-distribution performance) on these ImageNet tasks, especially without explicit knowledge of the distribution shift \citep{taori2020measuring,santurkar2020breeds,radford2021learning,miller2021line}.

Many other works demonstrate that pretraining on large datasets can improve robustness to common image corruptions, label shift, and label corruptions \citep{hendrycks2019pretraining,hendrycks2019selfsupervised};  to real-world spatial shifts in satellite imagery tasks~\citep{xie2021innout,kumar2022finetuning}; and to shifts across topics in natural language understanding tasks \citep{hendrycks2020pretrained,fisch2019mrqa,yogatama2019learning}.
As another example, diversifying the foundation model training data to include multiple languages (as in multilingual BERT~\citep{liu2020multilingual}) significantly improves performance in unseen language pairs.

\hypertarget{robustness-challenges}{\subsubsection{Persistent challenges}}
\label{sec:robustness-challenges}

Despite promising signs that foundation models will result in substantial improvements to robustness, we anticipate that foundation models are not a panacea for distribution shifts. We discuss this in the context of two broad categories of distribution shifts below.

\paragraph{Spurious correlations.} 
Spurious correlations are statistical correlations between features and labels with predictive power on the training distribution but not on the test distribution~\citep{heinze2017conditional, arjovsky2019invariant, sagawa2020group}.
Well-known examples include reliance on background color for object recognition~\citep{xiao2020noise}, surgical markers for medical diagnostics~\citep{winkler2019derm}, annotator biases in crowdsourced data~\citep{tsuchiya2018performance,gururangan2018annotation,poliak2018hypothesis,geva-etal-2019-modeling}, and demographic biases \citep{abid2021,stereoset,gehman-etal-2020-realtoxicityprompts}. 
Models learn these spurious correlations largely because the foundation model training and adaptation data exhibit these biases~\citep{nagarajan2020understanding,gehman-etal-2020-realtoxicityprompts}, and this issue cannot simply be addressed with larger models~\citep{sagawa2020overparameterization}.

Foundation models may exacerbate or mitigate the effects of spurious correlations, but this depends on the nature of the particular downstream task and its relation to the foundation model training data and algorithm. By training with a diverse dataset, foundation models may improve robustness to spurious correlations that are found only in a subset of the training data: \eg existing studies find that pretrained language models can avoid spurious correlations by quickly learning from counterexamples to the spurious correlations~\citep{tu2020empirical}.
However, foundation models can also exacerbate the issue by introducing biases present in the foundation model training data, as observed for demographic biases in GPT-3 and other NLP models~\citep{abid2021,stereoset,gehman-etal-2020-realtoxicityprompts}.
Moreover, training at scale alone need not fully address the root issue of identifying and not relying on the features that are predictive on the downstream training set but not on the downstream test set~\citep{heinze2017conditional}.
Addressing these challenges will require us to understand and manage the inductive bias from foundation model training and develop adaptation algorithms that are resistant to learning spurious correlations.

\paragraph{Extrapolation and temporal drift.} 

Finally, the few- and zero-shot capabilities of foundation models will mean that these models will increasingly be used far beyond the training distribution. While large-scale foundation model training can help with certain forms of extrapolation to new distributions ~\citep{papadimitriou2020learning}, there may be limits to their extrapolation capabilities. 
For example, existing language models cannot handle changes to world knowledge or language change without re-training~\citep{lazaridou2021pitfalls,dhingra2021time}, zero-shot transfer in CLIP suffers greatly in satellite image domains~\citep{radford2021learning}, and ImageNet pretraining does not substantially improve the performance of large models on medical images~\citep{raghu2019transfusion,ke2021chextransfer}.
We believe that foundation models cannot be assumed to automatically extrapolate within a given modality (\eg all images), and it will become increasingly important to define and separate the forms of extrapolation that are newly enabled by foundation models from those that remain out of reach.
Though existing taxonomies for distribution shifts have been proposed in generality~\citep{candela2009when,ye2021oodbench},
fully understanding and defining the types of distribution shifts for which foundation models are effective is a major open problem for robustness research.

\hypertarget{robustness-opportunities}{\subsubsection{Opportunities}}
\label{sec:robustness-opportunities}

Foundation models hold substantial promise as a general-purpose robustness intervention for distribution shifts and open new avenues for robustness research. We outline some opportunities and open questions below.

\paragraph{Understanding foundation model representations.}
Existing studies of the robustness of foundation models have been largely empirical, and there is little understanding of the mechanism behind gains in robustness.
~\citet{sun2019unsupervised} hypothesize that pretrained representations
bring disparate domains (such as ID and OOD distributions) closer together, which can in turn improve generalization from labeled ID data to OOD data~\citep{ben2010theory}. Controlled experimentation on measuring the distance between domain representations with and without pretraining can elucidate this effect. There are initial promising directions in characterizing foundation model training (\eg contrastive learning as a spectral graph decomposition~\citep{haochen2021spectral}) and their inductive biases~\citep{SaMa20mat,lee2020predicting,zhang20ont,xie2020selftraining}.
However these theories are limited and fail to address other empirically effective foundation models such as fully generative language models (\eg GPT-3~\citep{brown2020gpt3} and image-GPT~\citep{chen2020imagegpt}). Further understanding how these inductive biases are useful under distribution shift may lead to a more complete theory (\refsec{theory}) of how foundation models improve robustness. 

\paragraph{Data augmentation in foundation model training.} 
While foundation models trained without knowledge of the downstream tasks can avoid some task-specific biases and often improve robustness, certain statistical biases stemming from how the foundation model was trained may persist.
As a concrete example, many contemporary self-supervision algorithms are heavily dependent on choosing an appropriate set of data augmentations \citep{chen2020simclr}, which in turn confers different types of robustness in the adaptation phase. For instance, \citet{xiao2021contrastive} show that a foundation model for vision trained with contrastive learning on rotation augmentations may improve OOD performance on adaptation tasks with rotation invariance, but may not improve robustness for tasks where OOD generalization requires other invariances.
Further research into what types of data augmentations improve robustness for a wide range of downstream tasks\dash{}including data augmentations that are learned from data \citep{wong2020learningpert,Tamkin2021ViewmakerNL} or designed to be generally applicable across data modalities \citep{verma2021towards}\dash{}will inform better foundation model training algorithms (\refsec{training}).

\paragraph{Encoding structure in foundation model training.}
In general, exploring new ways of encoding known structure and invariances in the data is an important path forward for foundation model training. Many real-world tasks have additional metadata (\eg  spatial location coordinates, climate information from auxiliary satellites in our poverty prediction example), which may provide additional structure for OOD generalization (\eg across geographic areas)~\citep{xie2021innout,koh2021wilds}. For example,~\citet{xie2021innout} show that metadata can be used as targets for pretraining to improve downstream OOD accuracy. In language, modeling the tags in HTML data provides additional downstream-task-adjacent supervision, allows for new forms of prompting (\eg filling in \texttt{<title>} tags for title suggestion), and improves data efficiency~\citep{aghajanyan2021HTLM}. While current data augmentation methods encode hand-crafted knowledge, other avenues such as exploiting metadata could provide more automated ways of determining which structures and invariances to incorporate for foundation model training. 

\paragraph{Specialization vs.~diversity in foundation model training data.}
The choice of foundation model training data has downstream effects\dash{}training on a more diverse dataset is not always better for downstream performance than a more specialized foundation model~\citep{cole2021contrastive,chalkidis2020legal} (see \refsec{adaptation} for a more detailed discussion).
In some domains such as satellite images and specialized text topics, continued pretraining on the specialized domain improves the downstream performance significantly~\citep{reed2021selfsupervised,gururangan2020dont}.
This is a potential source of tension: on one hand, we might want to train the foundation model on a large, diverse dataset in order to have more robust performance under distribution shifts, while on the other hand, we might need to specialize the foundation model to improve its in-distribution and out-of-distribution performance on downstream tasks.
A better understanding of how specialization affects the in-distribution and out-of-distribution performance of foundation models will allow us to design and collect more effective foundation model training sets.

\paragraph{Adaptation methods.}
Although foundation models provide a strong starting point, how the adaptation method uses the pretrained information can affect robustness.
For instance, lightweight tuning methods for language models (\eg adapter/prefix/prompt tuning~\citep{houlsby19adapter,li2021prefix,lester2021power}), which adapt the model for a new task by optimizing a small set of parameters (such as a continuous prompt) while keeping the other foundation model parameters frozen, seem to give OOD performance benefits (\refsec{adaptation}).
~\citet{xie2021composed} explain this in a special case, where composing a learned model with a frozen foundation model can reduce the complexity of the learned model, improving generalization both ID and OOD.
On vision datasets,~\citet{wortsman2021robust, kumar2022finetuning} find that freezing a foundation model and training only the head can lead to better OOD performance than fine-tuning the entire model.~\citet{kumar2022finetuning} explain this theoretically by showing that full fine-tuning can distort pretrained features even in a simple setting (two layer linear networks).
However, it is still poorly understood in general why freezing parameters seems to improve OOD performance.
Finally, while current adaptation methods may suffice for good ID generalization, the methods do not explicitly account for distribution shift in their design. As a first step, we can investigate how methods for distribution shifts such as domain adaptation, domain generalization, and semi-supervised learning methods interact with foundation models when used for adaptation. Progress in these directions can lead to adaptation methods that can better leverage foundation models for robustness.
\newsection
\hypertarget{ai-safety}{\subsection{AI safety and alignment}}
\label{sec:ai-safety}
\sectionauthors{Alex Tamkin, Geoff Keeling, Jack Ryan, Sydney von Arx}

The field of Artificial Intelligence (AI) Safety concerns itself with potential accidents, hazards, and risks of advanced AI models, especially larger-scale risks to communities or societies. Current foundation models may be far from posing such risks; however, the breadth of their capabilities and potential applications is striking, and a clear shift from previous ML paradigms. 
While AI safety has historically occupied a more marginal position within AI research, the current transition towards foundation models and their corresponding generality offers an opportunity for AI safety researchers to revisit the core questions of the field in a new light and reassess their immediate or near-future relevance.\footnote{See \citet{amodei2016concrete} and \citet{hendryckssafety2021} for broader perspectives on open problems in AI Safety.}

\subsubsection{Traditional problems in AI safety}

A major branch of AI safety research concerns the implications of advanced AI systems, including those that might match or exceed human performance across a broad class of cognitive tasks \citep{everitt2018agi}.\footnote{This is referred to by some as AGI or artificial general intelligence, although terminology use varies \citep[\eg see][]{karnofsky2016potential}.} A central goal of safety research in this context is to mitigate large-scale risks posed by the development of advanced AI.\footnote{Note that this does not require a belief that building certain kinds of advanced AI is a desirable goal, nor even certainty that it is an achievable one.} These risks may be significantly more speculative than those considered in \refsec{misuse}, \refsec{robustness}, and \refsec{security}; however, they are of far greater magnitude, and could at least in principle result from future, highly-capable systems. Of particular concern are global catastrophic risks: roughly, risks that are global or trans-generational in scope—causing death or otherwise significantly reducing the welfare of those affected (\eg a nuclear war or rapid ecological collapse) \citep{bostrom2011global}. What AI safety research amounts to, then, is a family of projects which aim to characterize what (if any) catastrophic risks are posed by the development of advanced AI, and develop plausible technical solutions for mitigating the probability or the severity of these risks. The best-case scenario from the point of view of AI safety is a solution to the control problem: how to develop an advanced AI system that enables us to reap the computational benefits of that system while at the same time leaving us with sufficient control such that the deployment of the system does not result in a global catastrophe \citep{bostrom2011global}. However technical solutions are not sufficient to ensure safety: ensuring that safe algorithms are actually those implemented into real-world systems and that unsafe systems are not deployed may require additional sociotechnical measures and institutions. 

Reinforcement Learning (RL), which studies decision-making agents optimized towards rewards, has been a dominant focus in AI safety for the past decade. What is at issue here is the difficulty of specifying and instantiating a reward function for the AI that aligns with human values, in the minimal sense of not posing a global catastrophic threat.\footnote{See \citet{Hubinger2019RisksFL} for a discussion of some challenges that arise at the threshold between reward specification and reward instantiation.} While this problem, known as value alignment \citep{gabriel2020artificial, yudkowsky2016ai}, may seem trivial at first glance, human values are diverse,\footnote{See \citet{gabriel2020artificial} for an extended discussion of human diversity, ethics, and the value alignment problem} amorphous, and challenging to capture quantitatively. Due to this, a salient concern is reward hacking, where the AI finds an unforeseen policy that maximizes a proxy reward for human wellbeing, but whose misspecification results in a significant harm.\footnote{See \href{https://docs.google.com/spreadsheets/d/e/2PACX-1vRPiprOaC3HsCf5Tuum8bRfzYUiKLRqJmbOoC-32JorNdfyTiRRsR7Ea5eWtvsWzuxo8bjOxCG84dAg/pubhtml}{this spreadsheet} for a list of real-world examples of reward hacking, including an aircraft landing algorithm which achieved a perfect score by outputting large forces that exploited a flaw in the simulator.} Many efforts to combat the value alignment problem have focused on maximizing corrigibility, which is when errors in the design of a system can be corrected once the system is running \citep{soares2015corrigibility}. This can be far from straightforward—in the RL context, an agent with a specified goal would be incentivized to prohibit attempts to alter that goal, as any attempt to alter that goal would likely be suboptimal for the goal’s realization \citep{Omohundro2008TheBA}.

However, pure RL is not the only theorized route to advanced AI. Foundation models can also be trained with simple (self-)supervised objectives like next-token prediction, yet can still be used in interactive and goal-directed ways, with or without additional RL training. Moreover, it appears that many of these methods may result in increased capabilities through straightforward scaling of compute, number of parameters, and dataset size \citep{Hestness2017DeepLS, kaplan2020}. What concepts like value alignment and corrigibility amount to in the broader context of foundation models differ in several respects to the pure RL case, and must accordingly be carefully theorized.

\subsubsection{Current foundation models and AI safety}

Many of these risks in the RL setting result from models optimized to carry out goals. However, a key challenge for AI safety research on recent foundation models is that goal-directed behavior may emerge despite not being explicitly optimized for (see also \refsec{training}). As an example, large language models may be trained on corpora where agents use language in goal-directed ways, such as in persuasive text. To predict the next token well, a model may acquire a general capability to reason and produce arguments, which could emerge with suitable contexts. Foundation models trained on other kinds of human data may capture other kinds of goal-directed behavior present in the data; \eg robotic agents trained to mimic humans in videos may attempt to punch or knock-out their human operators if their training data includes videos of boxing matches. Recent work has also attempted to directly train agents to produce goal-directed behavior; for example, the Decision Transformer trains a sequence model on trajectories prepended with their returns \citep{Srivastava2019TrainingAU, Schmidhuber2019ReinforcementLU, Chen2021DecisionTR}. One can then generate high-return trajectories by ``prompting'' this model with a high return, which raises similar questions of reward hacking from the RL context. 

However, a major aim of safety research on goal-directed models is to gain more principled control and explainability over the actions being pursued by the agent, as opposed to relying on inscrutable decisions from a blackbox neural network.\footnote{For more on the relationship between understanding and semantics see \refsec{philosophy}} This makes current foundation models an exciting avenue of study for AI safety research, as aligning them may be a useful precursor for aligning more advanced models \citep{christiano2016prosaic, cotra2021narrow, Kenton2021AlignmentOL}. One challenge is the misalignment between the foundation model's training objective and the desired behavior; for example, a language model may be trained to predict the next word of all documents in the training corpus regardless of veracity, but users may want the model to only output true or helpful text \citep{tamkin2021understanding}. One potential way to steer goal-directed agents towards desired behavior may be to train them with natural language descriptions of actions\dash{}this may enable steering them with language as well as enabling them to output interpretable language describing the task they "believe" they are performing, similar to methods for controllable generation and source attribution \citep[\eg][see also \refsec{robotics}, \refsec{interaction}, and \refsec{interpretability}]{keskar2019ctrl}. However, further advances would be necessary to ensure the reliability and self-consistency of such models in the wild (\refsec{robustness}), as well as gaining a more mechanistic understanding of how these models operate \citep[also see \refsec{interpretability}]{cammarata2020thread}. And even if natural language-based control of future foundation models enables better task specification and monitoring, models may acquire deceptive or otherwise undesirable behavior from human data\dash{}identifying and neutralizing this behavior is another important direction for future study.

While the self-supervised objectives described in the previous paragraph train models to capture human behavior in the data, new training paradigms may produce goal-directed foundation models capable of carrying out a wide range of tasks in complex environments, and which exhibit capabilities superior to humans in different domains (see \refsec{training}). For example, goal-directed foundation models may be trained in an open-ended self-play setting, similar to AlphaGo, or in vast multitask single-agent RL setups. This might lead to emergent capabilities that complicate efforts to get agents to carry out goals, especially if many agents are trained together in a rich world-simulator that encourages the development of skills like deception, misdirection, dissimulation, persuasion, and strategic planning. Aside from countering deceptive behavior, it also remains unclear how to effectively evaluate and control the behavior of very capable models, known as scalable oversight or alignment \citep{amodei2016concrete, Leike2018ScalableAA}; \eg scoring novel reactions proposed by a chemical foundation model (see \refsec{evaluation}). New human-in-the-loop approaches for training, steering, monitoring, and understanding these models are thus exciting future directions.

Finally, even before any of these more advanced capabilities emerge, an important research area for AI safety in the near term is characterizing and forecasting the capabilities of current self-supervised foundation models. There are three aspects which make this challenging. First, the generality of foundation models means that they can be applied to countless different kinds of applications in unexpected ways. Enumerating current and planned applications of foundation models is not sufficient to capture the full range of ways they could be used. Second, even within a particular application, model capabilities are emergent: they grow and change in unexpected ways as models scale. For example, the ability to control GPT-3 via ``prompting" was an emergent phenomenon of which only the barest glimpses were evident in the smaller GPT-2 model \citep{radford2019language,brown2020gpt3}. What the emergent properties of future foundation models will look like is unknown. Third, even within a particular application and scale, a model's capabilities are not easy to characterize. For example, the ability of GPT-3 to perform addition improves dramatically once commas are added to the inputs \citep{branwen2020gpt, brockman2020math}. Similarly, small rewordings of prompts can have large impacts on task performance. Since the space of prompts is intractable to enumerate, it is challenging to definitely assert that any task is outside the reach of current prompt-based foundation models\dash{}this is a major challenge for reasoning about possible catastrophic risks from foundation models.

\subsubsection{Potential catastrophic risks from future foundation models}
The broad and quickly-growing capabilities of current models suggest the benefit of attempting to characterize possible catastrophic risks from more advanced systems. We see at least two ways in which advanced foundation models might contribute to such outcomes. 

\paragraph{Catastrophic robustness failures.} \refsec{robustness} discusses how models may behave in unexpected or harmful ways when confronted with new kinds of data \citep{amodei2016concrete, yudkowsky2008artificial}. These failures may be especially consequential if foundation models are integrated into important systems that leverage foundation models' ability to quickly adapt to many different tasks and situations. Failures could be catastrophic if they occur in warfare systems (resulting in unwanted discharge of weapons, possibly igniting a conflict), critical infrastructure (accidental destruction of critical energy or agricultural capabilities), or if they become essential to a large fraction of economic activity (whose unexpected failure could result in a sudden collapse in living standards and political instability; see also \refsec{economics}). Indeed, the threat of catastrophic robustness failures is particularly pertinent for foundation models in contrast to other kinds of AI. This is because a foundation model consists of a single model that may be adapted for many different use cases, such that robustness failures derived from the statistical associations learned by the model could in principle manifest in a correlated way across several different domains. If the same foundation model is integrated into multiple critical functions, then lack of robustness in the model could lead to correlated failures that span multiple critical functions or failsafes.

\paragraph{Misspecified goals.} The use of foundation models might increase the risks of optimizing misaligned yet easy-to-specify goals, often referred to as Goodhart’s Law \citep{Kenton2021AlignmentOL, goodhart1984}. A current-day example of these risks is the negative effects of some recommender systems (\eg polarization, media addiction) which may optimize simple engagement metrics rather than a harder-to-measure combination of societal and consumer well-being \citep{burr2018analysis, milano2020recommender}. Future institutions may leverage uninterpretable foundation models to maximize simple measures such as profit or GDP, due to these models' ability to adapt to the many different subproblems each of these metrics is dependent on. However, at larger scales optimizing for these proxy metrics instead of a more holistic goal designed for human welfare could inadvertently lead to environmental or geopolitical harms \citep{ gabriel2020artificial, creel2021}.

\subsubsection{Conclusion}
In sum, we argue that current and potential future emergent properties of foundation models make them ripe objects of study for the field of AI safety. We encourage future work on characterizing and forecasting the exact capabilities and risks of foundation models; developing new methods to align foundation models to human values and desired goals; and for states, research labs, and businesses to coordinate on proactive measures to mitigate salient risks.

\newsection
\hypertarget{theory}{\subsection{Theory}}
\label{sec:theory}
\sectionauthors{Aditi Raghunathan, Sang Michael Xie, Ananya Kumar, Niladri Chatterji, Rohan Taori, Tatsunori Hashimoto, Tengyu Ma}

\newcommand\lpre{\ensuremath {\ell_{\text{pre}}}}
\newcommand\emppre{\ensuremath {\widehat{L}_{\text{pre}}}}
\newcommand\popadapt{\ensuremath {L_{\text{adapt}}}}
\newcommand\empadapt{\ensuremath {\widehat{L}_{\text{adapt}}}}
\newcommand\poppre{\ensuremath {L_{\text{pre}}}}
\newcommand\datapre{\ensuremath {p_{\text{pre}}}}
\newcommand\trainpre{\ensuremath {\hat{p}_{\text{pre}}}}
\newcommand\datatask{\ensuremath {p_{\text{task}}}}
\newcommand\traintask{\ensuremath {\hat{p}_{\text{task}}}}
\newcommand\thetaum{\ensuremath {\hat{\theta}_{\textup{FM}}}}
\newcommand\gammatask{\ensuremath {\gamma_{\text{task}}}}
\newcommand\ladapt{\ensuremath {\ell_{\text{adapt}}}}
\newcommand\poptest{\ensuremath {L_\text{task}}}

Rigorous mathematical theory plays a foundational role in many engineering and science disciplines (\eg information theory in electrical engineering). We believe that theory of foundation models can be particularly beneficial in guiding technical decisions and innovations because of the huge computational costs associated with experimenting on foundation models. 
In addition, theoretical insights help elucidate fundamental limitations and explain surprising empirical phenomena.
However, the community currently has a limited theoretical understanding of foundation models, despite much recent progress~\cite{arora2019theoretical,haochen2021spectral,wei2021pretrained,wei2020theoretical,zhang2021inductive,saunshi2020mathematical,dao2019kernel,tosh2020contrastive,tosh2021contrastive,cai2021theory,lee2020predicting,zimmermann2021contrastive,bansal2020self,wang2020understanding,tsai2020self,tian2020makes,tian2020understanding, tripuraneni2020theory,du2020fewshot}. 

Deep neural networks form the backbone of foundation models. 
Even in the well-studied supervised learning setting, where the train and test scenarios have the same distribution, there are numerous open questions around deep nets such as understanding non-convex optimization, the implicit regularization effect of optimizers, and expressivity. Foundation models raise questions that significantly go beyond the supervised deep learning setting. 
The core problem in theoretically analyzing foundation models is understanding why training on one distribution with a possibly unsupervised/self-supervised loss leads to good adaptation performance on \emph{different} downstream distributions and tasks.\footnote{The theory for foundation models closely relates to, but also goes beyond the theory for transfer learning (which is itself an underexplored area): foundation models are possibly trained with unlabeled data and will be adapted to many or all natural tasks, whereas transfer learning typically studies labeled source tasks and a fixed number of target tasks.}

We will discuss an intuitive modularization to analyze foundation models that lays bare the connections between supervised learning and foundation models, concrete and core technical questions, and some promising theoretical tools to address these questions. These new core questions can provide useful insight into foundation models and can be studied in parallel to supervised deep learning theory. While we focus on analyzing the downstream performance, the proposed modularization and tools could prove useful to analyze other metrics of interest such as robustness to distribution shifts (\refsec{robustness}) and security (\refsec{security}).

\begin{figure}[!ht]
\centering
\includegraphics[width=\linewidth]{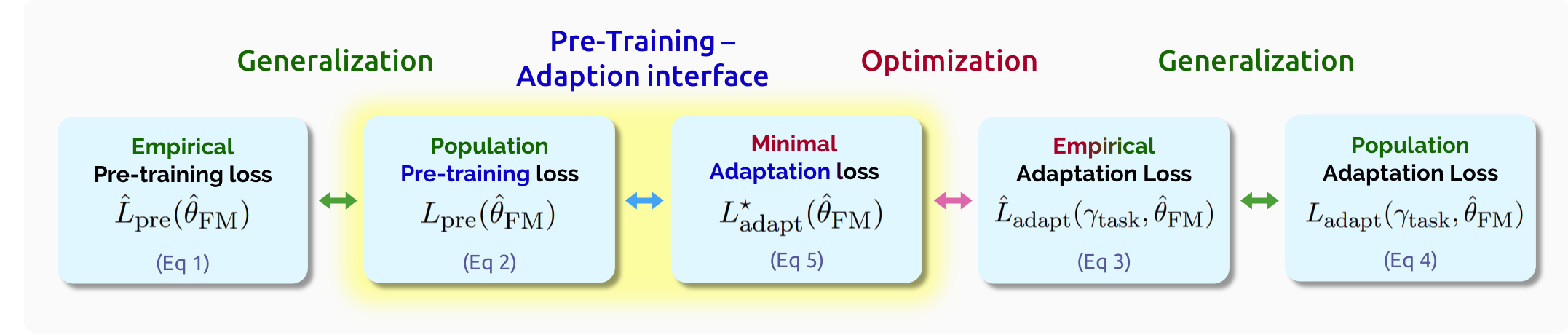}
\caption{\label{fig:theory} The analysis of foundation models from pretraining on diverse data to downstream performance on adapted tasks involves capturing the relation between different loss terms as shown above. The main challenge is to analyze the highlighted pretraining-adaptation interface which requires reasoning carefully about the population losses in addition to the model architecture, losses and data distributions of the pretraining and adaptation stages (\protect\refsec{theory-interface}). Analysis of generalization and optimization largely reduces to their analysis in standard supervised learning.}
\end{figure}

\hypertarget{theory-modularizations}{\subsubsection{Theoretical formulations and modularizations}}
\label{sec:theory-modularizations}
Recall that foundation models are trained on a large amount of raw data (\refsec{training})  then adapted to specific tasks (\refsec{adaptation}) and therefore can be decomposed naturally into training and adaptation phases. We identify interfaces between them and disentangle parts specific to foundation models from parts that require standard deep learning theory, so that they can be independently worked on. We introduce a modularized analysis framework, which has also been implicitly or explicitly employed in recent works, \eg~\citet{arora2019theoretical,haochen2021spectral,wei2020theoretical,tripuraneni2020theory}. The crucial component in this modularized analysis turns out to be the \emph{pretrain-adaptation interface.} We first describe the modularization, and discuss why we find this modularization promising and finally some limitations. 

We will refer to the training phase explicitly as ``pretraining'' to distinguish it from the adaptation phase that could also involve training on a few samples from a particular task. 

\paragraph{Pretraining phase.} 
The pretraining of foundation models often involves a data distribution $\datapre$ (\eg the distribution of natural text) and a \emph{pretraining loss} function $\lpre(x;\theta)$ that measures the loss (\eg language modeling loss in GPT-3) on an input $x$ for a model with parameters $\theta\in \Theta$. Let $\trainpre$ denote the empirical distribution over a large number of independent samples from $\datapre$. 

Pretraining minimizes the loss $\lpre$ on $\trainpre$, which we call \textit{the empirical pretraining loss}, and produces a model $\thetaum$:
\begin{align}
    \emppre(\theta) \eqdef \E_{x\sim \trainpre}[\lpre(x;\theta)],
        ~~\text{and}~~
\thetaum \eqdef \argmin \limits_{\theta \in \Theta} \emppre(\theta).
\end{align}

We consider the corresponding loss on the population distribution $\datapre$, called the \textit{population pretraining loss}, as a central concept: 
\begin{align}
   \poppre(\theta) \eqdef \E_{x\sim\datapre}[ \lpre(x; \theta)] .
\end{align}

\paragraph{Optimization-based adaptation phase.} 
We frame adaptation as a general constrained optimization problem that depends on $\thetaum{}$, abstracting away those adaptation methods that are based on optimizing certain loss functions such as fine-tuning and prompt-tuning (see, \eg ~\citep{houlsby19adapter,li2021prefix,lester2021power}, and \refsec{adaptation}).

Since different adaptation methods could modify different subsets of the model parameters, we denote the space of adapted model parameters by some $\Gamma$. Given a downstream task distribution $\datatask$ (\eg question answering in a particular domain) and a few empirical samples $\traintask$ sampled from $\datatask$, we model the adaptation phase as minimizing some \textit{adaptation loss} $\ladapt$ on $\traintask$ w.r.t adapted parameters $\gamma \in \Gamma$: 
\begin{align}
\gammatask(\thetaum) \eqdef \argmin \limits_{\gamma \in \Gamma, C(\gamma;\thetaum)\le c_0 }\empadapt(\gamma, \thetaum),
\end{align}
where $\empadapt(\gamma, \thetaum) \eqdef \E_{x\sim \traintask}[\ladapt(x;\gamma, \thetaum)]$ is the empirical adaptation loss, and $C(\gamma, \thetaum)\leq c_0$ is an optional constraint that controls the complexity of the adapted parameters, encompassing both explicit regularization (\eg model dimensionality and norm) and the implicit regularization of the adaptation process. 

We list some common adaptation methods and discuss the corresponding adapted parameter $\gamma$ and constraints $C(\gamma, \thetaum)\leq c_0$. 
\begin{enumerate}
    \item Linear probing: training a linear classifier on top of the representations from a foundation model. Here $\Gamma = \R^k$ is the set of linear classifiers on the representations of dimensionality $k$, and $C(\gamma, \thetaum)$ could be the $\ell_2$ or $\ell_1$ norm of $\gamma$. 
    \item Fine-tuning: optimizing a randomly initialized linear head for a few steps, and all other parameters $\theta$ from the initialization of $\thetaum$. Here $\gamma$ is the concatenation of $\theta$ and the linear head. Such a process could correspond to some implicit regularization of $\gamma$ towards the initialization $\thetaum$ captured by $C(\gamma, \thetaum) \leq c_0$. The exact term $C(\gamma, \thetaum)$ would depend on the optimization algorithm used, and such a characterization of the implicit regularization of optimization is an area of active research study ~\citep[\eg][and references therein]{gunasekar2017implicit, soudry2018implicit,gunasekar2018implicit,arora2019implicit, blanc2019implicit,woodworth2020kernel, wei2020implicit, haochen2021shape,damian2021label,kumar2022finetuning}.\footnote{
    It may not always be feasible to characterize the inductive bias of adaptation via an explicit constraint $C(\gamma,\thetaum)\leq c_0$. The modularization we propose is also applicable in these cases, but for notational simplicity, we focus on the case where implicit regularization can be approximated via an explicit constraint.
    }
    \item Prompt-tuning: optimizing a small set of continuous task-specific vectors that prepend the task inputs. Here $\gamma$ is the continuous prompt vectors which often has small dimensionality, and we may optionally have a constraint on the norms of $\gamma$. 
\end{enumerate}

One obvious limitation to note is that this formulation excludes adaptation methods such as in-context learning~\citep{brown2020gpt3} where there is no ``training'' (\ie the minimization of some empirical adaptation loss) during the adaptation phase. We discuss this and other limitations in \refsec{theory-incontext}. 

Two central quantities for the adaptation phase are the \textit{population adaptation loss}
\begin{align}
    \popadapt(\gamma, \thetaum) = \E_{x\sim\datatask}[ \ladapt(x; \gamma,\thetaum)]
\end{align}
and the \emph{minimal adaptation loss} 
\begin{align}
    \popadapt^\star(\thetaum) = \min_{\gamma \in \Gamma, C(\gamma;\thetaum)\le c_0}\popadapt(\gamma, \thetaum)
    \label{eqn:5}
\end{align}

\paragraph{Separate analysis for modularized phases.} Existing generalization theory for standard supervised learning aims to show that $\emppre\approx \poppre$ and $\empadapt\approx \popadapt$. 
Addressing these questions specifically for deep nets is an active research area. 
We can also leverage the standard learning theory decomposition to bound the final downstream task loss by the excess generalization error and the minimal adaptation loss as follows. 
 \begin{align}
    \popadapt(\gammatask, \thetaum) \le \underbrace{\popadapt^\star(\thetaum)}_{\textup{minimal adaptation loss}} + \textup{~~generalization error}
    \label{eqn:6}
\end{align}
where the generalization error captures the closeness between $\popadapt$ and $\empadapt$.\footnote{More precisely, the generalization error term is the sum of  $\popadapt(\gammatask, \thetaum) -  \empadapt(\gammatask, \thetaum)$ and $\empadapt(\gammatask^\star, \thetaum)-\popadapt(\gamma^\star, \thetaum) = \empadapt(\gammatask^\star, \thetaum)-\popadapt^\star(\thetaum)$, where $\gammatask^\star$ is the minimizer of \refeqn{5}. \refeqn{6} follows easily be using $\empadapt(\gammatask, \thetaum)\le \empadapt(\gammatask^\star, \thetaum)$.}
The decomposition and relationship between these key quantities are shown in \reffig{theory}. 
The generalization and optimization arrows, as argued above, largely reduce to deep learning theory in the supervised setting. 
What we are left with is the main challenge with foundation models, which is to understand why the minimal adaptation loss $\popadapt^*(\thetaum)$ can be small as a result of a small pretraining population loss, which study in \refsec{theory-interface}.

The work of~\citet{arora2019theoretical} pioneered the pursuit of this question by bounding from above $\popadapt^\star(\thetaum)$ by $\poppre(\thetaum)$ in the context of contrastive learning, and ~\citet{haochen2021spectral,tosh2020contrastive,tosh2021contrastive} relax the data assumptions. Other pretraining methods successfully analyzed under this framework (implicitly or explicitly) include pretraining with language models ~\cite{wei2021pretrained} or self-supervision~\cite{lee2020predicting}, with self-training algorithms~\cite{wei2020theoretical,cai2021theory}, and with multiple supervised tasks ~\cite{tripuraneni2020theory,du2020fewshot}. 

\hypertarget{theory-interface}{\subsubsection{Why is the pretraining-adaptation interface interesting?.}}
\label{sec:theory-interface}
As shown in \reffig{theory}, the main missing link beyond standard supervised theory is:
\begin{center}
    \emph{Under what conditions does a small population pretraining loss $\poppre(\thetaum)$ imply a small minimal adaptation loss $\popadapt^\star(\thetaum)$} and why?
\end{center}
The conditions that lead to a successful interface could depend on several quantities such as the pretraining and adaptation distributions, objectives and training methods, as well as the model architecture. This question is beyond the scope of standard generalization theory, but it does narrow us down to a few important factors specific to foundation models, and captures the essence of various important open questions on foundation models as we argue below.  

First, we note that this interface deals with population quantities that concern two \emph{different distributions}. Hence, the conditions for a successful interface are likely to involve special properties of the distributions, for example, the diversity of the pretraining distribution and structural shifts between the pretraining and adaptation data. This makes the analysis of the interface challenging (as discussed below in \refsec{theory-tools}) as we need to make careful modeling assumptions about how the two distributions relate to one another. However, this presents the possibility that tools and techniques developed to analyze such interfaces could be useful to understand the effect of distribution shifts and to predict when foundation models can improve robustness. 

Second, the population losses and possibly the conditions of a successful interface depend on the \emph{model architecture}. This raises the challenge of opening up the black-box of the neural nets. What does a small pretraining loss on a particular distribution tell us about the properties of the intermediate layers? Such analyses would also guide us in designing new adaptation methods that more carefully exploit different intermediate representations. 

Third, \emph{few-shot learning} or the sample efficiency of adaptation can be captured through the constraint on the complexity measure $C(\gamma, \thetaum) <c_0$ in the minimal adaptation loss. 
We need to formally characterize these complexity measures (\eg by understanding the implicit regularization effect of the adaptation process) and further understand why a small population pretraining loss would imply a low-complexity adaptation parameters $\gammatask$. 
A satisfactory answer to this question would likely allow us to improve the sample-efficiency of downstream adaptation.

Finally, and importantly, critical components of the interface are the choice of the \emph{pretraining and adaptation losses}. We want to understand how to best combine the pretraining and adaptation objectives for successful adaptation. It is possible that the pretraining objective that best guarantees successful adaptation differs from what is explicitly minimized during the pretraining process\dash{}the interface above allows one to use any surrogate population objective on the pretraining distribution. In addition, new surrogate objectives that provably lead to good adaptation across a broad set of tasks could shed light on the fundamental aspects that make foundation models successful. 

To summarize, the interface precludes the issue of generalization and allows us to formally reason about the interaction between several important quantities of the pretraining and adaptation phases that can guide practice in important ways.

\hypertarget{theory-incontext}{\subsubsection{Challenge: analysis of in-context learning and other emergent behavior}}
\label{sec:theory-incontext}
GPT-3~\citep{brown2020gpt3} demonstrates the power of in-context learning, an adaptation method that does not need any parameter optimization. In the adaptation phase, the pretrained language foundation model takes in a prompt\dash{}a sequence of tokens that concatenates input-output examples from the task\dash{}followed by a test example and simply generates the label of the test example by conditioning on the sequence seen thus far (prompt plus test example). 
In other words, there is no explicit training or change to the model parameters.  
What is the mechanism by which the model ``learns'' from the different examples by simply executing with the examples as inputs?
The previous modularization does not directly apply because we do not obtain new model parameters during adaptation, but rather we only use the generative capabilities of the foundation model by executing on structurally-designed inputs. 
However, the idea of separating pretraining with infinite data and pretraining with finite data can still be useful. For example, a recent work starts with the assumption of infinite pretraining data and sufficient model expressivity to study in-context learning~\citep{xie2021incontext}. These assumptions reduce the characterization of in-context learning to a matter of analyzing the pretraining distribution conditioned on in-context learning prompts, which are drawn from a different distribution than the pretraining data. In particular,~\citet{xie2021incontext} proposes that in-context learning emerges from long-term coherence structure in the pretraining distribution, which is described by a latent variable generative model with coherence structure. 
More broadly, while the modularization proposed in this section provides a nice framework to gain useful theoretical insights into foundation models, it is possible that some emergent behavior like in-context learning and other capabilities yet to be discovered would require going beyond the modularization, \eg by opening the black box of the architecture.

\hypertarget{theory-tools}{\subsubsection{Challenge: appropriate data assumptions and mathematical tools.}}
 \label{sec:theory-tools}
Understanding the interface between pretraining and adaptation phases requires a more careful study of data distributions than in traditional supervised learning. This is because the pretraining and task adaptation distributions are inherently different. By definition, foundation models are trained on raw data that is typically extremely diverse and task-agnostic, while the adaptation data depends  heavily on the task. Similarly, in-context learning emerges as a result of learning to generate data that looks like the pretraining distribution, and thereby understanding in-context learning requires careful modeling of the pretraining data. Hence answering the central questions around foundation models requires realistic and interpretable assumptions that are also amenable to analysis. 
Recent works either assume certain properties of the population data, \eg the expansion property in \citet{haochen2021spectral, wei2020theoretical}, or that the population data is generated from latent variable models with some structure~\citep{SaMa20mat,wei2021pretrained,arora2015latent,lee2020predicting,zhang20ont,tosh2021contrastive}.

We generally lack mathematical tools for relating properties of foundation models to the structure in the population data distribution.  ~\citet{haochen2021spectral} applies spectral graph theory to leverage the inner-class connectivity in the population distribution. More precise characterization of $\thetaum$ via probabilistic and analytical derivations is possible for latent variable models, but so far restricted to relatively simple ones. The community will significantly benefit from more systematic and general mathematical tools to address this question. 

It is also highly desirable to define simple toy cases so that theoreticians can precisely compare the strengths of various tools and analyses. For example, \citet{haochen2021spectral} and \citet{wei2020theoretical} consider the mixture of manifolds problem which might potentially be a good simplified test bed for vision applications. We need more interesting test beds for discrete domains such as NLP. 
We believe that tractable theoretical models which capture relevant properties of real datasets are a crucial step towards placing foundation models on solid theoretical footing.
\newsection
\hypertarget{interpretability}{\subsection{Interpretability}}
\label{sec:interpretability}
\sectionauthors{John Hewitt*, Armin W. Thomas*, Pratyusha Kalluri, Rodrigo Castellon, Christopher D. Manning}

Compared to most other machine learning models, foundation models are characterized by a vast increase in training data and complexity and the emergence of unforeseen capabilities: foundation models are able to do unforeseen tasks and do these tasks in unforeseen ways.
The increasing adoption of foundation models thereby creates growing desires, demands, and unprecedented challenges for understanding their behavior.

In contrast to task-specific models, foundation models are trained across vast and usually highly disparate datasets, potentially spanning many domains and modalities (see \refsec{training}).
Through this training, foundation models learn an exceptionally wide range of behaviors, which can vary profoundly between tasks and domains, as demonstrated by their ability to be adapted to different types of downstream tasks and to exhibit behaviors that are specific for each of these tasks (see \refsec{adaptation}).
Take GPT-3 as an example, which was trained as one huge model to simply predict the next word in a text.
While this is a very specific and simple-to-define learning task, it has enabled GPT-3 to gain capabilities that far exceed those that one would associate with next word prediction, by combining it with a vast training dataset that comprises all kinds of internet text.
As a result, GPT-3 can now adapt behaviors that are clearly outside of the scope of its original training task, such as simple arithmetic and computer programming, when provided with a few training samples.
This demonstrates that it is challenging to answer even the seemingly simplest question about a foundation model: what capabilities does it have?

\begin{figure}[!ht]
    \centering
    \includegraphics[width=\linewidth]{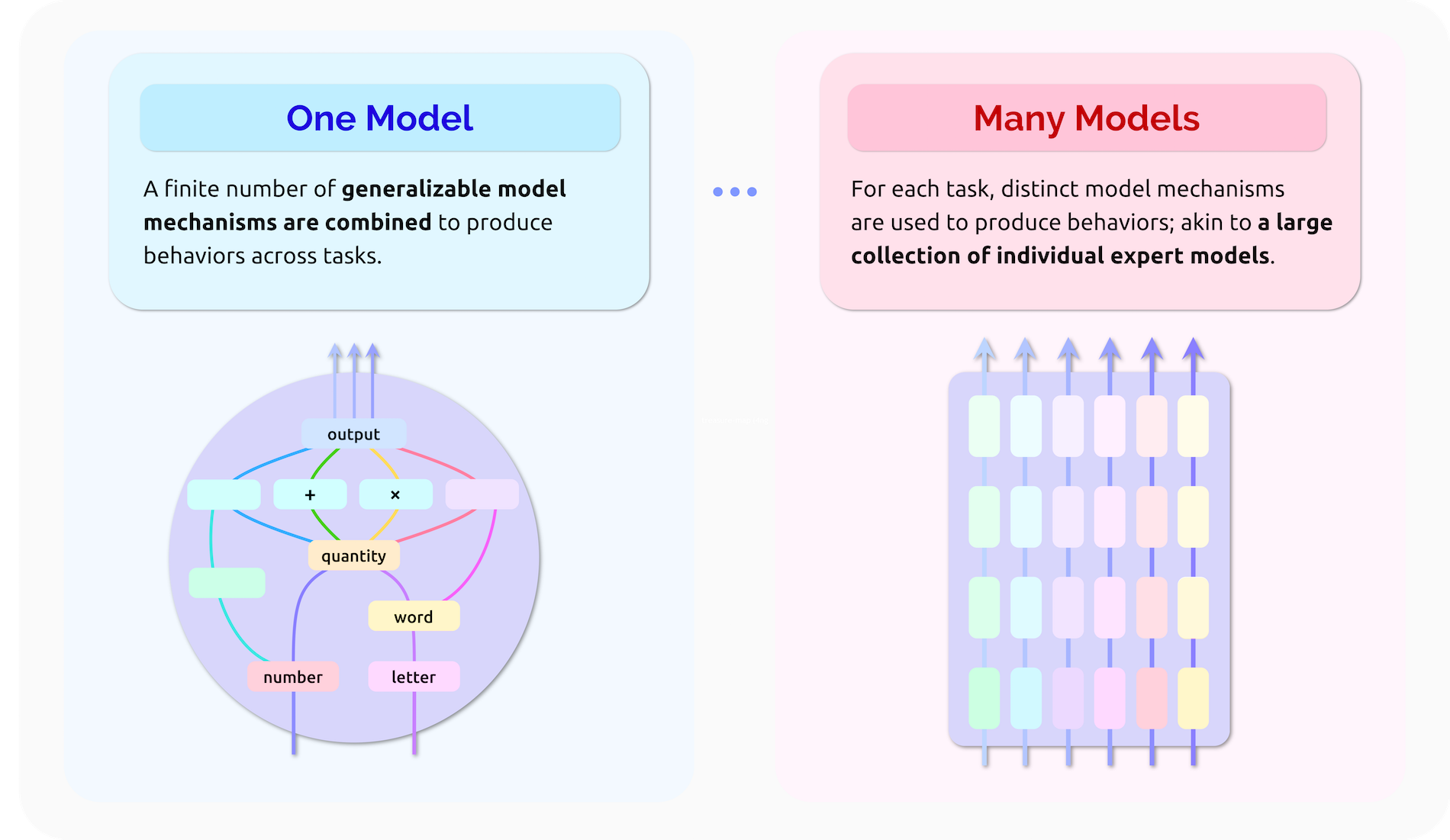}
    \caption{The one model--many model nature of foundation models: A central interpretability question is to understand where a foundation model lies on the spectrum between \textit{one model} and \textit{many models}. As one model, behavior can be made interpretable by identifying and characterising the finite number of generalizable model mechanisms used to produce behaviors across tasks (\eg mechanisms that assign meaning to words, compare quantities, and perform arithmetic). As many models, explanations of model behavior in one task are not necessarily informative about behavior in other tasks, thus requiring the independent study of behavior in each task.}
    \label{fig:intrp}
\end{figure}

Moreover, it is an open question to what extent these diverse capabilities rely on distinct or shared \textit{model mechanisms}, akin to algorithmic building blocks within the model.
On the one hand, foundation models can be interpreted as single models, which utilize some set of generalizable model mechanisms to perform well across tasks and domains.
In this case, a full understanding of their behavior can be gained by identifying and characterising these mechanisms.
On the other hand, the ability of foundation models to adapt profoundly distinct behaviors for different tasks suggests that they can also be understood as a large collection of independent expert models, each tailored to a specific task.
For example, it seems unlikely that the model parameters that GPT-3 uses to do arithmetic could have much to do with the parameters used to translate from English to French. 
In this case, explanations of model behavior in one task are therefore not necessarily informative about behavior in other tasks.
We refer to this as the \emph{one model--many model} nature of foundation models (see \reffig{intrp}) and argue that understanding where foundation models lie on this spectrum between one and many models will be central to understanding their behavior.

Toward systematizing this area of study, we present and discuss three levels of understanding foundation models
~\citep[inspired by][]{marr1982vision}:
we first discuss the challenges and opportunities in understanding \textit{what} a model is capable of doing, then \textit{why} it outputs certain behaviors, and lastly \textit{how} it does it.
Specifically, questions of \emph{what} aim to characterize the kinds of behaviors that a model can perform without peeking inside the model, while questions of \emph{why} aim to provide explanations of the model's behaviors in terms of potential causes in the data, and questions of \emph{how} aim to understand the internal model representations and mechanisms that produce these behaviors.
After presenting all three levels, we conclude by discussing potential consequences resulting from the non-interpretability and interpretability of foundation models.

\hypertarget{interpretability-behavior}{\subsubsection{Characterizing behavior}}
\label{sec:interpretability-behavior}

The simplest understanding of a technology is widely taken to be knowing \textit{what} the technology does.
This seemingly straightforward question is significantly challenging for foundation models, due to the myriad unforeseen behaviors and tasks that these models are capable of performing. 

Task-specific neural network models are trained to perform a single task in a single domain, \eg~image classification.
Their task and the input and output domains are therefore clear; yet even for these models it can be challenging to know exactly what the model will do, given a particular input. For instance, model behaviors can unexpectedly differ greatly for two perceptually similar inputs~\citep{garg2020bae,jin2020bert} or two subpopulations of the same data (stratified, for example, by race or gender~\citep{hovy2015,blodgett2016,tatman2017,buolamwini2018gender}).

This challenge of characterizing a model's behavior is amplified manyfold for foundation models.
The space of tasks that the model is able to perform is generally large and unknown, the input and output domains are often high-dimensional and vast (\eg language or vision), and the models are less restricted to domain-specific behaviors or failure modes.
Consider, for example, the surprising ability of GPT-3 to be trained on large language corpora and to subsequently develop the ability to generate mostly-functional snippets of computer programs. 
A key challenge for characterizing the behavior of foundation models is therefore to identify the capabilities that it has.
Even further, for each task that a foundation model can perform, and there may be many or infinitely many, all the challenges remain that one faces when trying to understand the behavior of much simpler, task-specific models.

Characterizing each `task' that a foundation model can perform is further complicated by their one model--many models nature (see \reffig{intrp}).
Again taking GPT-3 as an example, it was shown that it can be tailored to many tasks through simple prompting (see \refsec{adaptation}).
Yet, each task can be specified through many possible prompts and slight variations in prompts can result in meaningful changes of model behavior.
For instance, the task of sentiment classification of a movie review can be specified by presenting the movie review followed by `Her sentiment towards the film was...' or `My overall feeling was that the movie was...'; despite these prompts appearing to pose closely related tasks, GPT-3 will exhibit different response accuracies for each prompt~\citep{Zhao2021calibrate}.
Observations like these raise important questions regarding the relationship between the characteristics of prompts and the resulting model behaviors.
Specifically, can meaningfully different responses to seemingly similar prompts actually be considered as resulting from the same model or do they result from highly distinct model mechanisms, and does characterizing the behaviors of the foundation model (or its adapted derivatives) in one task truly aid in characterizing the behaviors of other possible adaptations of the model?

To identify the capabilities that a foundation model has and those it is missing, researchers can utilize controlled  evaluations.
Here, domain experts design prompts that are known to require a particular competence and then study the ability of a model to respond correctly to these prompts ~\citep{papadimitriou2020learning,DBLP:journals/corr/abs-2103-05247,kataoka2020pre,wu2021identifying,xie2021innout,koh2021wilds}.
For example, psycholinguists have designed prompts that require a language model to choose between a grammatically correct sentence and the same sentence with a specific  grammatical inaccuracy; knowing whether the model consistently prefers the grammatically correct sentence over its grammatically incorrect counterpart tells us whether the model has the particular grammatical competence required to identify this inaccuracy~\citep{linzen2016assessing}.

Given the huge range of possible capabilities of foundation models, and our current lack of any general method for determining a priori whether a foundation model will have a given capability, bespoke evaluations like these are crucial.
They allow exploring the range of behaviors that foundation models are capable of, while requiring minimal model access: we only need to present inputs and receive model outputs, and we need not depend on access to the implementation or parameters of a model.
Given the infinitely many desirable and undesirable tasks, subtasks, and behaviors that foundation models may be capable of (or incapable of), characterizing model behaviors and capabilities will be increasingly challenging and important.
We believe that instead of relying on a few experts to formulate and test for possible behaviors, it will be critical to extend these types of analyses to test for many more behaviors, in part by opening up this line of exploration to diverse communities and experts in many disciplines, as well as by increasing access to and scale of these evaluations.

\subsubsection{Explaining behavior}

In addition to characterizing what a foundation model is doing, one can try to characterize \textit{why} it performs certain behaviors by providing explanations of these behaviors in terms of potential causes in the data.
While current explanation approaches, which provide such explanations of behavior, can reveal qualities of inputs that affect a model's responses, they often require full access to the model to do so and are generally limited in their ability to elucidate any general model mechanisms, which foundation models use to respond to many inputs, tasks, and domains.

Current explanatory approaches can generally be understood as distinct models, which are designed to provide an explanation of particular behaviors of another \textit{black box} model.
Importantly, these approaches are separate from the model whose behavior is analyzed, which by itself is not interpretable.
This separation can be problematic, as the provided explanations can lack faithfulness \citep{jacovi2020towards}, by being unreliable and misleading about the causes of a behavior \citep[cf.][]{rudin2019stop}.
Even further, unsound explanations can entice humans into trusting unsound models more than they otherwise would (for a detailed discussion of trust in artificial intelligence, see~\citet{jacovi2021formalizing}).
These types of concerns grow as we transition from task-specific models towards the wide adoption of foundation models, as their behavior is vastly more complex.

Current explanatory approaches can largely be divided into either providing \textit{local} or \textit{global} explanations of model behavior~\citep{doshi2017towards}.
Local explanations seek to explain a model's response to a specific input, \eg by attributing a relevance to each input feature for the behavior or by identifying the training samples most relevant for the behavior \citep{simonyan2013deep,bach2015pixel,sundararajan2017axiomatic,shrikumar2017learning,springenberg2014striving,zeiler2014visualizing,lundberg2017unified,zintgraf2017visualizing,Fong_2017_ICCV,koh2017understanding}. Global explanations, in contrast, are not tied to a specific input and instead aim to uncover qualities of the data at large that affect model behaviors, \eg by synthesizing the input that the model associates most strongly with a behavior \citep{simonyan2013deep,nguyen2016synthesizing}. 

Local and global explanations have provided useful insights into the behavior of task-specific models~\citep[\eg][]{li2015visualizing,wang2015visual,lapuschkin2019unmasking,thomas2019analyzing,poplin2018prediction}.
Here, the resulting explanations are often taken to be a heuristic of the model mechanisms that gave rise to a behavior; for example, seeing that an explanation attributes high importance to horizontal lines when the model reads a handwritten digit `7' easily creates the impression that horizontal lines are a generally important feature that the model uses to identify all sevens or perhaps to distinguish all digits.

Given the one model--many models nature of foundation models, however, we should be careful not to jump from specific explanations of a behavior to general assumptions about the model's behavior.
While current explanatory approaches may shed light on specific behaviors, for example, by identifying aspects of the data that strongly effected these behaviors, the resulting explanations do not necessarily provide insights into the model's behaviors for other (even seemingly similar) inputs, let alone other tasks and domains.

Another approach could be to sidestep these types of post-hoc explanations altogether by leveraging the generative abilities of foundation models in the form of self-explanations~\citep[cf.][]{elton2020self,chen2018looks},
that is, by training these models to generate not only the response to an input, but to jointly generate a human-understandable explanation of that response.
While it is unclear whether this approach will be fruitful in the future, there are reasons to be skeptical: language models, and now foundation models, are exceptional at producing fluent, seemingly plausible content without any grounding in truth.
Simple self-generated “explanations” could follow suit.
It is thus important to be discerning of the difference between the ability of a model to create plausible-sounding explanations and providing true insights into its behavior.

\subsubsection{Characterizing model mechanisms}

Deep understanding of systems is generally taken to mean understanding \textit{how} a system performs: which knowledge and mechanisms does it contain, and how are these assembled to form the whole?

If this is indeed possible, characterizing the representations within foundation models and the mechanisms that operate on them will be central to satisfying the desire to thoroughly understand these proliferating models; and whether these mechanisms are many and specific or few and generalizable, they are at the core of the ability of foundation models to adopt a wide range of behaviors in varied tasks and domains.

To make the notions of model representations and mechanisms concrete, consider a simple behavior exhibited by GPT-3: 
It was quickly observed \textit{what} GPT-3 did when provided with examples of the addition of small numbers and then queried to perform addition of two new numbers: with high probability, it predicted the correct result of the addition \citep{branwen2020gpt,brockman2020math}.
When asking \textit{why} GPT-3 performed as it did, one could find evidence in the input, like aspects of its prompt that highly affected its response (these might be the two numbers to be added, though not necessarily), or aspects of GPT-3’s training data that affected its response (these might be examples of addition, though not necessarily).
Delving into the model, we may envision a deeper understanding of the mechanisms that GPT-3 uses to add a specific pair of numbers and the mechanism that it uses to add other arbitrary pairs of numbers.
We may also envision a deeper understanding of whether these mechanisms are similar to the mathematical notion of `addition' or merely correlated with this notion.

By understanding individual model mechanisms, we can build up a compositional understanding of complex behaviors of a foundation model.
A task slightly more complex than the addition of numbers is solving mathematical word problems, in which numbers come with units and the problem is presented in natural language. 
Once we understand the mechanism (or mechanisms) by which a model performs addition, we can investigate whether this mechanism is used as an intermediate step in solving word problems.
If the addition mechanism is used, we have built up our understanding of how the model solves word problems, we have increased confidence that the foundation model generalizes the notions of quantities and addition (not another correlation or heuristic), and, furthermore, we have increased confidence in our ability to predict the model's \textit{why} (which parts of the inputs it is attending to) and the output's \textit{what} (addition of two numbers).
If the addition mechanism is not used, we may retain a healthy skepticism that this is truly addition, and we can investigate which representations and mechanisms are used instead.

It is important to be aware that there are many potential cases of more complex and concerning model mechanisms, for instance, the estimation of race from the characters in a name, or the pixels in an image.
Establishing evidence of such a mechanism in a foundation model and its use can support a moral or legal responsibility to ban the model from tasks like predictive policing, marketing, loan applications, and surveillance at large.

A plethora of methods have emerged to investigate these internal aspects of neural network models. Typically, these approaches separate the model into nodes (\eg neurons, layers, or parts of layers), then interrogate either the representations captured in nodes or the mechanisms by which nodes are assembled. 
Some approaches are hypothesis driven: by hypothesizing that nodes may capture certain information (\eg a grammatical feature of a word, or the race of a person), one can probe all nodes to quantify how much of that information they make available \citep{alain2016understanding,veldhoen2016diagnostic,belinkov2017what,adi2017finegrained,conneau2018what,hewitt2019control,hewitt2019structural,voita2020informationtheoretic,pimentel2020information}. 
Other approaches build on explanatory methods, and, instead of identifying which data cause a certain behavior, they seek to identify which data cause a certain node to activate, or which nodes cause another node later in the model to activate, thereby uncovering collections of model representations and mechanisms
\citep{olah2020zoom,mu2020compositional,carter2019activation,goh2021multimodal}.
Taken together, these approaches inspect the interior of models and provide a 
basis for the ongoing explorations of the behavior of foundation models. 
Yet, the number of potential representations and mechanisms within foundation models is vast, particularly given their one model--many models nature, and these types of approaches often only capture a small slice of a model's interiority.
It is thus an open challenge to expand the discovery of representations and mechanisms and to elucidate those that are most relevant or general for model behavior.
As with many approaches to interpreting foundation models, these types of explorations will benefit from including and supporting more diverse and interdisciplinary investigators and from  more accessible, flexible, and scalable methods of discovery.

In summary, we believe that the one model--many models nature of foundation models (recall \reffig{intrp}) provides novel opportunities and challenges for current interpretability research: there are many adaptations of a single foundation model, and we simply do not know the extent to which they share common mechanisms.
To the extent that mechanisms are shared, understanding foundation models may be a tractable problem of characterizing these mechanisms and their relations.
To the extent that mechanisms are independent, each adaptation of a foundation model must be analyzed independently, leading to profound uncertainty about the nature of any new adaptation of the foundation model.

\hypertarget{interpretability-impacts}{\subsubsection{Impacts of non-interpretability and interpretability}}
\label{sec:interpretability-impacts}

Lastly, we would like to highlight that the wide adoption of foundation models is at odds with a recent plea of many interdisciplinary researchers not to use complex black box models for high stakes decisions~\citep[\eg][]{rudin2019stop}, but instead to focus on the long-standing development and application of more intrinsically interpretable models. 

In the midst of these pleas, work aimed at interpreting foundation models is a double-edged sword.
Large machine learning models, and now foundation models, are most often deployed by powerful corporations and institutions, and incremental advances in interpretability can be exaggerated to `ethics-wash' and continue use of models as though they have \textit{achieved} interpretability, belying the reality that they remain far below traditional standards of algorithmic interpretability.
Moreover, when approaches to interpretability regularly presume easy access to models and their implementation and parameters, interpretability can serve not only as cover for powerful institutions but also centralize model knowledge in the same hands.
For those working toward the interpretability of foundation models, it is a responsibility to consistently ask whether one is working toward 
making foundation models \textit{interpretable to researchers and model owners} or \textit{interpretable to everyone}. 

Simultaneously, to the extent that foundation models are already being deployed, work on interpretability presents unique opportunities to shift knowledge of foundation models, and thus power, back to datafied and evaluated peoples. 
Interpretation can facilitate the discovery of societally salient aspects of models.
More radically, work creating accessible methods that allow anyone to interpret the behavior of foundation models shifts power to diverse peoples, creating opportunities to investigate models, opportunities to discover aspects of models important to individuals or their communities, and opportunities to meaningfully consent to, improve, or altogether contest the use of foundation models.
Finally, it is important for researchers to view the interpretability of foundation models as not only a goal, but a question: research can explore and assess whether the lack of foundation model interpretability is intrinsic and should be deeply studied and widely known as a serious issue discouraging use (or increasing regulation) of these systems, or whether it is possible for future foundation models to uphold a high standard of interpretability for all.

\clearpage
\hypertarget{society}{\section{Society}}
\label{sec:society}

The societal impact of foundation models, referring both to the construction of the models themselves and their role in developing applications, requires careful examination.
Specifically, we anticipate that foundation models will have wide-ranging societal consequences that are challenging to understand: foundation models are intermediary assets that are not directly deployed, but rather serve as a foundation that is further adapted.
As a result, traditional approaches to reasoning about the societal impact of technology are likely complicated; societal impact is easier (but still difficult) to grasp for systems with well-specified purposes.
In this chapter, we discuss how we may grapple with and beginning to understand the complexity of the societal impact of models foundation models.
Specifically, we discuss (i) the harms with respect to inequity (\refsec{fairness}) and misuse (\refsec{misuse}), (ii) the impact with respect to the economy (\refsec{economics}) and environment (\refsec{environment}), and (iii) the broader considerations with respect to the law (\refsec{legality}) and ethics (\refsec{ethics}). 

\pl{Flesh this out more to talk about:
(i) overall idea of social benefit, as distinct from corporate economic benefit;
(ii) inequality as being a recurring theme;
and anything else that's recurring
}

\newsection
\hypertarget{fairness}{\subsection{Inequity and fairness}}
\label{sec:fairness}
\sectionauthors{Rishi Bommasani, Fereshte Khani, Esin Durmus, Faisal Ladhak, Dan Jurafsky}

\begin{figure}[!ht]
    \centering
    \includegraphics[width=\linewidth]{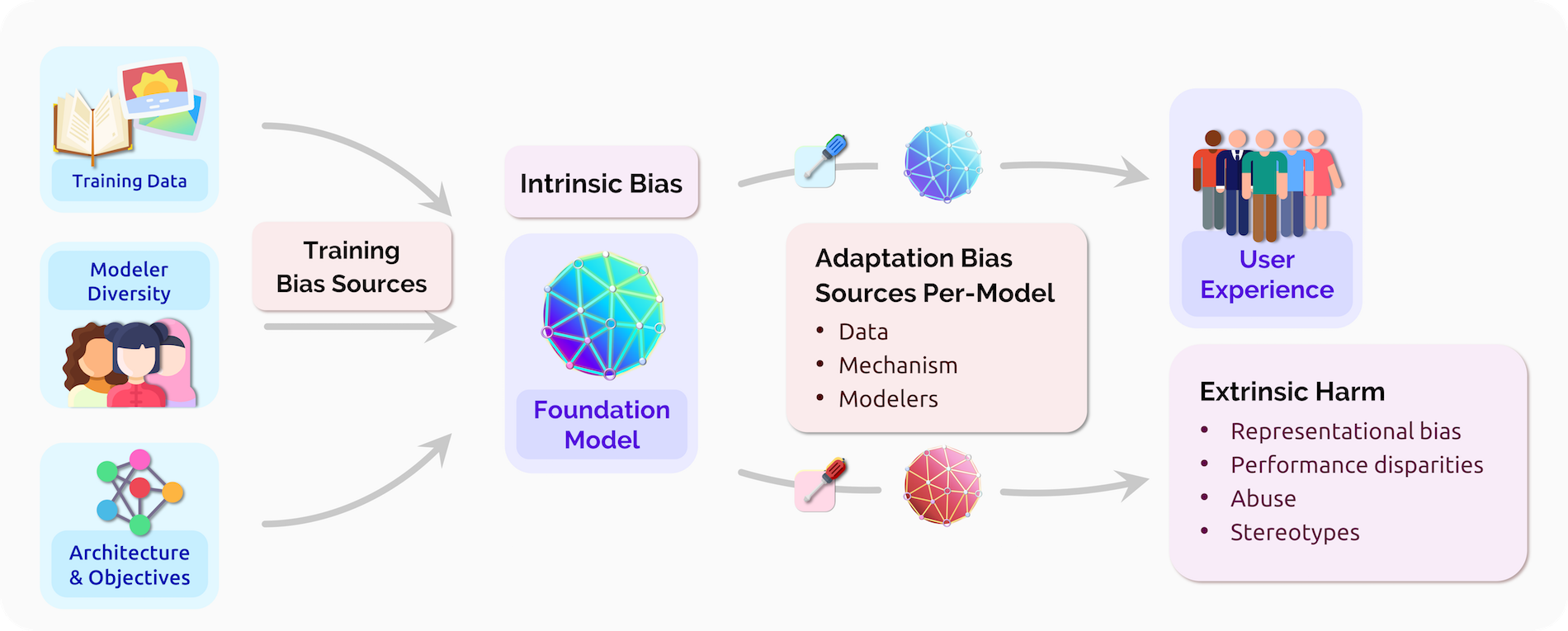}
    \caption{The \textit{intrinsic bias} present within foundation models is the byproduct of various training bias sources (\textbf{left}) which, alongside biases introduced during adaptation, determines the \textit{extrinsic harms} (\textbf{right}) experienced by users in the context of specific downstream applications. We emphasize that the same foundation model is the shared foundation for many different applications; its biases propagate to these many applications as a result. Further, since the harms experienced by users are the result of specific adapted models, attributing these harms to the various processes and sources depicted in this diagram is both crucial and challenging.}
    \label{fig:fairness}
\end{figure}

\hypertarget{fairness-introduction}{\subsubsection{Introduction}}
\label{sec:fairness-introduction}
Foundation models have the potential to yield inequitable outcomes: the treatment of people that is unjust, especially due to unequal distribution along lines that compound historical discrimination \citep{Hellman2021Big}. 
Like any AI system, foundation models can compound existing inequities by producing unfair outcomes, entrenching systems of power, and disproportionately distributing negative consequences of technology to those already marginalized \citep{sweeney2013,kay2015,buolamwini2018gender,benjamin2019,ajunwa2019paradox,datafeminism,crawford2021}. 
Here we ask what fairness-related harms relate to foundation models, what sources are responsible for these harms, and how we can intervene to address them. 
The issues we discuss here are related to broader questions of algorithmic fairness and AI ethics \citep{CorbettDavies2018, Chouldechova2020, Hellman2020, Johnson2020, fazelpour2021bias}, race and technology \cite{benjamin2019, Hanna2020, Gebru2021, field2021}, and the coexistence of society and technology \citep{Abebe2020}.

\hypertarget{fairness-harms}{\subsubsection{Harms}}
\label{sec:fairness-harms}
Foundation models are intermediary assets with no specified purpose before they are adapted; understanding their harms requires reasoning about both their properties and the role they play in building task-specific models. 
We delineate \textit{intrinsic} biases,\footnote{We use the word \textit{bias} to denote the properties of a foundation model that contribute to inequity; we follow \citet{blodgett_language_2020} in attempting, when possible, to delineate who is harmed and how they are harmed.} \ie~properties of the  foundation model that indirectly but pervasively affect downstream applications, and \textit{extrinsic harms}, \ie~harms that arise in the context of specific downstream applications \citep{galliers1993}.

\paragraph{Intrinsic biases.}
Properties of the foundation model can lead to harm in downstream systems.
As a result, these intrinsic biases can be measured directly within the foundation model, though the harm itself is only realized when the foundation model is adapted, and thereafter applied, \ie~these are \textit{latent} biases or harms \citep{DeCamp2020}.
We focus on the most widely studied form of intrinsic bias, \textbf{representational bias}, specifically considering misrepresentation, underrepresentation and overrepresentation. 
People can be \textbf{misrepresented} by pernicious stereotypes \citep{bolukbasi2016, caliskan2017, abid2021,stereoset,gehman-etal-2020-realtoxicityprompts} or negative attitudes \citep{hutchinson2020}, which can propagate through downstream models to reinforce this misrepresentation in society \cite{noble,benjamin2019}. 
People can be \textbf{underrepresented} or entirely erased, \eg~when LGBTQ+ identity terms  \citep{Strengers2020, Oliva2021, Tomasev2021} or data describing African Americans \cite{buolamwini2018gender,koenecke2020racial,blodgett17} is excluded in training data, downstream models will struggle with similar data at test-time.
People can be \textbf{overrepresented}, \eg~BERT appears to encode an Anglocentric perspective \citep{zhou2021} by \textit{default}, which can amplify majority voices and contribute to \textit{homogenization} of perspectives \citep{creel2021} or monoculture \citep{kleinberg2021} (\refsec{ethics}). 
These representational biases pertain to all AI systems, but their significance is greatly heightened in the foundation model paradigm.
Since the same foundation model serves as the basis for myriad applications, biases in the representation of people propagate to many applications and settings.
Further, since the foundation model does much of the heavy-lifting (compared to adaptation, which is generally intended to be lightweight), we anticipate that many of the experienced harms will be significantly determined by the internal properties of the foundation model.

\paragraph{Extrinsic harms.}
Users can experience specific harms from the downstream applications that are created by adapting a foundation model.  
These harms can be \textbf{representational}  \citep{barocas17,crawford17,blodgett_language_2020}, such as the sexualized depictions of black women produced by information retrieval systems \citep{noble}, the misgendering of persons by machine translation systems that default to male pronouns \cite{schiebinger13,schiebinger14}, or the generation of pernicious stereotypes \citep{nozza21,sheng-etal-2019-woman, abid2021}. 
They can consist of \textbf{abuse}, such as when dialogue agents based on foundation models attack users with toxic content \cite{dinan21,gehman-etal-2020-realtoxicityprompts} or microaggressions \citep{breitfeller-etal-2019-finding,jurgens-etal-2019-just}.
All of these user-facing behaviors can lead to psychological harms or the reinforcement of pernicious stereotypes \cite{spencer2016stereotype,williams2020psychology}.

In addition to harms experienced by individuals, groups or sub-populations may also be subject to harms such as group-level \textbf{performance disparities}.
For example, systems may perform poorly on text or speech in African American English \citep{blodgett17,koenecke2020racial}, incorrectly detect medical conditions from clinical notes for racial, gender, and insurance-status minority groups \citep{zhang2020hurtful}, or fail to detect the faces of people with darker skin tones \citep{wilson19,buolamwini2018gender}. 
As foundation models are more pervasively applied, including in high-stakes domains, these disparities can spiral into further, and more severe, harms. 
\citet{koenecke2020racial} discuss how if African American English speakers cannot reliably use speech recognition technologies (\eg~due to inequities in underlying foundation models), this may mean they cannot benefit from certain derivative products (\eg~voice assistants, assistive technologies) and will be disadvantaged if these technologies are used to conduct interviews for employment or transcribe courtroom proceedings.
More generally, characterizing these group-level harms (and working towards justice for those harmed) also requires the AI community to improve its understanding of group-based prejudice \citep{allport1954} and social groups: we point to relevant work in the social sciences and other communities on moving beyond binary treatments of gender \citep{lindsey2015, westbrook2015, richards2017, darwin2017, Keyes2018, hyde2019, cao2020, dinan2020}, more nuanced treatments of race \citep[\eg][]{penner2008, freeman2011,  saperstein2012, saperstein2013, penner2015, field2021}, better addressing intersectional identities \citep[\eg][]{crenshaw1989, nash2008, gines2011, penner2013, ghavami2013, bright2016, buolamwini2018gender, may2019, oconnor2019, guo2020}, and more modern treatments of disability \citep[\eg][]{batterbury2012, spiel2019, hutchinson2020}. 

\paragraph{Additional considerations.}
To more completely understand the harms of foundation models, further \textit{documentation} is required of both the intrinsic biases and extrinsic harms; future work should articulate the relationship between intrinsic biases and extrinsic harms \citep{blodgett_language_2020, blodgett2021, goldfarb-tarrant2021}.
This documentation requires centering stakeholders beyond academics and industry practitioners: the inequitable impact of foundation models will be experienced largely by minority populations, which are underrepresented in both academia and industry. 
For foundation models specifically, their creation and study likely will be conducted by those with the access and resources required, further emphasizing the importance of venues that center marginalized voices \citep[][\refsec{ethics}]{datafeminism}.
In particular, user studies of specific adapted models, when aggregated across applications, can provide compelling and individualized documentation of the harms that derive from the intrinsic biases of foundation models, all while centering individual users.
In this way, we imagine the methodologies in human-computer interaction (HCI), with some adjustment to accommodate the abstraction involved in foundation models, will help center the voices of marginalized communities (further discussion in \refsec{interaction}). 

\hypertarget{fairness-sources}{\subsubsection{Sources}}
\label{sec:fairness-sources}
In order to fully characterize and properly intervene on the harms of foundation models, we must be able to trace their source to the properties of the foundation model and the adaptation process, and further decompose to the roles of individual sources of biases \citep{friedman1996}. 
Source tracing is vital for attributing ethical and legal responsibility for experienced harm, though attribution will require novel technical research that foregrounds matters such as causality \citep{pearl2000causality} and influence \citep{koh2017understanding}.

\paragraph{Data.}
Data of several types shapes the behavior of applications, and the associated extrinsic harms, based on foundation models: the training data used to train the foundation model, the adaptation data used to adapt the foundation model, and test-time user data/interaction.
For all of these data sources, the properties of the data (\eg~toxicity and hate speech \citep{henderson17ethical},  abusive language \citep{waseem-etal-2017-understanding}, microaggressions \citep{breitfeller-etal-2019-finding}, stereotypes \citep{voigt-etal-2018-rtgender}) will manifest in the biases of the foundation model (and its adapted derivatives).\footnote{In adaptation, which involves labelled task-specific data, biases in the choices of the label space \cite{crawford2021} and biases in the annotators who label that data \citep{geva-etal-2019-modeling,sap-etal-2019-risk} can also contribute to extrinsic harms experienced by users.}
Since the training data is the key data source that determines the foundation model and the associated intrinsic biases, we focus on the training data here.
At present, the relationship between the training data, along with associated data practices (\eg~data curation, data selection, and data weighting \citep{paullada2020, bender2021, rogers2021}) and the intrinsic biases acquired by the foundation model remains unclear; future work is critically needed to clarify this relationship.
Since foundation models generally require training data of immense scale, which poses clear challenges not only to its documentation \citep{bender2021} but also comprehensive scientific exploration to articulate the relationship of data biases and model biases, we anticipate new protocols are required to address this scale.
Establishing scaling laws for bias, akin to those for accuracy metrics \citep{kaplan2020, henighan2020}, may enable systematic study at smaller scales to inform data practices at larger scales.

\paragraph{Modeling.} 
Modeling decisions (\eg~training objective (\refsec{training}), model architecture (\refsec{modeling}), adaptation method (\refsec{adaptation}))  influence the biases in foundation models and their derivatives, thereby affecting the experienced extrinsic harms.  
Existing work demonstrates that foundation models amplify training data biases, extending trends seen for machine learning and deep learning models \citep{zhao-etal-2017-men,wang2019balanced,jia-etal-2020-mitigating,hashimoto2018fairness}, though much still remains unclear about what and how model properties are responsible for this bias amplification.
Further, given that applying foundation models directly may be infeasible (due to their scale), efforts to compress these models or make them more efficient also appear to amplify bias \citep{hooker2020, renduchintala2021}.
Amplification may also be exacerbated by feedback loops, in which foundation models modify societal behavior and induce sociological changes, which modifies subsequent training data; feedback effects of this form tend to exacerbate inequity in other  ML applications \cite{lum2016predict,ensign2018runaway,hashimoto2018fairness}.
Beyond the explicit decisions made in training and applying foundation models, community values \citep{birhane2020} and norms (\refsec{ethics}) both indirectly and implicitly \citep{liu2021} shape decision-making in building models.
As a result, measuring biases in conjunction with work introducing foundation models \citep[\eg][]{brown2020gpt3} and in standard benchmarks \citep[][\refsec{evaluation}]{friedman1996}, as well as conducting user studies with diverse user groups to document experienced harm, are steps towards ensuring that best practices actively emphasize the consideration of bias and inequity.
\paragraph{Modelers.}
As with all algorithmic systems, poor representation and diversity of stakeholders and marginalized communities in decision-making bodies that develop or apply foundation models is inherently problematic, and may contribute to greater experienced harm for these communities.\footnote{We note that diversity, both with respect to disciplinary backgrounds and demographic identities, is of fundamental importance in these high-impact decision-making settings for reasons well beyond the potential improved recognition of fairness-related harms.} 
While difficult to document, existing efforts to develop foundation models suggest this as a possibility: 
\citet{caswell2021} demonstrate the flawed data handling of less-represented languages in the multilingual datasets used to train multilingual models and \citet{hutchinson2020} show that models often contain undesirable biases towards disabled persons. 
In both instances, these biases and harms may have been noticed earlier by better representation of these parties in developer teams. 
Further, since end-users are likely more diverse than developers and may notice these concerns earlier, allowing for user feedback to contribute to foundation model design (\refsec{interaction}) is an important direction forward.

\hypertarget{fairness-recourse}{\subsubsection{Interventions and recourse}}
\label{sec:fairness-recourse}

Addressing, mitigating, and rectifying the inequities associated with technology requires  integrating social and technical methodologies \citep{Abebe2020}.
 For foundation models specifically, we consider both proactive methods, which change how models are developed and deployed to prophylactically reduce harm, as well as reactive methods, which respond to harm and make changes for the future.
 At its core, the abstraction of foundation models complicates both aspects: knowing if interventions at the level of the foundation level are successful in reducing harm requires downstream observations at the level of specific deployed applications and recourse in the event of harm requires upstream propagation of both feedback and accountability to foundation model providers.
 
\paragraph{Intervention.}
General principles that govern intervention on technological systems apply to the foundation model setting: identifying which sources are most responsible for bias or harm provides the evidence required for targeted action.
For example, the urgency of calls for improved diversity in the teams that design, produce, and control technology (\eg~foundation models) and their applications \citep{Longino1990, Harding2015, Nielsen2017, oconnor2019, Hofstra2020, Katell2020} is further intensified if the lack of diversity is shown to relate to harm \citep{caswell2021}. 
In addition, transparent documentation \citep[\eg][]{gebru2018datasheets, bender2018data, Mitchell_2019} and auditing \citep[\eg][]{raji2019} are similarly critical in providing the impetus for intervention and change \citep{Burrell2016, Lipton2018, Creel2020, Raji2020, Wilson2021}.
The scale of foundation models, as well as the specifics of their accessibility, introduce new challenges for existing protocols for documentation and auditing that we discuss further in \refsec{ethics}.

To date, many of the interventions considered for reducing the inequitable impact of technology, including in the foundation model regime, are methods for technical mitigation that center the data (to obviate reflecting inequities or biases) and modelling decisions (to avoid amplifying data biases) involved.
Of specific importance in the foundation model regime is recognizing that these mitigation approaches may target different steps in the pipeline such as the training data \citep[\eg][]{lu2020}, modelling objectives \citep[\eg][]{zhao2018}), and adaptation methods and test-time use \citep[\eg][]{park2018,zhao2019}.
As a result, different approaches may not only be more or less effective, but require action from different entities (\eg~foundation model providers vs. application developers) and more or less intensively affect the expensive training process for these models (\eg~changing the process of creating a foundation model vs. altering it \textit{post hoc}).
Technical intervention of this form may also target different goals: some interventions, such as changing the training data, aims to reduce \textit{intrinsic bias}.
On the other hand, most work on mitigation in algorithmic/ML fairness instead considers reducing outcome disparities in terms of model behavior, \ie~the outputs of downstream systems that more directly relate to \textit{extrinsic harm}. 
Technical mitigation of all forms at present is severely limited: methods that measure or combat intrinsic bias are brittle or ineffectual \citep{gonen19, ethayarajh2019, bommasani2020, zhou-etal-2021-challenges, Antoniak2021}, methods that measure or combat extrinsic outcome disparities may not align with stakeholder goals \citep{saha2020}, and there is some evidence to suggest certain types of technical intervention may be simultaneously unsatisfiable \citep{CorbettDavies2018, kleinberg2017}, impossible \citep{lechner2021}, or may even exacerbate inequity \citep{xu-etal-2021-detoxifying}.
In spite of this state of affairs, we continue to believe technical methods will still play an instrumental role in addressing the harms that arise in the foundation model regime; in general, we advocate for transparency, especially given that technical mitigation methods may not be able to achieve the intended goals.
More broadly, claims of bias and bias mitigation must be made carefully to clearly communicate the status quo to various stakeholders with differing expertise (\eg~application developers building on top of foundation models and policymakers regulating the technology; \citep{nissim2020}). 

\paragraph{Recourse.}
Unfortunately, proactive intervention is unlikely to fully resolve all potential harm or inequity that may arise due to foundation models.
When harm arises, there is currently no widely-adopted (or legally required) framework for resolving the appropriate recourse for the harmed parties.
While certain protocols may exist for specific applications, the abstraction of foundation models again introduces a disconnect: harms likely are partially attributable to both the foundation model providers and the downstream application developers, but allocating this responsibility to either party remains challenging.
More simply, mechanisms are not in place to even communicate these harms to foundation model providers (even if feedback or complaints are raised to application developers).
As a result, new norms and standards are needed on  how  feedback from  application developers and end-users should reach upstream to the foundation model providers, how to determine the entities (\eg~foundation model providers, application developers) responsible for these harms, and the relationship to legal responsibility (\refsec{legality}).
To make progress on this matter, we encourage future work to consult the practices used in other domains (especially those with similar abstractions and multi-entity structures), and we anticipate any standards introduced will likely need to be reasonably dynamic, so that they can be synchronized with the rapidly changing status quo for these models and their applications.

\hypertarget{fairness-takeaways}{\subsubsection{Takeaways}}
\label{sec:fairness-takeaways}
Machine learning has an established trackrecord of inequitable impact, with much of the burden of its harms borne by marginalized communities.
Foundation models introduce new challenges to this calculus but, ultimately, for their societal impact to be equitable, significant research and change is required to understand the harms they cause and to meaningfully address and rectify these harms:
\begin{enumerate}
    \item 
    The one-to-many nature of foundation models, \ie~the same few foundation models being used across many applications, means the intrinsic properties of foundation models pervade to many downstream applications. 
    Pernicious biases in these models therefore have out-sized effect on the experienced harms.
    \item 
    Biases and harms in the foundation model regime originate from many sources (\eg~training and adaptation data, modelling and adaptation decisions, modeler diversity and community values). 
    Attributing the sources for bias and harm is fundamental for questions of intervention and responsibility; attribution requires new technical research to be done reliably.
    \item 
    The inequities of foundation models are not inevitable, but addressing them requires a multi-pronged approach comprised of both proactive intervention (\eg~data-centric and model-centric changes) and reactive recourse (\eg~mechanisms for feedback and accountability).
\end{enumerate}

\newsection
\hypertarget{misuse}{\subsection{Misuse}} 
\label{sec:misuse}
\sectionauthors{Antoine Bosselut*, Shelby Grossman*, Ben Newman}

\begin{figure}[!ht]
\centering
\includegraphics[width=\linewidth]{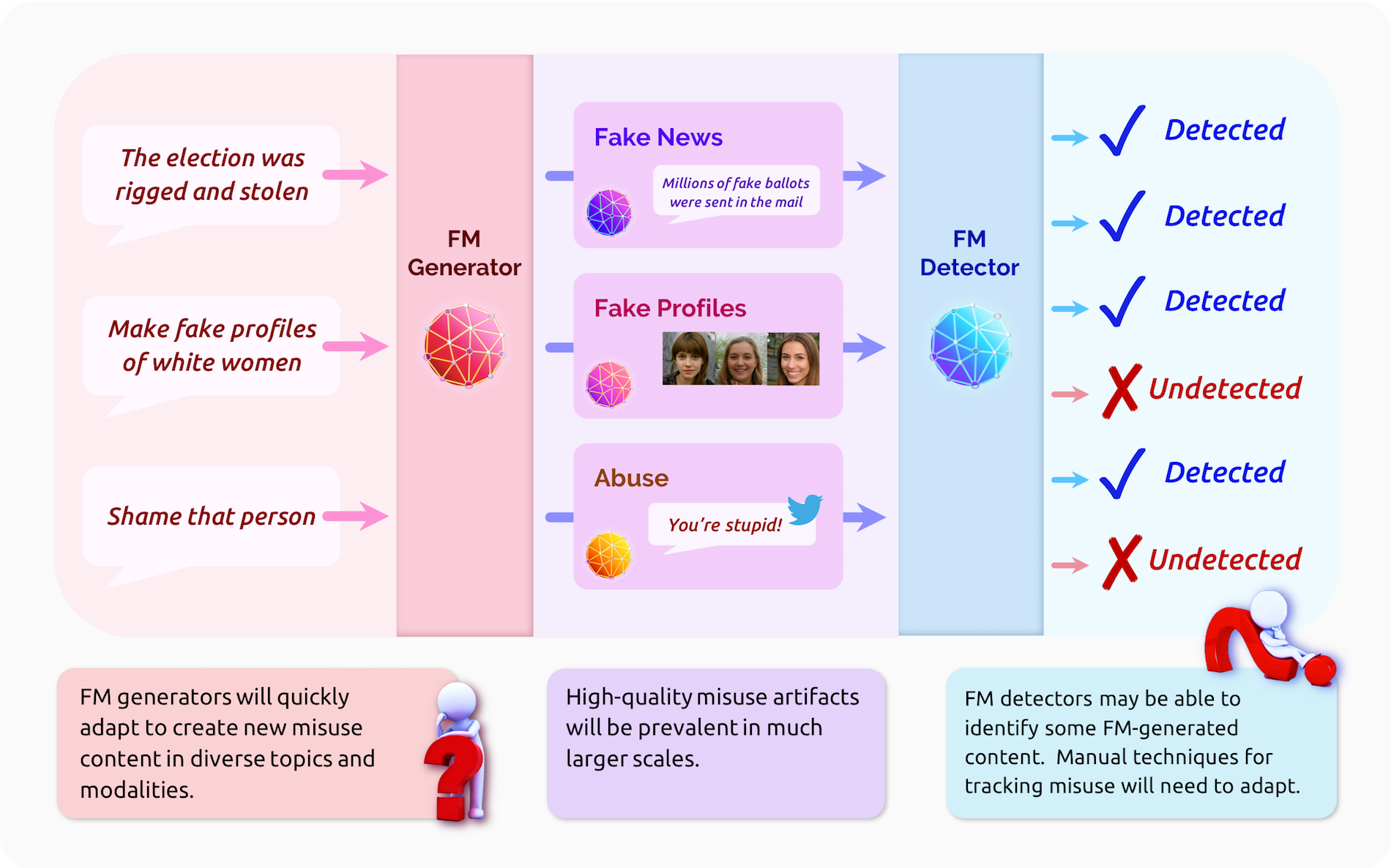}
\caption{\label{fig:misuse} This figure shows the effect foundation models will have on manipulative and harmful content generation, and the implications for detection. }
\end{figure}

In this section, we consider misuse of foundation models\dash{}situations where people use foundation models as they are intended to be used (\eg~to generate language), but where their capabilities are intentionally leveraged to cause harm to populations or individuals. This definition positions misuse concerns 
between those of inequity (where models can cause harm without bad intentions; \refsec{fairness}) and security (where bad actors exploit unintentional abilities or vulnerabilities in models to cause harm; \refsec{security}). 
Below, we outline how foundation models both enable new forms of misuse and support new tools for misuse detection and mitigation. 

\subsubsection{Foundation models will be misused for harmful purposes}
Advances in the scale (\refsec{training}),  multimodality (\refsec{modeling}), and adaptivity (\refsec{adaptation}) of generative foundation models will allow them to be misused to generate high-quality, cheap, and personalized content for harmful purposes. 
In this section, we discuss these three dimensions within the context of two examples of malicious activity: manipulative content creation and harassment. 

\paragraph{Content quality.}
Foundation models are capable of automatically generating much higher-quality, human-looking content than prior AI methods. 
They may empower disinformation actors, where states, for example, create content to deceive foreign populations without being transparent that the content is linked to a state. Currently, creating this content often requires hiring people who speak the language of the population being targeted. Governments may outsource content production to native speakers in the country they are targeting,\footnote{\url{https://www.lawfareblog.com/outsourcing-disinformation}}$^{,}$\footnote{ \url{https://fsi.stanford.edu/content/ira-takedown-20201215}} but this decision causes real risks for operational security. Foundation models will allow for the creation of content that is often indistinguishable from content created by humans \citep{kreps_mccain_brundage_2020, clark2021all}\dash{}and indeed it will be able to do this for a wide variety of languages\dash{}enabling both goals of creating content that resonates and maintaining operational security. 

In addition to deceiving foreign populations, foundation models’ ability to generate high quality synthetic images (deepfakes) or text may be abused to harass individuals. Deepfakes have already been used for the purpose of harassment. 
For example, Rana Ayyub, an Indian investigative journalist, was targeted by a high-quality deepfake that superimposed her face onto a pornographic video, leading her to leave public life for months.\footnote{\url{ https://www.huffingtonpost.co.uk/entry/deepfake-porn_uk_5bf2c126e4b0f32bd58ba316}}
Because foundation models are often multimodal (\refsec{modeling}), they could similarly impersonate speech, motions, or writing, and potentially be misused to embarrass, intimidate, and extort victims.\footnote{\href{https://www.wsj.com/articles/fraudsters-use-ai-to-mimic-ceos-voice-in-unusual-cybercrime-case-11567157402}{{ https://www.wsj.com/articles/fraudsters-use-ai-to-mimic-ceos-voice-in-unusual-cybercrime-case-11567157402}}}

\paragraph{Cost of content creation.}
Foundation models will substantially decrease the costs of content creation, further lowering the barrier to entry for malicious actors to carry out harmful attacks \citep{brundage2018malicious}. The budget for one 2017 influence operation that originated in Russia and targeted Americans was \$12.2 million \citep{DiResta2018TheT}.
More recently, individuals in Russia paid \$75-\$200 per article to American freelancers as part of a disinformation campaign.\footnote{\href{https://www.nytimes.com/2020/09/02/technology/peacedata-writer-russian-misinformation.html}{{ https://www.nytimes.com/2020/09/02/technology/peacedata-writer-russian-misinformation.html}}} 
Foundation models will lower these marginal costs. 
While foundation models, such as GPT-3, may make mistakes when generating content \citep{BuchananCSET2021}, it will be more feasible to hire a small number of editors to fix them than to hire content creators directly. 
Initial costs to train foundation models are more significant (\refsec{systems}), but these expenses should be manageable for most state actors \citep{BuchananCSET2021}.

In addition to monetary cost, foundation models require fewer technical skills to achieve high-quality results. 
Current tools, such as video editing software, can enable credible photo or video deepfakes, but require several hours of a skilled user’s time to yield quality content. 
Foundation models lower this barrier to use:~their few-shot adaptation capabilities (\refsec{adaptation}) enable new modes of interaction for application users (\refsec{interaction}) that will allow users to rapidly iterate for content creation. 

\paragraph{Personalization.}
Foundation models will reduce obstacles to creating personalized  content. 
For example, disinformation from Russian individuals that targeted the US in 2016 included highly customized content. 
Social media posts were crafted to push narratives about Syria (\eg~the U.S. should get out of Syria) that resonated with Black Lives Matter activists \citep{DiResta2018TheT} (\eg by suggesting that the U.S. should focus on issues facing the Black community in America, and not on issues in Syria). 
The same narratives were repackaged to resonate with Texas secessionists \citep{diresta2021}. 
Such a content creation endeavor is costly and time consuming. 
Foundation models will allow for similar activity, but at scale due to the low cost of adaptation (\refsec{adaptation}). 

In addition to foundation models allowing an actor to personalize content for niche audiences, they also allow an actor to personalize content to target a single individual\dash{}a capability that can be abused by harassers. 
Foundation models that condition their generations on personal attributes or information can create realistic personalized content, which could be more embarrassing, place victims in more danger,\footnote{\url{https://www.dw.com/en/social-media-uptick-in-honor-crime-in-middle-east/a-56370773}} and lead to more successful extortion attempts. 

\subsubsection{Foundation models will be powerful detectors of harmful content}

While the generative capabilities of foundation models will provide ample misuse opportunities, these same abilities may make them strong detectors of harmful content. While these capabilities are equally relevant for detecting human- and model-generated content, we focus on the detection of model-generated content in this section. First, we outline the challenges that current manual detection approaches will face in discovering harmful misuses of foundation model. Then, we propose how the interactive and multimodal representation capabilities of foundation models may make them powerful tools for automatic detection of harmful content. Finally, we discuss the risks associated with deploying automatic detection models in online settings to combat potential foundation model misuse.

\paragraph{Rethinking human interventions.} 

Currently, malicious practices are frequently uncovered (and on social media, sometimes removed) by humans searching the internet to uncover content origination.\footnote{\href{https://www.theatlantic.com/ideas/archive/2020/09/future-propaganda-will-be-computer-generated/616400/}{{ https://www.theatlantic.com/ideas/archive/2020/09/future-propaganda-will-be-computer-generated/616400/}}}
For example, fake social media profiles commonly steal profile photos from dating sites, which are discoverable through reverse image searches. 
Similarly, disinformation websites frequently use plagiarized content to mask deceptive content \citep{DiResta2019PotemkinP}, which is easily identified by conducting internet phrase searches. 
Foundation models will limit the efficacy of these detection strategies. 
Already, relatively unsophisticated disinformation campaigns have leveraged AI-generated photos\footnote{For a Middle East campaign example, see \url{https://www.thedailybeast.com/right-wing-media-outlets-duped-by-a-middle-east-propaganda-campaign}. \\For an example from Cuba, see \url{https://raw.githubusercontent.com/stanfordio/publications/main/twitter-CU-202009.pdf}} to remove the possibility of discovery through reverse image search. 
Tools for assessing whether these photos are AI-generated are available, but foundation models will complicate this work\dash{}for text and video as well\dash{}challenging manual human discovery techniques \citep{Ippolito2020AutomaticDO,clark2021all}. 

\paragraph{Foundation models as detectors.} 
The same abilities of foundation models that make them strong generators of creative content may make them strong detectors of model-generated content. Existing works demonstrate that foundation models can be adapted to detect disinformation from text generators \citep{zellers2019neuralfakenews}\dash{}which generate statistical textual artifacts \citep{Holtzman2020TheCC}\dash{}and that they can be used to evaluate the toxicity levels of their own generations using prompt questions \citep{Schick2021SelfDiagnosisAS}. Below, we describe how future foundation models will enable more powerful detection systems of machine-generated, harmful content.

Improvements in the interactive and multimodal interfaces of foundation models will provide new opportunities to improve detection of foundation model misuse for harmful content generation. Current statistical detectors must be retrained and re-deployed to integrate new knowledge about the textual content of misuse strategies \citep{Dinan2019BuildIB}. The rapid learning capabilities of foundation models (\refsec{adaptation}) may allow them to adapt from human feedback to new misuse strategies that the foundation model was not initially trained to recognize \citep{Lee2021TowardsFF}.

Simultaneously, the multimodal abilities of foundation models will enable more expressive representation of misuse ecosystems. Prior work has explored how misinformation spreads more rapidly across social networks than authentic content \citep{Starbird2018,Vosoughi1146}, yielding recognizable signatures when analyzed retrospectively. The multimodal capabilities of foundation models could allow them to jointly learn representations of harmful content and its typical dissemination signature on social networks. These joint representations could provide powerful tools for predicting whether certain types of automatically-generated content are indicative of misuse behavior. 

\paragraph{Risks of foundation models as automatic detectors.}
Improvements in automatic detection systems for both model-generated and human-generated harmful content will make these systems more prevalent online, yielding potential negative consequences. Any detection system will have false positive cases where human-generated fair content will be flagged as harmful \citep{sap-etal-2019-risk,xu-etal-2021-detoxifying}. 
The rate at which algorithmic false positives affect users (or groups of users) may cause downstream harm (\refsec{fairness}). The adaptive capabilities of foundation models should make systemic false positives easier to address as the model can be locally edited to re-classify those examples (\refsec{adaptation}). 
However, corner cases will likely not be prioritized and recourse will be challenging in these situations.

More broadly, wide-scale deployment of misuse detection systems may engender an ``arms race’’ between harmful content generators and detectors. Most content generators that use foundation models will lack the resources to develop them individually, and will use systems deployed by larger entities. While terms of use policies should outline acceptable uses of these systems (\refsec{ethics}), deployers of foundation models will also need internal detection systems to identify misuse of their products\footnote{\url{https://www.wired.com/story/ai-fueled-dungeon-game-got-much-darker/}} and mitigate them (\refsec{legality}). 
However, there will be fewer controls for misuse actors with the resources to develop their own foundation model-based content generators, putting pressure on platforms to curate the content shared through their distribution channels. Optimistically, content platforms encompass some of the most well-capitalized firms in the world. Their resources may enable the development of  detectors beyond the capabilities of most individual misuse agents. 
This resource advantage could disincentivize individual foundation model development due to the high costs of repeatedly training these systems at scale. However, many instances of foundation model misuse could still be successful even without the largest foundation models to power them, particularly as attackers may leverage the interactive capabilities of foundation models to rapidly generate content that can evade detection.

\newsection
\hypertarget{environment}{\subsection{Environment}}
\label{sec:environment}
\sectionauthors{Peter Henderson, Lauren Gillespie, Dan Jurafsky}

\begin{figure}[!ht]
\centering
\includegraphics[width=\linewidth]{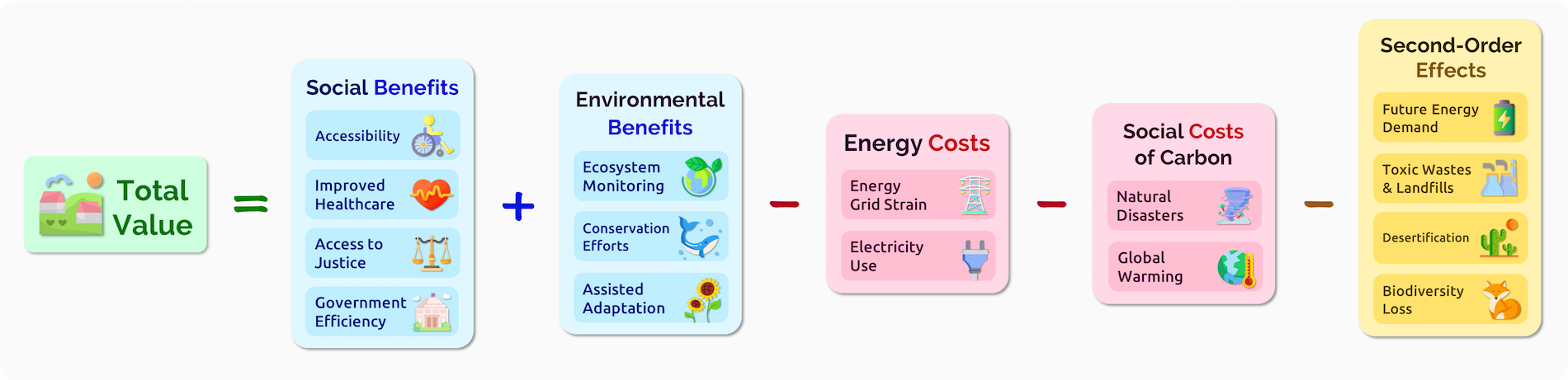}
\caption{\label{fig:environment} A visualization of a cost-benefit analysis for deploying a foundation model. The total value of a model can be approximated by first considering the net positive social benefits of the model, as well as any environmental benefits. Then, we subtract the negative energy costs to train and deploy the model, the social cost of the carbon emitted to train the model, and the secondary environmental effects. If the net costs outweigh the benefits, then foundation model developers and large-scale deployers should consider harm reduction strategies. This could include deploying a more efficient model or not deploying the model at all. }
\end{figure}

Foundation models can potentially lead to many social and environmental benefits, for example in legal domains (\refsec{law}), healthcare (\refsec{healthcare}), or even tackling climate change~\citep{rolnick2019tackling}. But because of their scale, they themselves can negatively impact the environment through increased carbon emissions if model creators are not careful~\citep{strubell2019energy,lottick2019nergy,schwartz2019green,lacoste2019quantifying,cao2020towards,henderson2020towards,bender2021,patterson2021carbon,lannelongue2021green,parcollet2021energy}. Addressing such emissions is an imperative: current forecasts show that climate change is occurring more rapidly than previously thought~\citep{ipcc2021}.

To understand where such emissions can occur in foundation models, we consider their lifecycle. First, they are trained on vast amounts of data, possibly for up to months of time and often distributed across hundreds to thousands of GPUs. Afterwards, they may be adapted to new domains or perhaps distilled into smaller models. All of this can be considered part of the training regime. Models used purely for research may not move beyond these steps. 
After models have been adapted and/or distilled, they might move on to be deployed into production. At this point many rounds of inference will run through the model until a new model is trained and the cycle repeats. 

Each one of these steps has the potential to utilize large amounts of energy and can contribute to carbon emissions. Foundation models can generate large, one-time energy costs and carbon emissions during the initial training phase. For example, the amount of emissions from training one BERT-base model, under some conditions, would only be offset by 40 trees grown for 10 years.\footnote{\citet{strubell2019energy} calculate carbon emissions for training BERT on an average energy grid in the U.S. and we use \url{https://www.epa.gov/energy/greenhouse-gas-equivalencies-calculator} to convert that to equivalent emissions in other domains. We note that this number can vary depending on the energy grid and other considerations~\citep{henderson2020towards,patterson2021carbon}.}
And if deployed at scale, foundation models can require substantial energy to service millions of requests\footnote{For example, transformers are already used at scale for search both at Microsoft and Google. See \url{https://www.blog.google/products/search/search-language-understanding-bert/} and  \url{https://azure.microsoft.com/en-us/blog/microsoft-makes-it-easier-to-build-popular-language-representation-model-bert-at-large-scale/}.}\dash{}translating to large carbon emissions if nonrenewable resources are used.

Therefore, the environmental impacts of certain design decisions for both training and deploying foundation models can be substantial. 
Even seemingly minuscule decisions, like reducing the number of layers a model has, may lead to significant environmental cost reductions at scale. 
For example, based on calculations from \citet{henderson2020towards}, a slightly more energy efficient translation model deployed at the scale of a commercial translation service could save between 78 kgCO2eq and 12,768 kgCO2eq of carbon emissions \textit{per day} depending on the energy grid used.
This is roughly equivalent to the carbon sequestered by 1 to 211 trees grown for 10 years, or the carbon sequestered by .35 to 57.4 acres of forest in one year.\footnote{Sequestration estimated via \url{https://www.epa.gov/energy/greenhouse-gas-equivalencies-calculator}, but may be larger depending on other estimation methods. More efficient energy grids will emit less carbon, resulting in wide estimated ranges of impacts.}
Thus the design, deployment, and post-deployment monitoring of foundation models should adequately reflect these risks.

There are of course uncertainties in calculating the amount of energy used or carbon emitted by any given model~\citep{henderson2020towards,cao2020towards,patterson2021carbon}, and other sources of emissions may currently be much greater than those generated by foundation models~\citep{mora2018bitcoin}. But if foundation models continue to scale and gain in popularity, they may very well become a significant contributor to carbon emissions. 
Our goal is to provide a framework for foundation model developers and large-scale deployers\footnote{We focus on model developers and large-scale deployers, like those who build production systems on top of foundation models, because they are most able to make meaningful changes to reduce energy use and carbon emissions. A single change by these actors\dash{}like using a more efficient model\dash{}can scale to massive carbon savings, which would otherwise require a massive campaign to reach all downstream model users.} to consider how they can mitigate any unnecessary carbon emissions and keep the net social impact of these models positive.
We recommend that: 
\begin{enumerate}
    \item Carbon impacts can and should be mitigated in many cases. This can be accomplished by training models in low-carbon intensity regions,
    or by using more efficient models and hardware (\refsec{environment-mitigation}).
    \item When all mechanisms for mitigation have been exhausted and mitigation is no longer possible, 
    the costs and benefits to society should be assessed 
    to determine if and when a larger foundation model should be deployed over a smaller, more efficient, model 
   \dash{}with the understanding that the up-front costs of a large foundation model may be amortized over the lifetime of the model (\refsec{environment-costs}).
    \item Energy, computational, and carbon costs\dash{}as well as any efforts taken to mitigate negative impacts\dash{}should be clearly reported to inform policymaking and research (\refsec{environment-reporting}).
\end{enumerate}

\hypertarget{environment-mitigation}{\subsubsection{Carbon impacts can and should be mitigated in many cases}}
\label{sec:environment-mitigation}
The  carbon impacts of training foundation models differ from the impacts of deploying them for inference.
Model training has no latency requirements, so  training  can be moved across energy grids with relative ease in cloud environments. 
Every energy grid has its own carbon intensity\dash{}the amount of carbon emitted per kilowatt-hour of energy used. For example, Qu\'ebec has an extremely low carbon intensity due to its reliance on hydroelectricity, while Estonia's energy grid has an extremely high carbon intensity due to its reliance on shale oil (though that is changing quickly)~\citep{henderson2020towards}.
Recent research has even suggested that the top 5\% of polluting power plants contributed 73\% of all electricity-based emissions~\citep{grant2021reducing}.
Thus, while training foundation models can be quite energy intensive, researchers have demonstrated that the carbon impacts of these models can be partly mitigated by \textit{selecting energy grids with minimal carbon emissions}~\citep{henderson2020towards,lacoste2019quantifying,patterson2021carbon}. 

\textit{Carbon offsets} have also been proposed as a stopgap until carbon-free renewable electricity is available at all data centers. This strategy involves reducing carbon emissions in one activity to offset the emissions from another. But most\dash{}if not all\dash{}carbon offsets are a strictly worse solution than not emitting CO$_{2}$ in the first place \cite{holl_tree_2020}. Some carbon offset programs can even have a negative impact. For example,  studies of forest planting campaigns (often a source of carbon offsetting) show that they can do more harm than good. They can yield monocultures (the use of one particular species of tree) that diminish the biodiversity of the region and reduce carbon storage in the forest soil~\citep{heilmayr2020impacts,hong2020divergent}. This could result in more carbon emissions when using carbon offsets than if the original carbon had never been emitted in the first place.
Therefore, when training or deploying a foundation model, we recommend designing for as little carbon emission as possible up-front, rather than simply relying on carbon offsets to cancel emissions.

When it is not possible to run in low-carbon regions, other mitigation strategies should be leveraged, reducing unnecessary energy usage. This includes:
\begin{itemize}
    \item using more efficient hardware,\footnote{Notably, California now regulates computers with inefficient GPUs for this reason, requiring that they stay below 30-100 kWhs/year, depending on the manufacturing date and computer type. \emph{See} Sections 1601-1608 of California’s Appliance Efficiency 
Regulations (Title 20).}
\item using mixed-precision training~\citep{micikevicius2017mixed} or quantization~\citep{gholami2021survey},
\item using more efficient architectures (\eg using an evolved transformer over a vanilla transformer architecture; or using sparse models)~\citep{pmlr-v97-so19a,patterson2021carbon,mostafa_parameter_2019},
\item distilling models and using distilled models (\eg~\citep{sanh2019distilbert}),
\item and utilizing other optimization strategies that will reduce energy costs (see more discussion in \refsec{systems}).
\end{itemize}

Maintainers of open source projects and cloud compute should strive to set their default settings to the most efficient possible, since ``green defaults'' are known to be the most effective mitigation strategies (see discussion in ~\citep{henderson2020towards}).
Other mitigation strategies can be found in recent literature~\citep{strubell2019energy,lacoste2019quantifying,schwartz2019green,henderson2020towards}.
We also note that reducing and mitigating energy usage also has the added benefit of making models more accessible to those with limited compute access (see \refsec{ethics} for more discussion).

However, when a model is mainly used for inference, \eg  deployed in a production application, it often cannot be moved to a less carbon-intensive energy grid for low-latency applications. 
In addition to using the mitigation strategies specified above, in this case it is important to weigh
the benefits of the proposed foundation model versus a more energy efficient alternative. We discuss this further in the subsequent section.

\hypertarget{environment-costs}{\subsubsection{Costs and benefits should be assessed before using foundation models}}
\label{sec:environment-costs}

After taking as many steps as possible towards mitigation (or where mitigation is not possible),
it is vital to assess the required size of a foundation model\dash{}or whether a foundation model should be used at all. This cost-benefit analysis should consider:

\begin{enumerate}
    \item Is the social cost and environmental cost from deploying the foundation model greater than the social benefit of the model?
    \item Would another, computationally simpler and cheaper approach achieve comparable social benefit (\eg a much more efficient foundation model, or perhaps simple baseline)?
\end{enumerate}

A simplified scheme for assessing this trade-off considers the overall impact of a model $M$ as:

\begin{equation}
\label{eqn:env}
    V(M) = S(M) - C(M) - E(M) - O(M).
\end{equation}

\reffig{environment} represents this equation and the costs and benefits that may enter each variable. Here, $M$ is the model and $S$ is the net social benefit, as well as environmental benefit, in dollars. $S$ can be increased by improving healthcare, access to justice, decreasing poverty, improving environmental monitoring, aiding ecosystem conservation efforts, and so on. 

$C$ is the social cost of carbon from energy use. This represents the future harm to society from the carbon released as a present-day monetary value. The upper bound U.S. Environmental Protection Agency (EPA) estimate from 2017 for the social cost of carbon was \$105 (in 2007 U.S. dollars) per metric ton of $\text{CO}_2$ emitted.\footnote{See \url{https://19january2017snapshot.epa.gov/climatechange/social-cost-carbon_.html}. But note that the social cost of carbon can be a contentious metric~\citep{stern2021social}. By using a favorable discount factor, one can reduce carbon costs. As such, it can the calculation of this metric can vary across methodologies.} 

$E$ is the energy cost of the model. For example, in April 2021, the average U.S. residential energy cost was about \$0.1376 per kWh.\footnote{\url{https://www.eia.gov/electricity/monthly/epm_table_grapher.php?t=epmt_5_6_a}} Added to this variable could be the costs from increased strain on the energy grid. For example, a recent study suggested that the cost per energy grid interruption event, normalized by average demand, could be as high as \$15.9 per average kW~\citep{sullivan_updated_2015}.\footnote{Like the social cost of carbon, calculation of these costs can fluctuate across modeling methodologies.}

$O$ is the social cost of other second order environmental effects. This could include:
\begin{itemize}
    \item The compounding carbon impacts from increased chip demand and chip production~\citep{gupta2021chasing}.
    \item Other environmental impacts of chip manufacturing, like the creation of toxic waste sites in 
 Silicon Valley, whose health effects are unequally distributed to socially vulnerable populations~\citep{stewart2014uneven}, or pollution from manufacturing in Taiwan that has been linked to chronic health problems~\citep{tu2009ineffective,lin2016increased}. 
 \item The compounding effects of climate change that are not already included in the SCC model. For example, these effects could include accelerated desertification~\cite{huang_accelerated_2016}, rapid ecosystem changes that put many species at risk of extinction~\cite{urban_accelerating_2015}, and increased carbon emissions due to melting permafrost~\citep{schuur_climate_2015}. 
 \item Unnecessary strain on chip production capacities. Recent chip shortages have led to work stoppages in automobile manufacturing.\footnote{\url{https://www.reuters.com/business/autos-transportation/ford-shut-some-n-american-plants-few-weeks-chip-shortage-2021-06-30/}} There is no evidence to suggest that increasing demand for ML-optimized chips led to this shortage.\footnote{Though recent reports have suggested that demand for datacenter chips have surpassed the gaming sector. \emph{See} \href{https://www.nextplatform.com/2020/08/21/the-local-maxima-ascension-of-datacenter-at-nvidia/}{https://www.nextplatform.com/2020/08/21/the-local-maxima-ascension-of-datacenter-at-nvidia/}.}
 But such considerations fall into second order effects, where researchers might weigh whether the risks, however slight, of contributing to such negative impacts are worth using or deploying a large model.\footnote{Like for other metrics described previously, there is uncertainty as to how these impacts might be calculated and attributed to models.}
\end{itemize}

It is important to consider in this analysis that the economic benefits and social costs of carbon could be distributed unequally across communities, with poorer communities being impacted more heavily by climate change and wealthier communities being benefited by a model~\citep{bender2021}.\footnote{See also, \url{https://www.un.org/sustainabledevelopment/blog/2016/10/report-inequalities-exacerbate-climate-impacts-on-poor/} and \url{https://blogs.imf.org/2020/12/02/how-artificial-intelligence-could-widen-the-gap-between-rich-and-poor-nations/}.} As such, when conducting the Equation~\ref{eqn:env}  analysis, one should consider the benefits and harms to society more broadly rather than for a given organization or country. In this case $V(M)$ can be viewed as a distribution and should ideally be evenly distributed across the population. In cases where the distribution is highly uneven\dash{}for example where all the benefits fall to the model designer while all the harms fall to populations that will never benefit from the model\dash{}the designer should spend substantially more effort on mitigation before deploying the model. 

There is, of course, some uncertainty in which methodology to use when valuing each component of Equation~\ref{eqn:env}. 
Empirical estimates for many of these terms can range in multiple magnitudes depending on the data source and modeling choice for the phenomena, such as the different mechanisms for evaluating the social cost of carbon. And of course additional externalities, that may be difficult to quantify monetarily, will continue to need to be considered. The key takeaway of this cost-benefit analysis, however, is not the dollar valuation of each term in the equation, but rather the \emph{existence of} and relative importance of each of these effects. 
Our goal is to provide a high-level framework for beginning to consider these trade-offs. 
Future research may give more guidance on how to quantify each of these values. 

\begin{figure}[t]
    \centering
    \includegraphics[width=.55\textwidth]{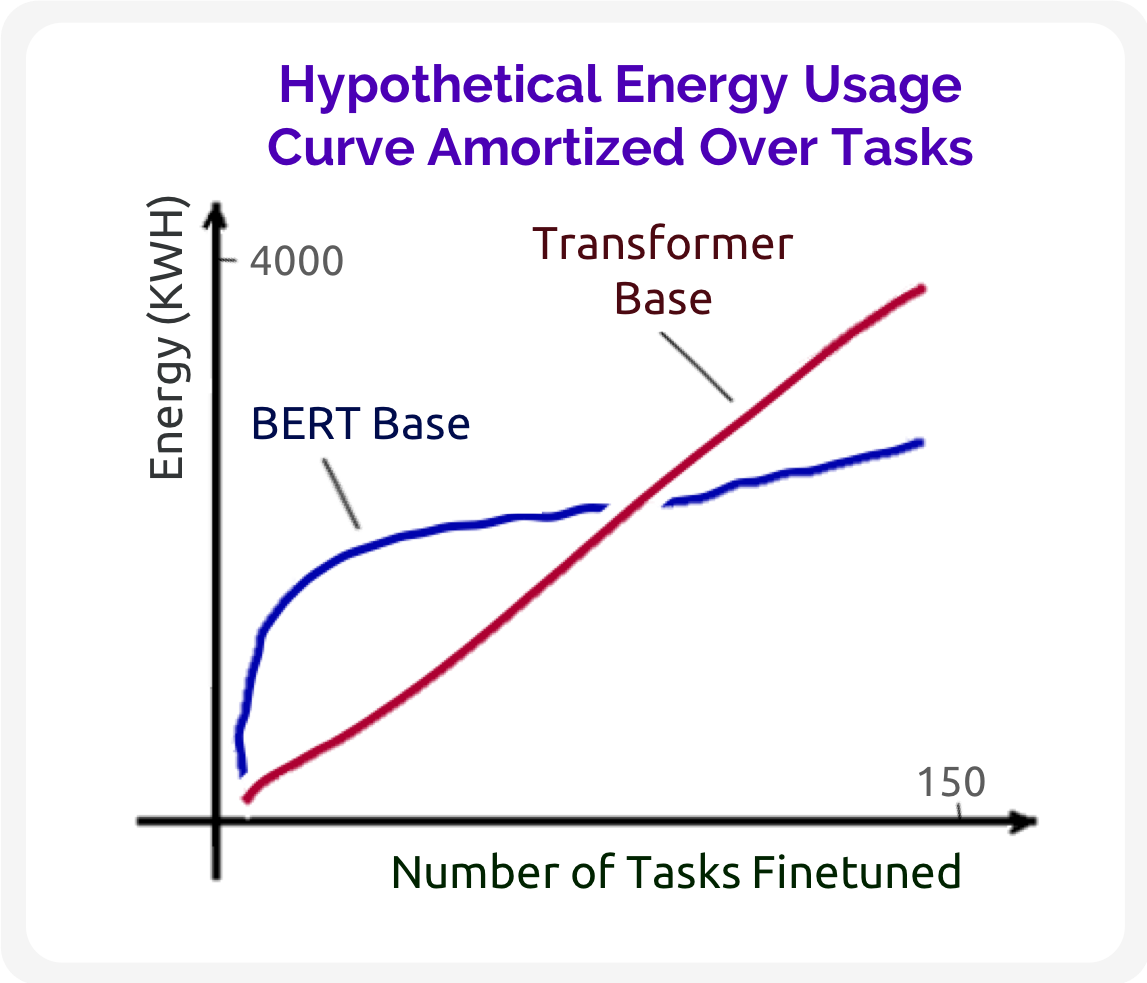}
    \caption{A hypothetical example of amortized fine-tuning showing the point at which a foundation model (in this case BERT Base) will have lower energy costs than a transformer model trained from scratch. We estimate the up-front energy cost for training BERT from~\citet{strubell2019energy}, and cost for fine-tuning a downstream task from~\citet{chaudhary2020topicbert}. We compare against the linearly increasing cost of training a transformer from scratch, from~\citet{strubell2019energy}. If BERT is used for less than $\sim$80 tasks, the up-front energy costs are not recovered. After that point, BERT is  more energy efficient than the model trained from scratch.}
    \label{fig:env_bert_fine-tune}
\end{figure}

Finally, we note that these factors should also be evaluated over the lifetime of the model, not on a per-run basis. Consider an alternative baseline model that must be trained from scratch for every new task. The baseline may well require an expensive hyperparameter search to achieve equivalent performance on downstream tasks. In contrast, the foundation model places the brunt of the costs on the initial pretraining procedure, with fine-tuning perhaps being much simpler and more energy efficient. Over the lifetime of the foundation model, it could be more carbon efficient than the baseline  (\reffig{env_bert_fine-tune}). 
Even more efficient adaptation mechanisms could improve this amortization further (see \refsec{adaptation}).

The efficiency of adaptation, however, is not guaranteed. It may be true that some foundation models will never be more efficient than a particular baseline, even when amortized over many tasks.
For example, it cannot be assumed that a smaller model with fewer parameters will translate to energy efficiency improvements. Due to increased hyperparameter tuning costs or other optimizations, the number of parameters has been shown not to correlate with energy efficiency in some cases~\citep{zhou2020hulk,henderson2020towards}.
Therefore, foundation model developers should rigorously assess the efficiency of their models and adaptation mechanisms before beginning large-scale training efforts.

The framework in this section is meant to guide the reader in thinking about the environmental and societal trade-offs in training and deploying their model, but there are other substantial social justice considerations involved in deploying a foundation model, discussed in \refsec{ethics}.
\refsec{economics} also discusses in more detail the dynamics of social welfare from algorithm deployment.

\hypertarget{environment-reporting}{\subsubsection{Carbon/energy impacts should be systematically reported}}
\label{sec:environment-reporting}

A cost-benefit analysis cannot be conducted unless researchers and engineers working on foundation models report the computational, energy, and carbon costs of their models.
We encourage foundation model developers, providers, and curators to report these metrics, as well as what carbon reduction strategies were used in the making of the foundation model. 
See \citep{henderson2020towards,lottick2019nergy,lacoste2019quantifying,codecarbon,anthony2020carbontracker} for examples of a Carbon Impact Statement and for tools that can facilitate this reporting.
For researchers, such reporting can occur at publication time, but we also encourage industry actors to adopt transparency mechanisms to report these metrics for their deployed models.\footnote{A small step toward this has been taken by some cloud compute providers that identify the most carbon friendly cloud regions. See, for example, \href{https://cloud.google.com/blog/topics/sustainability/pick-the-google-cloud-region-with-the-lowest-co2}{https://cloud.google.com/blog/topics/sustainability/pick-the-google-cloud-region-with-the-lowest-co2}.}
This will help set policy recommendations within industry and academia, as well as help downstream users identify carbon-friendly usage patterns.
Standardized reporting will also aid in determining which models are accessible to those with limited compute access (see \refsec{ethics} for more discussion on accessibility).

To encourage more reporting of energy and carbon impacts, we suggest, among other strategies: giving green badges at conferences, requiring reporting of relevant metrics for submission to conference venues, lobbying large-scale deployers of foundation models to provide more transparency, and generally shifting professional norms in academia and industry towards standard reporting of these metrics (see more discussion on professional norms in \refsec{ethics} and more discussion on reporting mechanisms by \citet{henderson2020towards}).
\newsection
\hypertarget{legality}{\subsection{Legality}}
\label{sec:legality}
\sectionauthors{Neel Guha, Peter Henderson, Lucia Zheng, Mark Krass, Daniel E. Ho}

In this section, we describe how US law may influence, constrain, or foster the creation and use of foundation models.\footnote{Our perspective here centers on US law and legal frameworks. Discussions of the implications of foundation models with respect to other countries may consequently take different perspectives.} We note that the legal landscape surrounding algorithmic tools remains uncertain. We highlight issues pertaining to (1) model training, (2) liability for model predictions, and (3) protections for model outputs.

Though understanding how the law affects foundation models is crucial, it is important to recognize that the law cannot be the only lens through which we evaluate the construction, maintenance, and use of foundation models. Ethical frameworks are necessary to understand where legally permissible applications of foundation models may still be ill-advised for the harms they inflict and are discussed in more depth in \refsec{ethics} and \refsec{fairness}. Studying the potential for misuse and possible security concerns (see \refsec{misuse} and \refsec{security}) is critical for preventing harmful outcomes \textit{ex ante}, as opposed to the \textit{ex post} treatment that legal mechanisms often provide.

\subsubsection{Training}
Training foundation models will require accumulating vast amounts of multi-modal data, raising questions around data collection and data use. 

First, the ability for model creators to grow datasets via web scraping will be governed by the manner in which courts will interpret terms of service provisions and, notably, the U.S.  Computer Fraud and Abuse Act (CFAA), which criminalizes accessing a server ``without authorization''~\citep{cfaa}. Courts are in conflict on these questions, and recent cases have sought to clarify the circumstances under which web scraping may be barred.\footnote{\textit{Van Buren v. United States}, 141 S.Ct. 1648 (2021).} The restrictiveness of data access would fundamentally affect the diversity of data practitioners can use to train foundation models~\citep{levendowski2018copyright}.

Second, much of the data contained in training sets will be copyrighted and potentially protected by intellectual property law. However, copyright law recognizes exceptions when individuals may be permitted to use copyrighted material.\footnote{\textit{See, \eg}, 17 U.S.C \S 107 to 112.} Some scholars believe that the legal permissibility of training datasets will largely rest on whether courts interpret the process of model training as ``transformative'' under fair use doctrine~\citep{lemley2020fair}. Though the question of what qualifies as transformative is highly context dependent, the general rule is that transformative uses are those ``that add something new, with a further purpose or different character, and do not substitute for the original use of the work"~\citep{usco}. Already, the recently released Github Copilot tool is bringing these arguments to the fore~\citep{verge_copilot}.

Finally, some training datasets may run afoul of privacy laws. Illinois, for instance, enables individuals to sue for improper collection or use of biometric data (\eg retina or iris scans, fingerprints, voiceprints, or scans of hand or face geometry).\footnote{IBM is the defendant in a current class action alleging that IBM's collection and use of this data (including for machine vision purposes) violates this statute. See Class Action Complaint at 2, Vance v. Int'l Bus. Machines Corp., No. 20 C 577 (N.D. Ill. filed Jan. 24, 2020).} Foreign privacy laws like the E.U.’s General Data Protection Regulation (GDPR)\dash{}which will affect American model creators if datasets contain information from E.U. citizens\dash{}would require data subjects to be informed about the purpose of data collection. Further issues could arise for laws like the California Consumer Protection Privacy Act (CCPA), which provide individuals with a ``right to be forgotten,'' raising questions as to whether model creators will need to ``remove'' training data from models~\citep{villaronga2018humans, ginart2019making}.

\subsubsection{Output liability.} 
Though foundation models themselves are task agnostic, fine-tuned models\dash{}or the representations learned by foundation models themselves\dash{}may be used for traditional prediction tasks.   
Where these tasks form components of larger decision-making systems, foundation models will thus influence actions, decisions, or policies. When these result in harm, model creators\dash{}and the individuals operating them\dash{} may be legally responsible. 

Embedding foundation models in physical systems (\eg self-driving cars, electric grid management, medical diagnostics, etc.) may result in physical harm to individuals. Here, courts will likely resolve questions of liability under tort doctrine~\citep{lemley2019remedies, selbst2020negligence}. 
Key open questions include the interplay between the liability of users, foundation model providers, and application developers, as well as the standards courts will use to assess the risk profile of foundation models. Deployments in particularly sensitive domains (\eg medicine) will require regulatory approval, and the development of standardized processes to assess safety~\citep{wu2021medical}.

Fine-tuned foundation models that classify individuals in ways that correlate with protected attributes (\eg~race, gender) may face challenges under civil rights laws. Scholars have noted that claims for disparate treatment resulting from foundation models may be brought in the context of hiring, housing, or credit lending ~\citep{gillis2019big, scherer2019applying}. Exactly how courts will adjudicate these issues is far from clear.  Scholars have noted for instance, that the courts’ traditional views on ``discrimination'' would actually prevent machine learning practitioners from implementing many algorithmic fairness techniques~\citep{xiang2021reconciling, ho2020affirmative}.\footnote{For more information on how models may embed certain biases, see \refsec{fairness}.} 

U.S. law recognizes special privileges and limits on governmental entities. Thus, the use of foundation models by governmental entities\dash{}at a local, state or federal level\dash{}will implicate special considerations, in addition to equal protection claims. 
The use of models for risk assessment\dash{}or in other settings which result in a deprivation of life, liberty, or property\dash{}will invite procedural due process claims.\footnote{Procedural due process recognizes that plaintiffs usually have certain basic rights during any deliberation that will deprive them of life, liberty, or property (\eg the right to cross-examine adverse witnesses).} When models are used by administrative agencies (\eg the~Environmental Protection Agency) for instance, plaintiffs may allege that such use violates basic standards of due process, reasonableness / non-arbitrariness, and transparency.

\subsubsection{Legal protections for outputs}
Model outputs\dash{}and by extension the model creators responsible for the models\dash{}may also be afforded certain legal protections. 
First, content produced by generative models may implicate free speech issues. 
The extent to which courts will find First Amendment protections for machine generated content is unclear.
Scholars have discussed a number of open questions, including whether ``AI speech'' is protected~\citep{massaro2016siri} or if model outputs are in effect the human programmer’s speech~\citep{kajbaf2019first}. Others have noted the possibility of disclosure requirements (akin to safety disclosures for pharmaceutical drugs or other substances), also implicating speech doctrine, under which models would be forced to share with listeners that their content is machine generated~\citep{lamo2019regulating}. 
These issues could have wide ranging consequences, affecting whether individuals can use foundation models to mass produce speech, or whether model creators could be held liable for content generated by foundation models.

Second, there is uncertainty regarding who may assert ownership over model outputs.  Existing copyright law does not recognize computer programs as authors, and hence, does not afford copyright protection to ``work'' created by computer programs~\citep{grimmelmann2015there}. As a result, scholars have advocated for a variety of approaches. Some have argued that, depending on the circumstances, both the human creator of a program and its human user may have viable claims to being the ``author'' of the program’s output~\citep{ginsburg2019authors}. 

As models are increasingly used in the process of ``creation''\dash{}from artistic endeavors to more mundane settings like news filings\dash{}disputes over the ownership of machine generated content will become more commonplace.

While our analysis above only skims the surface of the legal issues implicated by foundation models, the resolution of these questions will be critical to the construction, use, and deployment of foundation models, or, to borrow Larry Lessig’s phrase, how “code is law”~\citep{code_is_law}.

\newsection
\hypertarget{economics}{\subsection{Economics}}
\label{sec:economics}
\sectionauthors{Zanele Munyikwa, Mina Lee, Erik Brynjolfsson}

Foundation models have the potential to substantially improve overall living standards by increasing productivity and innovation. 
These models can be deployed to substitute for human labor, augment humans, or help in the discovery of new tasks and opportunities, 
which can lead to increased concentration of ownership and power, or more decentralization. 
On a broader level, the result can be either increased inequality due to potential centralization (\refsec{fairness}, \refsec{ethics}), or more broadly shared prosperity due to the easier adaptation of foundation models for a wide range of applications (\refsec{introduction}).
The ultimate outcomes on all these dimensions are not dictated solely by technology or economics, 
but by the choices and actions of technologists, policymakers, managers, workers, and other members of society.

Foundation models can be thought of as what economists refer to as a \emph{general-purpose technology} \citep{Bresnahan1995}. 
General-purpose technologies refer to technologies like the steam engine and electricity,
which drive waves of transformation and productivity growth due to their pervasiveness, improvement over time, and ability to spawn complementary innovations (a host of products and services that revolve around one core product). 
While foundation models may not be pervasive at the moment, they seem poised to be the basis of widespread technological innovations, 
and have the key hallmarks of a general-purpose technology.
As a result, these models are likely to be economically important.
In considering the impact of foundation models on the economy, 
we will focus on three broad areas of impact: productivity, wage inequality, and ownership.

\subsubsection{Productivity and Innovation}
Foundation models are likely to substantially increase both productivity and innovation. 
Productivity growth is one of the main contributing factors to boosting living standards, as it increases the wealth of nations and addresses a host of challenges from poverty and healthcare to the environment and education.

Productivity is defined as output per unit input.\footnote{Note that when properly measured, productivity is not just a matter of counting units produced or hours work, but also accounts for quality changes. Therefore, an increase in quality for a given amount of labor, such as more interesting fiction, also counts as an increase in productivity.}
One way to boost productivity is to reduce the denominator; for instance, enabling a company's advertisements to be written with fewer copywriters or fewer labor hours per copywriter lowers the number of units of input. 
Productivity can also be boosted by increasing the numerator, for instance by enabling a software developer to write more code in a given time.
If the growth in the numerator is great enough, this can lead to more people developing software, not fewer ~\citep{david2015there}.  
In many tasks, we have already observed machine learning systems increasing productivity. 
For instance, an autocomplete system for clinical documentation reduces keystroke burden of clinical concepts by 67\% \citep{pmlr-v126-gopinath20a}.
Likewise, the potential for foundation models to affect productivity spans almost every industry and many occupations. 
Considering language alone, an analysis of U.S. occupations using the US Department of Labor's O*NET database shows that many occupations involve the types of language-related work that could be affected by foundation models. 
Approximately 13\% of occupations have a \textit{primary} task that is related to writing, and the total wage bill of these occupations (annual salary multiplied by the number of individuals employed in the occupation) is over 675 billion dollars. 
However, the potential impact of foundation models extends beyond language. 
They will also have effects on diagnostic imaging in medicine, graphic design, music\footnote{https://www.landr.com/},
and many other tasks where people are creating something that is similar to something else that already exists~\citep{winkler2019derm,ramesh2021zeroshot}.

Perhaps the most profound, if still speculative, effect of foundation models is their potential to enhance creativity and boost the rate of innovation itself.
For instance, DALL·E ~\citep{ramesh2021zeroshot} could transform the market for illustrations much as inexpensive cameras revolutionized photography. 
If these models enable humans to develop new ways to write new songs and novels (\refsec{interaction}), discover variants of drug molecules (\refsec{healthcare}), extend patents (\refsec{law}), build innovative software applications, or develop new business processes, then not only the \textit{level} of productivity, but the \textit{rate} of growth of productivity would be increased.
In this way, foundation models have some of the characteristics of the ideas or blueprints in Paul Romer's growth models~\citep{romer1990endogenous}, or even meta-ideas (ideas about ideas) which, unlike most other goods, are non-rival, thus speeding growth.

It is worth noting that changes in productivity are not always visible in the official statistics, because many aspects of input and output are difficult to measure ~\citep{BrynjolfssonErik2019Hswm}.
As a result, the benefits and costs of foundation models will not be fully captured by traditional productivity metrics, nor by related metrics like gross domestic product (GDP) or price levels (the average of current prices across the entire spectrum of goods and services).
This is especially true for general purpose technologies historically, since they are catalysts for a cascade of secondary innovations that often transform the set of goods and services in the economy, 
and even the nature of production and innovation over a period of years or even decades.

\subsubsection{Wage inequality}
Even if foundation models increase average productivity or income, there is no economic law that guarantees everyone will benefit.
In part, this is because not all tasks will be affected to the same extent.
More importantly, the effects of foundation models on the demand for labor (and thus employment and wages) can be either positive or negative, regardless of productivity growth~\citep{brynolfsson2011race,brynjolfsson2017can}. 
When a technology substitutes for human labor in completing tasks, it tends to reduce demand for the workers doing those tasks. 
This depresses employment and wages. 
However, when a technology complements labor, or facilitates the creation of new opportunities or tasks, it tends to increase labor demand~\citep{acemoglu2019automation}. 
Employment can (and often does) go up, even as productivity increases.
For instance, the invention of the airplane created the demand for an entirely new occupation, the airline pilot.
In turn, the development of jet engines was complementary to human pilots, further increasing demand for them.
Similarly, the effects of foundation models on employment, wages, and income inequality will differ depending on how they are used.

While the industrial revolution mainly transformed physical work, foundation models are likely to transform tasks involving cognitive work, like content creation and communication. 
In general, since foundation models are intermediary assets that often possess strong generative capabilities, we envision that they will be able to augment humans in many creative settings, rather than replace humans as there are still significant limitations in using these models stand-alone for open-ended generative tasks \citep{see2019}.
As we describe in \refsec{interaction}, foundation models may also power systems that users can leverage to co-construct novel forms of art or more efficiently prototype new applications.
Fluid human-machine and human-in-the-loop interaction will require advances in interface design (\refsec{interaction}) as well as fundamental improvements in the interpretability (\refsec{interpretability}) and robustness (\refsec{robustness}) of these models, so that humans can understand model behavior and expect models to perform well in diverse contexts.

\hypertarget{economics-centralization}{\subsubsection{Centralization}}
\label{sec:economics-centralization}
Another key determinant of foundation models' economic impact is who owns data and models. 
In particular, pushing the frontier of foundation models has thus far primarily been the purview of large corporate entities. 
As a result, the ownership of data and models are often highly centralized, leading to market concentration (\refsec{ethics}).  
In turn, this can lead to significant centralization of decision rights and power, reducing income and opportunities for those who don't have ownership. This centralization of power can lead to an equilibrium where fewer people have social and economic mobility and opportunity, a situation that   \citet{Brynjolfsson2022inpress} calls "The Turing Trap". To counterbalance this centralization, there have been grassroots efforts to open source AI research such as Masakhane, EleutherAI, and HuggingFace, or build foundation models through distributed training. 
However, it likely that the gap between the private models that industry can train and the ones that are open to the community will remain large due to foundation models' dependence on massive amount of data and computational resources (\refsec{environment}).\footnote{Lambda Lab estimates that GPT-3 training costs over \$4.6M, research and development costs between \$11.4M and \$27.6M, hardware required to run GPT-3 costs between \$100K and \$150K without factoring in other costs (electricity, cooling, backup, etc.), and running costs a minimum of \$87K per year. (\url{https://bdtechtalks.com/2020/09/21/gpt-3-economy-business-model})}

\subsubsection{Other considerations}
This short chapter is not meant to be comprehensive of all the economic effects of foundation models. 
In addition to affecting productivity, wage inequality, and ownership, 
foundation models may also have significant effects on job quality and job satisfaction.
For instance, they may increase job satisfaction by automating repetitive, uninteresting parts of work, or decrease satisfaction by increasing the pace of work, thereby inducing more frequent burnout.  
As discussed in \refsec{fairness} and \refsec{ethics}, they can also amplify and perpetuate bias, often in unexpected ways, or be used as a tool for reducing it.  
Foundation models can facilitate global trade and remote work,
just as earlier uses of machine translation systems had significant effects in these areas \citep[\eg][]{Brynjolfsson2019}.
There may also be significant environmental effects (\refsec{environment}), as well as unexpected and unanticipated effects on the rate and direction of occupational change and business transformation in an economy. 
More broadly, given the emergent capabilities of foundation models, we should expect new unknown unknowns to arise that are difficult to predict, and which may have substantial follow-on effects.\footnote{As an example of a secondary effect, consider that the invention of the automobile influenced the development and expansion of the suburbs.}
 
In summary, foundation models are poised to be an important general-purpose technology of our era.
They have potential to increase living standards substantially, but also pose risks of increasing inequality and concentrating power.
The economic implications of these technologies are not predetermined, but rather depend on how technologists, policymakers, managers, workers, and other stakeholders answer challenges such as:
\begin{itemize} 
\item   How can we harness the potential of foundation models to boost productivity?
\item  	Can we develop models that enhance creativity and boost the rate of innovation?
\item   Will the benefits and control rights be limited to a few or widely shared?
\end{itemize}
Understanding the economic potential of these systems is the first step to guiding them in directions that match our values.

\newsection
\hypertarget{ethics}{\subsection{Ethics of scale}}
\label{sec:ethics}
\sectionauthors{Kathleen Creel, Dallas Card, Rose E. Wang, Isabelle Levent, Alex Tamkin, Armin W. Thomas, Lauren Gillespie, Rishi Bommasani, Rob Reich}

The widespread adoption of foundation models poses ethical, social, and political challenges in addition to concerns about increased inequity, the subject of \refsec{fairness}. 
In this section, we discuss social, political, and ethical risks  related to the scale of application of foundation models, such as homogenization and the concentration of power, the norms and release strategies appropriate to address them, and concerns about the broader political economy in which foundation models are developed and deployed. 

\subsubsection{Homogenization and scale} 
 
If the same model is used across a variety of domains with minimal adaptation, the strengths, weaknesses, biases, and idiosyncrasies of the original model will be amplified (\refsec{fairness}). 
This is true of the widespread adoption and reliance on any standardized technology. 
Similar to how a failure in the manufacturing of a part used in many cars or airplanes could have widespread and severe consequences across sectors, a bias or failure of service \textit{intrinsic} to a foundation model could ripple outwards. 
However, the current uninterpretability (\refsec{interpretability}) of foundation models and their task-agnostic training makes predicting, understanding, and addressing these weaknesses challenging.  
If, as seems likely, foundation models become widely adopted, foundation model developers bear greater responsibilities of care than standard model developers, as their choices in design and deployment  have widespread implications \citep{Arendt1987}. 

The defining feature of foundation models\dash{}their capacity to be usefully adapted for a multiplicity of tasks\dash{}is what makes them likely to be widely adopted for a vast range of socially consequential tasks. 
In contrast to the current distributed and varied model of decision making, employing many adaptations of the same foundation model for multiple automated decision-making tasks means that decision subjects may face a more homogeneous set of judgments rooted in the underlying foundation model. 

This algorithmic monoculture \citep{kleinberg2021} could lead to consistent and arbitrary rejection, mis-classification, or ill-treatment of individual decision subjects \citep{Gandy2021}. 
We will call this \textit{homogenization} \citep{creel2021}.  
For example, \refsec{data-solutions} discusses data quality issues that lead to undesirable behavior on subpopulations of data, where subpopulations can be produced by any filter that stratifies the data, including by social group  (see related discussions in \refsec{interpretability-behavior} and \refsec{robustness-advantages}). 
Until improvements are made in data quality tooling (\refsec{data-solutions}) and the ability to identify slices of data on which the model under-performs  \citep{chung2019slice, goel2021robustnessgym}, 
a foundation model might consistently fail to provide accurate information or services to a subgroup of people (see also \refsec{robustness}.  

Homogenization has the potential to amplify bias; to standardize bias, compounding injustices rather than distributing them; and to amplify arbitrary exclusion \citep{creel2021, Gandy2021}. 
For example,~\citet{zhou2021} have argued that BERT encodes an Anglocentric similarity metric \textit{by default}, one that could be harmful if applied across contexts where foundation models are applied. 
The application of foundation models across domains has the potential to act as an epistemically and culturally homogenizing force, spreading one implicit perspective, often a socially dominant one, across multiple domains of application. 

Existing trends in standardization of training corpora are likely to be exacerbated  in foundation models due to the massive scale of both unlabeled and labeled data needed. 
To the extent that models train on similar data, they are likely to acquire similar patterns of behavior, biases (\refsec{fairness-sources}), and errors. 
Previous high-effort data curation and labeling efforts such as ImageNet have standardized training corpora. 
In doing so, they have also standardized errors: models trained on ImageNet often rely on the same ``spurious cues'' and ``shortcuts'', for example using background textures like green grass to predict foreground object classes such as cows \citep{geirhos2020shortcut, hendrycks2021natural}.  
Despite their increased robustness to many types of distribution shifts (\refsec{robustness-advantages}), foundation models and other large models have been no less likely to learn spurious correlations (\refsec{robustness-challenges}), and are therefore likely to learn similar errors if trained on the same datasets. Similar effects may arise due to the choice of publicly available unlabeled data.  
Many foundation models are trained on unlabeled corpora that are chosen for their convenience and accessibility, for example public internet data \citep{caswell2021}, rather than their quality. 
However, publicly accessible data, whether labeled or unlabeled, is often outweighed by proprietary data in the training corpora of many proprietary foundation models, as discussed in \citep{marr2017} and \refsec{data-desiderata}. 
Therefore more research is needed on the extent to which training on similar data homogenizes correlations within foundation models and the extent to which this homogenization might cause uniform failures in adapted derivatives of the model (unless constraints are applied to eliminate the behavior during each adaptation, as discussed in \refsec{adaptation-usecases}). 

Homogenization is not inevitable.  
As model developers intentionally broaden the range of perspectives represented in their datasets (\refsec{fairness-sources}), more research is needed on the capacity of foundation models to deliver a diversity of perspectives when used for generative tasks. 
For example, \citet{sheng2021revealing} have demonstrated that dialogue systems that adopt ``personas'' of specific demographic groups behave differently on measures of social bias. In addition to choosing between ``personas'' with the goal of avoiding bias, ``personas'' that are diverse along a variety of cognitive and demographic axes could also be used to generate a broader range of coherent outputs for generative tasks.  
There remain many open questions about how to balance diversity of outputs with relevance and utility to an individual user.\footnote{For possible approaches to implementation, see the discussions of controllable generation in \citep{keskar2019ctrl} and \refsec{adaptation-usecases} and general discussions in \citep{dinan21}.}

\hypertarget{ethics-surveillance}{\subsubsection{Surveillance, exclusion, and power}}
\label{sec:ethics-surveillance}

A key premise of foundation models is that massive unlabeled datasets can be combined with vast computational resources to create a basis from which numerous products can be derived for a variety of applications. This paradigm shift has the potential to alter social structures and shift power, establishing or entrenching the influence of model creators \citep{zimmerman_2020}. 
We discuss three potential implications below.

\paragraph{Mass data collection and surveillance.} 

Whereas collecting a labeled dataset typically requires working with domain experts and understanding the problems with and limitations of such data, the need for exceptionally large amounts of data in training foundation models has encouraged some researchers to emphasize \emph{quantity} rather than quality.\footnote{For example, \citet{ding.2021} collected 30 million text-image pairs, chose not to address artefacts such as watermarks and white edges, despite their impact on model quality.} 
Though preprocessing can help improve the quality of this data \citep[\eg][]{brown2020gpt3}, the scale involved necessitates automated approaches, which may be blunt or poorly documented \citep{dodge.2021.documenting}.

Although there is an evolving landscape of data protection legislation (\eg GDPR in Europe), a variety of questionable practices continue to be used in acquiring data, from opaque policies \citep{obar.2020} and the use of ``dark patterns'' (\ie manipulative interfaces \citep{narayanan.2020}) to outright violation of terms of service. 
Indeed, this was essentially the strategy taken by Clearview AI\dash{}a company which scraped photos from social media, without user consent, and in violation of platforms' terms of service, for the purpose of developing facial classification software. 
The company was nevertheless able to sell this technology to police departments and other organizations, in many cases without the knowledge of state lawmakers or department heads \citep{mac.2021}. 
To the extent that the paradigm of foundation models increases the value of being first to have the largest possible dataset for any particular domain, this may further encourage actors to pursue aggressive data collection, even when that pursuit is legally questionable or contrary to user expectations \citep{nissenbaum.2009, zuboff.2018}.

The importance of data to foundation models also means that organizations already engaged in widespread data collection will be in a strong position to develop such models, and will likely have incentive to maintain this advantage. 
To the extent that derivative products could themselves be used to collect additional data (\eg in surveillance or health diagnostic applications), developers of foundation models may seek to ensure that they obtain ownership of such data. 
Thus, even though a key advantage of the foundation model paradigm is the ability to generate adapted derivatives, the developers of foundation models might seek to license their work in a way that ensures that data flows back to them from all adapted derivatives.\footnote{As a less sophisticated example, consider the credit scoring industry, which has been able to position itself such that information flows back to central data brokers as people use its products (as in vetting loan applications), and individuals have little choice but to participate \citep{lauer.2017}.}

\paragraph{Concentration of power.} 

Although the absolute cost of computation has become dramatically cheaper over time, the training of the largest foundation models currently requires computational resources that put their development beyond the reach of all but a few institutions and organizations (\refsec{environment}). Thus, the question of who has access to the relevant computational resources and data will likely determine who is able to produce cutting-edge foundation models in the coming years (see also \refsec{economics-centralization}).

GPT-3 was at least partly an experiment in scale, showing that major gains could be achieved by scaling up the model size, amount of data, and training time, without major modeling innovations. 
Although there is extensive ongoing research into reducing the amount of resources required in training such models (see \refsec{training}), OpenAI's work suggests that there are still gains to be had from even larger scale efforts \citep{kaplan2020}, and it seems plausible that other organizations may seek to follow this path in other domains (for example, see \citep{jurassic1}).

If scale does turn out to be critical to success, the organizations most capable of producing competitive foundation models will be the most well-resourced:  venture-funded start-ups, already-dominant tech giants, and state governments.  This raises potential concerns about market concentration, and might indicate the kind of incumbent monopoly or oligopoly that currently exists in extreme capital-intensive industries such as defense and semi-conductor manufacturing \citep{carril.2020}.

Moreover, this centralization of power raises concerns about the ability of currently-marginalized individuals and communities to participate in the foundation model development process \citep{kalluri.2020}.  Especially within the realm of government services, the adoption of foundation models could further transfer decision making power from governments to corporate service providers, and introduce additional barriers to due process and accountability \citep{citron.2008}. Nevertheless, more grassroots efforts (\eg Masakhane, EleutherAI, HuggingFace) provide encouraging alternatives, and there is extensive work on ways to incorporate participatory or value-sensitive design  \citep{friedman.2019, prabhakaran.2020}.

\paragraph{Fueling widespread automated decision-making.}

Recent years have seen a dramatic expansion in the use of automated decision-making systems in industry and government \citep{oneil.2016, engstrom2020government}. 
Although many of the concerns over such automation are not specific to foundation models, the generative abilities of models such as GPT-3, as well as the impressive performance on benchmark tasks
(\eg \citet{devlin2019bert}), have the potential to prompt a less-than-careful adoption of this technology by, for example, administrative agencies, many of which lack the expertise necessary to understand sophisticated ML systems \citep{calo.2021}. 
As such, it is especially important to communicate clearly about the realistic capabilities and limitations of foundation models.

Most automated decision-making systems will exist as parts of broader sociotechnical systems in which humans play key roles \citep{selbst.2018}.\footnote{For an extended study of how humans interact with automated judgements, including discussion of both positive and negative automation biases, see \citet{hidalgo2021how}.} 
As such, there is no guarantee that even large improvements in performance on standardized evaluations will translate into the desired outcomes in the real world (especially if systems are deployed without careful consideration or ongoing evaluation).
For example, research has shown that judges may re-impose racial prejudice in interpreting the outputs of a risk assessment system \citep{albright.2019}, or otherwise impose their own biases \citep{stevenson.2021}. 
Ongoing evaluation with proper ecological validity \citep{vries2020ecological} will be critical in this regard, but may not stop potentially dangerous or costly systems from being adopted without adequate evidence \citep{ferguson.2017}. 
Research is ongoing on methods of refusal: ways for individuals to opt out of participation in foundation models and their adapted derivatives, either as data or decision subjects, without repercussions   \citep{benjamin_ruha_bioethics}.

In short, the existing problems with algorithmic decision making will be seen in the functioning of foundation models once they are deployed. And to the extent that adopting foundation models accelerates a shift from human to machine decision making, foundation models accentuate the concerns with automation. 
Although there are not obvious solutions to these challenges, it is important to make questions about how foundation models will impact power part of the conversation about their creation; to communicate with civil society organizations, policy makers, and citizens about the capabilities and limitations of such systems; and to strive for broader dialogue among diverse segments of society about the adoption of such models.

\subsubsection{Norms} 

Public policy and formal regulation by law (\refsec{legality}) play an essential role in creating the infrastructure for technological innovation as well as mitigating the potentially harmful effects of widely disseminated technologies. 
As illustrated by the decades-long gap between the Tuskegee Syphilis experiments and the development of research protocols and institutions like the IRB, public policy to protect human subjects and stakeholders tends to lag behind public awareness and evidence of harms to them \citep{Grady2015, stark_IRBs, belmont_report}. 
As a result, society relies upon professional norms for responsible development and deployment and the establishment of best practices.  

Norms exist on a continuum between \textit{recommendation} and \textit{requirement}. 
As a nascent technology, the norms for responsible foundation model development and deployment are not yet well established at either strength of recommendation \citep{lawfare_norms}. 
In what follows we will discuss norms for deployed models, as models for research have a wider latitude.

Those who wish developers of foundation models to adopt certain norms might lead by example, allowing their own conduct and statements to \textit{recommend} the norm. 
As discussed in  \refsec{ecosystem}, we believe that universities and other nonprofit institutions have an important role in modeling norms for foundation models. 
As educational institutions, universities are in the unique position to encourage the next generation of theorists and practitioners to consider the issues raised in this report and also to foster interdisciplinary conversation between researchers and students~\citep{rogers2021}. 
Universities and colleges may also contribute to the establishment of norms by auditing existing foundation models and publishing their findings, instituting ethics review boards \citep{bernstein_esr_2021}, and developing their own foundation models. 

To create and adopt norms will require institutionalization in funding structures, model repository, release practices, conference submission, and grant proposal requirements.\footnote{For helpful discussion of partial compliance with ``non-compulsory fairness-conscious policy'' such as the norms under discussion here, see \citet{Dai2021}.}  
For example, HuggingFace’s interface currently encourages the posting of data and model cards, including discussions of bias and social impact.\footnote{\url{https://huggingface.co/docs/datasets/master/}} 
Since it is not required, and perhaps since data quality work is undervalued relative to its importance \citep{sambasivan2021everyone}, few are filled out. 
Bias and social impact are included in ethics statements for conferences and some forms of standard evaluation (as discussed in \refsec{evaluation}), but otherwise treated as optional considerations by some researchers. This must change.

For some socially consequential use cases, we recommend legal standards be established that \textit{require} adapted derivatives to provably exhibit certain properties (\refsec{legality}).
Domains of special concern should be democratically decided but are likely to include allocating and distributing government services, medical diagnosis and monitoring, hiring, and lending: 
all contexts in which opportunities or even lives of people rest on the proper functioning of an adapted derivative. 

What norms should we promote, institutionalize, or require? 
We recommend a few here, but aim primarily to encourage dialogue about appropriate norms for the development and use of foundation models. 
Prior work has often focused on norms that advocate documentation \citep{gebru2018datasheets, bender2018data, Mitchell_2019, dodge2019work}. 
Because many of the negative social consequences that appear in a downstream context may initially appear to be \textit{extrinsic} or particular to a use case (\refsec{fairness}), documentation and transparency are especially important for foundation models. 
Currently, those who adapt foundation models that document the biases or other negative features of their adapted derivatives have no automatic mechanism to report their findings to the developers of the foundation model.  Compiling multiple reports of related problems in adapted derivatives may allow the model development team to discover an \textit{intrinsic} property of the model that spans multiple use cases. 
Because creators of adapted derivatives often represent different entities than from foundation model developers or providers, additional reporting structures and norms or regulation would be needed for this type of feedback to reach foundation model developers. 
Such feedback could also be made available to the general audience of model auditors, thereby making auditing and pursuing recourse more accessible.
 
Public commitment to norms, standards, and creation of reporting mechanisms could also allow downstream users to submit feedback to foundation model providers.
In order to enable this, adapted derivatives should be consistently labeled in a way that allows impacted parties to trace problems to their source. 
Significant technical and social barriers may impede this tracing in practice, such as privacy considerations and the proprietary nature of many foundation models, but without labeling it would be impossible.  

It is important that model developers and providers create mechanisms for such reporting.  Reporting mechanisms could be informed by similar structures on current platforms, such as issue tracking on open source projects on GitHub. In particular, the submitted issues should be public so that other users can identify trends even if changes have not yet been made and so that developers and providers can be held accountable for unaddressed issues. Additional mechanisms are needed to escalate trends upwards to foundation model providers. 
Similar suggestions regarding tracking issues in training data are discussed in \citet{dinan21} and \refsec{data}.

\citet{holland2018dataset} suggest 
the nutrition label as a helpful model, drawing from labeling discussions in consumer privacy \citep{kelley2009nutrition}.  A nutrition label includes both a list of the ``raw’’ ingredients and the full nutritional information of the processed food. So too a model card \citep{Mitchell_2019} or nutrition label for an adapted derivative could include both a list of the ``raw materials’’ such as training data and foundation models used, and the full ``nutritional content'' of the adapted derivative such as its known capacities, weaknesses, and biases.  

Reporting of the full pipeline is necessary in order for data subjects and impacted parties to trace harms to their sources. However, without the ability to \textit{attribute responsibility} for the harm to either the adapted derivative, the foundation model, or both, and without a framework for recourse once harm has been attributed, even a successful tracing of a harm will be unlikely to lead to changes in the model (see also \refsec{fairness-recourse}). Thus, significant technical, policy, and legal work is needed in order to develop frameworks for communicating data, model, and derivative contents to other experts and eventually to the public; to attribute responsibility for harms; and to create avenues for recourse. 

\subsubsection{Release and Auditing}

In February 2019, OpenAI embarked on an experiment.  By releasing a reduced 124M parameter GPT-2, sans datasets, they hoped to buy time: time to test for bias, time to prepare for misuse, and time for society to adapt to the presence of large language models \citep{solaiman_release_2019}. Eight months later, when OpenAI released the full $\sim$1.5 billion parameter version, testing had exposed some but by no means all of the model’s capabilities and limitations. 
When considering similar questions today, the possible harms of release, centering primarily on misuse (\refsec{misuse}),\footnote{For analysis of harms related to misuse, see \citep{Rini2017-fakenews} on fake news and \citep{Rini2020deepfakes} on deepfakes.} must be weighed against the benefit of transparency that no closed-door testing can replicate, namely broader and independent auditing and access.  

\paragraph{Auditing}  
Auditors probe the limitations of current models and suggest paths to fixing them, as well as testing the model's adapted derivatives in a wide variety of natural settings. 
A policy of open access for auditing allows more numerous and diverse researchers to investigate any model’s biases, limitations, and security vulnerabilities, better informing acceptable uses of the models and calibrating \textit{appropriate trust} in them \citep{Danks2019, Baier1986}.\footnote{Calibrating trust may require an explanation capable of illuminating features of the model relevant to trust, such as ``discriminatory use of a sensitive feature'' \citep{dimanov2020you}.}  In order to support independent audits of foundation models, model developers or third-party intermediaries could host open API access for auditors, including gradient access, and allow access to training data \citep{raji2019, Raji2020}. 

Foundation models trained on proprietary data in industry are unlikely to be released, and those trained on private data (as in a medical context) should not be. In order for proprietary models to benefit from independent audits, and for model subjects to benefit from improvements prompted by an auditing process, we recommend that audits occur during a staged release.  While staged release may not illuminate all possible model use cases, one way to broaden the range of uncovered use cases is to enlist a neutral third party to decide which individuals or organizations should receive early access in the staged-release program.
When model developers decide who should receive staged access, they open themselves up to charges of favoritism, selective distribution, and manipulating public perception of their product. 
A neutral “staged release board”, or federal auditors, could provide a backstop against these failure modes and ensure that a wide range of auditors and users are provided access in order to capture a range of disciplinary expertise and sectors of society. 
A staged release board could also mitigate any perception that auditors would be at risk of losing their early access to the model if they share unflattering outputs, as they might be in a standard staged release process. 

\paragraph{Access and adaptation.} 

To the extent that there are social benefits to foundation models, release of models holds the potential to further distribute them. Large language models such as BERT and M-BERT are capable of cross-lingual transfer, which\dash{}when the models are open-sourced\dash{}may allow for adaptation to languages which otherwise would have too few texts available \citep{wu-dredze-2019-beto, wang_extending_2020}. 
Given the number of languages not currently well served by commercial providers, such a benefit alone could be substantial.   

Release is not sufficient to democratize access to foundation models, as the barrier of compute power still precludes many from modifying or even loading foundation models, let alone developing their own. 
However, on each of these points we have seen significant recent technical  improvement. 
Memory techniques such as the zero redundant optimizer (ZeRO) allow researchers to run and train very large models on a simple setup \citep{rasley2020deepspeed, rajbhandari2021zeroinfinity}.
Techniques such as distillation could allow the release of smaller, more tractable models that recoup much of the performance of their parent model while being much easier to train \citep{li2020train}. 
Development of less energy-intensive training methods, as discussed in \refsec{environment}, could further spread the ability to work with released models. 
Increases in efficiency such as the co-design of hardware and software are needed to train yet larger models, as discussed in \refsec{systems}, but could also be used to lower the price of access to current models.  
 
The most powerful of the harms, by contrast, are not obviously fueled by release.   
Sophisticated or institutional actors with the capacity to embark on large-scale disinformation, cyberwarfare, or targeted phishing also are likely to have the capacity to create a similar model if none were released.  
Although potentially significant, these harms should not therefore weight heavily on a release calculus \citep{solaiman_release_2019, shevlane_offense-defense_2020}.  
The harms to be weighed against the benefits are those from less well-resourced actors who would not be able to create their own foundation model but may be motivated to generate spam or abuse, fake reviews, or cheat on tests.
Does the benefit of release outweigh the potential for harm from actors sophisticated enough to use a released model or API but not sophisticated enough to create their own? We believe that the answer is yes. 
Research teams with the resources and connections necessary to develop foundation models are few in number.  Even collectively, we are unlikely to be numerous or diverse enough to imagine all possible beneficial use cases or all possible probes that could illuminate the capability surface of a foundation model.

\hypertarget{ethics-nottobuild}{\subsubsection{When not to build}} 
\label{sec:ethics-nottobuild}

The development and deployment of powerful technologies is not like gravity, an external force that acts upon us.
Technologies reflect a set of choices made by humans; human agency shapes the technological frontier.
It follows that technologists can choose when not to build, design, or deploy foundation models \citep{zimmermann_stop_2021}. 
This decision need not be binary; instead, one can refuse to engage in the default way by subverting embedded values, challenging assumptions, and shaping research agendas \citep{audra_simpson}. 
Technical artifacts, foundation models included, are inherently political, so the research about them has a socio-political context, not solely a technical one. Developers and researchers should be cognizant of which problems they seek to address, \eg~how to scale up a foundation model versus how to make it more computationally accessible; how those problems are formulated; and who their solutions ultimately empower \citep{rogaway_moral_nodate, winner_artifacts_1980, Passi2019}. We should value research that seeks to make foundation models more interpretable, accessible, sustainable, and fair (see \refsec{interpretability}, \refsec{environment}, \refsec{fairness}). 

By asking when not to build a foundation model or adapted derivative, we are implicitly asking not only ``What should we build or not build?'' but also, ``Under what conditions should a model be built?'' and ``What criteria and principles govern building?'' The first question stems from the model view; the following questions from the ecosystem view (\refsec{introduction}).

An invitation to consider refusing to build is not tantamount to saying, “Do nothing.” It is an invitation to make deliberate and judicious choices about what is worth the time, financial resources, expertise, and energy use to build, design, and deploy. Ultimately, this is a difficult, moral question rooted in context and values. 
There are cases in which the application of adaptive derivatives (and algorithms and machine learning more generally) is inappropriate, because the community impacted protests or because the adaptive derivative naively exacerbates systemic issues that are better addressed with public policy, additional funding, or interdisciplinary collaborations \citep{COMPAS_propublica}.

The Belmont Report, as applied to machine learning in \citet{Floridi2018}, provides one possible framework for this question. Drawing from the principle of "beneficence" \citep{belmont_report}, we can identify cases to reconsider building when an adaptive derivative or a research avenue might cause more harm than good or even provide no benefit at all. Alternatively, there may be cases in which an adaptive derivative is better at a task on a metric of efficiency, performance, and generalization, values prioritized in the machine learning community \citep{birhane2020}, but an individual, community, or organization might choose to prioritize an existing solution that highlights other values such as human connection and  interpretability \citep{benjamin_ruha_bioethics}.\footnote{See also \refsec{interpretability-impacts} for relevant discussion of impacts of uninterpretability.} In doing so, they exercise their autonomy\dash{}as explained in the Belmont Report's "respect for persons"\dash{}in deciding that this is not an appropriate context in which to build \citep{belmont_report}. 

Answering the question of when not to build is a matter of individual responsibility as well as a broader professional responsibility. 
The decision not to build something by one person, or one team, or one company, invites the reply, “But if we don’t build this, someone else will, and they may likely do it worse.”  
A simple utilitarian weighing of comparative harms of the outcomes of the two models misses the importance of integrity. 
It matters very much whether \textit{we} are the ones building the bad model or whether someone else is \citep{Williams1973}.
Individuals have reasons not to build something that goes against their values or that they cannot endorse as right to build \citep{Korsgaard2009}. However, the structural environment so created is different. If even one company decides to build the most effective version of an ethically-dubious model, they open the door for other companies to consider similar avenues of research; they make it competitively disadvantageous not to pursue the research \citep{askell_role_2019}. 
When not to build is then a collective question as much as it is an individual one, requiring the community to adhere to codes of professional ethics and responsibility. 

In the AI/ML community this infrastructure is underdeveloped compared to other fields such as the medical field. Although professional bodies like the Association for Computing Machinery (ACM) have ethics statements, both industry and academia lack widely used and accepted professional oaths (\eg the Hippocratic Oath or the the Obligation of the Engineer), regulatory bodies involved in deployment and research (\eg the FDA for drugs), and official protocols for ethics review (\eg the IRB for research involving human subjects; \citep{bernstein_esr_2021}). 
The ability to opt-out can be incorporated into the foundation model ecosystem at many stages, including during data production, adaptation, and deployment. As the norm veers towards collecting larger and larger swaths of training data (\refsec{data}), we should endeavor to maintain a "respect for persons," \citep{belmont_report} emphasizing privacy and consent as part of the data life cycle. 
This would require innovation in data management and a more concrete understanding\dash{}technically and philosophically\dash{}of informed consent online, ways of documenting and ensuring that consent is respected, and privacy (see \refsec{data} for a specific data management proposal; \citep{changingtherules_ohm}). 
Although data and foundation models are diverse in their applications, data participants should be able to indicate how they do not want to have their data used. An opt-out consent model favors developers, as it does not require them to to get consent for each new, unexpected use case. Important then is the right to revoke consent given vacuously for applications that are now being pursued, but were not when consent was originally given. 

\subsubsection{Conclusion} In this section, we have surveyed some of the risks to society that accompany the widespread adoption of foundation models, such as the homogenization of outcomes and centralization of power. Developers of foundation models should adopt norms regarding development, auditing, and release of foundation models in order to address these risks, aided by legislative requirements, and individuals should be able refuse to be data or decision subjects of foundations models without repercussion.

Many implications of foundation models' generative and interactive capacities remain unsurveyed here.  
For example, \refsec{economics} discusses the potential gains to economic productivity from the automation of creative and design work.  
However, in virtue of their generative nature, foundation models may replace work that many people find meaningful and fulfilling, such as graphic design and writing. 
We hope that the scope of this report will aid others in their pursuit of the questions of ethics and society unaddressed here. 

\clearpage
\hypertarget{conclusion}{\section{Conclusion}}
\label{sec:conclusion}
In this report, we have endeavored to comprehensively discuss many of the most critical aspects of foundation models, ranging from their technical underpinnings to their societal consequences.
In this way, we acknowledge the unusual approach taken: we have attempted to clarify the nature of a paradigm that may only have just begun, rather than waiting for more to unfold or the dust to settle.
Therefore, much still remains unclear in spite of our efforts and we reiterate that this is just the beginning of a paradigm shift: foundation models have only just begun to transform the way AI systems are built and deployed in the world.
Moving forward, we view this document as serving an important role in orienting and framing dialogue on these models and this new paradigm in AI. 
That said, to ensure the responsible development and deployment of these models on durable foundations, we envision collaboration between different sectors, institutions, and disciplines from the onset to be especially critical. 

\begin{acks}
We would like to thank the following people for their valuable feedback:
Mohit Bansal,
Boaz Barak,
Yoshua Bengio,
Stella Biderman, 
Su Lin Blodgett,
Sam Bowman,
Collin Burns,
Nicholas Carlini,
David Chalmers,
Jack Clark,
Jeff Dean,
Jesse Dodge,
Jarred Dunnmon,
Gabe Dupre,
Jason Eisner,
Iason Gabriel,
Dan Hendrycks,
Avery Hill,
Yacine Jernite,
Gabbrielle Johnson,
Sarah Kreps,
Jay McClelland,
Preetum Nakkiran,
Julian Nyarko,
Fernando Pereira,
Vinodkumar Prabhakaran,
Colin Raffel,
Marten van Schijndel,
Ludwig Schmidt,
Yoav Shoham,
Madalsa Singh,
Megha Srivastava,
Jacob Steinhardt,
Emma Strubell,
Qian Yang,
Luke Zettlemoyer,
and
Ruiqi Zhong.
In addition, we would like to especially thank Vanessa Parli for helping to organize this effort.
\end{acks}

\section*{Conflict of Interest}
This report was authored by the Center for Research on Foundation Models (CRFM), a center at Stanford University borne out of the Stanford Institute for Human-Centered Artificial Intelligence (HAI).
CRFM receives funding from Google, Microsoft, and the McGovern Foundation as of July 2022, though this funding was not directly related to this report.
Authors of this report may also be affiliated with other institutions beyond Stanford: their contributions reflect only their views and not those of these institutions. 

\bibliography{main_without_refdb,refdb/all}

\bibliographystyle{ACM-Reference-Format}



\end{document}